# Continual Learning with Deep Learning Methods in an Application-Oriented Context

# Dissertation

*for the attainment of the degree*
**Doctor rerum naturalium (Dr. rer. nat.)**

from
Benedikt Simon Pfülb, M. Sc.

Accepted as dissertation from the Promotionszentrum Angewandte Informatik of the Hessian Universities of Applied Sciences

**Supervisor:**  Prof. Dr. Alexander Gepperth
Fulda University of Applied Sciences
Leipziger Straße 123
36037 Fulda
Germany

**Co-Supervisor:**  Prof. Dr. Sebastian Rieger
Fulda University of Applied Sciences
Leipziger Straße 123
36037 Fulda
Germany

**First Reviewer:**  Prof. Dr. Ute Bauer-Wersing
Frankfurt University of Applied Sciences
Nibelungenplatz 1
60318 Frankfurt am Main
Germany

**Second Reviewer:**  Prof. Dr. David Filliat
U2IS - ENSTA Paris
828, boulevard des Maréchaux
91762 PALAISEAU CEDEX
France

**Partner:**  Hochschule Fulda
Frankfurt University of Applied Sciences
Hochschule Darmstadt
Hochschule RheinMain

# Abstract

Abstract knowledge is deeply grounded in many computer-based applications. An important research area of Artificial Intelligence (AI) deals with the automatic derivation of knowledge from data. Machine learning offers the according algorithms. One area of research focuses on the development of biologically inspired learning algorithms. The respective machine learning methods are based on neurological concepts so that they can systematically derive knowledge from data and store it. One type of machine learning algorithms that can be categorized as *deep learning* model is referred to as Deep Neural Networks (DNNs). DNNs consist of multiple artificial neurons arranged in layers that are trained by using the *backpropagation* algorithm. These deep learning methods exhibit amazing capabilities for inferring and storing complex knowledge from high-dimensional data.

However, DNNs are affected by a problem that prevents new knowledge from being added to an existing base. The ability to continuously accumulate knowledge is an important factor that contributed to evolution and is therefore a prerequisite for the development of strong AIs. The so-called "catastrophic forgetting" (CF) effect causes DNNs to immediately loose already derived knowledge after a few training iterations on a new data distribution. Only an energetically expensive retraining with the joint data distribution of past and new data enables the abstraction of the entire new set of knowledge. In order to counteract the effect, various techniques have been and are still being developed with the goal to mitigate or even solve the CF problem. These published CF avoidance studies usually imply the effectiveness of their approaches for various continual learning tasks.

This dissertation is set in the context of continual machine learning with *deep learning* methods. The first part deals with the development of an application-oriented real-world evaluation protocol which can be used to investigate different machine learning models with regard to the suppression of the CF effect. In the second part, a comprehensive study indicates that under the application-oriented requirements none of the investigated models can exhibit satisfactory continual learning results. In the third part, a novel deep learning model is presented which is referred to as Deep Convolutional Gaussian Mixture Models (DCGMMs). DCGMMs build upon the unsupervised approach of Gaussian Mixture Models (GMMs). GMMs cannot be considered as deep learning method and they have to be initialized in a data-driven manner before training. These aspects limit the use of GMMs in continual learning scenarios.

The training procedure proposed in this work enables the training of GMMs by using Stochastic Gradient Descent (SGD) (as applied to DNNs). The integrated *annealing* scheme solves the problem of a data-driven initialization, which has been a prerequisite for GMM training. It is experimentally proven that the novel training method enables equivalent results compared to conventional methods without iterating their disadvantages. Another innovation is the arrangement of GMMs in form of layers, which is similar to DNNs. The transformation of GMMs into layers enables the combination with existing layer types and thus the construction of deep architectures, which can derive more complex knowledge with less resources.

In the final part of this work, the DCGMM model is examined with regard to its continual learning capabilities. In this context, a replay approach referred to as Gaussian Mixture Replay (GMR) is introduced. GMR describes the generation and replay of data samples by utilizing the DCGMM functionalities. Comparisons with existing CF avoidance models show that similar continual learning results can be achieved by using GMR under application-oriented conditions. All in all, the presented work implies that the identified application-oriented requirements are still an open issue with respect to "applied" continual learning research approaches. In addition, the novel deep learning model provides an interesting starting point for many other research areas.

# Zusammenfassung

Abstraktes Wissen ist in vielen computergestützten Anwendungen fest verankert. Ein wichtiger Forschungsbereich der Künstlichen Intelligenz (KI) beschäftigt sich mit dem automatischen Ableiten von Wissen aus Daten. Maschinelle Lernverfahren bieten diesbezüglich die grundlegenden Algorithmen an. Ein Forschungsbereich befasst sich mit der Entwicklung von biologisch inspirierten Lernverfahren. Derartige maschinelle Lernverfahren basieren auf neurologischen Konzepten, um Wissen aus Daten systematisch ableiten und speichern zu können. Eine Art davon, die unter die Kategorie der *deep learning* Modelle fällt, wird als Künstliches Neuronales Netz (KNN) bezeichnet. KNNs bestehen aus mehreren künstlichen Neuronen, die in Schichten angeordnet sind und mit Hilfe des *Backpropagation* Algorithmus trainiert werden. Derartige deep learning Verfahren weisen erstaunliche Fähigkeiten auf, um komplexes Wissen aus hochdimensionalen Daten ableiten und speichern zu können.

Nichtsdestotrotz sind KNNs von einem Problem betroffen, welches das Hinzufügen von neuem Wissen zu bestehendem verhindert. Die Fähigkeit, kontinuierlich Wissen anhäufen zu können, war ein wichtiger Faktor für die menschliche Evolution und demnach eine Voraussetzung für die Entwicklung starker KIs. Der sogenannte "catastrophic forgetting" (CF) Effekt führt bei KNNs dazu, dass bereits abgeleitetes Wissen nach wenigen Trainingsiterationen auf einer neuen Datenverteilung unmittelbar verloren geht. Lediglich das energetisch aufwendige erneute Training mit der vereinten Datenverteilung von vergangenen und neuen Daten ermöglicht die Abstraktion des gesamten Wissens. Um dem Effekt des Vergessens entgegenzuwirken, wurden und werden verschiedenartige Ansätze vorgeschlagen, die das katastrophale Vergessen abmildern oder sogar lösen sollen. Veröffentlichte Studien derartiger Modelle bestärken die Wirksamkeit der Lernverfahren für unterschiedliche kontinuierliche Lernaufgaben.

Diese Dissertation steht im Kontext des kontinuierlichen maschinellen Lernens mit *deep learning* Verfahren. Der erste Teil befasst sich mit der Entwicklung eines anwendungsorientierten Evaluationsprotokolls, mit dessen Hilfe verschiedene Modelle auf die Unterdrückung des CF Effekt untersucht werden können. Im zweiten Teil folgt eine umfassende Untersuchung, die zeigt, dass unter den anwendungsorientierten Anforderungen keines der untersuchten Modelle zufriedenstellende kontinuierliche Lernergebnisse vorweisen kann. Im dritten Teil wird ein neuartiges deep learning Modell vorgestellt, was als Deep Convolutional Gaussian Mixture Models (DCGMMs) bezeichnet wird. Grundsätzlich bauen DCGMMs auf dem unüberwachten Ansatz der Gaussian Mixture Models (GMMs) auf. GMMs zählen nicht zu den deep learning Modellen und müssen für das Training zunächst datengetrieben initialisiert werden. Diese Nachteile erschweren den Einsatz von GMMs in kontinuierlichen Lernszenarien.

Das vorgeschlagene Trainingsverfahren ermöglicht das Training von GMMs mittels Stochastic Gradient Descent (SGD) (wie bei KNNs). Zudem löst die *annealing*-Methode das Problem der datengetriebenen Initialisierung, welche bisher für das Training von GMMs vorausgesetzt wird. Es wird explorativ gezeigt, dass das neuartige Trainingsverfahren im Vergleich zu herkömmlichen Methoden gleichwertige Ergebnisse ermöglicht, ohne dessen Nachteile. Ein weiteres Novum besteht in der Anordnung von GMMs in Schichten, ähnlich wie sie bei KNNs aufzufinden sind. Die Transformation von GMMs in Schichten ermöglicht die Kombination mit existierenden Schichtarten und somit die Konstruktion von tiefen Modellen, welche komplexeres Wissen mit weniger Ressourcen ableiten können.

Im abschließenden Teil der Arbeit wird das DCGMM Modell auf die kontinuierlichen Lernfähigkeiten hin untersucht. Hierfür wird ein replay-Ansatz vorgeschlagen, der als Gaussian Mixture Replay (GMR) bezeichnet wird. GMR beschreibt das Generieren und Wiedereinspielen von Datenpunkten durch die Nutzung der DCGMM Funktionalitäten. Vergleiche mit existierenden CF Unterdrückungsmodellen zeigen, dass mittels GMR ähnliche kontinuierliche Lernergebnisse unter anwendungsorientierten Bedingungen erzielt werden können. Die vorliegende Dissertation verweist abschließend darauf, dass die identifizierten anwendungsorientierten Anforderungen noch ein offenes Thema in Bezug auf Forschungsansätze zum kontinuierlichen Lernen darstellen. Darüber hinaus bietet das neuartige Modell einen wesentlichen Ansatzpunkt für weitere Anwendungsbereiche und zukünftige Forschungsfragen.

# Table of Contents

# 1.  Introduction

## Chapter Contents

Human's ability to learn has been one of the decisive factors for evolution. Men and women learned to speak, use tools, invented writing systems and used punched cards to process data. In today's society which is coined by technology, the challenge is to transfer this ability to the computer. Computers are used in almost all areas of our daily life: They have to make decisions in the shortest time possible on the basis of constantly increasing amounts of data. However, several questions arise regarding the knowledge basis of decisions and how knowledge is obtained. In the past, teams of experts developed problem-specific algorithms as a solution, whereas the algorithms were based on selected criteria. Increasingly complex problems and huge amounts of data result in the economic inefficiency of this approach. This includes, for instance, biological vision and interpretation, genome sequencing and more. Even though it may be easy for humans to distinguish the entities illustrated in figure 1.1, describing them in terms of an algorithm by hand can be difficult.

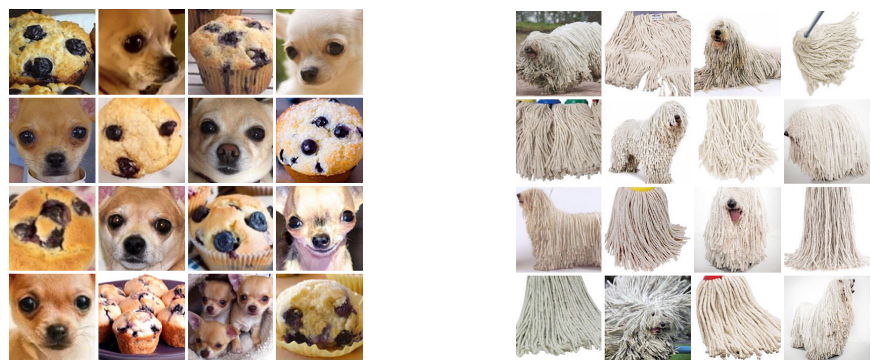

Figure 1.1: Example of visual classification problems.

Today's *machine learning* (ML) methods are predestined for this task as they can automatically derive knowledge with the help of exemplary data. The ML approach is part of the *artificial intelligence* (AI) research area. ML partially replaces the expert teams by deriving abstract models from data. Thus, knowledge is no longer encoded by hand in the algorithms but instead via an (abstract) knowledge base. It does not matter what the data looks like initially and what problem it is supposed to represent. Huge amounts of data need to be processed in form of numbers which can be considered too big a challenge for humans.

One area of AI research is Artificial Neural Networks (ANNs), which has been subject to research since the early 1940s. However, it was the development of faster and highly parallel hardware that enabled the efficient use of this technology. The idea of ANN results from observations in nature. ANNs mimic the biological nervous system of animals (including humans) and usually consist of many individual connected artificial neurons. These types of ML models have many advantages over conventional methods and are particularly powerful. They take their performance from stacking multiple layers which they refer to as "deep" learning methods. Deep Neural Networks (DNNs), for example, are among this method. The efficiency of DNNs has already been proven in a wide range of applications. Due to their structure, highly complex functions can be approximated. At the same time, the used training procedure (Stochastic Gradient Descent (SGD)) results in its attractiveness for many application scenarios.

Yet another crucial factor for evolution has been the ability to accumulate knowledge. Learning step by step and thus the gradual accumulation of knowledge without forgetting prior concepts is fundamental. This is one of the foundations of the present work: The so called *continual learning* (CL) or life-long learning paradigm. In addition to reproducing biological mechanisms for learning, the challenge for many technological applications is to adapt existing and add new knowledge. The same applies to ML models and in particular DNNs.

## 1.1   Problem Statement and Motivation

A well-known and common problem with some ML models is the gradual addition of knowledge. Once a learning process with a set of data is complete, it is difficult for many models to add new and somehow different knowledge. A simplified example is a student who learns the letter *A* in the first lesson and then the number *1* in the second. In a figurative sense, the brain as one and the same model first constructs the knowledge of the letter. As a next step, the knowledge regarding the number is added. It must be taken into account that 1) The knowledge is stored in one and the same medium – the brain – and 2) The previous knowledge is available after having learning the second piece of knowledge. Transferring this learning process to ML, the following aspects can be noted: 1) The existing (not more) memory must be used to store all knowledge and 2) The required computational capacity must not be dependent on prior knowledge. The described scenario is known as CL or life-long learning scenario.

Many deep learning models such as DNNs can address very complex problems. However, models based on DNNs suffer from various disadvantages. They are affected by the so called *catastrophic forgetting* (CF) effect, which describes the loss of all existing knowledge as soon as new knowledge is added (see French 1997; McCloskey and Cohen 1989). The CF effect is often related to the *stability-plasticity dilemma*. The latter describes the necessity of plasticity to be able to add new knowledge. At the same time, the ML model must provide stability in order to preserve existing knowledge. As an example, the first sub-problem of figure 1.1 is reconsidered. Here, objects in images should be discriminated. Chihuahuas and muffins are depicted and it is interesting to note that they have amazing similarities from the human perspective. It is, therefore, the first task to train an ML model with regard to the distinction of these two types of objects. Figure 1.2 shows an initially empty ML model on the left-hand side. The model is trained with labeled data, i.e., the Chihuahua class in red and muffin class in blue. The result of the training process is a trained model that should be able to distinguish these two types of entities.

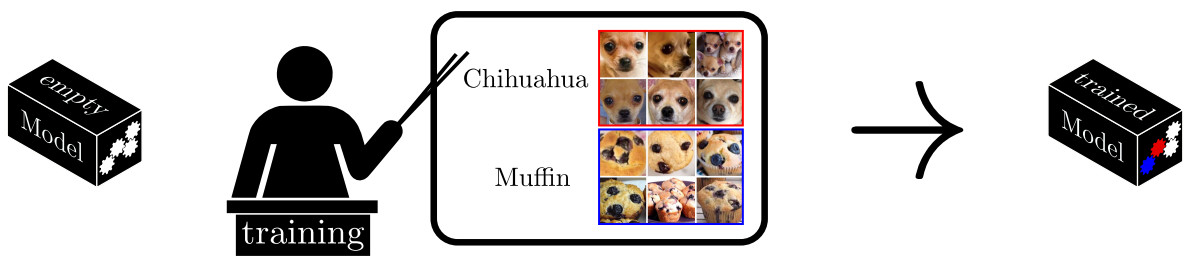

Figure 1.2: Initial training of an ML model.

In order to evaluate the functionality of an ML model, test data is used. Test data are often taken from the original set of training data, but are separated for testing. As a result, the model can be verified with previously unknown but authentic data. As illustrated in figure 1.3, the object type assigned by the ML model matches the reference value. Thus, the model can distinguish Chihuahuas from muffins.

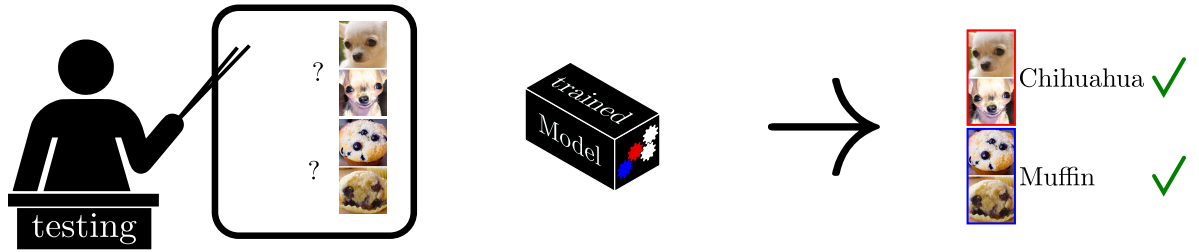

Figure 1.3: Evaluation of the initially trained ML model.

Even though the present model works very well, two additional object types should be distinguishable. This is comparable to the step by step addition of new knowledge in the human brain. For this purpose, new data is collected and again, the pre-trained model is re-trained with the new data (see figure 1.4). Images of sheepdogs are supposed to be distinguished from the presented images of mobs.

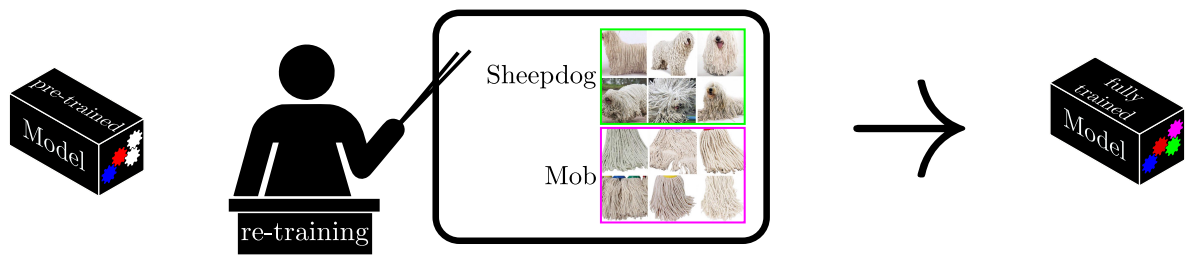

Figure 1.4: Adding additional knowledge to a pre-trained ML model.

Again, the model is evaluated with test data showing that the model has learned the difference. Figure 1.5 depicts the test results. It is obvious that the model works well, which is consistent with the CL paradigm. Consequently, it is assumed that knowledge can be accumulated task by task within the same ML model.

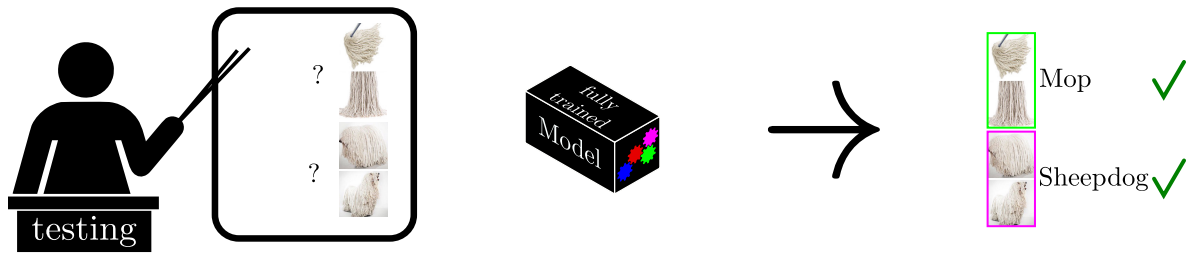

Figure 1.5: Evaluation of the re-trained ML model.

However, when the performance of the fully trained model is re-evaluated with regard to the first task, all test samples are incorrectly classified (see figure 1.6). This is also true when using the original training data for testing. In fact, the retrospective addition of new knowledge during the second training process leads to the loss of the previously learned knowledge. For this reason, continual learning processes become impossible.

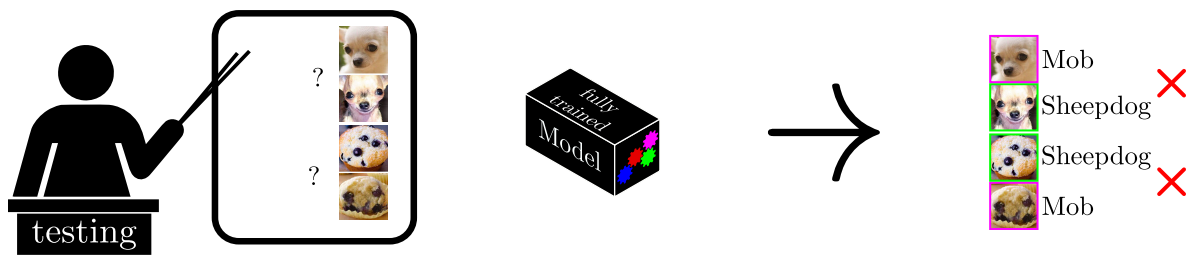

Figure 1.6: Re-evaluation of the re-trained ML model.

This problem can be circumvented by combining the training data of both tasks in the given example and (re-)training the model (see figure 1.7 and figure 1.8). Interestingly, a high quality ML model can be derived without adding additional resources, i.e., memory. However, the joint training approach has some significant disadvantages. First of all, the initially learned data has to be available and it needs to be combined with the new data. Furthermore, the time- and energy-consuming training process must be carried out from scratch again. This process can take up to several weeks, depending on the complexity of the ML model, the amount and structure of the data.

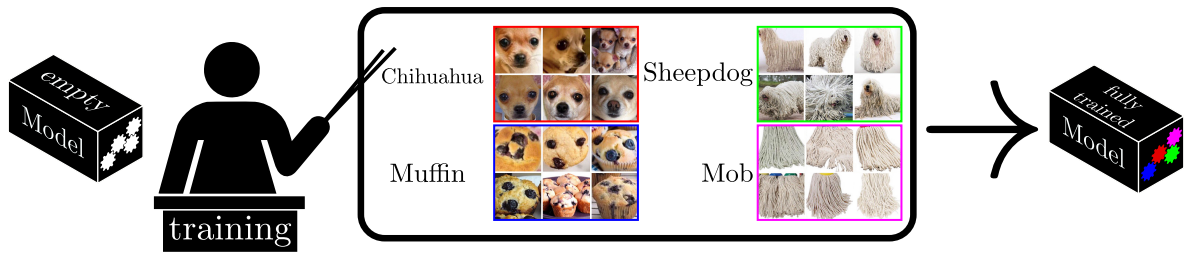

Figure 1.7: Joint training of an ML model.

The joint training approach becomes even more problematic when the size of the datasets grows infinitely, which usually applies to data streams. Devices such as embedded systems with very limited memory are particularly affected. The same is true for very large amounts of data that would otherwise require excessive memory capacities. In many application scenarios, the joint training is impossible or undesirable for data protection reasons. One example of a future application is an autonomous robot that is equipped with a human/pet recognizer by default. After being delivered to the customer the robot should then learn to distinguish recognize family members. However, privacy issues may also prohibit the joint training of an ML model. Thus, knowledge needs to be added in a consecutive training step.

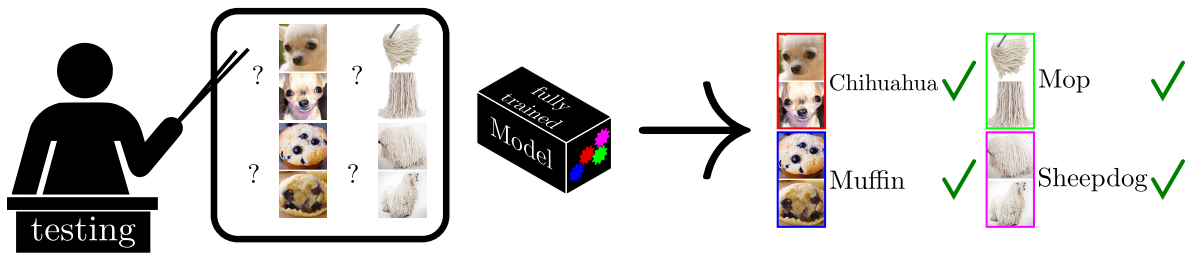

Figure 1.8: Testing of joint trained ML model.

A large scientific community has been addressing this very problem of the CF effect for several decades. CF is often evaluated from a theoretical perspective, resulting in the presentation of new approaches to circumvent the CF effect. Unfortunately, "clean-room conditions" are often assumed, under which the effect is greatly simplified or easily bypassed. This is especially true for the evaluation of the theoretically CF-immune models. The main problem that arises is that requirements from real-world scenarios are often neglected or ignored. Looking into future data is, for example, not possible in a real-world application. Furthermore, so called oracles are sometimes used as integral part of CF-immune models. Oracles simply provide additional information that is unavailable or not existent in certain scenarios. Consequently, the CF effect cannot be avoided without the application of an oracle.

To conclude, it is the fundamental motivation of the present work to investigate the CF effect in ML models such as DNNs under real-world conditions. Avoiding the CF effect can result in the application of these ML models in more real-world CL scenarios. In addition to new application areas, training time and the associated amount of energy can also be reduced. Yet another benefit of intentionally controlling the forgetting is related to ethical aspects of ML. This way, challenging, outdated or incorrect knowledge can be removed in a systematic manner.

## 1.2   Research Questions

The following research questions result from the initial problem description and motivation stated in section 1.1. Some MLs models, and DNNs in particular, are subject to the CF effect. DNNs and their variants are very powerful, and they can model very complex problems. They do, however, fail as soon as new knowledge is added after an initial training phase. The consequences of this effect are particularly serious in real-world applications. The so called *continual learning* (CL) scenarios are among them, as new knowledge is usually added step by step. Considering this scenario and the related challenges, the following research questions can be derived.

**RQ 1:**   *How can an application-oriented validation protocol be modeled to detect catastrophic forgetting?*
The answer to this research question aims at providing a valid CL investigation protocol, which makes the CF effect detectable under real-world conditions. As a next step, different deep learning models are examined with regard to the occurrence of the CF effect by using one and the same procedure. This ensures comparability between the models and allows for a ranking.

**RQ 2:**   *To what extent can existing machine learning models avoid the catastrophic forgetting effect?*
The second research question examines the validity of existing CF avoidance models by using the evaluation method developed as a result of RQ 1. In order to answer RQ 2, empirical research is required to investigate the specialized CF avoidance models under the CL protocol. In this context, various learning problems will be investigated by using different datasets. Conducting numerous experiments and parameter variations is assumed to prevent model parameters from affecting the CL performance. At the same time, all of the presented learning problems have to be "solved" in order to ensure that this particular procedure avoids the CF effect.

**RQ 3:**   *How can a novel deep learning model be designed so that it avoids catastrophic forgetting?*
This question aims to extend/adapt an existing model or develop a novel ML model so that the CF effect becomes controllable. The findings from RQ 2 can be used as a basis for the further development of extensions or enhancements. Alternatively, a new approach may emerge as an answer to RQ 3. In this respect, existing non deep learning models are considered helpful. The crucial question is whether they can be effectively layered/stacked in order to develop a true deep learning method with an appropriate CL performance.

**RQ 4:**   *How does the novel model perform when compared to existing continual learning models?*
The final question is dedicated to the impact of the resulting/new model on the CF effect. In this context, the CL performance is compared with those of other models. For this purpose, an (re-)evaluation will be performed by using the evaluation protocol developed for RQ 1.

## 1.3   Innovative Aspects

An important part of this work is the development of a universal and unified protocol to detect the CF effect. The main focus is on the uniformity and traceability of the protocol and the resulting comparability of individual procedures. At the same time, potential weaknesses in existing detection methods are considered for real-world applications. Evaluations of novel CF avoidance models often present results indicating that the CF problem is "solved". While this may be true for individual investigations, it does not mean that the problem is generally solved, or solved under application-oriented conditions. This distinction of scenarios makes the comparison of models and the associated areas of applications very difficult, if not impossible. This work aims to address these requirements for real-world applications resulting in a framework to investigate further CF avoidance models. At the same time, the theoretical investigation is shifted towards a more application-oriented scenario and the respective requirements. The goal is to derive implications and solutions of the CF problem in real-world scenarios. This perspective may be considered crucial for industry or other application contexts in which ML methods are relevant for CL scenarios. Likewise, a set of requirements can be beneficial in order to align certain capabilities to the respective ML models. This overview can help increase the comparability among the various CF avoidance models and, if necessary, support the selection of an adequate model.

The novel deep learning model is yet another innovation which is supposed to address the CF effect. With a higher or equal CL performance compared to other models, the newly developed model may be suitable for other and new application areas. Moreover, this model can be used as a basis for further improvements and adjustments. A novel deep learning model generally offers many starting points for new research projects of, both, theoretical and practical nature. Even though a direct comparison of the new model's performance with DNNs may be out of proportion, the research potential is quite considerable. The possible applications in non-CL scenarios may be of particular interest.

## 1.4    Structure of the Work

The first part of this dissertation is the chapter on fundamentals (chapter 2). This chapter introduces the generally used notations in this work. Moreover, relevant definitions of terms are briefly elaborated. The fundamentals are followed by chapter 3 which addresses related work. Current research is presented which is dedicated to "catastrophic forgetting" (CF) and "continual learning" (CL). This includes, for example, research with corresponding machine learning models addressing these problems. Chapter 4 clarifies the general procedure used to answer the four main research questions. It also identifies the connections between the research questions stated in section 1.2 in detail. Accordingly, the main questions are specified. In addition, the tools used for the conduction of the experiments are described. As part of chapter 5, an exemplary real-world scenario is presented. This exemplary real-world scenario is taken from the research field of computer communication networks. As a result, the requirements of CL applications in this scenario will be derived. Upon the introductory chapters, the specific research questions are examined. Chapter 6 discusses the first research question (RQ 1). The first question to be answered focuses on how the CF effect can be detected and confirmed. In order to develop the investigation protocol, other protocols from related works are considered (chapter 3). In addition, the requirements from the real world scenario (chapter 5) serve as basis. The result is a valid test protocol for detecting and comparing the CF effect. Chapter 7 addresses the second research question (RQ 2). For this purpose, a large-scale investigation is carried out based on the test protocol described in chapter 6. Some of the models presented in chapter 3 are examined with this evaluation protocol. The aim of this study is to draw conclusions about the individual model's performance with respect to the CF effect. At the same time, the study may identify promising approaches or techniques that can be used for further improvements. Chapter 8 serves as an answer to the third research question (RQ 3), Therefore, the novel model or technique that is supposed to control the CF effect is introduced in its basic version. That is due to the fact that the model is initially not reviewed in the context of CL. In chapter 9 an extension of the newly developed model in the context of CL is presented. Moreover, the novel model, among others, is re-evaluated using the investigation method from chapter 6. As the results of each research step are presented and discussed chapter wise, chapter 10 serves as a final discussion. In addition, the findings are summarized in order to answer the research questions. The dissertation is closed with a brief summary of contents in chapter 11. In addition, final conclusions from the entire work are presented. Last but not least, emerging aspects for future research are described. These open issues can be used for the continuation of the present work.

# 2.  Foundations of Continual Learning

## Chapter Contents

This chapter introduces the fundamental concepts used in the present work, as several disciplines and research areas prefer different notations. Furthermore, the basic principles of Deep Neural Networks (DNNs) and deep learning are introduced. The chapter concludes with the definition of relevant terminology. Changing data distributions, incremental learning, the *continual learning* paradigm, *catastrophic forgetting* (CF) effect, and continual learning approaches are among these basic concepts.

## 2.1  Mathematical Notation

In order to ensure a uniform mathematical notation, the following representations are used in this work. The definitions correspond to a subset from Goodfellow, Bengio, et al. (2016).

<div align="center">

**Number and Vectors**

</div>

| | |
|---|---|
| $a$ | a scalar (integer or real) |
| $\boldsymbol{a}$ | a vector |
| $\boldsymbol{A}$ | a matrix |
| $\mathbf{A}$ | a tensor |
| $\boldsymbol{I}$ | identity matrix |
| $diag(\boldsymbol{a})$ | a square, diagonal matrix with diagonal entries given by $\boldsymbol{a}$ |
| $\mathrm{tr}(\boldsymbol{A})$ | trace of a square matrix $\boldsymbol{A}$ |

<div align="center">

**Sets**

</div>

| | |
|---|---|
| $\mathbb{A}$ | a set |
| $\mathbb{R}$ | a set of real numbers |
| $[a, b]$ | interval including $a$ and $b$ |
| $(a, b]$ | interval excluding $a$ but including $b$ |
| $(0, 1)$ | a tuple containing 0 and 1 |
| $\{0, 1\}$ | a set containing 0 and 1 |
| $\{0, 1, \ldots, n\}$ | a set of all integers between 0 and $n$ |

**Indexing**

| | |
|---|---|
| $a_i$ | element $i$ of vector $\boldsymbol{a}$ |
| $\boldsymbol{A}_{ij}$ | element $i, j$ of matrix $\boldsymbol{A}$ |
| $\boldsymbol{A}_{i,:}$ | row $i$ of matrix $\boldsymbol{A}$ |
| $\boldsymbol{A}_{:,j}$ | column $j$ of matrix $\boldsymbol{A}$ |
| $\mathsf{A}_{i,j,k}$ | element $(i, j, k)$ of a 3D tensor $\mathsf{A}$ |

**Linear Algebra Operations**

| | |
|---|---|
| $\boldsymbol{A}^\top$ | transpose of matrix $\boldsymbol{A}$ |
| $\boldsymbol{A}^{-1}$ | inverse of $\boldsymbol{A}$ |
| $\det(\boldsymbol{A})$ | determinant of $\boldsymbol{A}$ |

**Probability Theory**

| | |
|---|---|
| $p(a)$ | a probability distribution over a continuous variable |
| $a \sim P$ | random variable $a$ has distribution $P$ |
| $\mathrm{Var}(f(x))$ | variance of $f(x)$ under $P(x)$ |
| $\mathrm{Cov}(f(x), g(x))$ | covariance of $f(x)$ and $g(x)$ under $P(x)$ |
| $\mathcal{N}(\boldsymbol{x}; \boldsymbol{\mu}, \boldsymbol{\Sigma})$ | Gaussian distribution over $\boldsymbol{x}$ with mean $\boldsymbol{\mu}$ and covariance $\boldsymbol{\Sigma}$ |

**Calculus**

| | |
|---|---|
| $\frac{\partial y}{\partial x}$ | partial derivative of $y$ with respect to $x$ |
| $\nabla_{\boldsymbol{x}} y$ | gradients of $y$ with respect to $\boldsymbol{x}$ |

**Functions**

| | |
|---|---|
| $f : \mathbb{A} \to \mathbb{B}$ | function $f$ with domain $\mathbb{A}$ and range $\mathbb{B}$ |
| $\log(x)$ | natural logarithm of $x$ |

**Datasets and Distributions**

| | |
|---|---|
| $p_{\mathrm{data}}$ | the data generating distribution |
| $\hat{p}_{\mathrm{data}}$ | the empirical distribution defined by the training set |
| $p_{\mathrm{model}}$ | the distribution defined by the model |
| $\mathbb{X}^{train}$ | a set of training examples |
| $\mathbb{X}^{test}$ | a set of test examples |
| $\boldsymbol{x}_n$ | the $n$-th example (input) from a dataset |
| $\boldsymbol{y}_n$ | the target associated with $x_n$ for supervised learning |

## 2.2   Deep Learning

Our today's world is increasingly dominated by the support of computers, algorithms and programs. Humanity is outsourcing tasks to machines, which are supposed to be either faster and/or better. The available resources and capacities keep growing, and with them the amount of available data. However, the enormous quantity of data contains information, pattern or knowledge that cannot be extracted without machines, as it would be unfeasible. *Machine learning* (ML) is a sub-field of *artificial intelligence* (AI) is concerned with techniques that allow for the extraction of knowledge from data. Mitchell (1997) defines ML as an automated learning process that is able to identify rules within the data:

> "*Machine Learning is the study of computer algorithms that improve automatically through experience. Applications range from datamining programs that discover general rules in large data sets, to information filtering systems that automatically learn users' interests.*" Mitchell 1997

Generally speaking, ML algorithms try to highlight generic statistical features of the analyzed data. This process usually results in a model that can recognize the patterns and regularities of the input data. The data should not be learned by memorizing them. Instead, derived rules are used to predict unknown samples. Currently, various approaches are used, depending on the application area.

The first categorization of ML models refers to the type of knowledge representation. One of the classes comprises models that represent the knowledge by means of symbols (also called symbolic or knowledge-based models). The derived and stored information of this type of model is comprehensible to humans. This includes, for example, the storage and representation of rule sets derived from the data. With the help of such models, decisions can be reconstructed, whereas the algorithm derives the rules. Another type of model is based on statistical probabilities. In contrast to knowledge-based models, it is not possible to immediately comprehend why a decision was made. The group of models based on statistical probabilities includes, for example, DNNs, which can appear in numerous variations.

Further types of ML algorithms can be distinguished based on the data's properties. In the following, a selection of the most relevant ones (for this work) is briefly introduced: Supervised learning, unsupervised learning (and reinforcement learning).

**Supervised Learning**   In supervised learning scenarios, the training algorithm is presented with the data $\mathbb{X} = \{(\boldsymbol{x}_1, \boldsymbol{y}_1), ..., (\boldsymbol{x}_N, \boldsymbol{y}_N)\}$ with $N$ denoting the number of samples. The dataset $\mathbb{X}$ consists of both, the input signal $\boldsymbol{x}_n$ and the expected output $\boldsymbol{y}_n$. The expected output is referred to as target values or labels. These can take a single value of the continuous range ($y_n \in \mathbb{R}$), e.g., a real number. Considering the described circumstances, a regression problem is assumed (Arbib and Bonaiuto 2016; Fausett 1994).

Categorical values, however, do assign one or multiple classes to a sample (classification problem). Assigning values to multiple classes indicates a multi-class classification problem. A single class mapping is a special case. The so-called *one-hot* encoding is usually used for classification problems. See equation 2.1 where $C$ constitutes the number of available classes. For a problem with 10 classes to choose from, a label takes the form $\boldsymbol{y} \in \{0, 1\}^C$. A sample $\boldsymbol{x}_n$ of class 4 is represented by the vector $\boldsymbol{y}_n = (0, 0, 0, 0, 1, 0, 0, 0, 0, 0)$. In addition to these types, the binary classification is used for the distinction of precisely two classes (Heaton 2015).

$$\mathbb{1}\boldsymbol{y}_n := \begin{cases} 1 & \text{if } y_n \in C \\ 0 & \text{if } y_n \notin C \end{cases} \tag{2.1}$$

Equation 2.1: One-hot encoding for single-class classification problems.

**Unsupervised Learning**   In contrast to supervised learning, no labels $\boldsymbol{y}_n$ are available in unsupervised (or self-supervised) learning scenarios (Trappenberg 2009). Therefore, it is not the goal to predict a target value, but rather to learn the structure of the data. Clustering can be considered as an example. In order to cluster the data, the dataspace is divided into areas. Subsequently, a cluster center can be assigned (hard or gradual) to each sample. At the same time, measuring the group membership of data points can be the goal. A well-known example is the k-means algorithm.

A combination of supervised and unsupervised learning is referred to as semi-supervised learning. In this scenario, many unlabeled and a few labeled samples are available.

**Reinforcement Learning**   Reinforcement learning is defined by an agent's actions in a given environment. Labeled samples are not required for reinforcement learning. Instead, a reward function assigns values to the outcomes of actions. Reinforcement learning is not further discussed in this work. Nevertheless, it may be relevant for *continual learning* (CL) scenarios (Keller, Liu, et al. 2016).

## 2.2.1   Deep Neural Networks

Deep Neural Networks (DNNs) (also known as Neural Networks (NNs) with multiple hidden layers) are one of the research areas of ML. This type describes a machine learning model inspired by biology. DNNs consist of artificial neurons that are arranged in a network. In the following, a simplified version of a biological neuron is presented. Artificial neurons in DNNs are derived from this biological blueprint. In addition, the perceptron is introduced as a basis for DNNs.

#### 2.2.1.1   Biological Neuron

The simplified biological neuron as depicted in figure 2.1 illustrates the dendrites that can transmit input signals to the cell body (the soma). The soma is responsible for processing the input signals and generating the output signal, which is transmitted by the axon. At the end of the axon, one or several terminals can be used to transmit the output signal to other neurons. The nerve terminals (displayed as circles in figure 2.1) can be connected to dendrites and serve as connectors. The representation of knowledge is determined by the processing of signals and their output. Moreover, several other types of nerve cells with different properties and purposes exist (Alberts, Bray, et al. 2002).

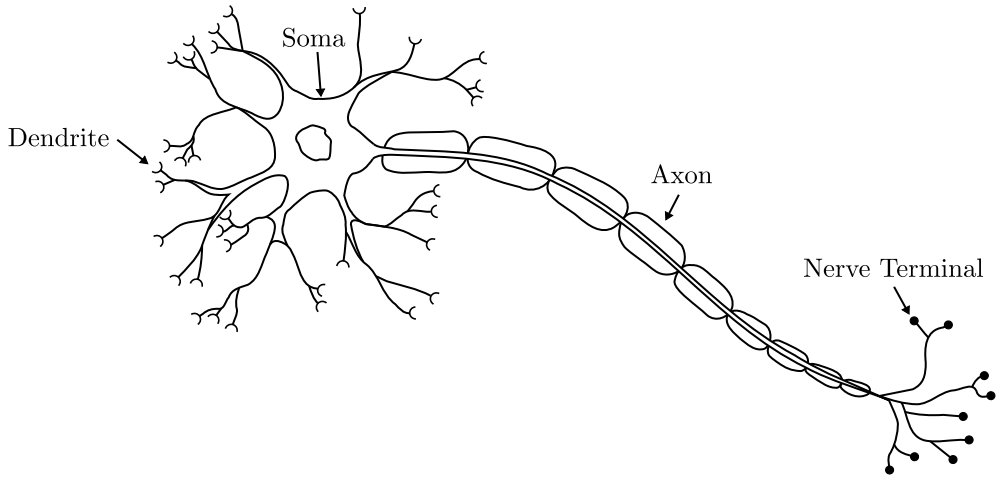

Figure 2.1: A simplified visualization of a biological neuron.

#### 2.2.1.2   Artificial Neuron

If the concept of the biological neuron is transferred to information technology operations, a model as in figure 2.2 can be derived. It is fundamentally based on the McCulloch-Pitts neuron proposed by McCulloch and Pitts (1943). The input signals $x_i$ (left, $i \in \{1, ..., n\}$) are comparable to the dendrites of the biological template. The two functions displayed in the center (activation $\sum$ and output $f_\varphi$) represent the logic contained in the nerve cell body (soma). These functions are described in greater detail below. The output $y$ of the artificial neuron is displayed on the right-hand side of figure 2.2. $y$ corresponds to the axon including the nerve terminals and can also be passed on to several other artificial neurons. The latter reflects connections between nerve terminals and dendrites of the biological model.

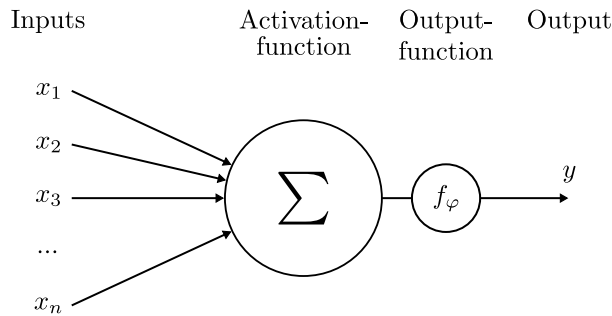

Figure 2.2: Illustration of the components of a single artificial neuron.

**Activation Function**   The illustration of the components of an artificial neuron represents the activation function $\sum$ for pre-processing the input signals. As there is no consensus on the activation- and output function in the literature, the following definition is used in this work. Duch and Jankowski (1999) describe the activation function as follows: "The activation function determines the total signal a neuron receives.". Hence, the activation function aggregates all input signals and results in a single scalar value (Hagan, Demuth, et al. 1997; Fausett 1994).

The most common activation function is indicated in equation 2.2. In this example, the weighted sum of all input signals is calculated and a bias $b$ is added. Equation 2.2 may be replaced with other types of activation functions (e.g., distance-based). However, equation 2.2 is frequently used due to its simple calculation of the derivative. Moreover, all customizable parameters of an artificial neuron are included in this function and correspond to the weight vector $\boldsymbol{w}$ and the bias $b$.

$$\sum(x) \sum_{i=1}^{n} x_i \cdot \boldsymbol{w}_i + b \tag{2.2}$$

Equation 2.2: Example activation function.

**Output Function**   The second component of the artificial neuron is the output function $f_\varphi$, also referred to as transfer function (Hagan, Demuth, et al. 1997). These two terms are often mixed and interchanged in the literature. In order to prevent this misunderstanding, the definition by Duch and Jankowski (1999) is used.

In general, the output function tries to imitate the behavior of a biological neuron: Only if the input intensity of all inputs exceeds a certain threshold value, the neuron forwards an output signal. If the function $f_\varphi$ corresponds to the identity function ($f_\varphi(x) = x$), the neuron is considered a linear unit. The explicit addition of the output function is due to the fact that several linear functions (like the activation function $\sum$) can be combined to a single linear function by transformation. It is therefore the goal to prevent this combination by adding a (partially) non-linear monotone output function. Similarly to the activation function, various output functions can be applied.

Output functions should fulfill a number of properties. Some of them are briefly presented in the following. First of all, the output function should be non-linear. Second, the universal approximation theorem states that at least one approximation of the function is possible (see Cybenko 1989). By limiting the range of values, the training process becomes more stable. This can be traced back to, among other things, the value ranges that the parameters can assume. Moreover, the output function should be continuously differentiable. Otherwise, problems can arise when calculating the derivative. In the following, two frequently used output functions are represented: The sigmoid and the rectifier output function.

**Sigmoid Function**   The sigmoid function is one of the commonly used output functions. By definition (equation 2.3), it is a monotone non-linear function which is continuously differentiable. At the same time, it can be parameterized, as illustrated in figure 2.3. By adjusting the parameter $a$, it is possible to achieve an approximately similar step function. Additionally, the range of output values $y \in [0, 1]$, which has further advantages with regard to computability. The output can be evaluated as the probability of class membership in a two-class problem (Hagan, Demuth, et al. 1997).

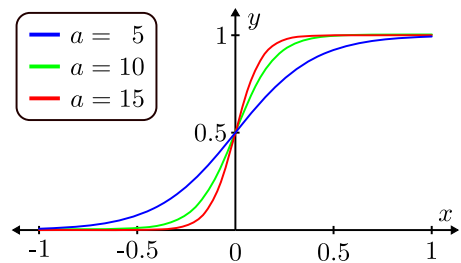

$$f_{\varphi_a}(x) = \frac{1}{1 + \exp(-ax)} \tag{2.3}$$

Equation 2.3: Sigmoid output function.

Figure 2.3: Plot of sigmoid output function.

**Rectifier Function** Another frequently used output function is the rectifier function. Neurons equipped with this function are also referred to as Rectifier Linear Units (ReLUs). As defined in equation 2.4 and indicated by its name, this output function is non-linear and not continuously differentiable. This can cause neurons to enter an inactive state during the training process. However, this can be compensated with different variants (e.g., Leaky ReLU). At the same time, the range of values is not bounded from above, which can lead to new problems. Nevertheless, the rectifier function is successfully applied to many problems (e.g., computer vision or speech recognition) and network types (e.g., DNNs or Convolutional Neural Networks (CNNs)). As shown by Glorot, Bordes, et al. (2011), better gradients result from the use of ReLU for visual problems (Heaton 2015).

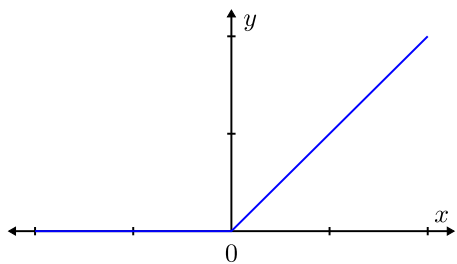

$$f_\varphi(x) = \max(0, x) \qquad (2.4)$$

Equation 2.4: Rectifier output function.

Figure 2.4: Plot of rectifier output function.

### 2.2.1.3 Perceptron

Due to the historical development, the perceptron will be introduced before DNNs at this point (McCulloch and Pitts 1943; Graupe 2013). While the perceptron is technically classified as a subset of DNNs, it is also used as a synonym. In this work, a perceptron is defined as a network consisting of one or multiple artificial neurons. In fact, the term defines properties of the networks. One of them is that the perceptron is always a feed-forward network. In terms of graphs, it is a finite acyclic graph without loops.

One of the most simple perceptrons consists of multiple neurons arranged in a single layer (single layer perceptron). This special form can only solve linear problems. The function that can be used for its representation is in the form of $y = x_1 w_1 + x_2 w_2 + \ldots + x_n w_n$, where $x_i$ is the input feature and $w_i$ is the corresponding weight of the $i$-th neuron. Therefore, it is often referred to as a linear layer.

The far more common and also more powerful variant of a perceptron is the Multi Layer Perceptron (MLP) (Graupe 2013). As suggested by its name, the artificial neurons are arranged in multiple layers (see figure 2.5), which eliminates the problem of linearity among other things. In general, an MLP consists of an input layer (highlighted in blue), one or more so-called hidden layers (green) and an output layer (red).

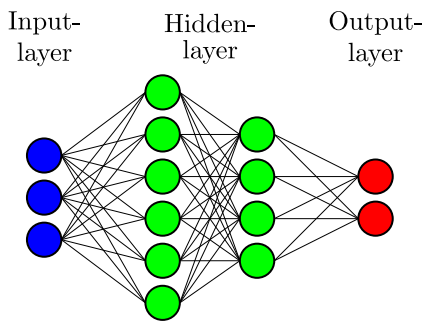

Figure 2.5: Layer structure of an MLP.

### 2.2.2 Deep Learning

Deep Learning is commonly used, and it is often associated with DNNs and supervised learning. However, the basic technologies, developments or algorithms used in this context have been known for many years or even decades. In general, deep learning is understood as the training of DNNs (mostly variants of MLPs, see section 2.2.1.3) by means of the "backpropagation" algorithm. The adjective "deep" refers to the number of hidden layers of the DNN (Goodfellow, Bengio, et al. 2016; Fausett 1994).

Deep learning and the associated models and algorithms derive a function from data. This highly complex function contains knowledge about the provided data, for example, in form of labels (supervised). The data dimensionality can result in a complexity that makes it challenging, if not impossible, for humans to develop a function from scratch. Visual problems related to images or videos can reveal this high-dimensionality. In this case, deep learning algorithms adjust the set of free model parameters. The resulting function is supposed to describe either the problem in a generic way, or a considerable solution for the problem (Heaton 2015).

### 2.2.2.1   Loss Function

The loss function $\mathcal{L}$ (also known as error-, cost-, or objective function) provides a measure of how well the model describes the data. It is used in a later step to optimize the parameters of a DNN. Similarly to the other introduced functions, there is no suitable one for all problems. In general, a distinction can be made between two types of supervised learning problems: loss functions for classification, or regression problems. In the following, two frequently used loss functions are briefly introduced (Heaton 2015).

**Mean Squared Error**   The Mean Squared Error (MSE) (also known as quadratic- or L2 loss) is a commonly used loss function for regression problems. It penalizes large errors by squaring them. At the same time, outliers in the data lead to increased error values. This loss function can also be used for classification problems. In equation 2.5, the loss is calculated for $N$ samples, whereby $\hat{\boldsymbol{y}}_n$ is the output of the model (Heaton 2015).

$$\mathcal{L}_{MSE}(\mathbb{X}) = \frac{\sum_{n=1}^{N}(\boldsymbol{y}_n - \hat{\boldsymbol{y}}_n)^2}{N} \tag{2.5}$$

Equation 2.5: MSE loss function.

**Cross-Entropy Loss**   One of the most commonly used loss functions for multi-class problems is the Cross-Entropy (CE) (or negative log-likelihood). It can also be applied to binary classification problems, which is facilitated by an additional simplification (omitted here). A prerequisite for their use is that the labels are available in a one-hot format (see section 2.2). In equation 2.6, $C$ represents the number of classes, $\boldsymbol{y}$ is the given label vector and $p$ of $\boldsymbol{y}$ is always 1 for the correct label, otherwise zero. Before the loss is determined by cross-entropy, the output of the network is usually normalized by the softmax function $\varsigma$ (see following section 2.2.2.1). Therefore, $p_j$ (output of the DNN) behaves like a probability distribution (Brownlee 2019; Heaton 2015).

$$\mathcal{L}_{CE}(\mathbb{X}) = \sum_{n=1}^{N} -\sum_{j=1}^{C} p(\boldsymbol{y}_{nj}) \log(\hat{\boldsymbol{y}}_{nj}) \tag{2.6}$$

Equation 2.6: CE loss function.

**Softmax**   The softmax function $\varsigma$ is a normalization function that transforms a $K$-dimensional vector $\boldsymbol{z}$ into the value range $(0, 1]$ (Dayan, Abbott, et al. 2003). Similarly to probability distributions, the transformed components sum up to 1.0 (see equation 2.7).

$$\varsigma : \mathbb{R}^K \rightarrow \left\{ z \in \mathbb{R}^K | z_i \geq 0, \sum_{i=1}^{K} z_i = 1 \right\}$$

$$\varsigma(\boldsymbol{z})_j = \frac{e^{z_j}}{\sum_{k=1}^{K} e^{z_k}} \quad \text{for } j = \{1, \dots, K\}$$

(2.7)

Equation 2.7: Softmax function and its properties.

#### 2.2.2.2   Gradient-Based Optimization

Gradient-based optimization is often used in conjunction with deep learning algorithms. The function that needs to be optimized is the loss function (see section 2.2.2.1, Goodfellow, Bengio, et al. 2016; Heaton 2015). The objective of the optimization is to adjust the parameters $\boldsymbol{x}$ of the function $f(\boldsymbol{x})$ so that the result is smaller or larger. The function can be used for maximization as well, but it has to be negated, i.e., $-f(\boldsymbol{x})$.

**Gradient Descent**   Gradient descent is based on the derivative of a function. The general derivative of a function shows how changing the parameters affects the output. This is represented as either $f'(x)$ or $\frac{df}{dx}$. So, the assumption is that for a sufficiently small change of $\epsilon$, $f(x + \epsilon) \approx f(x) + \epsilon f'(x)$ (see Cauchy et al. 1847; Goodfellow, Bengio, et al. 2016; Heaton 2015).

The following is a simple example with the function $f(x) = x^2$ (see figure 2.6). By applying the derivative rules, the gradient can be determined for any value of $x$ for this quadratic function. The first derivative of $f(x)$ is $f'(x) = \frac{df}{dx}(x^2) = 2x$ (visualized in figure 2.7). Based on the sign of the first derivative, a direction can now be determined at which the result $y$ becomes smaller or larger for a certain value $x$. In figure 2.7, a positive sign (indicated by the green arrows) is used to make the $x$ smaller, or vice versa (red arrow).

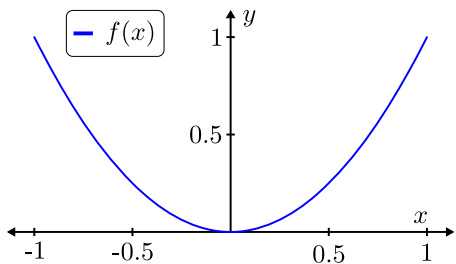
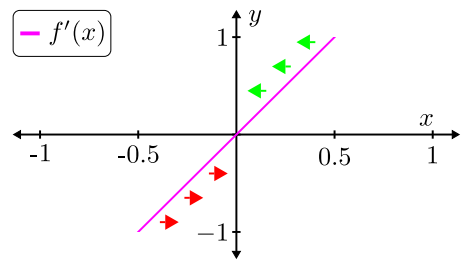

Figure 2.6: Plot of $f(x) = x^2$.            Figure 2.7: Plot of $f'(x) = 2x$.

In both of these simple illustrations (figures 2.6 and 2.7) one optimum exists (in this example $x = 0$). The best possible value for $x$ should be located from an arbitrary starting point or initialization $a$. For this purpose, the parameter value $x$ is adjusted step by step so that the optimum is (nearly) reached. The step size is given by the $\epsilon$ where the direction is determined by the sign of the derivative $f'(x)$ (see figure 2.7). Figure 2.8 illustrates the iteration steps resulting in an algorithm. It becomes obvious that the optimum cannot be reached with this fixed $\epsilon$ and that a circulation between two points may occur. The only way is to decrease the step size $\epsilon \approx 0$, but this is impractical in terms of the number of gradient descent steps that need to be performed. More intelligent strategies for adjusting the step size $\epsilon$ can be applied. This is realized by different optimizers, as described in section 2.2.2.4. If a cycle is detected as shown in figure 2.8, the step size can be reduced (Heaton 2015).

However, the method also poses challenges, e.g., with regard to more complex functions. If the derivative equals 0 (vanishing gradients), no optimization direction can be derived from it. This especially applies to saddle points. Likewise, the first derivative cannot be used to determine whether a local or global minimum has been reached. Figure 2.9 shows a simple function $f(x)$ with a poor local minimum, the global minimum and an acceptable local minimum. Usually, the functions to be optimized have many of these local minima and saddle points. It is often satisfactory to find an acceptable local minimum, as indicated by the minimum on the right-hand side in figure 2.9 (Goodfellow, Bengio, et al. 2016; Wani, Bhat, et al. 2020).

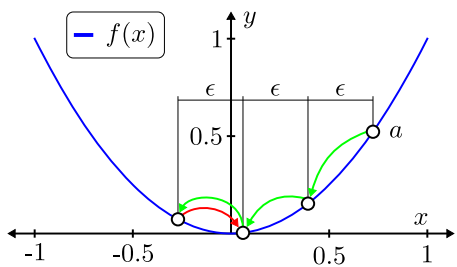

Figure 2.8: Visualization of gradient descent steps.

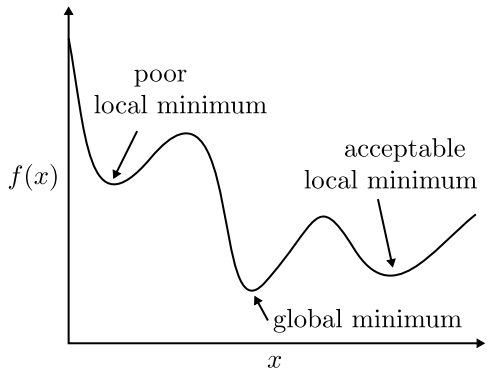

Figure 2.9: Exemplary function $f(x)$ and its different types of minima (Goodfellow, Bengio, et al. 2016).

Nonetheless, functions rarely depend on a single parameter. The same is true for the preceding example, whereby the result of the function is still a scalar ($f : \mathbb{R}^n \to \mathbb{R}$). Since the input vector $\boldsymbol{x}$ consists of several parameters, the derivative of the function must be determined for each component $x_i$. This concept is known as partial derivative and it is represented by $\frac{\partial}{\partial x_i} f(\boldsymbol{x})$. All resulting gradients for $f(\boldsymbol{x})$ are combined and denoted as $\nabla_{\boldsymbol{x}} f(\boldsymbol{x})$ (Goodfellow, Bengio, et al. 2016).

As a next step, a descent direction must be determined from the gradient vector. The most trivial procedure is to subtract the determined gradients ($\nabla_{\boldsymbol{x}} f(\boldsymbol{x})$) from the parameters $\boldsymbol{x}$. Many other methods and optimization steps are available for the determination of the descent direction, even though they are not discussed in this work. Anyway, the increment must be specified for each gradient step. A scalar is used for this purpose, which is also referred to as learning rate $\epsilon$. The value is usually problem-dependent and can be determined, for example, by trial-and-error. Again, many different methods for finding and adjusting the learning rate can be used (Heaton 2015).

For the sake of completeness, a trivial example is provided. Let the function $f(x, y) = x^4 + \sin(y)$ be defined by the corresponding two parameters $x$ and $y$. The resulting face is roughly rasterized in figure 2.10. For the determination of the partial derivative, other parameters are assumed as constants. Thus, they are omitted by the derivative 0. Taking into account the derivation rules, the derivation to $\frac{\partial}{\partial x} f(x, y) = 4x^3$ and $\frac{\partial}{\partial y} f(x, y) = \cos(y)$ is determined. On this basis, the corresponding descent direction can be defined for both directions, $x$ and $y$. A randomly chosen point $a$ (here $a = (2.2, 1.0)$) is used as a starting point for the gradient descent. Putting the point into the parameters of the function $f(x = 2.2, y = 1.0)$, this yields the value $\approx 24.2$. By means of this point, the descent direction for $x$ and $y$ (blue respectively red line, see figure 2.10) can be determined. As for the one-dimensional gradient descent, a learning rate is required, which is $\epsilon = 0.4$ (corresponds to the grid step size) in the present example. If the step size is applied while taking into account the direction, a new point $a' = (x = 1.8, y = 0.6)$ on the surface is obtained. The new function value for $a'$ is $f(x = 1.8, y = 0.6) \approx 11.0$, which corresponds to a minimization.

**Stochastic Gradient Descent**   A further development of gradient descent (see section 2.2.2.2) is Stochastic Gradient Descent (SGD). SGD adapts the gradient descent method for large amounts of data by determining the gradient for a single or a subset of samples. Since the gradient for a single sample is not necessarily representative, e.g., for an outlier, several samples are used. This means that the data needs to be divided into smaller chunks, so-called *(mini-)batches*. Due to the smaller chunks, the expression "mini-batch gradient descent" is sometimes used to refer to SGD. In this work, the processing of mini-batches is always assumed when referring to SGD. This is due to more representative gradients and the processing speed. A step by step solution can be derived from the determined gradient batches. This allows the method to process enormous amounts of data. The low amount of required memory is especially advantageous, as it does not take up the entire storage for all data samples. At the same time, this allows for the processing of potentially infinite data streams. Another advantage of this method is that the cost of a training step is constant, regardless of a dataset's size. In theory, an infinite amount of data can be processed (Keller, Liu, et al. 2016; Hagan, Demuth, et al. 1997; Wani, Bhat, et al. 2020).

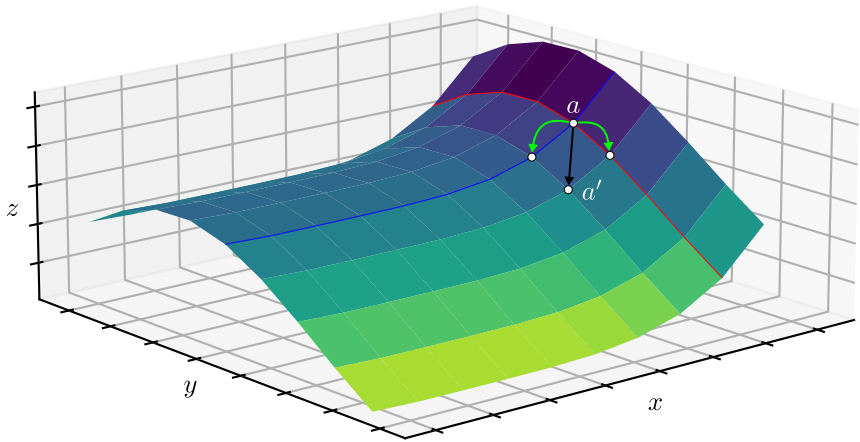

Figure 2.10: Visualization of a 2D gradient descent step.

Despite the advantages, several disadvantages are related to SGD. On the one hand, the size of the batches ($\mathcal{B}$) has to be specified. It is often set to a few hundred samples. The larger $\mathcal{B}$ is chosen, the more representative gradients are derived. On the other hand, a step size $\epsilon$ must be defined, which determines to what extent the gradients from a batch are weighted. Thus, the parameters of a function are adjusted step by step, depending on $\epsilon$: $\boldsymbol{x}' = \boldsymbol{x} - \epsilon \nabla \boldsymbol{x} f(\boldsymbol{x})$. The selection of $\epsilon$ is problem-dependent and difficult to generalize (Flach 2012).

### 2.2.2.3   Backpropagation

The *backpropagation of error* or short *backpropagation* (see Fausett 1994; Haykin 2009; Graupe 2013) is an algorithm for the adjustment of free parameters in an Artificial Neural Network (ANN) (or DNN). This algorithm brings all the pieces of the puzzle together, which is why the term *deep learning* is often used in this context. A DNN is used as the basis for backpropagation, as presented in section 2.2.1.3. The DNN consists of several artificial neurons (see section 2.2.1.2), which are arranged in one (NN) or multiple layers (DNN). Each neuron receives an input signal, which is at the same time the output signal of the neurons of the previous layer. Similarly, each neuron provides an activation function whose parameters are free but initialized with small random values. In addition, the neuron is terminated by an output function, which ensures the non-linearity of the model. Finally, the quality of the model has to be measured by a loss function (see section 2.2.2.1).

The backpropagation algorithm is summarized in algorithm 2.1. As a first step, the data is forwarded through the DNN. Thus, a prediction of the dataset or a batch of data is performed. Based on the outcome, the current quality of the model is determined by the loss function. The specified target values of the data are used for this purpose. The result of the loss function is a single scalar which can be considered as quality measure. As a next step, the "backward" part of the algorithm follows. Hence, the free parameters of the DNN are adjusted. In this context, the state of all model parameters $\theta$ is considered a starting point for the optimization. Subsequently, the partial derivatives of all free network parameters are determined. The chain rule is applied in order to help calculate the derivatives for all neurons in all layers. The goal is to minimize the loss by adjusting the parameters. Thereby, the quality of the model is supposed to improve. The adjustment intensity of each parameter is determined by the gradients and the learning rate $\epsilon$. Thus, $\theta^* \leftarrow \theta - \epsilon \nabla_\theta$ is obtained. The described steps are usually repeated several times until either a fixed number of training iterations or a convergence criterion is fulfilled. Various criteria can be used related to the desired convergence, e.g., gradients, accuracy scores or the value of the loss function. Often, *optimizers* take over this process (Haykin 2009; Wani, Bhat, et al. 2020). The concept of optimizers will be briefly described in the following.

---

Algorithm 2.1: Steps of backpropagation algorithm.

---

**Data:** training data: $\mathbb{X} = \{(\boldsymbol{x}_1, \boldsymbol{y}_1), \ldots, (\boldsymbol{x}_N, \boldsymbol{y}_N)\}$, model parameter: $\theta$, learning rate $\epsilon$

**Result:** trained model

**1**  add input data $(\boldsymbol{x}_n)$ to feed forward through the network

**2**  determine the network output $\hat{\boldsymbol{y}}_n$

**3**  calculate the loss-value $\mathcal{L}$ from real $\boldsymbol{y}_n$ to predicted label $\hat{\boldsymbol{y}}_n$

**4**  adapt model parameters $\theta$ based on error value by gradient descent

---

#### 2.2.2.4   Optimizers

Optimizers are an enhancement of the plain SGD process. They are intended to address the challenges associated with SGD. These well-known problems include, for example, locating unsatisfactory local minima or adjusting an inappropriate learning rate.

A commonly used optimization technique is derived from physics – the momentum optimizer proposed by Qian (1999). The idea of this technique can be compared to a ball rolling down a hill. Depending on the gradient, the ball becomes faster and faster. In the real world, the increase of the ball's speed is limited by air resistance. In the context of optimizers, the parameters serve the same purpose. This allows for faster runs in steeper regions and small step searches in flat areas. Momentum and another optimized version of it, e.g., Nesterov accelerated gradient (NAG), are well-known optimizers (Nesterov 2013). Above all, the convergence time is accelerated by these optimizers. In general, the same result can be achieved by plain SGD with a constant learning rate (depending on the initialization and other local minima).

The Adagrad optimizer (Duchi, Hazan, et al. 2010) and yet another version of it, Adadelta (Zeiler 2012), adjust the learning rate individually for each parameter. The frequency and degree of change within each parameter is tracked. Thus, individual parameters are adjusted more strongly in case they do not substantially change. This can be advantageous if the data is unevenly distributed. RMSprop (Hinton, Srivastava, et al. 2012a) and Adaptive Moment Estimation (Adam) (Kingma and Ba 2015) operate in a similar manner.

Generally speaking, optimizers should lead to either a faster (local) optimum or a better one. The advantage is that this reduces the influence of a fixed learning rate which may be unfavorable. One of the disadvantages is the addition of complexity. Moreover, new hyper-parameters are introduced. Although default values are available, hyper-parameters need to be selected according to the problem (Wani, Bhat, et al. 2020).

#### 2.2.2.5   Deep Neural Networks

This section presents the classification of DNNs along with various details (see Heaton 2015; Kelleher 2019; Wani, Bhat, et al. 2020). First of all, the expression DNN is composed of "Deep" and "Neural Network". As described in section 2.2.1, the underlying concept is constituted by ANNs. The addition "deep" points out that ANNs consist of multiple hidden layers (i.e., MLPs).

DNNs are used in many application-oriented areas due to their excellent performance. These include, for example, visual object-, speech-recognition, translation and many others. The underlying reason is the ability to approximate particularly complex functions, which includes non-linear relationships. Many different variations are used for this purpose. They are often summarized under the term DNN. An overview of the related advantages and disadvantages is presented below.

DNNs are considered sufficient function approximators, as long as they provide enough capacity (neurons and layers). In fact, the underlying problem is irrelevant under the condition that the input format is adapted. Two-dimensional image data can, for example, be converted into a one-dimensional input vector. When compared to other models, DNNs often achieve equal or even better results. At the same time, the advantages provided by the training procedure SGD remain. This is particularly true for the processing of immense amounts of data, as required in the context of big data or streaming. Another advantage is the high degree of parallelization as a result of the current developments in the field of parallel processing. In addition to highly specialized hardware like Tensor Processing Units (TPUs), commercial graphics cards with Graphic Processing Units (GPUs) can also be used to accelerate training and inference. This allows for the efficient use of very large and computationally intensive models. Additionally, the simplicity of DNNs is an important advantage. By now, many different

ML frameworks are available for different platforms and programming languages. These frameworks facilitate the implementation and reduce the respective workload. The usability of DNNs is simplified by the adapted abstraction of parallel programming and knowledge of special hardware. Available frameworks include, for example, PyTorch, Keras, TensorFlow and Caffe. In order to implement a model, it is often sufficient to define its structure at an abstract level by means of a few lines of code (Goodfellow, Bengio, et al. 2016; Heaton 2015; Nguyen, Dlugolinsky, et al. 2019).

At the same time, the use of DNNs involves many challenges, for example with regard to the so-called hyper-parameters. These include the network architecture among other things, and thus the number of layers and their neurons. If a DNN is "too small", the function describing the problem may not be approximated. If the network consists of too many layers or neurons, it may cause the effect that the data is simply memorized. This effect is known as *overfitting*. Figures 2.11 and 2.12 illustrate an exaggerated example of an adequate function approximation (left, blue) and an overfitted function (right, red). The overfitting effect becomes particularly obvious with outliers arising from measurement errors or corrupt data. This effect can be avoided by various approaches, such as the application of dropout (Hinton, Srivastava, et al. 2012b; Wani, Bhat, et al. 2020). Based on fixed random variables, connections within the network are intentionally switched off in order to achieve an improved generalization.

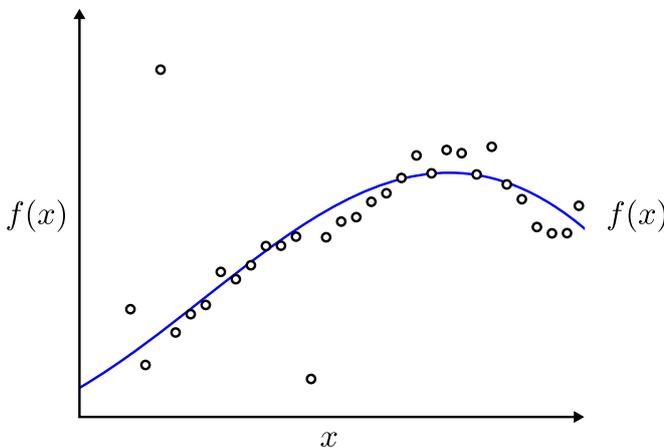
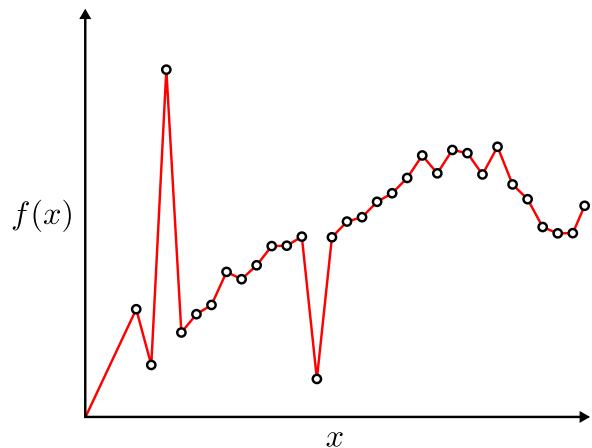

Figure 2.11: Example of generalization.        Figure 2.12: Example of overfitting.

Another decisive parameter is the learning rate $\epsilon$ which is used for the training of a DNN. If the learning rate is not adequately adapted to the problem, it becomes impossible to find a reasonable parameter configuration during the training process. The problem of hyper-parameter determination is often solved by an exhaustive grid-search in the parameter space. This procedure can be compared to a brute-force search for an adequate parameter configuration. At the same time, the final result depends on the initialization of the model, respectively its weights. Various heuristics are offered for the choice of initialization. The proposal of Glorot and Bengio (2010) is, for example, based on a normal distribution that is parameterized by the number of inputs of the previous layer (Xavier initialization). Nevertheless, random initialization can also lead to worse model performances (Goodfellow, Bengio, et al. 2016).

Another disadvantage can be constituted by the data or its properties. If the data distribution for a multi-class problem is uneven, it can cause effects that distort a realistic evaluation of the model (Japkowicz and Stephen 2002; Buda, Maki, et al. 2018; Kubat, Matwin, et al. 1997; Li, Hu, et al. 2017). An extreme example of this effect is a binary classification problem where 9 999 examples come from the first class and only 1 originates from the second class. Without an appropriate weighting, the ML model would only learn to predict the first class, regardless of the actual input. Thus, the error rate would be very low and the single example would hardly have any influence due to the training method (e.g., SGD).

One more problem that affects all ML models is whether the available data is representing the entire problem (Karahoca 2012). Even though a well-generalizing function is derived for the training data, a ML model may fail in real-world applications. This challenge especially arises when synthetic data is used instead of "natural" data. If synthetic data is used, the model merely learns how the data generator works, which is insufficient. The same effect can occur during the collection of real-world data, for example, related to seasonal events. Further problems can arise when converting the data into an appropriate machine-compatible format (not a DNN specific problem). This requires looking at how the data can be structured into a valid format. The question here is how to represent an infinite continuous number, e.g., milliseconds since 1970. Alternatively, such a number can be deconstructed into its components, as otherwise the information may be lost. A further problem can arise when the data originates from different sensors and has very differing value ranges. These ranges can have a negative effect on the training process, as convergence times may increase. If values suddenly become significantly larger than the average, the gradients abruptly increase. In this context, the used data types along with their properties can cause numerical problems (Najafabadi, Villanustre, et al. 2015).

Furthermore, DNNs and their parameter configurations can be difficult to interpret. They are often referred to as "black box". In fact, a slightly different initialization of the model parameters hardly ever results in the same parameter configuration. There are approaches that allow for an interpretation of model parameters. But these are only applicable to data that can be interpreted by humans, e.g., images. It is crucial for certain use cases to clearly state *why* the model classified a certain sample. In this context, the present research is connected to the issue of adversarial attacks. This research area is dedicated to the manipulation of input data and its effects. If the model is not robust against adversarial attacks, the application in critical areas should be avoided. An example is a seemingly insignificant sticker on a stop sign which causes the entire traffic sign to be recognized as a right-of-way sign (Eykholt, Evtimov, et al. 2017). Another example shows that adding noise can also change the classification result of images (Goodfellow, Shlens, et al. 2015).

To name a last challenge, DNNs are subject to the effect of *catastrophic forgetting* (CF), as already described in the introduction (see section 1.1). Trained DNNs can achieve excellent classification accuracies. However, as soon as a new data distribution is presented and knowledge is supposed to be extracted, prior knowledge is lost. The knowledge vanishes so quickly (only a few gradient descent steps) that the effect is denoted as *catastrophic*. This particular problem will be addressed in detail in section 2.3.4 of the present work.

## 2.3  Definitions of Terms

In order to assure a consistent use of terminology, relevant terms for this work are defined in this section. First of all, the relevance of changing data distributions is discussed. Second, a definition for incremental learning is provided, as it is often used differently in related work. Moreover, detailed definitions of *continual learning* (CL) and *catastrophic forgetting* (CF) are presented. Last, some general categories for the classification of CF approaches are introduced.

### 2.3.1  Changing Data Distributions

Changing data distributions may occur in many problems, especially in real-world scenarios. This means that the source generating the data changes over time. The detection and handling of changing data distributions is in theory a different subject area, but it can also be transferred into the context of CL. Thus, CL scenarios can also be described by various changes in data. In the literature, technical terms are often not defined in the same way (e.g., Wang, Schlobach, et al. 2010; Tsymbal 2004; Lesort, Caccia, et al. 2021; Widmer and Kubát 2019; Žliobaité 2010; Wang and Abraham 2015). The following expressions are often combined in order to specify the changing data distributions: real, virtual, domain, criterion, population, concept, data(set), model, covariate, drift, shift, and more.

**Concept Change**  The first type of change distinguishes various changes within the data (concept). These changes can refer to both the features and/or the labels. The problem is defined as follows and it is visualized in figure 2.13. At time $t$, samples $\boldsymbol{x} : \boldsymbol{x}^t \sim \hat{p}_{\text{data}}^t$ from the generating data distribution $p_{\text{data}}^t$ are used to derive a model $p_{\text{model}}^t$ (see figure 2.13a). If new samples $\boldsymbol{x}^{t+1} \sim \hat{p}_{\text{data}}^{t+1}$ are drawn at time $t+1$ from a changed data distribution $p_{\text{data}}^{t+1}$, the derived function of $p_{\text{model}}^t$ does not fit. This can be due to either the features or the labels having changed. In case the features changed, this is referred to as *concept drift* in this work, because the data points move away and keep their label (see figure 2.13b). If the labels (in this example all of them) change, it is referred to as *concept shift*. In addition, a combination of both cases can occur.

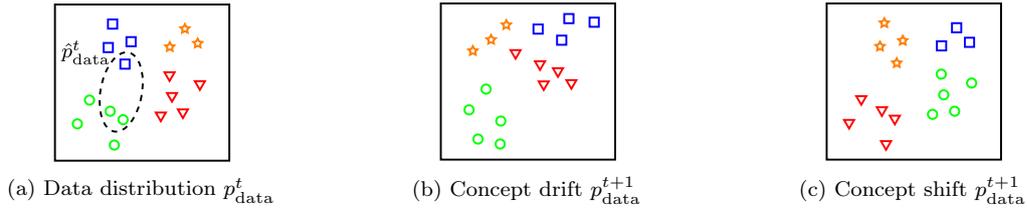

Figure 2.13: Changes in the data distribution (concept drift/shift).

**Context Change** The second type distinguishes changes in the context of the observed data distribution $\hat{p}_{\text{data}}$ (see figure 2.14). This type refers to different views on the data as a whole. Assume a fixed data distribution $p_{\text{data}}$ (of the entire data distribution) that is independent of $t$ and thus constant. The initial observed distribution $\hat{p}_{\text{data}}^t$ is represented in figure 2.14a. If a change does not lead to a new concept, this is referred to as *virtual concept drift*, as illustrated in figure 2.14b. In case the change causes a completely new concept, which is represented in figure 2.14c, it is known as *real context drift*. A *context shift* is represented in figure 2.14d, including the old and new observed distribution.

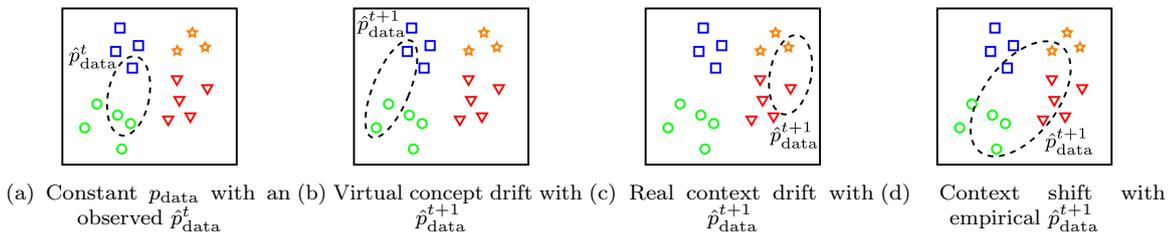

Figure 2.14: Changes in the empirical distribution (context drift/shift).

**Changes over Time** In the context of real-world applications it is helpful to detect statistical changes within the data. However, the two variants (concept and context change) can occur in different combinations and temporal progressions (see figure 2.15). A sudden change is usually caused by a specific event, e.g., by an occurring hardware defect (see figure 2.15a). This does not apply to incremental changes, as represented in figure 2.15b. Incremental changes can be detected when values slowly change over time, for example, related to a deteriorating machine or tool. It is similar to the gradual change (see figure 2.15c), but the transition is not as smooth. Gradual changes keep jumping back and forth until the change remains constant at some point. Figure 2.15d depicts the last type of change over time – the re-occurring concept change. This type is due to a rhythm, such as the days of the week, the calendar, etc. A common problem that affects the detection of these events is the occurrence of outliers. These outliers can be found in real-world applications and lead to an unwanted detection of concept or context changes.

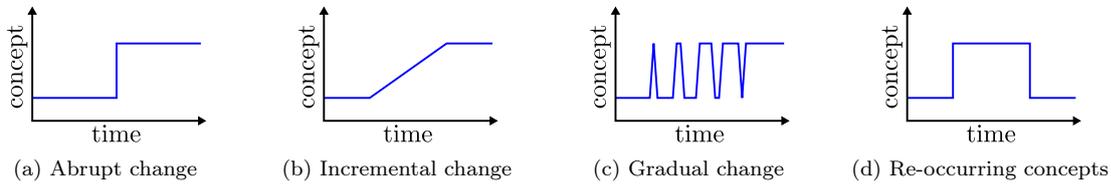

Figure 2.15: Types of changes over time.

## 2.3.2 Incremental Learning

Incremental learning is often used as a synonym for *continual learning* (CL) (see related work in chapter 3). In this work, however, these terms are clearly distinguished from each other. When referring to incremental learning, it is important to note the ability of an ML model to be trained in an online manner. Online learning hereby describes the successive provision of individual samples or subsets of training data. Thus, one sample after the other is available for training and cannot/does not have to be stored (Goodfellow, Bengio, et al. 2016; Bishop 1995).

Even though several models and techniques do not support CL, they do allow for incremental learning (as introduced in section 2.3.3). DNNs, for example, are already able to incrementally learn on a stream of data. At the same time, they are subject to the condition of (catastrophic) forgetting. Therefore, the most simple form of DNNs can be referred to as incremental learners. However, they are not automatically continual learners.

Nonetheless, not every type of model can be trained in an incremental manner. These ML models require intensive modification before they can process streaming data. The standard variants of Support Vector Machines (SVMs) (Scholkopf and Smola 2001) are among them. Usually, adding a single sample requires a complete training of SVMs (Schlag, Schmitt, et al. 2019). Furthermore, the training of large datasets is challenging. This is due to the fact that distances between all data points must be determined in order to separate the support vectors (Graf, Cosatto, et al. 2005). As a result, the standard version of SVMs becomes unsuitable for incremental learning.

### 2.3.3   Continual Learning

The following sections describe the difference between what we as humans perceive as *continual learning* (CL) and what we, in contrast, expect from machines. Accordingly, the CL paradigm is defined as a basis for this work (Chen and Liu 2016).

#### 2.3.3.1   Biological Continual Learning

Humans have the ability to continuously acquire knowledge without (or at least linear) losing previously learned knowledge (Shadmehr and Mussa-Ivaldi 2012). The effort to learn continuously is represented by the survival of the species until today. Adaptability, including the accumulation of experience and knowledge, is the key to success. Even in today's post-knowledge society, CL is applied every single day. In a student's school career, for example, knowledge is accumulated (see figure 2.16). However, this learning process is not a sequence of facts without context. Instead, knowledge, skills and dispositions are interrelated in the context of a task (Raj, Sabin, et al. 2021) and thus enable the construction of new knowledge or ideas. At the same time, only one organ is available for the processing of data and its capacity is limited. In summary, the human being is able to learn continuously by creating and separating connections in the brain.

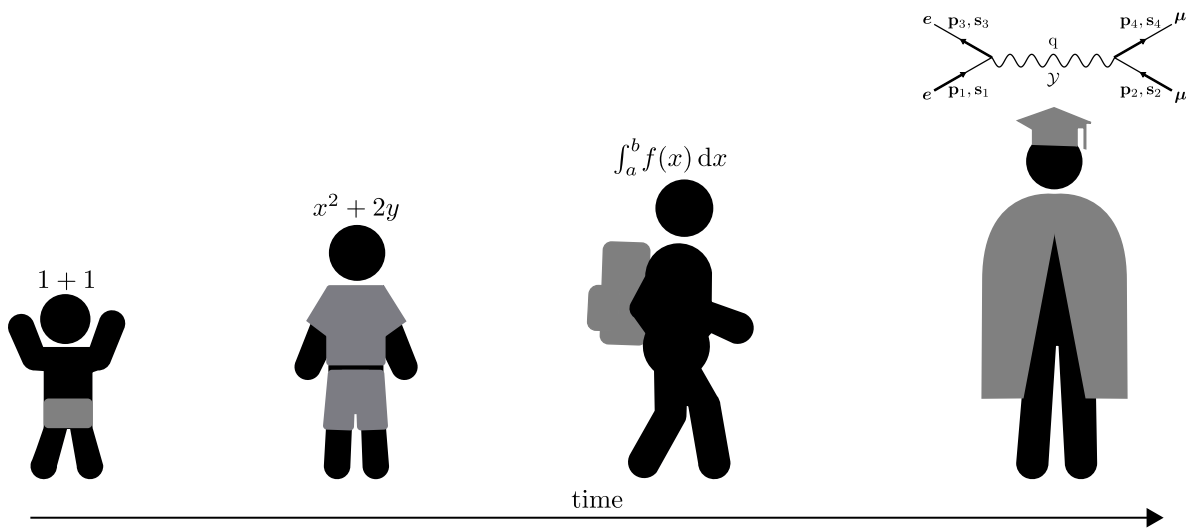

Figure 2.16: The life-long learning paradigm.

Interestingly, the *stability-plasticity dilemma* is also recognizable in humans. At a young age, the human brain is very receptive to new knowledge, which is due to its plasticity. At an older age, learning new things is usually much more difficult, indicating stability.

#### 2.3.3.2  Continual Machine Learning

In the context of *continual learning*, ML is an attempt to imitate the biologically inspired abilities. CL is also referred to as *life-long learning* or *sequential learning* (or even *incremental learning*). Even though the CL is biologically inspired, current models are not yet able to accumulate and combine knowledge. Instead, the goal is to find a model and its parameters, which represents a generic function of the CL problem. Some work related to continual or lifelong learning will be examined in more detail below. All of the presented works represent a similar definition of the CL paradigm (e.g, Mermillod, Bugaiska, et al. 2013).

Thrun (1996a), for instance, describes the learning process inspired by the human being. Thrun defines the ability to learn as a "stream". Moreover, the learning process is lifelong. Thrun's work (Thrun 1996b) describes a "simple version" of lifelong learning problems as "concept learning tasks". Tasks are functions ($f : f_n \rightarrow \{0, 1\}$) which must be learned sequentially based on the presented data. Moreover, Thrun views functions as a binary distinction. Chen and Liu (2016) define the "lifelong supervised learning" process based on successive tasks $\mathcal{T}_1, \mathcal{T}_2, ..., \mathcal{T}_N$. Knowledge must be stored in a "knowledge base" and it needs to be updated based on new data. Some machine learning methods only have this ability under certain conditions, or not at all.

The work of Parisi, Kemker, et al. (2019) comprehensively presents challenges and requirements of the CL paradigm. Lifelong learning is viewed from different perspectives, such as the human brain as bio-inspiration. Despite the many advances in CL, current models are still far away from the capabilities of biological systems. In this context, the CL problem can also be represented as a concept or context change (see section 2.3.1). In terms of the student's school career and learning process, a virtual concept drift takes place. The generated data distribution usually remains constant, whereas the view of the data $\hat{p}_{\text{data}}$ changes constantly (comparable to an abrupt change). Nevertheless, knowledge accumulates and assembles into a whole, even if individual parts are independent. Delange, Aljundi, et al. (2021) present a taxonomy in the context of the "task incremental learning setting". Moreover, Delange, Aljundi, et al. (2021) define requirements for the CL scenario that occur in real-world applications. Their approach also considers the CL paradigm from the CF effect's perspective, which is particularly efficient in sequentially trained tasks. The definition is based on a potential stream of data that repeatedly originates from different domains. Similarly, Hayes, Kemker, et al. (2018) provide a definition of *continual learning* (CL) as "streaming learning". A dataset $D$ is partitioned into individual tasks $T$, assuming that the individual tasks $T$ are independent and identically distributed. The classes of the individual data points are represented as discrete labels. In addition, various constraints are defined, which result from real-world scenarios, such as the use of robots.

**Definition of a CL *task*:**  The following definition of CL is used in this work. A CL *task* is defined by a sequence of sub-tasks $T_i$. Each sub-task $T_i$ defines a subset of samples from a dataset $\mathbb{X}$, where $T_i \subseteq \mathbb{X}$. The index $i$ defines the order of the sequence $i \in (1, ..., I)$, where $I$ is the last task. Each task $T_i$ consists of $N$ individual samples, which are defined by the input vector $\boldsymbol{x}$ and by its class label $y$ (see section 2.2). Likewise, the data within a task is divided into training $T_i^{train}$ and test $T_i^{test}$ data. Each sub-task is sequentially presented to a model $m$. It is assumed that each task contains only individual classes (non-overlapping classes).

**Definition of the CL *goal*:**  A model $m$ should be able to reproduce all derived knowledge from all sub-tasks after the completion of the last sub-task $T_I$. For the supervised learning context the unified test dataset $T_{All}^{test} = \cup_i T_i^{test}$ is the benchmark. Even though this benchmark does not have to be applied to all scenarios, it is an initial quality measure that should be achieved.

### 2.3.4  Catastrophic Forgetting

*Catastrophic forgetting* (CF) or *catastrophic interference* is an effect that occurs in DNNs (Chen and Liu 2016). It has been the focus of research since the early 1990s. CF describes the effect of a model losing all previous knowledge, as soon as new knowledge is added. Although ML models are biologically inspired, they are affected by the problem. This seems true for every model based on artificial neurons.

The work of McCloskey and Cohen (1989) identified the problem by conducting the following experiment. Their training scheme was inspired by elementary school students who should learn the addition of two numbers. As a first task, the model should learn how to add 1 to another single-digit number (i.e., 1+1, 1+2, ..., 1+9 and commutative variants). The second task was similar, but with the number 2 (i.e., 2+1, 2+2, ..., 2+9 and commutative variants). Both tasks were trained until the model could solve them. An interesting effect appeared during testing on the test data of the first task and having trained the second task: In each case, the model was off by 1 whenever calculations from the first task were performed, e.g., 5+1=7 or 6+1=8. Although both tasks included the operations 1+2 and 2+1, a high error rate was observed. To be precise, the errors occurred at the beginning of the second training sequence.

Ratcliff (1990) started to address the CF problem and investigated further models. In addition to an extended test scheme, first solution approaches were presented. In this context, a simple buffer is used as a solution to enable "rehearsal" or "replay". For this purpose, samples from the first task are retained and included in the next training session.

A visualization of the CF effect is depicted in figure 2.17. The figures display the training and testing of a ML model on two sequential tasks. During the first training phase, knowledge is extracted from the data of the first task (task 1, white background). In the second training phase, the knowledge from the second task has to be learned (task 2, gray background). Both tasks are mutually disjunctive, which means they contain different knowledge. The blue lines represent the test performance on the test data of task 1. The test performance of task 2 is presented in green and starts with the training of the second task. Performance is defined by the degree to which individual test samples are correctly assigned to their origin/class. As depicted in figure 2.17a, the performance on task 1 catastrophically drops as soon as learning on the second task begins.

The trend displayed in figure 2.17b is different and shows an ML model that is slowly losing knowledge. After extracting the knowledge from task 1 (as before), samples from task 2 are presented in the second training phase. But this time, the prior knowledge from the previous task disappears in a linear manner. In theory, the linear forgetting of knowledge from previous tasks is considered an attractive solution. Obviously, the complete elimination of the forgetting effects is desired.

In general, the ML model must be able to derive the joint knowledge from task 1 and task 2. The joint training trend is depicted in figure 2.17c. The red line represents the performance of the ML model while it is trained on a combination of task 1 and 2. The so-called "baseline" is thus the reference value and represents the upper limit that should be achieved even in CL scenarios.

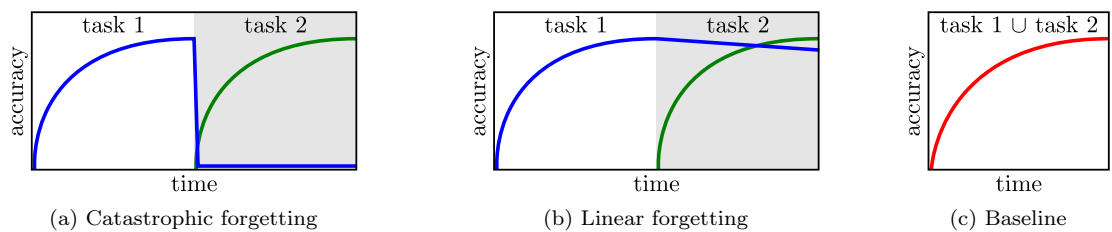

(a) Catastrophic forgetting          (b) Linear forgetting          (c) Baseline

Figure 2.17: Visualization of the CF effect, a linear forgetting and the baseline.

#### 2.3.4.1   Reason for Catastrophic Forgetting

The main reason for the occurrence of the CF effect in connectionist models is the way the information is stored – namely distributed (French 1997). However, this distribution of knowledge is the reason why these models have their beneficial properties. Although DNNs are biologically inspired (see section 2.2.1), their simple abstraction does not seem to result in the same properties as the more complex original. This raises the question whether the brain itself may be subject to this effect. Studies with animals show that although a forgetting effect occurs, it is not considered catastrophic and rather linear (McClelland, McNaughton, et al. 1995). Nevertheless, this effect seems to be less problematic in highly developed species. This is probably due to the different areas of the brain such as the hippocampus, which is particularly responsible for memory consolidation (McClelland, McNaughton, et al. 1995). Consequently, some ML methods are trying to address the CF problem based on a similar mechanism.

The brain seems to be able to accumulate knowledge over time through multiple mechanisms. One of the basic conditions is that there is a limited capacity available to integrate knowledge. Due to this integration, existing knowledge is not immediately lost. Moreover, the brain is particularly adaptive in certain phases of human development, although this characteristic slowly decreases over time.

### 2.3.5   Continual Learning Approaches

In this section, several known categories are presented that intend to mitigate or remedy the CF effect. Accordingly, all ML models can be grouped into a category. It is, however, possible that a model can be classified into more than one category. In addition, the challenges of the various CL approaches are indicated in this section. Likewise, the basic concepts of the methods are briefly introduced. Further methods can be used for CL, the presented methods thus do not represent a complete overview, but a selection. Transfer learning is a corresponding example, which transfers knowledge from one model into another.

**Regularization Methods**   The goal of regularization methods is depicted in figure 2.18 (Solinas, Rousset, et al. 2021; Delange, Aljundi, et al. 2021; Chen and Liu 2016). The $\Theta_{T_1}$ (yellow area) represents all parameter configurations of the model which are acceptable for possible solutions of the first task $T_1$. $\Theta_{T_2}$ (blue area) denotes acceptable solutions for the problem in $T_2$. The points $\theta^*_{T_1}$ and $\theta^*_{T_2}$ illustrate parameter configurations that provide a solution for the corresponding task. It is the goal of regularization to find a parameter configuration that is in the green range, e.g., influenced by the loss function. Therefore, it is not the solution $\theta^*_{T_2}$ (red dashed line) that should be obtained, but a parameter configuration which is valid as a solution for $T_1 \cup T_2$ (black dashed line). The existence of the joint parameter space can be shown by training the model with a joint dataset from $T_1$ and $T_2$.

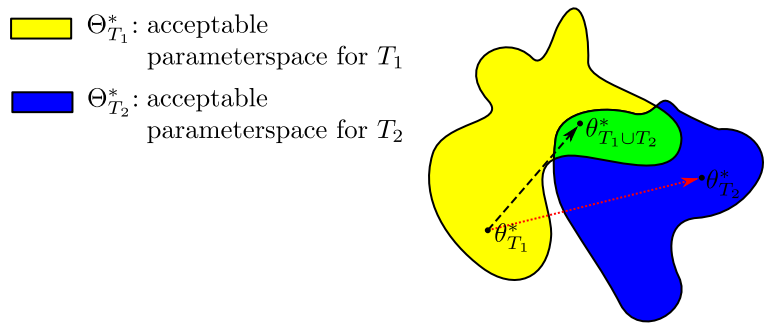

Figure 2.18: Visualization of the objective of regularization methods.

The regularization approach tries to keep certain parameters from changing by determining their "importance" (Farquhar and Gal 2018). As there are usually many parameters available in an DNN, the determination of their dependencies is usually very complex. Depending on the size, the determination is too computationally complex so that it becomes unfeasible. Therefore, the parameters are often individually considered. Other challenges can arise, such as the determination of the importance of individual parameters. At the same time, the influence on the re-training process needs to be determined.

**Replay Approaches**   Replay approaches avoid the CF effect by simulating a joint model training. For the training of the second task, samples of the previous task are added in order to rule out the CF effect (Solinas, Rousset, et al. 2021; Delange, Aljundi, et al. 2021; Chen and Liu 2016). The general idea is equivalent to a joint training. The difference to an actual joint training is the selection and quantity of samples used for the replay. In some application areas, however, the question of data privacy issues arises. Three basic replay methods are distinguished: *rehearsal*, *pseudo-rehearsal* and *constrained optimization*.

The *rehearsal* method uses samples that are stored in a separate memory. The basic rehearsal concept is illustrated in figure 2.19. The challenge is to use as little memory as possible, preferably a constant amount. Models determine samples that are particularly representative and thus have a high added value for the replay. However, this approach has limitations and challenges. Open questions are related to the representative nature of samples, the scope of available memory and how to keep enough samples for all future tasks. If there are only a few samples available, it remains unclear how to avoid overfitting (Solinas, Rousset, et al. 2021; Delange, Aljundi, et al. 2021; Chen and Liu 2016).

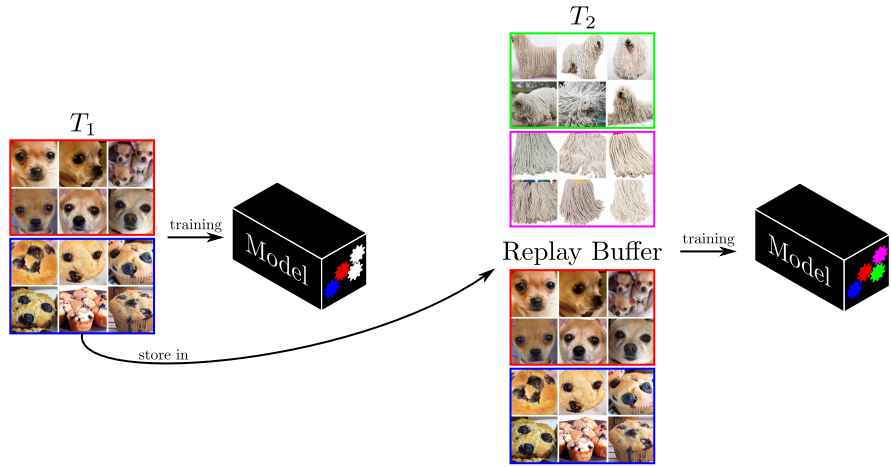

Figure 2.19: Outline of rehearsal methods (replay).

*Pseudo-rehearsal* follows a similar procedure, except that the samples do not have to be explicitly selected and stored. Instead, they are generated (see figure 2.20). As a consequence, the problem of the selection and the storage can be ignored. Nevertheless, the question *how* to generate data arises. In this context, it is important how the generating model derives the distribution from the data. Moreover, the approaches need to describe in which ratio old data is generated and how the generated samples are labeled. Another problem for pseudo-rehearsal methods is the quality and variability of the data (Chen and Liu 2016).

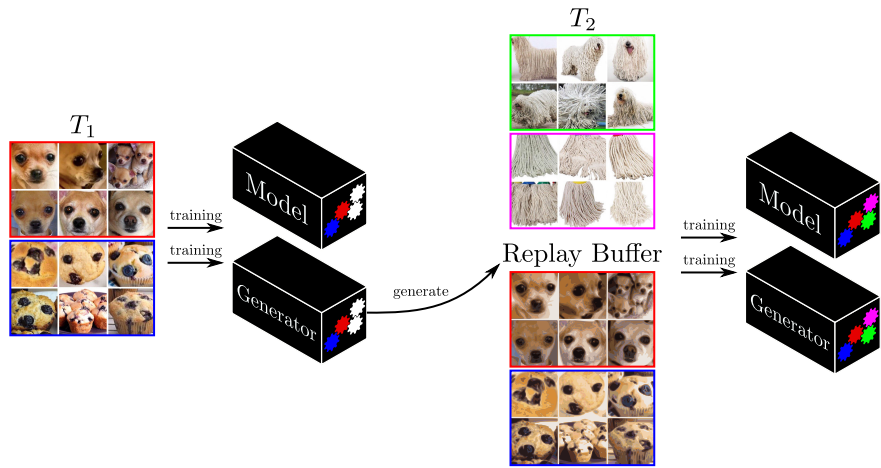

Figure 2.20: Delineation of pseudo-rehearsal methods (replay).

Another methodology of this category is the *constrained optimization*. Re-training is not directly influenced by replaying (generated) samples. Instead, their influence on the training process and thus on the model and its parameters is analyzed and stored. For re-training, the influencing factors are combined with the factors influencing the current training process. In order to provide an example, the following process is assumed: The gradients generated during the first training process are used to modify the gradients generated in the re-training process.

**Parameter Isolation**   Approaches using parameter isolation exploit the structures or architecture of the underlying model (Solinas, Rousset, et al. 2021; Parisi, Kemker, et al. 2019; Delange, Aljundi, et al. 2021). As part of this strategy, certain components within the network are reserved for sub-tasks. Basically, a distinction is made between two types of architecture: *dynamic* and *static*.

   *Dynamic* architectures are able to store the new knowledge in additional components of the network. Thus, the knowledge for a corresponding sub-task is stored separately. Challenges that can arise include choosing a component for a task or sample. In terms of storage technology, the scalability of such a system remains unclear. Assuming that a CL problem consists of an infinite number of tasks, memory limits are restrictive (Solinas, Rousset, et al. 2021; Parisi, Kemker, et al. 2019).

This is in contrast to the *static* models that do not add new parameters, but "freeze" existing ones. For re-training, parameters from the previous tasks are not changed. The decisive factor is which component of the network is responsible for which task. Furthermore, questions related to each parameter's role for a particular task remain open (Solinas, Rousset, et al. 2021; Parisi, Kemker, et al. 2019; Delange, Aljundi, et al. 2021).

# 3. Related Work

## Chapter Contents

This chapter presents the current state of research with regard to deep learning in context of *continual learning* (CL). First of all, *catastrophic forgetting* (CF) evaluation schemes are introduced. However, related work that merely constructs a new *machine learning* (ML) model is not included. Instead, the focus is only on research that presents an evaluation strategy for the CF effect and, if necessary, conducts the respective investigations. Related work is usually presented by providing the definition of its evaluation methodology, the metrics, datasets and the description of the investigated scenarios. Moreover, related work with regard to CF avoidance models are presented. The number of available ML models and algorithms for this purpose is extensive, and new proposals are constantly being added. The respective papers usually include an evaluation scheme. In order to summarize the related works, a brief discussion is provided.

## 3.1   Training and Evaluation Schemes for CL Scenarios

In this section, the training and evaluation schemes used for evaluations in the CL context are presented. The challenge is that related works usually introduce an evaluation scheme along with a new ML approach or model. As the evaluation results tend to show the superiority of the newly developed model, these works are excluded from a detailed review. Moreover, this decision is due to the lack of detailed specifications of the evaluation scheme.

Chen and Liu (2016) describe the used evaluation methodology as a four-step procedure: 1) sequential training on the data of old tasks; 2) training on the data of the current/new task; 3) creating benchmarks by running baseline experiments on the merged data or comparing with other algorithms; and 4) the final analysis and interpretation of the results. Besides, three "kinds of evaluations" are described, first of all, the number of tasks and the information content for current task. For the evaluation, the already determined previous knowledge in relation to the effect on the new task is taken into account. Second, the effect of the task's order is addressed, which can lead to different results. The third kind of evaluation is "progressive experiments". This type determines the influence of the number of previous tasks, assuming that more knowledge could already be derived.

Hsu, Liu, et al. (2018) criticize the methods used for evaluation in the context of CL. Their work introduces requirements that apply to real-world applications: Determining constant memory usage, prohibiting the use of oracles at test time, and not specifying task limits. Considering these requirements, 10 of the 14 CF avoidance models are excluded from the investigation of Hsu, Liu, et al. (2018). In addition, various types of tasks are described. Task Incremental Learning (Task-IL) and Class Incremental Learning (Class-IL) show how datasets are divided into tasks. Overlaps with regard to the contained classes are not used. The difference is that Task-IL specifies from which task the sample originates. This is different compared to Class-IL, as this particular information is missing.

The latter task type is therefore represented as "hardest scenario". Furthermore, Domain Incremental Learning (Domain-IL) is described, which applies a "task-dependent transformation" to the input. In this context, a permutation and rotation is applied once. The same types of tasks are described in other works, such as Ven and Tolias (2019). Fixed models with 2 hidden layers and 400 neurons (including Rectifier Linear Unit (ReLU)) are used for the evaluation (Hsu, Liu, et al. 2018). The selection of the hyper-parameters is realized by a grid-search where the highest accuracy measured at the end is used as criterion.

Farquhar and Gal (2018) denounce that conventional CL evaluation schemes do not reflect the fundamental challenges. Accordingly, a separate evaluation procedure is presented based on the example of the Mars rover. The following constraints are assumed. *Cross-task resemblances* describes the distinctness/difficulty of the tasks. *Shared output head* must be specified, which describes whether a separate output vector is used depending on the task (also known as multi-head). *No test-time assumed task labels* prohibits the use of oracles, which denotes the specification of the task from which the sample is taken. *No unconstrained retraining on old tasks* restricts random access to data from all tasks. *More than two tasks* are proposed to be investigated. Even though the basic feasibility can be shown with two tasks, this does not have to apply to several tasks.

In addition to these constraints, other requirements are imposed on the models (Farquhar and Gal 2018). Regarding the tasks and their limits, three types are distinguished: *Unclear task demarcation* – it is often assumed to know these task boundaries; *continuous tasks* – assumption is that there are hard task boundaries that are not subject to natural variability; *overlapping tasks* – denotes that single or disjoint tasks are examined. In addition, Farquhar and Gal (2018) outline *time/compute/memory constraints*. In this context, fixed memory is assumed to be more useful than large computational overheads. The last requirement is *strict privacy guarantees* with special emphasis on intermediate individual tasks.

Several evaluations are presented as a critical analysis (Farquhar and Gal 2018): *Pixel permutation* describes the shuffling of features based on a random but fixed pattern. In the case of an image, this means that pixels are exchanged with others. After performing this process, images usually become unrecognizable by means of visual inspection. In contrast to that, *split classification* tasks split the dataset into several partial datasets. These tasks are associated with the requirements and constraints, where the impact is enormous.

Lesort, Caselles-Dupré, et al. (2019) describe a number of different CL scenarios. Unlike other works, the focus is not on disjoint class scenarios, but drift scenarios. Accordingly, different types of drift are described at first: Real concept drift, virtual drift, virtual concept drift and domain drift. The evaluation protocol describes two distinct objectives. The maximum performance at the end of the training ("final performance") process and the performance measured over the entire course of the training. In general, both of them are associated with the *current performance*. In contrast to that, the "cumulative performance" describes the performance during the training process and is thus related to the currently possible level of knowledge. Furthermore, various stationarities that occur in CL scenarios are described, e.g., concept stationarity, target/reward stationarity, finite world, etc. Moreover, drift intensity and patterns are discussed. Besides, incremental learning (virtual concept drift) is distinguished from lifelong learning (domain drifts) and real-concept drift scenario (real concept drifts).

Kemker, McClure, et al. (2017) introduces three metrics that allow a comparison between datasets. Accordingly, the performance of the model after the first session $\Omega_{base}$ has to be determined. $\Omega_{new}$ indicates how models directly react to new tasks. In addition, $\Omega_{all}$ indicates to what extent a model can represent both, old and new knowledge. Furthermore, various experiments are described, such as the permuted, incremental class learning and multi-modal experiments. The latter describes tasks from different datasets with diverse dimensions. A hyper-parameter investigation is also carried out for the studied models. One of the conclusions is as follows: "We demonstrate that despite popular claims, catastrophic forgetting is not solved." (Kemker, McClure, et al. 2017).

## 3.2 Continual Learning Models/Methods

This section presents related models claiming to mitigate or avoid the CF effect. All approaches use different techniques to mitigate the effect and should therefore be suitable for the use in CL scenarios. The following summary represents only an subset of all existing models addressing the CF effect. This is especially true, as novel and better methods are being presented on an almost daily basis. Thus, the importance of the CL paradigm is highlighted. In this work, the categorization of the approaches according to the method to address the CF problem is attempted (see section 2.3.5).

**Parameter Isolation**    The basic idea of the subsequent approaches is to modify the network architecture so that the CF effect does not occur. In any approach the Deep Neural Network (DNN) is extended or only certain parts of the DNN are used for a given task.

Srivastava, Masci, et al. (2013) present a model referred to as Local Winner Takes All (LWTA), which addresses the CF effect. The idea is based on the fact that every neuron consists of two or more units, so-called blocks. Depending on the input, one of the corresponding neurons "wins". This procedure is realized as an activation function. Thus, sub-networks are created which can be selected and are responsible for different tasks. This model is described in section 7.1.1 in more detail.

Goodfellow, Mirza, et al. (2013) illustrate in their empirical study that the application of dropout is "beneficial" in relation to the CF effect. Depending on a factor, dropout leads to the random non-forwarding (or setting to 0) of neurons activities. Smaller DNNs, which can be trained by using the dropout mechanism, are more resistant to CF than large networks. Furthermore, it is observed that different activation functions, such as sigmoid, ReLU, LWTA and maxout, have fewer effects with regard to CF. This proposal will be described in section 7.1.1 in more detail.

Rusu, Rabinowitz, et al. (2016) propose Progressive Neural Networks (PNN), a model that counteracts the CF effect at the architectural level. The model extracts useful features for the different tasks from several models. Thus, a new DNN (or column) is created for each task. The transfer is then achieved via connections to other networks. Each column is connected to its predecessor columns (lateral connections), where parameters are "frozen" after training.

Aljundi, Chakravarty, et al. (2017) propose Expert Gate, a life-long learning system. The procedure is based on a network of experts. Thus, a corresponding model is derived for each task. For the selection of the respective expert, a "gating" function is used, which decides on the basis of the characteristics of each task. The encoder part of an autoencoder is used for this purpose. The autoencoder takes over the task of the oracle, which selects the expert.

PathNet is suggested by Fernando, Banarse, et al. 2017. The idea is to reuse parts of the network for specific tasks. The selection of parameters is based on a genetic algorithm characterized by replication and mutation. Path determination is initially working with a random selection, where the "fitness" (negative classification error) is calculated. The same is true for other random paths, whenever the better/best path wins. Finally, the best path is mutated by a random selection of elements.

The model proposed by Mallya and Lazebnik (2018) is referred as PackNet and follows the approach of iterative pruning and re-training, Thus, knowledge from several tasks is to be "packed" sequentially into a DNN. A fixed percentage of neurons is selected for pruning. The selection is based on the method suggested by Han, Mao, et al. (2017), which defines neurons with small weights as "unimportant connections". After releasing the unimportant neurons, the remaining ones are "frozen".

Serra, Suris, et al. (2018) describe the Hard Attention to the Task (HAT) procedure in order to overcome the CF problem. The mechanism referred to as "task-based hard attention" maintains task-specific relevant information from others. This is realized by a gating mechanism, which activates or deactivates neurons depending on the task. Similarly to PathNet (Fernando, Banarse, et al. 2017), this results in different paths through the neural network for individual tasks. The binary attention masks (gates) are already parameterized by the training process, whereby the selection is realized by the task ID.

Joseph and Balasubramanian (2020) propose the Meta-Consolidation for Continual Learning (MERLIN). The underlying assumption is that a meta-distribution (namely the latent space) exists for each task $t$. This distribution is learned and consolidated as an online variant. Several base networks are trained (for each task). The learned weights are used to learn task-specific parameters (VAE-like strategy). In the consolidation phase, "task-specific prior" is generated to "refine" the VAE.

**Regularization Methods**    Regularization methods affect the parameters of a model so that the CF effect is mitigated. In the following, some well-known methods are briefly presented.

Elastic Weight Consolidation (EWC), as proposed by Kirkpatrick, Pascanu, et al. (2017), is based on the idea of penalizing changes in parameter that are important for previous tasks. For this purpose, a Gaussian distribution of the information over the parameters is approximated using the Fisher information matrix. EWC is described in section 7.1.1 in greater detail. Huszár (2018) presents an adaptation of the method based on a Laplace approximation, which makes it a "multi-penalty version" of EWC. This adaption is used as the basis for the Progress & Compress (P&C) framework (Schwarz, Luketina, et al. 2018).

Zenke, Poole, et al. (2017) present the Synaptic Intelligence (SI) approach in their work. Each unit, or "synapse", is trained to be "important" for a specific task. Unlike EWC, this method is an online variant and determined on the basis of the "entire learning trajectory in parameter space" (Zenke, Poole, et al. 2017).

Lee, Kim, et al. 2017 introduce Incremental Moment Matching (IMM) as a transfer technique. For each task, Bayesian Neural Networks (BNNs, Bishop 1995) are trained one after another. The distinct feature of IMM is that the parameters $\theta$ are specified according to a certain distribution, depending on the data (posterior parameter distribution). Merging the networks is based on the calculated posterior distributions of the weights. For this purpose, IMM uses an approximation of mixture of Gaussian posteriors. Two variants are distinguished: *Mean*, which averages the parameters while minimizing the KL-divergence; and *mode*, which tries to maximize the mixture of Gaussian posteriors.

The approach Riemannian Walk (RWalk) was introduced by Chaudhry, Dokania, et al. (2018). It "generalizes" Elastic Weight Consolidation (EWC) (Kirkpatrick, Pascanu, et al. 2017) and Synaptic Intelligence (SI) (Zenke, Poole, et al. 2017). Moreover, an efficient version of EWC is presented – namely EWC++. The KL-divergence as a distance measure for the Riemannian Manifold is fundamental for this method. The combination of the Fisher information matrix and KL divergence determines the parameter importance. However, a single-head output leads to "poor test accuracy". An additional replay procedure is used to counteract this disadvantage.

Learning without Forgetting (LwF), as proposed by Li and Hoiem (2018) is based on Convolutional Neural Networks (CNNs). It can be classified as a knowledge distillation approach. The LwF approach uses some model parameters for all tasks $\theta_s$, special ones for individual old tasks $\theta_o$, and $\theta_n$ for the current one. For the training process, $\theta_n$ are initially adjusted. The knowledge distillation loss determines how the other parameters are adjusted. Precisely, network outputs (pseudo labels) are specified as previous network states. Kim, Kim, et al. (2018) suggest a combination of EWC (Kirkpatrick, Pascanu, et al. 2017) and LwF (Li and Hoiem 2018), known as EWCLwF. An adapted loss function is introduced for this purpose.

**Replay-Based Approaches**   Replay-based approaches use stored or generated samples to bypass the CF effect by joint training. Ratcliff (1990) describes the Experience Replay (ER) model as a very early version of the replay process. Since then, many variations have been introduced. However, the principle is similar in all replay approaches (see section 2.3.5). Samples from past tasks are stored and replayed for training with samples of the new task. Self-generated samples can also be used, which is referred to as pseudo-rehearsal. Further approaches do not directly replay the samples but compute their influence on the model.

Gradient Episodic Memory (GEM) is suggested by Lopez-Paz and Ranzato (2017). GEM stores samples for each task in an *episodic memory* (e.g., Nagy and Orban 2016). However, the samples are not replayed, but the loss is influenced by their loss gradients (depending on the angle between gradients). Chaudhry, Marc'Aurelio, et al. (2019) propose further developments, namely Averaged GEM (A-GEM). The crucial optimization is realized via the determination and replay of the averaged gradients from the previous tasks.

The work of Chaudhry, Gordo, et al. (2021) presents the Hindsight Anchor Learning (HAL) procedure. The approach is based on so-called "anchors", which are used as representatives for individual classes per task. These anchor points are determined by gradient descent and used to estimate the level of forgetting for further training. The anchors are usually positioned on the task borders, which allows for a better protection. Likewise, the anchors can be used to enable "experience-replay".

The procedure coined by Aljundi, Lin, et al. 2019 is labeled Gradient based Sample Selection (GSS). It describes a procedure to create the best possible selection of samples representing all past problems/tasks. For the determination of the set, the gradients or their angles to each other are used. For this purpose, new samples in a "recent" buffer are compared with those in the "replay" buffer. If necessary, they are exchanged. Furthermore, the greedy version of the algorithms represents the problem of quadratic dependency of the buffer size. With this variant, the re-processing of all stored samples is omitted.

Rebuffi, Kolesnikov, et al. (2017) propose the incremental Classifier and Representation Learning (iCaRL) model. Their approach includes a classifier which works with the *nearest-mean-of-exemplars classification*. This is based on a compressed set of sample data. A learner uses these samples in combination with a knowledge distillation process. Furthermore, a herding procedure is suggested, which is responsible for the selection of samples.

Shin, Lee, et al. (2017) introduce a pseudo-rehearsal procedure based on short-term memory referred to as Generative Replay (GR). A Generative Adverserial Network (GAN) (Goodfellow, Pouget-Abadie, et al. 2014) is applied as generator, whereas the special variant Wasserstein GAN (Arjovsky, Chintala, et al. 2017) is selected. In addition to the generator a solver is implemented. For each task, both parts are (re-)trained. After the completion of a task, the generator creates samples, which are then replayed for a joint training. A detailed representation is given in section 9.2.1.2.

Meta-Experience Replay (MER) by Riemer, Cases, et al. (2019) enforces gradient alignment using samples. By means of this customization, the new gradients are appropriately adjusted via the old ones. As a result, the interference between tasks decreases.

Another rehearsal method including knowledge distillation mechanisms is known as Dark Experience Replay (DER) (Buzzega, Boschini, et al. 2020). DER can be used in streaming scenarios, because it addresses real-world requirements. These include constant memory consumption, the lack of a test time oracle and task boundaries. The procedure is based on a hyper-parameter $\alpha$, which is consolidated based on the "teacher-student approach". In order to achieve that goal, data from a replay buffer is used. In addition, another version is specified that selects the data in the buffer based on a higher likelihood.

Interpretable Continual Learning (ICL) is a CL approach presented by Adel, Nguyen, et al. (2019). It is based on a framework called Variational Continuous Learning (VCL) (Nguyen, Li, et al. 2018) which uses a replay buffer to gain CL performance. This model is extended by the introduction of "saliency map". Saliency maps define the most relevant properties (parts) of an image (Zintgraf, Cohen, et al. 2017) (also known as Prediction Difference Analysis (PDA)). For each sample, a saliency map is created. Moreover, it is used for the training of the learner, which in turn controls the focus.

Kamra, Gupta, et al. (2017) present the architecture referred to as Deep Generative Dual Memory Network (DGDMN). Like the work of Shin, Lee, et al. (2017), the DGDMN is based on a generator and solver. The main difference is the provision of several generators. Generators are labeled Short-Term Task Memory (STTM) and realized by Variational Autoencoders (VAEs) (Kingma and Welling 2013). In contrast, the solver represents the Long-Term Memory (LTM). Similarly to the processes during human sleep, the knowledge of the STTMs is transferred to the LTM.

## 3.3   Discussion

Initially, the discussion of related work focuses on the presented evaluation strategies. Subsequently, a critical discussion with regard to the emerging CF avoidance models follows.

### 3.3.1   Evaluation Strategies

First of all, the presented evaluation strategies describe the evaluation process insufficiently and with few details. Instead, the type of data/tasks used for the evaluation is described. To be precise, related work usually explains how the data is re-structured for CL tasks. Other important aspects concern the real-world requirements and constraints, which are represented in several different works. The presented real-world scenarios, however, tend to be rather fictional than relevant use cases. In general, the related works do not describe the entire CL training process including hyper-parameter selection and the training/evaluation process. This can be considered as an open issue that is often neglected when describing the experimental setup. The selection and justification of hyper-parameters is often stated. In contrast to that, the process of determining the parameters is not. These details of the process can be interpreted very differently, and the implementation leads to very different outcomes. Thus, the evaluation scheme is crucial for the review of ML models in CL scenarios. Even though the obtained results are correct, the specific circumstances remain unspecified. To conclude, the presented criticism of other evaluation schemes constitutes one of the main desiderata and, thus, the basis for this work.

### 3.3.2   Existing Catastrophic Forgetting Avoidance Models

CL as a research area offers a great number of applications by now and therefore great opportunities. In the last few years, this trend has been on the increase. The gained traction is clearly illustrated by the number of newly developed deep learning models. More and more of them emerge, and every single one is improving the CL performance.

In each of the related works, the reviewed procedure is almost the same: A novel model is introduced, followed by an experimental setup with the evaluation scheme, and finally it is shown that the CF effect decreases compared to other models. Many related works refer to the CF problem as "resolved" (Lee, Kim, et al. 2017), "controlled" (Serra, Suris, et al. 2018), "mitigated" (Chaudhry, Dokania, et al. 2018), or "avoided" (Nguyen, Li, et al. 2018) – just to provide a few examples. Accordingly, the fundamental question is why should the CF problem still be relevant at all? Interestingly, new models are being introduced, as mentioned. Every one of them claims to improve the CL performance compared to previous models. But why does the work on new, improved models continue, if the CF problem is already solved and the "infinite accumulation of knowledge" is possible? Accordingly, the existence of such a model or mechanism would result in the application in numerous scenarios.

Another issue with many of the related works does not concern the presentation of the models or new mechanisms, but often the poorly detailed circumstances under which the models are evaluated. The reproducibility of results is difficult in many cases – or simply impossible. Without the specification of the exact requirements/constraints/circumstances, a statement concerning the CF behavior is difficult to judge. In addition, the code base, if presented at all, only includes the model, but not the training or evaluation scheme. This may be due to a lack of the community's interest, or to space limitations. Unfortunately, the seemingly unimportant, precise description of the evaluation procedure, is too often omitted in the literature.

# 4. Research Design

## Chapter Contents

This chapter specifies the research procedures applied to achieve the objectives of this work. The starting point is constituted by four research questions, which are divided into more specific sub-questions. They define the structure of this dissertation. In the following, the detailed steps and processes of the investigation are presented. The same is true for the interrelation of the research questions. As several conclusions are based on empirically validated results, the setup of the conducted experiments is presented. This includes a brief description of the experimental environment and how the experiments are performed.

## 4.1 Research Questions in Detail

The previously introduced research questions (see section 1.2) are presented in the following along with the respective sub-questions. The research design is derived from the research question and it is represented in this work's structure.

This research examines the *continual learning* (CL) paradigm for *machine learning* (ML) models, especially Deep Neural Networks (DNNs). In this context, the *catastrophic forgetting* (CF) effect in deep learning methods is of special interest. The CF effect describes that all stored knowledge is lost, as soon as new knowledge is tried to be added to an already trained ML model. Some works claim to have solved this problem, but this only accounts for special conditions. However, these conditions do not reflect the real-world, application-oriented requirements of CL scenarios. As a result, the application of available approaches may cause problems in real-world scenarios.

Therefore, one of the goals of this work is to develop an evaluation protocol that reflects real-world requirements. This will be used to investigate ML models that are intended to prevent the CF effect. In response to the problem, a novel method is to be developed. This method will be evaluated according to the application-oriented protocol. Taking these goals into account, the following detailed research questions are derived:

**RQ 1:** *How can an application-oriented validation protocol be modeled to detect catastrophic forgetting?*
The aim of this first research question is to develop a verification method to detect the avoidance of the CF effect. Requirements and constraints derived from a real-world CL scenario are included. Through the use of a uniform and application-oriented evaluation protocol, a consistent and comprehensive investigation is targeted. In order to answer this first question and develop the application-oriented validation protocol, the following sub-questions are raised:

**RQ 1.1:** *Which requirements can be derived from a real-world application scenario?*
In order to answer this question, requirements are identified by examining a real-world scenario. The applicability of deep learning models will be investigated along with the emerging restrictions related to CL behavior.

**RQ 1.2:** *How do machine learning models behave that are affected by catastrophic forgetting?*

The first step involves an investigation of how models respond when presented with a CL task. The goal is to make the occurrence of the CF effect detectable and measurable. In order to achieve this, a metric has to be defined. The metric is supposed to measure what happens once new knowledge is added to models that already derive existing knowledge from data. The question can be answered by conducting first experiments and by considering the literature.

**RQ 1.3:** *How are existing models examined for the catastrophic forgetting effect?*

The investigation of other evaluation methods from related works begins with their identification and classification. As a part of this process, promising methods or ideas are extracted and adapted for further steps of this research.

**RQ 2:** *To what extent can existing machine learning models avoid the catastrophic forgetting effect?*
The second question has the goal to examine existing CF avoidance models for their suitability in CL scenarios. For this purpose, the application-oriented protocol is applied, which has been developed as an answer to RQ 1. Well-known models are evaluated by means of different learning problems in various variants. Since the models and the resulting performance usually depend on a wide range of parameters, a large-scale investigation is conducted. The aim of the study is to show to what extent the CF effect is suppressed by one or more models. Again, the following sub-questions are supposed to specify the main RQ 2:

**RQ 2.1:** *Which adapted deep learning models avoid the catastrophic forgetting problem?*
This question will be answered by a systematic literature investigation in the context of CF (see chapter 3). A prerequisite for a balanced and reproducible evaluation is, however, the provision of a code base by the authors of the respective research. Hence, all provided data and information needs to be processed in the same, or comparable way, by all investigated models.

**RQ 2.2:** *How can machine learning models be benchmarked under real-world conditions?*
This research question aims at the development of an evaluation framework. The goal is to investigate existing ML models in a consistent manner and environment. By developing a new interface for training, testing and evaluation, all of the respective models are examined for the occurrence of the CF effect and its CL performance.

**RQ 3:** *How can a novel deep learning model be designed so that it avoids catastrophic forgetting?*
The aim of this question is to present a new or adapted ML method, if none of the investigated models provides satisfactory results under real-world conditions. Elements and methodologies of existing models (RQ 2) can serve as a basis. Alternatively, a completely new method in the field of deep learning may be developed. RQ 3 is composed of two sub-questions:

**RQ 3.1:** *How do existing deep learning models attempt to control the catastrophic forgetting effect?*
After the evaluation of models in RQ 2, the best ones are examined in greater depth with regard to the used techniques and methods. The goal is to identify the most effective methods and, if necessary, transfer them to other models.

**RQ 3.2:** *Which machine learning models are not deep, but still avoid catastrophic forgetting?*
Since CF results from the distributed representation of knowledge, alternative models are examined. This aims at finding an answer to the question whether parts or entire models can be transferred into a layered structure, which allows them to be considered deep learning models.

**RQ 4:** *How does the novel model perform when compared to existing continual learning models?*

This question aims at the evaluation and comparison of the novel model and the already investigated models. The CF detection protocol from RQ 1 is applied for this purpose.

## 4.2   Research Structure

The research structure reiterates the relation between research questions and the structure of the work. For this purpose, the questions including specified sub-questions are summarized and displayed in figure 4.1.

The first research question RQ 1 focuses on the development of an application-oriented evaluation protocol. With the help of this protocol, existing models are going to be examined with respect to the CF effect. The goal is to show whether certain models can control the effect and if they can be compared.

Since the focus of this work is on real-world applications, requirements of realistic CL scenarios need to be gathered at first. In order to understand CL scenarios, an exemplary application is presented. The identified requirements of the real-world scenario constitute the answer to *RQ 1.1* (see chapter 5).

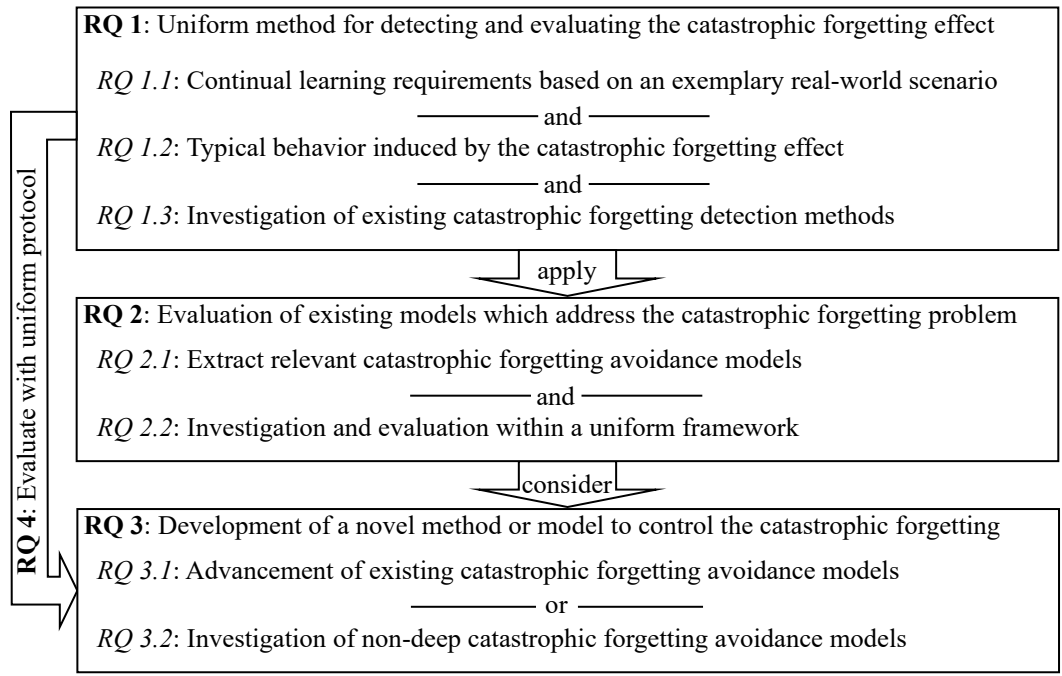

Figure 4.1: Research structure and interrelations of research questions.

For the development of a research protocol, CF has to be quantified. Although the effect is generally known, the question arises how CF can be measured and which conditions are required for its measurement. While an experiment illustrates how the CF effect manifests (*RQ 1.2*), the effect is described in the literature. CF avoidance models are usually evaluated in the respective research papers. However, a detailed description of how these models are evaluated is often unavailable. This makes it difficult to compare the existing results, which is why this work argues for a standardization of evaluation protocols.

The respective models along with the corresponding evaluation methodologies are presented in chapter 3 (*RQ 1.3*). As a result for RQ 1, a detailed evaluation protocol is proposed in chapter 6. It is one of the important contributions of the present work.

The second research question **RQ 2** aims at the investigation of a subset of the presented CF avoidance models from chapter 3. The objective is to evaluate the CL capabilities of the models under a uniform protocol. For this purpose, the application-oriented protocol from chapter 6 is applied. RQ 2.1 and RQ 2.2 address whether the models are compatible with the evaluation protocol and how a unified validation can be implemented. Results of the study and lessons learned are presented in chapter 7 as an answer to the second research question.

Research question **RQ 3** focuses on the development of a novel model in order to address the CF effect. Based on the results of the previous study (RQ 2, see chapter 7), an existing method may be developed further (*RQ 3.1*). Alternatively, the development of a novel ML model may be more beneficial (*RQ 3.2*). Presumably, models and techniques that inherently do not exhibit the CF effect can be used. The transfer of an existing model into the deep learning context is a prerequisite for fulfilling the given requirements. This should result in a new model (chapter 8) answering **RQ 3**.

The present work is concluded by answering the last research question **RQ 4** in chapter 9. For this purpose, the newly developed deep learning model is transferred into the CL context. At the same time, the evaluation of the new model (chapter 8) is performed by means of the CF detection protocol as described in chapter 6. In order to answer **RQ 4**, the results are compared to the evaluation of other CF avoidance models.

All research questions are revisited and answered in chapter 10. In addition, the contents and findings of the individual chapters are briefly presented. An analysis of the individual components of the overall work concludes the discussion.

## 4.3   Exhaustive Grid Search

Empirical experiments are conducted to draw conclusions about the investigated ML models. Since many parameters influence the results of ML methods, the significance of single experiments requires a critical assessment. One crucial factor refers to hyper-parameters, another one to the random initialization states. Both of them can heavily influence the results. In order to guarantee the validity of statements, a large number of parameter configurations has to be evaluated. This in turn requires a corresponding amount of computing capacity, so that a sufficient number of experiments can be conducted. Compliance with a realistic time scale is assumed, as a 20 year duration exceeds the limit. One of the prerequisites is the independence of experiments, which allows for parallelization. Another prerequisite is the availability of sufficient computing capacities, as it is a large-scale investigation. Due to the limited conditions, a special system was developed that allows for the computing of massive experiments by using further computational resources.

The software developed for this purpose is specially designed for the performance of distributed experiments and the collection of results. Additionally, a flexible option for the generation and evaluation of experiments was implemented. A brief description of the environment is presented in the following. It outlines the selected experiments and the distribution strategy. Finally, the module providing the different datasets is presented.

**Experiment Distributor**   As the present research requires immense processing capacities, the parallel processing of experiments seems an adequate solution. The basis for the parallel execution of experiments is referred to as *experiment distributor* in this work. The individually parameterized experiments are distributed to available compute nodes. Interestingly, the used nodes are regular computers located in the university's laboratories. These computer labs are usually used by students and lecturers during week days. On weekends and at night, these resources are available and can therefore be used for research.

The experiment distributor is a lightweight tool and follows a simple server/client architecture. The server starts a web server via Secure Shell (SSH) on the client side, which is used for further communication. Subsequently, experiments can be launched in the form of independent processes by a Representational State Transfer (ReST) interface. The state of an experiment is monitored by its process ID. Once an experiment is completed, the resulting data/files are collected. Since the available nodes contain different hardware components (see table 4.1), the number of experiments can be limited individually per node.

Table 4.1: Used computational resources.

| #  | GPU Type        | CUDA Cores | Memory (GB) | Clocking (MHz) | CPU ($\# \times$ GHz) | RAM (GB) |
|----|-----------------|------------|-------------|----------------|-----------------------|----------|
| 1  | TITAN Xp        | 3840       | 12          | 1582           | $24 \times 2.3$       | 64       |
| 1  | GTX 1080 Ti     | 3584       | 11          | 1480           | $24 \times 2.3$       | 64       |
| 10 | RTX 2080        | 2944       | 8           | 1515           | $8 \times 3.4$        | 16       |
| 2  | GTX 980 Ti      | 2816       | 6           | 1000           | $24 \times 2.3$       | 64       |
| 20 | Quadro P5000    | 2560       | 16          | 1600           | $16 \times 3.2$       | 32       |
| 40 | RTX 2060 Super  | 2176       | 8           | 1470           | $8 \times 3.6$        | 16       |
| 20 | Quadro K2200    | 640        | 4           | 1046           | $24 \times 2.4$       | 32       |
| 90 | Quadro P620     | 512        | 2           | 1354           | $16 \times 3.6$       | 48       |

Despite the heterogeneous hardware, various operating systems are used (i.e., Windows and Linux) Due to the tailored interface, the invocation of a remote method is generic. Therefore, heterogeneous operating systems can be utilized. As some operating system-specific differences need to be considered, they are implemented individually by the distribution tool. Moreover, there is another prerequisite for the use of a node. This is due to the used ML library TensorFlow, which is designed for a specific hardware manufacturer. Thus, an Nvidia GPU and the ML framework need to be available.

Before an experiment can be executed on a node, all prerequisites have to be met. This includes the required software modules (e.g., Python), libraries (e.g., TensorFlow) and, of course, the experimental code. These requirements are checked and resolved before the actual experiments are executed. In order to perform an individual execution of the experiments, it is assumed that the underlying code base can be controlled by command line parameters. These can be transferred easily to the individual nodes/clients and finally be executed. Result files can be traced back from an experiment ID and mapped to the command line parameters. Additionally, the parameter configurations are recorded as part of the result files.

The distribution proceeds as follows: Potential nodes are identified on the basis of a configuration file, which checks for accessibility and requirements. Likewise, the experiments are imported and prepared specifically for different operating systems by adjusting the command line parameters. If the processing time of a node is within a free time slot, one or more experiments are started depending on the node definition. In order to do this, the selected parameters are transmitted and a corresponding process is started. As soon as the processing ends or a timeout is triggered, the result files are collected and stored. However, a node may fail during its designated time slot, or it may not work due to the choice of parameters. In both of these cases, the experiment is repeated or marked as failed.

**Experiment Generator**   Another component of the experiment distribution tool is the *experiment generator*. It allows for the flexible creation of experiments and their command line parameters. On the one hand, the experiment generator combines single possible hyper-parameters. On the other hand, it implements additional features. Accordingly, dependent parameters can be added, where the parameter $b$ is set depending on parameter $a$. Furthermore, experiment IDs are assigned, the result files are specified and the number of experiment repetitions can be determined.

**Experiment Evaluator**   Another component of the developed software is responsible for the evaluation. In this context, command line parameters can be specified according to the aggregated and evaluated experiments. The repeated execution of individual parameter configurations can, for example, be used for an aggregation. In this case, the Comma-separated values (CSV) or JavaScript Object Notation (JSON) files are imported, converted, aggregated and evaluated according to the specified method. Finally, plots can be generated or results can be compiled for the generation of tables.

**Experiment Dataset**   Various types of benchmark datasets are used for evaluation. These are usually available in the form of public datasets. For the distributed application itself, the datasets need to be available on each computation node. This is provided by the *experiment dataset* component. If a dataset does not exist on the node, it is automatically downloaded and stored in a pre-processed format on the local node. After these pre-processing steps, the pre-processed datasets can be loaded into different experiments via an interface. As a last step, they are further processed/adjusted according to the specified command line parameters. The command line parameters can be used to define different types of dataset divisions for each individual experiment.

# 5. Exemplary Real-World Scenario

## Chapter Contents

In this chapter, an applied scenario of *continual learning* (CL) is introduced. It will be used to derive realistic requirements that are imposed on a typical application (based on the publications Pfülb, Hardegen, et al. 2019; Hardegen, Pfülb, et al. 2019; Hardegen, Pfülb, et al. 2020).

The scenario represents a section of today's standard computer networks where routing algorithms are responsible for path finding. When a network packet is sent, the computer behaves as if sending a letter, a package or even something bigger. Likewise, a number of decisions has to be made in order to successfully deliver the packet. Moreover, the context needs to be considered. In this example, the sender is not concerned with the transportation route of the package. This decision is made by a carrier or a delivery chain. The transportation route is optimized based on various criteria. If the package is too large, e.g., a wind turbine, a heavy load transport needs to be organized. This in turn limits the selection of the route, e.g., with regard to the maximum street width or the load capacity of bridges.

Transferring these concepts to the computer network scenario, the following challenge arises. The scope of a communication (e.g., number of packages or duration) between computers of a network is unknown prior to its initiation. In conventional networks, this information cannot be used to determine the path. Therefore, as a rule, the shortest path is selected, which is not necessarily the best/fastest one. The shortest path, however, does not take into account whether the route is suitable, for example due to bottlenecks or traffic jams (depicted in figure 5.1). Thus, a significant opportunity to improve the process is constituted by the prediction of a communication's scope.

Nevertheless, the explained scenario is subject to temporal changes. Various events may influence the behavior of transmitters, e.g., a pandemic or recurring events like holidays. New applications, updates or simply the users' behavioral changes transfer this problem into the CL problem.

**Contributions**   This chapter presents an application-oriented scenario that helps derive requirements for CL scenarios. The resulting requirements serve as an answer to research question *RQ 1.1* (see section 4.1).

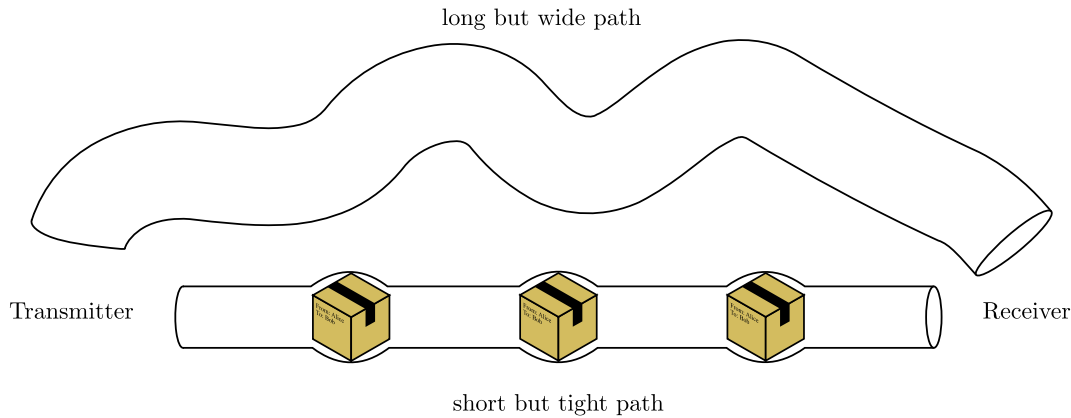

long but wide path

Transmitter                                                                    Receiver

short but tight path

Figure 5.1: Outline of the exemplary real-world problem.

The most important contribution is the definition of real-world requirements and constraints. These have to be addressed by *machine learning* (ML) applications in the context of CL scenarios. Accordingly, it is crucial to include them into the evaluation protocol for measuring the CL performance. The resulting protocol therefore attempts to guarantee the validity of the investigated models in an application-oriented context.

**Structure**  The structure of the chapter is as follows: First, the detailed problem description is presented including a simplified use-case (see section 5.1). Moreover, problem-specific related work with regard to traffic-classification is summarized. Subsequently, an overview of the data acquisition and pre-processing steps is provided (see section 5.2). In section 5.3 a data analysis is performed with regard to the exemplary CL problem. Also, this chapter outlines the results of all related experiments in section 5.4. The present chapter is concluded by deriving and summarizing requirements (see section 5.5) from this real-world scenario for the CL context.

## 5.1  Problem Description

Computer networks are used to exchange information between connected nodes. In order to enable communication within a computer network, systems must be identifiable so that a message exchange is possible. The connections between individual components of a network are known as *links*, where the complete route represents the *path*. The path is determined via the destination address. For the sake of simplicity, the term path is used in this work, even though technically a link is denoted (path of length 1). As a metric for determining the "best" path, the number of *hops* (routers on the way to the destination) is generally used.

Problems may occur in case a path is already utilized or even overloaded. This challenge can be compared to a traffic jam. Typically, it is advisable to consider bypassing the concerned section of the route. The same solution is usually suggested by navigation systems. However, this requires the deviation from the original route. The pitfall in terms of network communication is that the decision about the path is made as soon as the connection is established. Thus, redirecting the communication is hardly possible. This lack of flexibility is due to the stateful network components (so-called middleboxes) on the way to the target, such as firewalls (Brim and Carpenter 2002; Iyengar, Raiciu, et al. 2011). The described phenomenon is particularly true for connection-oriented network protocols such as Transmission Control Protocol (TCP).

One approach to address the problem is determining the size of an emerging communication in advance. However, two pieces of information need to be available as a prerequisite: First, the maximum possible bandwidth and the current utilization of the path is required. The utilization information can be obtained from the network components by means of telemetry data. Second, the bandwidth required for the network communication has to be available. The bandwidth (bits per second) is constituted by the ratio of transmitted data volume (bits) to their transmission time (duration in seconds). Part of the challenge is that meta-data of a communication are usually available after its termination. A collection of meta-data is referred to as *network flows* or in short – *flows*.

The prediction of information, such as the bit rate of a flow, is not trivial. This determination can be addressed by ML techniques. By using an ML-based prediction, an optimal path can be selected before the connection is established. The determination of the path can be realized by using Software Defined Network (SDN) techniques (Kreutz, Ramos, et al. 2015). These allow for specific paths to be programmed for individual flows based on the underlying network components. Besides the efficient utilization of the network resources, other application scenarios may be considered.

Conventional computer networks consist of a large number of different types of components and autonomous systems. Such systems usually follow fixed rules so that predictions are somewhat easier. Backups are, for example, often scheduled during night-time, resulting in a load on the network. A more complex load of the network is caused by humans. The respective traffic results from applications or other online activities. The latter are influenced by many factors, e.g., time of the day and year, the used applications, new trends, hashtags, habits, etc. Especially the load caused by humans leads to a high variance and constantly changing flow patterns. Due to these continuous changes, this exemplary scenario is set in the context of CL.

In order to train a supervised ML model, data needs to be collected for training. Meta-data of terminated/closed network connections can be used for this purpose, in this case the training of Deep Neural Networks (DNNs). The relevant meta-data primarily includes the so-called 5-tuple consisting of a source/destination IP address, the source/destination port and the protocol. Since the flows are already completed, additional meta-data is available, such as the required bandwidth of the flow. The goal is to use an ML model to predict the target values (e.g., bit rate) only based on the data stored in the 5-tuple. A visualization of the prediction problem is available in figure 5.2. The figure displays a transmitter on the left, as well as a receiver on the right-hand side. The flow meta-data (the 5-tuple), which is available once a communication is initiated, is classified using an DNN. In this example, the bit rate serves as the label which has to be predicted.

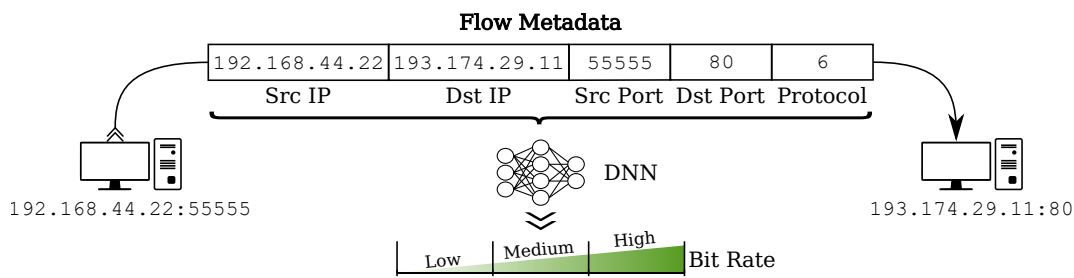

Figure 5.2: Use of ML to enable the classification of network flows.

#### 5.1.0.1   Use Case

In the following, the outlined application scenario is illustrated in form of a use case. The use case indicates the disadvantage of path determination with the standard decision metric, i.e., the number of hops. Using only the number of hops can, however, lead to the overloading of a path, so that other paths are slightly utilized. The uneven utilization of the network can cause interferences which affect all flows on the route. This in turn can lead to an increased latency or even to packet loss/drop. Ultimately, the application and user experience can be affected.

Figure 5.3 shows a simple computer network that consists of 7 routers $\{R_1, \ldots, R_7\}$. There are 3 paths ($P_x$) between $R_1$ and $R_7$, which serve as source and destination. $P_1$ consists of routers $R_1$, $R_2$ and $R_7$. The second path $P_2 = (R_1, R_3, R_4, R_7)$ and the third $P_3$ is defined by $(R_1, R_5, R_6, R_7)$. At path $P_2$ and $P_3$, four routers have to be traversed. The path $P_1$ is the shortest one, with three router hops and therefore the preferred path in a classic routing scenario. It is assumed that each path (link) offers the same bandwidth (capacity of 100 %). In this scenario, three different load levels are distinguished: low ($<10\,\%$, green), medium ($10\,\text{-}\,90\,\%$, orange) and high ($>90\,\%$, red). The challenge is to route three flows from source $R_1$ to destination $R_7$. Flow $f_1$ and $f_3$ requires a bandwidth capacity of 40% each, whereas flow $f_2$ only requires 30%. Since $P_1$ is the shortest path, all three flows are routed via $P_1$. This results in a congestion of $P_1$ with 110% (over)load. At the same time, the paths $P_2$ and $P_3$ are not utilized at all. The scenario displayed in figure 5.3 shows that routing one flow via a free path would solve the congestion. However, this is not possible due to the used metric (minimum hops). In order to solve this problem, the ahead-of-time estimation of the required bandwidth of a flow can be used as a more "intelligent" metric.

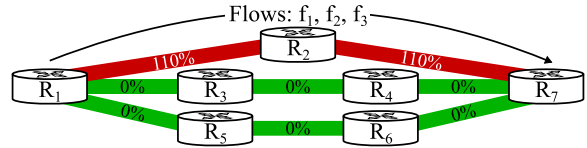

Figure 5.3: Use case scenario with congestion on a single shortest path.

Assuming that it is possible to predict the bit rate of a flow in advance, the scenario can be modified. In this case, the structure of the network, as shown in 5.3, does not change. Based on the prediction, however, the expected bit rate of the flows is (perfectly) known: $f_1$, $f_3$ = 40% and $f_2$ = 30%. The three flows occur sequentially, so that a path for $f_1$ is determined first. Since the utilization for the shortest path $P_1$ ($R_1$, $R_2$, $R_7$) is not yet at full capacity (0% + 40% < 100%), $f_1$ is routed over it. The same applies to $f_2$. For both paths, the current load is unproblematic (40% + 30% < 100%). Only the attempt to route the last flow $f_3$ over the shortest path can lead to a predicted overload and detection. Thus, an alternative path is selected, which is longer but less utilized. $P_2$ and $P_3$ are available for selection. A random choice between both options is sufficient, since both are equally utilized and have the same length.

The result of the prediction-based routing can be seen in figure 5.4. It is obvious that no congestion occurs on the shortest path and that the network is therefore equally utilized. Similarly, the flows are not suffering from congestion, which can result in a lower latency. To conclude, the application of prediction-based routing can lead to a higher quality of an application.

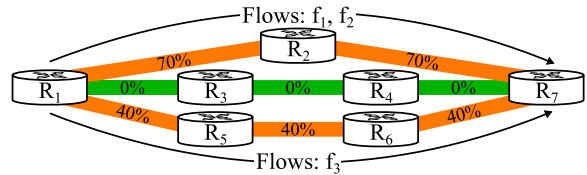

Figure 5.4: Preventing path congestion by throughput prediction.

### 5.1.1   Related Work

The scenario is located in the research area of computer networks, especially traffic engineering. This specific area has been the focus of research for many years (Awduche 1999; Agarwal, Kodialam, et al. 2013). In recent years, the application of ML models has been established (Wang, Cui, et al. 2018; Fadlullah, Tang, et al. 2017a; Yao, Mai, et al. 2019; Rusek, Suárez-Varela, et al. 2019; Zhuang, Wang, et al. 2019). This is due to the fact that the hardware required for machine learning has become so powerful. Another aspect is the performance that can be achieved with ML models. Since most computer networks are still routed in a destination-oriented manner, the use of ML is a promising area of research and advancements. An interesting application is the prediction of traffic characteristics, e.g., the classification into "mice" and "elephant" network flows (Mori, Uchida, et al. 2004). This approach may achieve an optimized utilization of capacities (Valadarsky, Schapira, et al. 2017). The same is true for predicting the duration of a flow (Jurkiewicz, Rzym, et al. 2018). Basic summaries of the entire subject area are outlined in the literature ( Nguyen and Armitage 2008; Fadlullah, Tang, et al. 2017b; Boutaba, Salahuddin, et al. 2018; Mohammed, Mohammed, et al. 2019; Zhang, Huang, et al. 2019 and Wang, Cui, et al. 2018).

Recent developments in the ML area have advanced their application in the research area of computer networks. However, there is very few related work that addresses network flow prediction in terms of *continual learning*. Another issue of related work is the type of data being used – either synthetic, or real-world data. For synthetic data, a generator is required. The problem of generating processes is that ML models may easily derive the generating function. This is mostly due to the lacking complexity of the generator so that it cannot necessarily represent a real-world problem.

**Data Gathering**   The viable alternative is to collect real-world data, which in turn entails challenges. A comprehensive collection of real-world data is supposed to include at which points within the network the data is collected. Moreover, the question arises whether the data is representative at all. Another technical challenge during data acquisition concerns the used components. Even though data has to be captured by the hardware components, information extraction is not their primary function. At the same time, the amount of generated data is too large to store. For this reason, processing the data in a stream is the only option. Furthermore, it is often difficult to obtain datasets from network traffic because they are either outdated, contain too few data elements, or are simply not published for privacy reasons. In the works of Poupart, Chen, et al. (2016); Azzouni, Boutaba, et al. (2017) and Xiao, Qu, et al. (2015), data collection is performed by using OpenFlow (McKeown, Anderson, et al. 2008). Poupart, Chen, et al. (2016) criticize the scaling of the collection strategy, as it does "not scale well". This is mainly due to the available memory of the used hardware (switches or routers). The hardware limitation sometimes leads to packages not being included in the statistical analysis. As a result, the data may contain a bias that represents a distortion of reality.

Other works capture all or parts of the network traffic and select "elephant" flows, which are labeled with Quality of Service (QoS) class labels in the second step (Wang, Cui, et al. 2018). Deep packet inspection is used for this purpose. In the work of Shi, Li, et al. (2017) real-world data is used. However, the data was extracted from a PC room with about 20 computers. Moreover, the resulting dataset is not accessible. In contrast to the already presented works, simulators are applied (Rastegarfar, Glick, et al. 2016; Valadarsky, Schapira, et al. 2017). Benson, Akella, et al. (2010) analyze network traffic from multiple data centers and conclude that this data is very complex and data center specific.

**Binary Classification**   A technical detail presented in almost every paper is the classification between "mice" and "elephant" flows. The regression problem is thus transformed into a binary classification problem (Poupart, Chen, et al. 2016; Xiao, Qu, et al. 2015; Valadarsky, Schapira, et al. 2017). In this context, the question of the ground truth of data arises (see section 6.1.1, Wang, Cui, et al. 2018). The choice of the threshold is problem-specific and must be adapted to the scenario. A study by Poupart, Chen, et al. (2016) examines different threshold ranges. Likely, Rastegarfar, Glick, et al. (2016) and other authors (e.g., Jurkiewicz, Rzym, et al. 2018) encounter the large imbalance in the distribution of data within these two classes. Benson, Akella, et al. (2010) highlight this as a problem in their research. In contrast to other works, no size of the flow is determined by Reis, Rocha, et al. (2019), but an additional step is used to directly make a path-based routing decision.

**Features for ML**   Likewise, related work deals with the question of which features/attributes are collected and how they are transformed. The question *which* features are used is often answered with the 5-tuple (source/destination IP, source/destination port and transport protocol) (Xiao, Qu, et al. 2015). However, there are studies focusing on the first three transmitted packets, such as Poupart, Chen, et al. 2016. The work of Wang, Cui, et al. (2018) considers as many as 20 packets. It is their goal to determine whether the flow is a client request or a server response. In order to evaluate the feature transformation performed by related works, it is often necessary to study the program code (Poupart, Chen, et al. 2016) – if it is published (Xiao, Qu, et al. 2015). Although the normalization of data is mentioned in Azzouni, Boutaba, et al. 2017 and Reis, Rocha, et al. 2019, it is not clear how exactly the individual features are converted into the value range $[0, 1]$.

**Offline vs. Online Learning**   A key factor for the online or offline training of the ML models is the data processing strategy. In this context, the memory ($\mathcal{M}$) and time complexity ($\mathcal{O}$) for inference and training is the subject of discussion. This especially refers to the used ML models. A study by Poupart, Chen, et al. 2016 uses online methods (DNNs, Gaussian Process Regression and Online Bayesian Moment Matching), allowing for the processing of theoretically infinite large data streams. This ability mainly concerns the incremental training of models. Other works do not address this issue or directly refer to the offline learning paradigm (e.g., Wang, Cui, et al. 2018).

## 5.2   Flow Data Stream Pipeline

The example scenario presented in section 5.1 is implemented by means of an experimental application. The application's core is constituted by the so-called Flow Data Stream Pipeline (see figure 5.5). This pipeline describes the processing stages of the application step by step. It includes the collection, the pre-processing as well as the learning and evaluation process of the data by an ML model. In order to increase flexibility, processing components are logically separated. As a result, all processes that do not depend on parameters are implemented as a server application (Flow Data Stream Server). The Flow Data Stream Client can be adapted to different conditions by parameterization. This allows for the performance of a large number of experiments without the necessity of major adjustments.

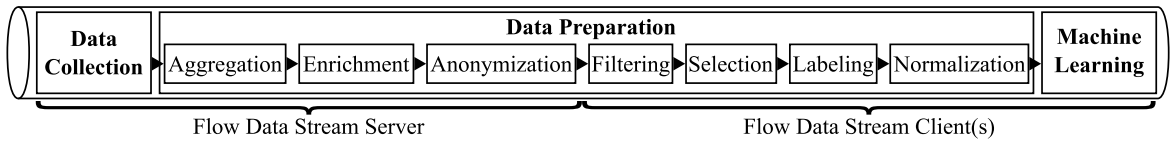

Figure 5.5: Flow Data Stream Pipeline.

## 5.2.1  Flow Data Stream Server

The first half of the Flow Data Stream pipeline is implemented as a server application. It is responsible for collecting data (as described in section 5.2.1.1) and performing initial processing steps (detailed in section 5.2.1.2). The division into a server-client architecture has several advantages. On the one hand, processing steps that apply to all clients receiving data from the Flow Data Stream Server are combined. On the other hand, the executed processing steps are subject to data protection. As a consequence, the Flow Data Stream Server needs to be hosted by an authorized institution (e.g., automated processing of IP addresses).

### 5.2.1.1  Data Collection

The data collection is one of the most challenging tasks with the goal of providing valid data. Since most of the publicly available datasets are either outdated, synthetic, or unrealistic, a new dataset is considered useful (see section 5.1.1). Therefore, real-world data from a productive university network is collected.

**Network Architecture**  The selected campus network is a commonly used network structure, namely a core distribution access model. The network connects approximately 30 buildings. Each building represents an element of the distribution layer, which is connected to the core routers (see figure 5.6). The network connects sub-nets: The data center, laboratories, administration, research facilities and WiFi. The network provides access to approximately 1 000 staff members and 10 000 students. Essential IT services, such as mail or Domain Name System (DNS) services are provided by the university data center. The presented network structure model allows for the collection of network flow data at central points (at the two core routers: CatE and CatM). Thus, the resulting dataset consists of realistic samples of real-world network traffic. However, flows that are routed within one and the same building cannot be included into the analysis.

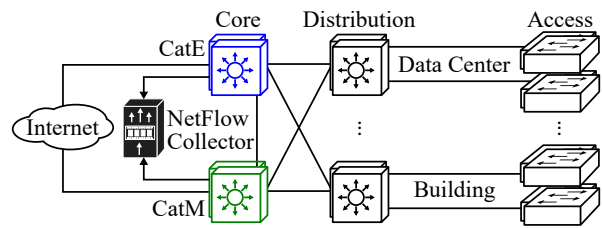

Figure 5.6: Infrastructure model of the campus network in which flow data is collected.

A NetFlow collector is used for the collection of the flow data. It is integrated into the Flow Data Stream Server and allows for the reception of exported flow data from the two primary routers in the Cisco Flexible NetFlow format (Cisco 2008; Claise 2004).

**Flow Export**   As previously mentioned, the two core routers continuously export the flow data via the NetFlow protocol. Thus, it is possible to collect realistic network traffic within the campus network structure. Individual packages of a flow are identified by matching criteria. The matching criteria comprise all elements of the 5-tuple. This includes the information if a flow is complete (TCP connection terminated, Claise 2004) and exported. If this is the case, additional meta-data is provided.

The so-called collection criteria include timestamps for the start and end of each flow, the number of packets and bytes transferred, and, if present, the TCP flags. Moreover, two available timeouts are used to trigger a flow export. If no packets are forwarded for a flow within a certain period, the inactive timeout (e.g., 30 s) triggers the export of the flow. An active timeout (in this case 600 s) is responsible for longer lasting flows. The latter may lead to incomplete flow exports. A device's cache size, however, may limit the maximum of the currently traced flow records. Due to the cache size limitation and timeouts, exported flows may not describe a complete communication. Accordingly, unfinished flows are referred to as *flow records*. Completed flows are labeled as *flow entries*.

The export of the corresponding devices is configured for incoming traffic only. This setup prevents flows from being exported twice by one component. However, flows that are routed through both components (switches) can still be exported twice. Only IPv4 and unicast flows are investigated, as the amount of other traffic (e.g., multicasts or broadcasts) is very low in this network. Disregarding the differences between daytime and nighttime leads to an average of about 2 000 processed flows per second.

### 5.2.1.2   Server Side Data Preparation

Before the exported flows can be processed by the ML model, they have to be prepared. Parameter-independent pre-processing steps are implemented in a server application. As a first step, flows are collected until a certain number (100 000) is reached. The collection of 100 000 flows is referred to as *block*. In case a block is completed, the flow records are pre-processed in the following steps: Data aggregation, enrichment and anonymization.

**Data Aggregation**   Due to hardware limitations (timeouts and cache sizes), potentially non-completed flow records are exported. Flow records need to be combined into flow entries prior to further processing steps. This is performed individually for each block (100 000 flow records). The meta-data of each flow (5-tuple and timestamp, as well as flags) are used for this purpose. The goal of this procedure is to recalculate the meta-data such as duration, number of packets and bytes, as well as the bit rate of completed flows (flow entries). The disadvantage of this method is that long-lasting flows crossing the boundaries of a block are dropped. Approximately 2 500 flows per block are excluded. At the same time, the aggregation step removes duplicate entries of both exporters ($\varnothing \approx 4\,200$ per block). This mechanism reduces the number of flows to approximately 75% within a block.

**Data Enrichment**   A data enrichment step is performed in order to add internal and external information to the flows. This allows, for example, the addition of a private or public prefix to an IP address. As an example of internal information, the Virtual Local Area Network (VLAN) tag is added. This is realized by using a lookup table and the address prefix of the flow. Using the same principle, the Autonomous System Number (ASN) tag is added among others. A lookup service (MaxMind 2019) is used to add external (public) information. As a result, more information related to a communication's context and level can be added (e.g., the country).

**Data Anonymization**   Data anonymity is an important aspect in the present real-world scenario. First of all, the processing of personal data is sensitive. The collected IP addresses are considered as personal data, as they are assigned to a person at least for a certain period of time. Second, personal data should be protected before the dataset can be published. Furthermore, the internal network structure should not be revealed for security reasons.

The anonymization of IP addresses is realized by parameterized substitution tables. A cryptographically hashed password (seed), which is defined by the data center administrators of the university, is used for this purpose. An individual substitution table is applied to each octet of an IP address. Thus, the semantics of the addresses remain the same despite their adjacency. Additionally, the relation to the network address remains, even though the neighboring addresses are separated. This is the last server-side pre-processing step before the data is transferred to a client.

### 5.2.2   Flow Data Stream Client

The second part of the Flow Data Stream Pipeline (see figure 5.5) is implemented as a client. Hence, the client receives the pre-processed data from the server application. The client is responsible for the parameterized part of the data processing steps (section 5.2.2.1) and the machine learning stage (section 5.2.2.2).

#### 5.2.2.1   Client Side Data Preparation

Before the data flows into the machine learning model, further pre-processing steps need to be performed. Some of them are optional. In this work, optional means that several different parameter configurations are used for the investigation.

**Data Filtering**   The (optional) filter mechanism allows to exclude flows that match one or more filter criteria. These may represent sub-problems, which might be interesting for a corresponding real-world application. This allows, for example, the exclusion of User Datagram Protocol (UDP) traffic, so that only TCP flows are learned and predicted.

**Data Selection**   The (optional) data selection stage reduces the number of features. This constitutes a contrast to the enrichment stage, but supports the investigation of an individual feature's influence. Thus, the quality improvement due to the enrichment process (5-tuple vs. all additional features) becomes measurable.

**Data Labeling**   Target values (labels) are required for supervised machine learning (see section 2.2). In this study, most of the target values for the routing scenario are real-valued numbers. Due to the imbalance of data, the implementation of a regression problem is challenging. Therefore, the regression problem is transformed into a classification problem.

   As part of the data labeling, each flow is assigned to a specific class by means of parameterizable predefined class boundaries. The selection of class boundaries is described in section 5.3.1. The following labels are useful for the given routing scenario: number of bytes or packets, a flow's duration or bit rate. Due to the assignment of class labels, a critical discussion of the ground truth of data (see section 6.1.1) is required. Nevertheless, an approximately equal distribution of the flows within the class is possible (and recommendable, see section 2.2.2.5). All in all, the routing problem shown in section 5.1 can be addressed by the classification of flows. The result of the data labeling is a one-hot encoded (see section 2.2) class label.

**Data Normalization**   Before the ML model can be fed with flows, two last pre-processing steps are performed: The normalization of value ranges and a feature encoding. The normalization is supposed to ensure that no extreme gradients arise and that a convergence of the model is achieved more quickly. Moreover, the appropriate encoding should help ensure the independence of features (LeCun, Bottou, et al. 2012; Bishop 1995).

   With regard to the coding, 3 different formats are offered: *float*, *bit pattern*, *one-hot* (e.g., table 5.1). Simple floating point values are transformed into a given range of values ([0.0, 1.0]) by a min-max normalization. This can be realized easily, as the value ranges for all features are known. The bit pattern format is used to convert real values into their binary representation. It is important to know the range of values the numbers can assume in advance (e.g., for IP addresses). Categorical values are represented by means of one-hot encoding (also known as 1-of-c coding). This avoids proximity, which would be the case with continuous variables. In addition to these standardized transformations of the data, problem-specific modifications are executed. The replacement of dynamic port numbers ($\geq 2^{15}$) by 0 belongs to this data-specific modification.

Table 5.1: Flow feature normalization and encoding examples.

| Feature | Raw Data | Data Type | Output Data |
|---|---|---|---|
| IP address | 81.169.238.182 | Float | 0.317, 0.662, 0.933, 0.713 |
| Protocol | 6 | Bit pattern | 0, 0, 0, 0, 0, 1, 1, 0 |
| Locality | Private \| Public | One-hot | 0, 1 \| 1, 0 |

Table 5.2 represents the possible flow features. Additionally, the three data formats plus the size of output vectors for the individual features are included into the table. The used formats for the analyses and experiments are highlighted in gray. Features marked with the symbol $\rightleftarrows$ do exist twice in the data. In the table, they are used to describe the source (src) and the destination (dst).

Table 5.2: Details of the flow features in the data stream.

| Feature | | Data Format | | | Src | Feature | | Data Format | | | Src |
|---|---|---|---|---|---|---|---|---|---|---|---|
| | | Float | Bit | One-hot | | | | Float | Bit | One-hot | |
| month | | 1 | 4 | 12 | | network | $\rightleftarrows$ | 4 | 32 | ✗ | |
| day | | 1 | 5 | 31 | | prefix len | $\rightleftarrows$ | 1 | 5 | ✗ | |
| hour | | 1 | 5 | 24 | Data Collection | ASN | $\rightleftarrows$ | 1 | 16 | ✗ | Data Enrichment |
| minute | | 1 | 6 | 60 | | longitude | $\rightleftarrows$ | 1 | ✗ | ✗ | |
| second | | 1 | 6 | 60 | | latitude | $\rightleftarrows$ | 1 | ✗ | ✗ | |
| protocol | | 1 | 8 | ✗ | | country code | $\rightleftarrows$ | 1 | 8 | 240 | |
| IP address | $\rightleftarrows$ | 4 | 32 | ✗ | | VLAN | $\rightleftarrows$ | 1 | 12 | ✗ | |
| port | $\rightleftarrows$ | 1 | 16 | ✗ | | locality | $\rightleftarrows$ | 1 | 1 | 2 | |

### 5.2.2.2   Machine Learning

The machine learning module is responsible for the training and inference process of a given model. The concrete ML model can be exchanged easily due to an interface. This allows for a quick and easy implementation of different models, which might be an interesting approach for further research. In the present work, DNNs as a type of ML model are used for network flow characteristic prediction.

First of all, a block of data is provided to the ML module, which consists of a chronologically ordered set of flow entries. In order to perform an evaluation, the block is divided into 90 percent training and 10 percent test data. The chronological order of the flow entries before the division is preserved in order to allow an approximately realistic evaluation. Thus, the test data are part of the future, as far as the model is concerned. The training on a data block takes place until a new block arrives, which depends on the speed of the exported flows. The prediction output of the model depends on the selected label (see section 5.2.2.1).

Depending on the class boundaries, a range for the classification of flows (e.g., duration or bit rate) is specified. A detailed specification of the used ML model (i.e., DNNs) is presented in conjunction with the experiments (see section 5.4).

### 5.2.3   Data Flow of Network Flow Data

The block diagram in figure 5.7 summarizes the flow of data through the individual stages of the Flow Data Stream Pipeline (see figure 5.5). Network flows exported by multiple switches/routers are collected centrally in the Flow Data Stream Server by the NetFlow Collector unit.

This part of the application is based on network sockets, which are used to transmit or receive the flow data ①. Flows are collected until a certain number is reached. The collection is referred to as *block*, whereas a block size of 100 000 is chosen. This approximately equals 1 min if many flows are received. A block (see section 5.2.1.2) consists of flow records, which represent potentially unfinished network flows.

As a next step, the block is forwarded to the Flow Processor ②. It divides the block into several approximately equal chunks based on the 5-tuple ③. Thus, corresponding flow records are placed in the same chunk. The number of chunks $n$ depends on the number of flow processors. These, in turn, are implemented as independent processes in order to take advantage of multi-core processors ④. Thereby, the complex processing steps become scalable and ensure that processing ends before a new block is available.

Each of the Flow Processors performs the same three processing steps ④. First of all, an aggregation is executed so that the potential flow records are converted into flow entries. In this context, the 5-tuple is used to aggregate multiple flow records into a single flow entry by evaluating timers and flags. This is how the meta-data can be recalculated, e.g., the duration. The second step is the data enrichment.

Both, internal and external information resources are used for this purpose. As an example for internal additional information, the network prefix is added by using the IP address and a lookup table. It is also possible to determine the VLAN tag. The latter helps separate a physical network into several logical networks, e.g., data center, WiFi, administration network. A lookup service is used to add further external information (MaxMind 2019). Additional external information includes, for example, the ASN or the country code (see table 5.2). The download of the provided database enables very short query times.

The third and final pre-processing step of the Flow Processors is anonymization. Accordingly, personal data, i.e., IP and network addresses, are masked. An individual substitution table is available for every single octet of an address. The table is initialized based on a password. Thus, the semantics of an address remains while the adjacency is dissolved.

The results of the flow processors are chunks of flow entries ⑤. They are choronologically merged back into a block. The block is then passed to the connection handler ⑥ and transmitted to all registered Flow Data Stream Clients in a compressed form ⑦. It is optional to export the pre-processed block as a dataset file ⑧.

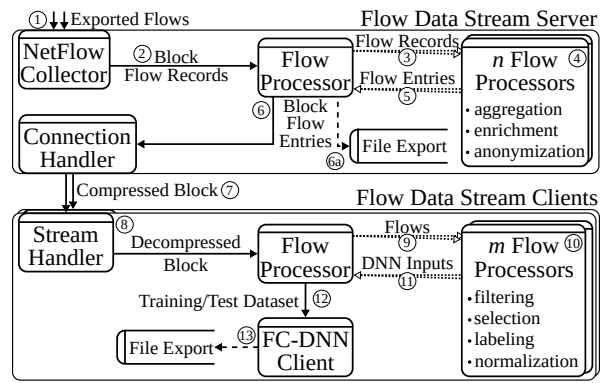

Figure 5.7: Data flow diagram of network flow data.

As illustrated in figure 5.7, the pre-processed compressed block is unpacked on the client side by the Stream Handler and passed on to a client Flow Processor ⑧. The client's Flow Processor divides the block into chunks ⑨, again, to perform processing steps by $m$ Flow Processors ⑩. The number of processes is adapted to the used hardware in order to minimize the pre-processing time. Filtering is the first optional step. It allows for the removal of individual flow entries from the dataset based on filter criteria. This is how a type of protocol, e.g., DNS, can be removed from the data. A second optional step is the feature selection. By means of this mechanism, individual features can be removed from the data. Likewise, the influence of the individual features on the prediction results can be investigated. Removing all features except the 5-tuple, for example, helps measuring the influence of enrichment.

As a third step, the data is categorized into classes. Two decisions are required for this step. First of all, label criteria need to be selected. The used flow data contains various meta-information (possible labels) relevant for routing: duration, number of bytes or packets and the bit rate. Second, class boundaries have to be specified. At the current processing step, flows are divided into classes based on predefined boundaries. How to determine boundaries is explained in more detail in section 5.3.1.

The last processing step of the Flow Processors is normalization and encoding. The goal of this step is to specify the format of each feature (see section 5.2.2.1). At the same time, a min-max normalization is applied. All of the performed steps can be parameterized individually. Therefore, different experiments can be performed simultaneously. Each experiment is represented by an individually parameterized Flow Data Stream Client.

As a result, the Flow Processor receives several chunks of data, which are merged into one block ⑪. The block is divided into training (90%) and test data (10%), while preserving the chronological order of the flows ⑫. Thus, future data needs to be predicted by the ML model. This data-driven time shift merely simulates the real time difference of the prediction of a new flow and the current state of the model. A data block can be used until a new block of data is available. As shown in figure 5.7, a DNN consisting of fully-connected artificial neurons (Fully-Connected (FC)-DNN) is used for the experiments ⑬. At this point, both, the evaluation results and the pre-processed data (optional) can be exported.

An advantage of this architecture is the full implementation of the parameterized part on the client side. This way, several clients can simultaneously connect to the server. Thus, many different experiments can be performed. Another benefit is that experiments can be repeated, assuming that the pre-processed dataset can be stored.

## 5.3   Traffic Analysis

Before describing experiments with flow data, the results of a data analysis are summarized. The entire dataset contains ≈480 million flow entries which are distributed over 6 800 blocks. The dataset represents one week of network flow data. Different analyses are performed on a single extracted block of 100 000 flow records, whereas blocks that are temporally close to each other produce similar results.

In order to use representative data, a block at 2:00 p.m. is selected. At that time, the peak of students on-campus (and within the network) is assumed. It has to be noted that the distribution within the collected blocks differs depending on the daily rhythm or the days of the week. The corresponding effects are illustrated by the experiments later (see section 5.4). To help ensure the validity of the processing steps and the analyses, network flows are manually generated, e.g., in form of downloads, and traced back.

### 5.3.1   Label-based Data Distribution

The first step is to analyze the data distribution in relation to the most interesting labels, e.g., bit rate, duration, number of packets and number of bytes. To be precise, a single block of 100 000 flows records is investigated. A histogram (25 bins, note the logarithmic scale) for each type of label is shown in figure 5.8. As depicted in figure 5.8a, most flows are very short while transferring very few bytes or consisting of few packets (see figures 5.8c and 5.8d). This result indicates a low bit rate for most flows (see figure 5.8b).

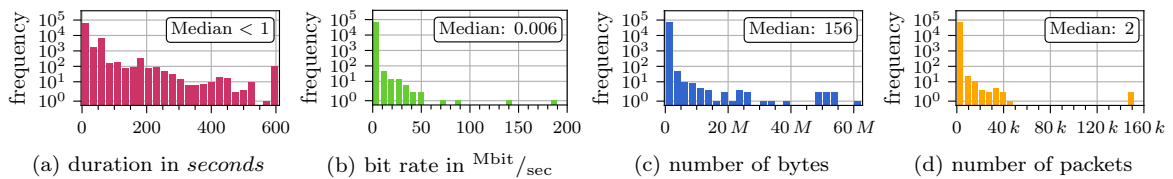

(a) duration in *seconds*   (b) bit rate in $^{\text{Mbit}}/_{\text{sec}}$   (c) number of bytes   (d) number of packets

Figure 5.8: Histograms of a block for all flow labels (log scale).

The histograms clearly present the uneven distribution based on the possible labels. The same result is implied by the median values given shown in figure 5.8. At this point, converting a regression into a classification problem reveals a challenge, namely the definition of the class boundaries. The choice of class boundaries depends on the given problem, in this case it is the routing decisions. This problem has been simplified in many related works (section 5.1.1) by making it a binary classification problem (distinction between "mice" and "elephant" flows). In the present work, however, the data is divided into three classes, even though the definition of specific boundaries remains a challenge. This is particularly due to the used ML model (here DNNs) and its difficulties with unbalanced data. Even if the chosen class boundaries are data-driven, it becomes obvious that a classification with several classes is possible.

The determination of class boundaries is realized as follows: First, the division into three classes is chosen in order to show that a fine-grained classification of flows is feasible. Second, the uniform class distribution, which is required by the used DNNs models, is increasingly difficult to achieve with a higher number of classes. In order to determine class borders, the complete dataset is exposed to an equal-depth frequency partitioning method (also known as equal frequency binning). The derived class boundaries allow for a maximum balance of the data distribution for the chosen label in the dataset. This procedure is, however, only applicable in an offline scenario where the data is available in advance. For online variants, a block-wise adjustment of the boundaries is a better option. This offline approach is, however, not problem-specific, but data-driven.

An example of using the bit rate in $^{bit}/_{sec}$ as a label is presented below. The class boundaries derived from the entire dataset are as follows: class 0 = [0, 4169[, class 1 = [4169, 12 288[, class 2 = [12 288, ∞]. Applying these bounds results in the data distribution within the classes as shown in figure 5.9. The black line illustrates the number of all flows after pre-processing within the Flow Data Stream Server (see section 5.2.1). The reduction of 100 000 to about 75 percent is mainly due to the aggregation (see section 5.2.1.2). In addition, flow records that are not completed within a block are removed. The three colored lines in figure 5.9 indicate the data distribution for the three classes (red, green and blue). It is approximately the same for ≈6 800 blocks. The slightly unequal distribution can be compensated more easily by a weighting method (e.g., by adjusting the learning rate) within the ML model. Alternatively, over- or under-sampling could be used as a balancing method.

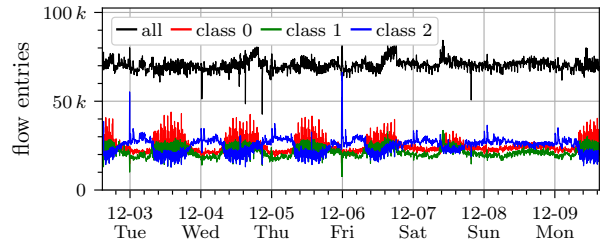

Figure 5.9: Class distribution for all blocks of the dataset.

## 5.3.2   Structural Data Patterns

Recognizing structures in unknown data is difficult. This is particularly true if their properties cannot be represented in a visual manner. In the present scenario, the 5-tuple is one of them. In order to visualize the structures within the data, t-distributed Stochastic Neighbor Embedding (t-SNE) (Maaten and Hinton 2008) is used. Based on the Euclidean distance, t-SNE reduces high-dimensional data to two-dimensional points. This procedure tries to preserve the neighborhood of the individual 2D data points. Therefore, the absolute position of individual points is meaningless.

The runtime behavior ($\mathcal{O}(n^2)$) of the standard method does not allow to reduce the dimension of all data points in an adequate time. The same is true for an entire block or online visualization. Therefore, an optimized tree-based approximation proposed by Maaten (2014) is used to process 10 000 flows. For a better understanding of the correlations for one and the same t-SNE output within the data, different properties are tagged. For the application of t-SNE, 2 000 iterations are performed at a perplexity of 150.

The tags in figure 5.10 refer to the used transport protocol. Flows belonging to UDP (●) are disproportionately frequent (85.5 %). Only 13.5 % of 100 000 flows are part of the TCP traffic (▲). The remaining flows take about 1.0 % (■) including, for example, ICMP. Figure 5.10 clearly illustrates the symmetric separations emerging between the used protocols.

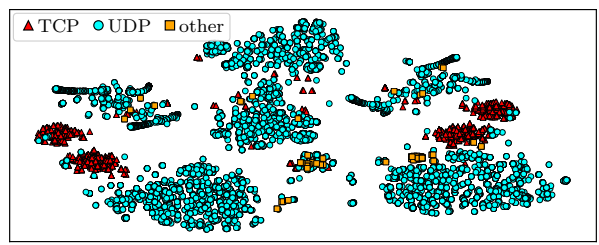

Figure 5.10: Transport protocol-based network flow tagging of the t-SNE results.

Another type of tagging that refers to the locality of a communication is presented in figure 5.11. The IP is used to distinguish between local and public addresses. In figure 5.11, the outward and return direction becomes visible. Only 3.4 % of the communication takes place exclusively between local systems (■). This is due to the network architecture (core-distribution-access model, see section 5.2.1.1), which also means that flows within buildings cannot be captured. That is the reason why the number of exclusively internal flows is lower.

The largest proportion of flows (36.6 % ✚ respectively 34.8 % ●) is constituted by communication with one public system (Internet traffic). Since the data center offers IT services with public addresses, the respective number of flows is considerably high with 25.2 % (▲). The direction of communication is clearly visible through the symmetry. This fact is strengthened by the highlighted WiFi (46.3 %) traffic.

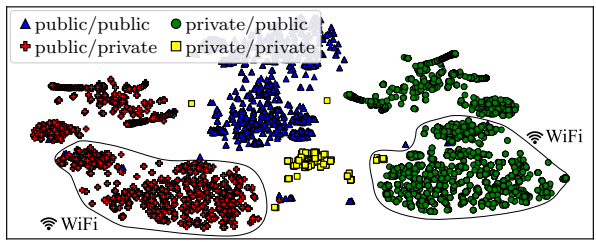

Figure 5.11: Locality-based network flow tagging of the t-SNE results.

Considering the application protocol, it is clear that the majority (79.8 %) of flows are assigned to DNS (●). 13.2 % are HTTP flows (◊) and the remaining 7.0 % are assigned to other application protocols (▲). The high proportion of DNS traffic is due to two facts. First, almost every type of network communication requires address resolution. A web page request, for example, does not only trigger a single name resolution, but any dynamically content that is loaded. The second reason is the structure of the DNS resolver arrangement. Resolving an address comprises the request of an internal DNS server which, if necessary, forwards the request to an external server. Both flows are captured, causing an increase in DNS traffic. The present analysis reinforces the statement concerning the uneven distribution of data presented in section 5.3.1. One of the reasons is that DNS requests comprise very few packets. This results in the highly unequal distribution in terms of duration and data volume.

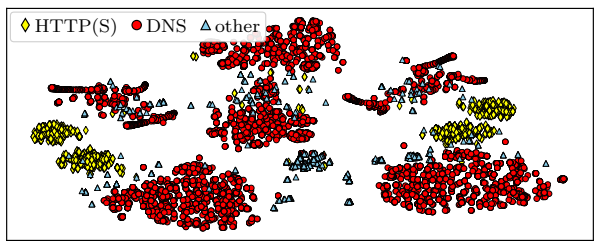

Figure 5.12: Application-based network flow tagging of the t-SNE results.

## 5.4   Flow Prediction Experiments

In this section, the conducted flow data experiments are presented. As described in section 5.4.1, a DNN architecture for the used model needs to be determined first. Since the DNN architecture has to be defined in advance, a parameter search is performed on a small portion of the collected data. Based on these results, further experiments related to different contexts within the network flow data are performed (stated in section 5.4.3). As a next step, experiments are conducted with the full dataset. It consists of one week of flow data.

### 5.4.1   Deep Neural Network Architectural Experiments

The research approach is based on DNNs, as applied in other related works (see section 5.1.1). Many properties of DNNs cause them to be interesting for flow prediction application scenarios. This includes the ability to process batches of data with a constant runtime $\mathcal{O}(1)$ for both, training and inference. This is in contrast to other models, such as Gaussian Process Regression (GPR). For the latter, a runtime behavior of $\mathcal{O}(n^3)$ is expected without further optimization (Poupart, Chen, et al. 2016). Even though approximations or heuristics may be able to improve the runtime behavior, additional hyper-parameters are introduced (Rasmussen and Williams 2006).

In general, the comparatively good performance of DNNs for higher dimensional data is crucial. Additionally, DNNs are suitable for the processing of infinitely long data streams. This is due to their ability to train/learn incrementally (see section 2.3.2). Other methods, such as vanilla Support Vector Machines (SVMs), are not applicable as they lack the incremental learning capability. Some properties of DNNs are, however, disadvantageous. These include, for example, the difficulties related to unbalanced data distributions in regression and classification problems. Although various techniques can mitigate the effects, the method remains vulnerable.

**Hyper-Parameter Grid Search**   Before a detailed investigation of the data can be executed, the architecture of the DNNs needs to be defined. For this purpose, a grid-search (see section 4.3) is performed on a subset of the collected data. The subset consists of 10 blocks (each about 75 000 flow records), which are pre-processed by the Flow Data Stream Pipeline. In addition to the learning rate $\epsilon$, the standard DNN hyper-parameters are varied. The latter include the number of layers $L$ and the contained artificial neurons $S$. More complex functions can be approximated, if DNNs are deeper and have more neurons. However, larger DNNs increase the probability of the data being learned by heart, which is known as overfitting effect (see section 2.2.2.5).

A commonly used technique for the avoidance of overfitting is the application of dropout (Hinton, Srivastava, et al. 2012b). Dropout specifies that input signals are randomly passed (or set to 0) to the next layer based on a probability. One probability is defined for the input layer $d_i$ and one for the hidden layers $d_h$ (a factor of 1.0 corresponds to dropout being disabled). The application of dropout is supposed to result in an improved generalizability of DNNs.

In order to address the unbalanced data problem, two class balancing methods $\mathcal{W}$ are applied. This includes the standard method of proportionally adjusting the learning rate for a class, as well as under-sampling. The latter leads to the reduction of the samples in all classes, so that all of them maintain an equal number. The class with the fewest samples serves as a basis.

The flow data are used with and without the enrichment (indicated by $\mathbb{F}$). The goal is to avoid that the enrichment influences the choice of a network architecture. Table 5.3 summarizes the available values for the hyper-parameters. Combining the hyper-parameters results in the performance of 5 400 experiments. The last step is the determination of the best DNN architecture.

Table 5.3: Overview of the varied hyper-parameters for the initial DNN architecture grid-search.

| Parameter | Variable | Values |
|---|---|---|
| Dropout (input, hidden) | $(d_i, d_h)$ | $\{(1.0, 1.0), (0.9, 0.6), (0.8, 0.5)\}$ |
| Layers | $L$ | $\{3, 4, 5\}$ |
| Neurons per layer | $S$ | $\{200, 400, 600, 800, 1\,000, 1\,500\}$ |
| Learning Rate | $\epsilon$ | $\{0.01, 0.001, 0.0001\}$ |
| Features | $\mathbb{F}$ | $\{\text{5-tuple, all}\}$ |
| Class balancing method | $\mathcal{W}$ | $\{0 \text{ (under-sampling)},$ $1 \text{ (class weighting)}\}$ |

In addition to the tuning of hyper-parameters, the DNN structure is defined as follows. For the experiments, fully-connected DNNs are used. In contrast to sparsely connected models, the complex pruning of DNNs can be omitted. Rectifier Linear Units (ReLUs) are used (see section 2.2.1.2) as an output function $f\varphi$. Weights are updated by the Adam optimizer (see section 2.2.2.4). Cross-Entropy (CE) loss with softmax is optimized (see section 2.2.2.1). In this application context, the examination of the bit rate only is used as an initial label. It is assumed that this value is the most interesting for potential applications. Both, the batch size $\mathcal{B} = 100$ and the number of training epochs $\mathcal{E} = 10$ are fixed for each training iteration and block. Performance in terms of accuracy is measured after every $50^{\text{th}}$ training iteration. For testing purposes, each block is divided into 90 % training and 10 % test data while adhering to the chronological order. Accordingly, a complete epoch is evaluated on the test data.

**Result of the Architectural Experiments**   The goal of the grid-search is to determine an DNN architecture with a reasonable performance. The respective architecture should be used to execute experiments on the entire dataset in a subsequent step. The collected accuracy values at the measurement points constitute the basis for the evaluation of the 5 400 experiments. The maximum measured accuracy within an experiment is used as a metric to determine the best architecture. For long-term experiments, this criterion is questionable, but sufficient to derive an adequate DNN architecture. The maximum measured accuracy is 86.7 %, if all features are included in the training process. Without the enrichment (using only the 5-tuple), the maximum is about 2 % lower. Table 5.4 shows the hyper-parameters used for the best experiments.

Table 5.4: Results of the architectural hyper-parameters experiments.

| Features $\mathbb{F}$ | Layers $\mathcal{L}$ | Sizes $\mathcal{S}$ | Learning Rate $\epsilon$ | Dropout Probability $(d_i, d_h)$ | Weighting $\mathcal{W}$ |
|---|---|---|---|---|---|
| all | 3 | 1 000 | 0.0001 | (1.0, 1.0) | 1 |
| 5-tuple | 5 | 1 000 | 0.001 | (0.9, 0.6) | 0 |

### 5.4.2   Concept Drift/Shift Experiments

Many problems do not need to be considered within a specified time interval or treated as CL problem. In general, an ML model's parameter configuration is derived from the training data. Then it is fine-tuned and deployed, e.g., to recognize faces in images. The question arises whether this approach can be applied to the prediction of network flow data. Accordingly, conducting the following experiment aims at detecting a changing data distribution (see figure 5.13) in this scenario.

In order to investigate the changing distribution, an ML model is trained on a single block for 20 epochs (blue line). Subsequently, the quality of the model is measured at regular intervals (each 50<sup>th</sup> block, red line). The model remains unchanged after the initial training.

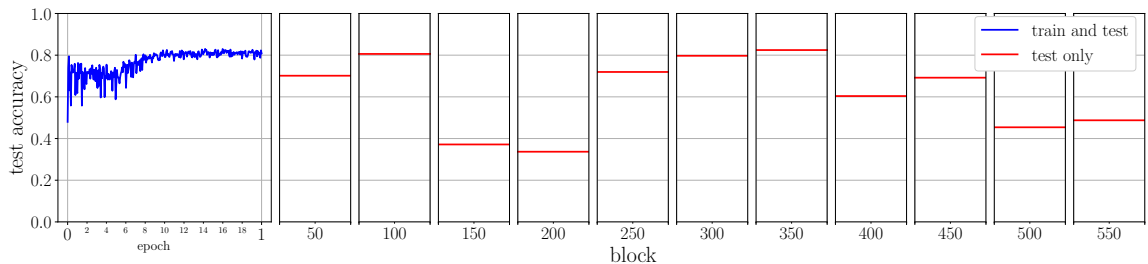

Figure 5.13: Concept drift/shift experiment.

The different test accuracies of blocks reveals that the data change over time (concept drift and/or shift, see section 2.3.1). It should be noted that a three class problem, as well as an equal distribution within the class, has a 33 % guessing chance. This limit is almost reached for block number 150 and 200, which indicated that the model does not fit the data anymore. For the subsequent blocks, e.g., block number 350, similar accuracies are achieved as during training. The following conclusion for further experiments can be drawn: The training should be conducted incrementally for each incoming block in order to counteract the changing data distribution.

### 5.4.3   Streaming Experiments

New challenges arise for a wider investigation of different parameters and contexts. Other useful communication contexts are enabled by the enrichment of the data, among other things. Considering them leads to 10 possible combinations. At the same time, 4 labels can be assigned for each context: Bytes, packages, duration and bit rate. The influence of the used features or feature groups is significant (6). To conclude, 240 experiments can be performed on the dataset with the resulting architecture described in 5.4.1. On a single machine, this amount would take 4.6 years of computation time.

Since each experiment requires exactly one week of computing time, the processing is performed in parallel. The used hardware components are specified in table 4.1. Different hardware specifications (number and clocking of CUDA cores), however, cause difficulties when comparing parallel online experiments. These differences would lead to more training iterations on faster hardware. Figure 5.14 represents the comparison between the number of training iterations/epochs. The slower Nvidia Quadro P620 is depicted in green and the faster Nvidia RTX 2080 in blue. The day-night rhythm also becomes visible with significantly fewer flows at night, and more training iterations per block. During daytime it is vice versa.

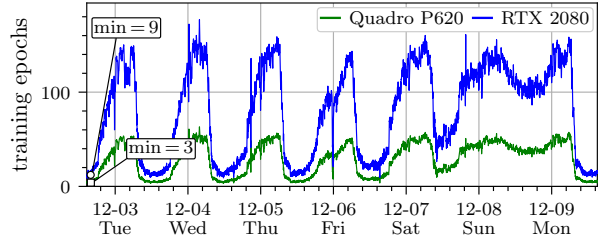

Figure 5.14: Training epochs comparison of slowest and fastest GPU.

The number of training iterations with the slowest available hardware is recorded first. It is used subsequently for all further experiments. This choice is supposed to rule out the advantages of the faster hardware components. The factor between the slowest and fastest hardware amounts to the constant number 2.75. Comparing the high number of training iterations to the lower number of iterations, a small impact is indicated. The difference between the faster and slower hardware causes a minor difference of 0.7 percent accuracy on average.

As shown in section 5.3.1, the best possible class boundaries are determined for each context prior to the experiments. For this purpose, the data must be available as an offline dataset. As a result of this pre-processing step, the boundaries in table 5.5 are derived for the assignment of class labels. Only the intermediate values are given, where the lower limit is 0 and the upper limited is $\infty$. These limits for all contexts and labels cause the distribution of data within the dataset to be less application-oriented (also see section 5.3.1). Nevertheless, the distribution within the classes is more even and therefore beneficial for the training of DNNs.

Table 5.5: Context-specific label boundaries for flow streaming experiments.

| Communication Context | Label Boundaries | | | | | | | | | | | |
|---|---|---|---|---|---|---|---|---|---|---|---|---|
| | KBytes | | | Packets | | | Duration ($s$) | | | Bit Rate ($Kbit/s$) | | |
| all contexts | 0.11 | 0.35 | | 1 | 3 | | 0.14 | 0.53 | | 4.1 | 12.2 | |
| only WiFi | 0.15 | 0.45 | | 2 | 3 | | 0.21 | 1.67 | | 4.2 | 10.8 | |
| exclude WiFi | 0.09 | 0.33 | | 1 | 2 | | 0.11 | 0.24 | | 4.1 | 12.9 | |
| exclude DNS | 0.51 | 2.94 | | 5 | 12 | | 0.18 | 2.38 | | 3.5 | 24.1 | |
| only TCP | 1.50 | 4.49 | | 9 | 15 | | 0.25 | 4.92 | | 6.3 | 45.8 | |
| only UDP | 0.09 | 0.17 | | 1 | 2 | | 0.12 | 0.57 | | 4.0 | 9.7 | |
| multi-packet flows | 0.28 | 1.41 | | 2 | 8 | | 0.21 | 1.39 | | 5.6 | 20.8 | |
| public/public | 0.09 | 0.28 | | 1 | 2 | | 0.11 | 0.23 | | 4.4 | 12.0 | |
| private/private | 0.09 | 0.44 | | 1 | 4 | | 0.10 | 0.28 | | 3.9 | 14.4 | |
| private/public | 0.01 | 0.50 | | 1 | 3 | | 0.17 | 0.92 | | 4.0 | 12.8 | |

In addition to considering the contexts, the impact of individual features is examined. In order to minimize the number of performed experiments, they are divided into groups. The various groups along with their number of features are represented in table 5.6. For the investigation of the experiments, the Adam optimizer is not used for the results presented in the following. Instead, a plain Stochastic Gradient Descent (SGD) is applied. During the first experiments, several side effects were observed, which is why the optimizer was changed. Figure 5.15 presents the measured accuracies of an entire experiment. Moreover, the learning rate is specified. It becomes clear that strong fluctuations occur during the experiment when Adam is used (blue line). Sometimes, after several hundreds of processed blocks, the training process stabilizes. This effect was observed in almost all experiments and is independent of time or respectively training iteration. However, the effect cannot be reproduced by repeating the experiment, as at least the first and last block changes. Moreover, the accuracy within individual blocks changes dramatically when using Adam. This is similar to the effect of a too high learning rate. The respective blocks, however, do not reveal any irregularity. An investigation regarding the internal variables of the Adam optimizer remains without new findings. The present work does not follow this thread any further.

Table 5.6: Feature groups for the flow data experiments.

| Feature Group | Features in the Data Stream | Inputs |
|---|---|---|
| all | each flow feature from table 5.2 | 247 |
| 5-tuple | IP address $\rightleftarrows$, port $\rightleftarrows$, protocol, timestamp | 109 |
| internal | network + prefix length $\rightleftarrows$, VLAN $\rightleftarrows$ | 98 |
| external | network + prefix length $\rightleftarrows$, ASN $\rightleftarrows$, country code $\rightleftarrows$, geo coordinates $\rightleftarrows$ | 112 |
| 5-tuple + internal | | 207 |
| 5-tuple + external | | 221 |

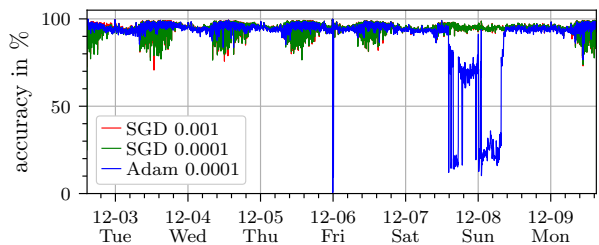

Figure 5.15: Visualization of the optimizer's side effect (plain SGD vs. Adam).

Table 5.7 (left) summarizes the results of the 240 one-week-lasting experiments. Combinations of communication context, used feature groups and the predicted label are distinguished. All collected measurement points are reduced per experiment to the maximum (white), the average (light gray) and the median (dark gray) of the accuracy. As a consequence, many details are lost as part of the information compression. The trend, however, is still recognizable. Accordingly, the challenging prediction of the bit rate and the duration is clearly visible. The same is true for the prediction of the number of transmitted packages and bytes, even though it is easier to predict. The right part of table 5.7 illustrates the proportions of the distribution within the collected flows. To be precise, the percentage of TCP, UDP and, in particular, DNS flows is presented. In addition, the median of the four potential labels is stated. The reason for appropriately selected class boundaries (see table 5.5) becomes obvious: Smaller median values illustrate an uneven distribution of target values.

In the following, eight individual experiments are selected. All of them represent the trend for the entire week (see section 5.4.3). Only TCP flows are selected as context for all experiments. Each sub-figure of section 5.4.3 illustrates two experiments by using all features (blue) and the 5-tuple as input features (orange). During the training on a single block, the accuracy is measured repeatedly after 50 training iterations.

In order to visualize the results, all test values measured on a block have to be combined. For this purpose, the maximum accuracy value is used for the representation of a block. The day and night rhythm is evident, while the performance improves at night. This is due to an increased variability of flows during the day, as more students are present on the campus. The same applies for the weekend. Enrichment further influences the result, which is clearly visible in section 5.4.3. All of the visualizations indicate the following ranking. Using only the 5-tuple (orange) hardly ever leads to better results than using all features (blue). The lower limit supports this conclusion, especially during working hours. The horizontal line in the figures, which represents the average of the measured values, indicates a relatively small improvement of about 2 %.

Table 5.7: Overview of the flow prediction experiment results (left) and flow volume metrics (right) for each context.

| Communication Context | | all | | | | 5-tuple | | | | internal | | | | external | | | | 5-tuple + internal | | | | 5-tuple + external | | | | Flow Volume Metrics (Median) | | | | | |
|---|---|---|---|---|---|---|---|---|---|---|---|---|---|---|---|---|---|---|---|---|---|---|---|---|---|---|---|---|---|---|---|
| | | Bytes | Packets | Duration | Bit Rate | Bytes | Packets | Duration | Bit Rate | Byte | Packets | Duration | Bit Rate | Bytes | Packets | Duration | Bit Rate | Bytes | Packets | Duration | Bit Rate | Bytes | Packets | Duration | Bit Rate | | Proportion (%) | KBytes | Packets | Duration (s) | Bit Rate (KBit/s) |
| all contexts | max | | 100 | 90 | 93 | 100 | 100 | 90 | 93 | 98 | 99 | 99 | 98 | 99 | | 95 | 91 | 100 | 100 | 90 | 94 | 100 | 100 | 90 | 92 | TCP | 21 | 2.6 | 12 | 1.0 | 14.9 |
| | mean | also 96 | 98 | 67 | 74 | 95 | 97 | 67 | 73 | 85 | 88 | 61 | 67 | 85 | 88 | 61 | 67 | 95 | 97 | 67 | 73 | 95 | 97 | 67 | 74 | UDP | 79 | 0.1 | 1 | 0.2 | 6.1 |
| | median | 97 | 98 | 68 | 75 | 95 | 98 | 68 | 73 | 85 | 88 | 61 | 67 | 85 | 88 | 61 | 67 | 95 | 98 | 68 | 74 | 95 | 98 | 68 | 74 | DNS | 71 | 0.1 | 1 | 0.2 | 6.0 |
| only WiFi | max | 100 | 100 | 96 | 95 | 100 | 100 | 94 | 95 | 98 | 100 | 93 | 95 | 98 | 100 | 97 | 95 | 100 | 100 | 96 | 95 | 100 | 100 | 94 | 95 | TCP | 24 | 3.3 | 13 | 3.3 | 9.0 |
| | mean | 96 | 99 | 67 | 69 | 94 | 98 | 64 | 66 | 76 | 81 | 58 | 62 | 94 | 98 | 65 | 68 | 94 | 98 | 66 | 68 | 94 | 98 | 66 | 68 | UDP | 75 | 0.2 | 2 | 0.2 | 6.2 |
| | median | 97 | 99 | 69 | 69 | 95 | 98 | 66 | 66 | 82 | 86 | 60 | 62 | 82 | 86 | 60 | 62 | 95 | 98 | 67 | 68 | 95 | 98 | 67 | 68 | DNS | 73 | 0.2 | 2 | 0.2 | 5.9 |
| exclude WiFi | max | 100 | 100 | 94 | 96 | 100 | 100 | 93 | 95 | 98 | 99 | 88 | 90 | 98 | 99 | 88 | 90 | 100 | 100 | 93 | 96 | 100 | 100 | 93 | 96 | TCP | 19 | 2.2 | 11 | 0.8 | 17.7 |
| | mean | 97 | 98 | 71 | 77 | 96 | 98 | 71 | 76 | 86 | 88 | 65 | 68 | 86 | 88 | 65 | 69 | 96 | 98 | 71 | 76 | 96 | 98 | 71 | 76 | UDP | 81 | 0.1 | 1 | 0.1 | 6.1 |
| | median | 98 | 99 | 72 | 77 | 97 | 99 | 72 | 76 | 86 | 89 | 66 | 69 | 87 | 89 | 66 | 69 | 97 | 99 | 72 | 76 | 97 | 99 | 72 | 76 | DNS | 71 | 0.1 | 1 | 0.1 | 6.0 |
| exclude DNS | max | 96 | 95 | 91 | 90 | 95 | 95 | 90 | 89 | 93 | 92 | 88 | 88 | 93 | 92 | 88 | 88 | 95 | 95 | 91 | 90 | 95 | 95 | 90 | 90 | TCP | 70 | 2.6 | 12 | 1.0 | 14.9 |
| | mean | 85 | 84 | 72 | 72 | 84 | 82 | 71 | 71 | 74 | 71 | 64 | 65 | 74 | 71 | 65 | 65 | 84 | 82 | 71 | 71 | 84 | 82 | 72 | 71 | UDP | 27 | 0.1 | 1 | 0.1 | 4.9 |
| | median | 85 | 84 | 73 | 73 | 83 | 81 | 72 | 71 | 74 | 70 | 64 | 65 | 74 | 71 | 65 | 65 | 83 | 81 | 72 | 71 | 83 | 82 | 72 | 72 | DNS | 0 | 0.0 | 0 | 0.0 | 0.0 |
| only TCP | max | 95 | 93 | 87 | 89 | 94 | 92 | 87 | 89 | 92 | 89 | 83 | 89 | 93 | 90 | 83 | 89 | 94 | 92 | 87 | 90 | 94 | 92 | 87 | 89 | TCP | 100 | 2.6 | 12 | 1.0 | 14.9 |
| | mean | 82 | 78 | 76 | 72 | 80 | 76 | 74 | 71 | 70 | 65 | 67 | 65 | 71 | 65 | 68 | 65 | 80 | 76 | 75 | 71 | 80 | 76 | 75 | 71 | UDP | 0 | 0.0 | 0 | 0.0 | 0.0 |
| | median | 80 | 77 | 76 | 72 | 79 | 75 | 74 | 71 | 70 | 64 | 67 | 64 | 70 | 64 | 67 | 65 | 79 | 75 | 75 | 71 | 79 | 75 | 75 | 71 | DNS | 0 | 0.0 | 0 | 0.0 | 0.0 |
| only UDP | max | 97 | 100 | 100 | 99 | 97 | 100 | 100 | 99 | 96 | 100 | 100 | 97 | 97 | 100 | 98 | 97 | 97 | 100 | 100 | 100 | 97 | 100 | 100 | 100 | TCP | 0 | 0.0 | 0 | 0.0 | 0.0 |
| | mean | 75 | 83 | 56 | 54 | 75 | 83 | 56 | 55 | 71 | 80 | 54 | 53 | 71 | 80 | 54 | 53 | 75 | 83 | 56 | 54 | 75 | 83 | 56 | 54 | UDP | 100 | 0.1 | 1 | 0.2 | 6.1 |
| | median | 74 | 82 | 56 | 53 | 74 | 83 | 56 | 53 | 71 | 80 | 54 | 51 | 71 | 80 | 54 | 51 | 74 | 82 | 56 | 53 | 74 | 82 | 56 | 53 | DNS | 100 | 0.1 | 1 | 0.2 | 6.2 |
| multi-packet-flows | max | 98 | 97 | 93 | 93 | 97 | 96 | 92 | 94 | 95 | 93 | 90 | 91 | 94 | 94 | 91 | 91 | 97 | 96 | 92 | 94 | 97 | 97 | 93 | 94 | TCP | 40 | 2.6 | 12 | 1.1 | 16.5 |
| | mean | 90 | 91 | 76 | 76 | 88 | 89 | 75 | 74 | 78 | 78 | 67 | 67 | 78 | 78 | 68 | 68 | 88 | 89 | 76 | 75 | 88 | 89 | 76 | 75 | UDP | 60 | 0.2 | 2 | 0.2 | 6.7 |
| | median | 89 | 91 | 77 | 76 | 88 | 89 | 76 | 74 | 78 | 77 | 68 | 67 | 78 | 78 | 68 | 67 | 88 | 89 | 76 | 75 | 88 | 89 | 76 | 75 | DNS | 54 | 0.2 | 2 | 0.3 | 6.0 |
| public/public | max | 100 | 100 | 99 | 98 | 100 | 100 | 99 | 98 | 98 | 100 | 96 | 99 | 99 | 100 | 97 | 95 | 99 | 100 | 99 | 98 | 100 | 100 | 99 | 98 | TCP | 11 | 2.2 | 11 | 0.9 | 18.0 |
| | mean | 86 | 94 | 72 | 61 | 86 | 94 | 72 | 61 | 87 | 94 | 73 | 62 | 96 | 99 | 78 | 73 | 96 | 99 | 78 | 73 | 96 | 99 | 78 | 73 | UDP | 88 | 0.1 | 1 | 0.1 | 6.2 |
| | median | 88 | 100 | 79 | 75 | 97 | 100 | 79 | 73 | 88 | 95 | 72 | 61 | 89 | 95 | 73 | 62 | 97 | 100 | 79 | 74 | 97 | 100 | 79 | 74 | DNS | 87 | 0.1 | 1 | 0.1 | 6.2 |
| private/private | max | 100 | 100 | 97 | 98 | 100 | 100 | 97 | 97 | 99 | 99 | 97 | 97 | 99 | 99 | 97 | 97 | 100 | 100 | 99 | 98 | 100 | 100 | 99 | 98 | TCP | 25 | 1.2 | 7 | 0.2 | 29.0 |
| | mean | 93 | 96 | 63 | 76 | 92 | 95 | 62 | 76 | 78 | 80 | 56 | 68 | 78 | 80 | 56 | 68 | 93 | 96 | 63 | 76 | 93 | 96 | 63 | 76 | UDP | 71 | 0.1 | 1 | 0.1 | 4.9 |
| | median | 94 | 97 | 63 | 77 | 93 | 96 | 62 | 76 | 79 | 80 | 55 | 68 | 79 | 80 | 55 | 68 | 93 | 96 | 63 | 76 | 93 | 96 | 63 | 76 | DNS | 7 | 0.7 | 6 | 5.7 | 2.9 |
| private/public | max | 100 | 100 | 94 | 94 | 100 | 100 | 91 | 93 | 96 | 99 | 90 | 91 | 96 | 99 | 88 | 91 | 100 | 100 | 90 | 97 | 100 | 100 | 90 | 98 | TCP | 25 | 3.1 | 13 | 2.5 | 10.1 |
| | mean | 95 | 98 | 68 | 71 | 94 | 97 | 67 | 70 | 82 | 86 | 59 | 64 | 82 | 87 | 59 | 65 | 94 | 97 | 68 | 71 | 94 | 97 | 68 | 71 | UDP | 24 | 0.3 | 2 | 0.2 | 6.1 |
| | median | 96 | 99 | 70 | 72 | 95 | 98 | 69 | 70 | 83 | 87 | 60 | 64 | 83 | 87 | 60 | 64 | 95 | 98 | 69 | 71 | 95 | 98 | 69 | 71 | DNS | 71 | 0.2 | 2 | 0.2 | 6.0 |

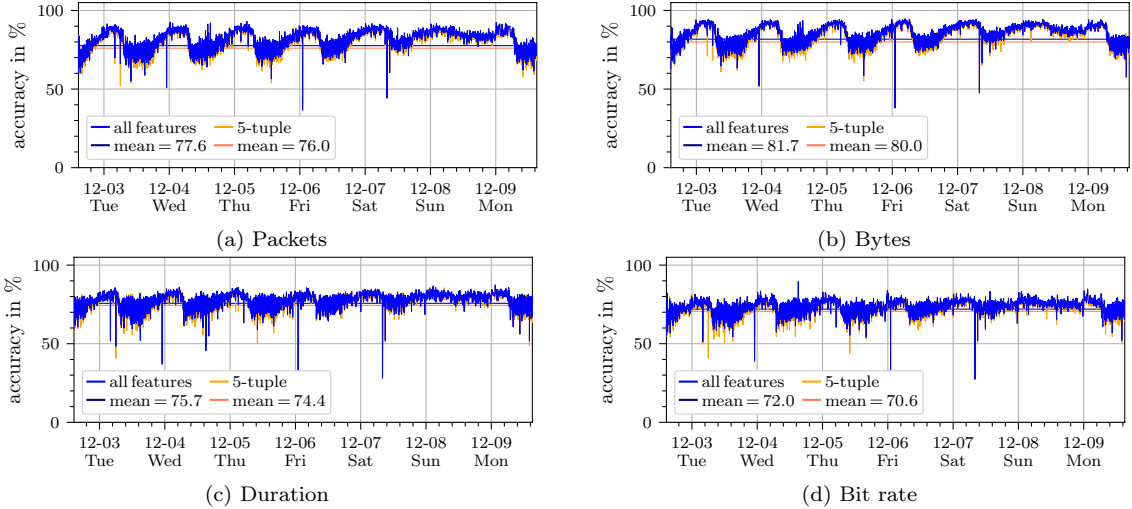

(a) Packets      (b) Bytes

(c) Duration      (d) Bit rate

Figure 5.16: Trend for eight selected experiments (only TCP flows and different labels).

## 5.4.4 Discussion

This section discusses the real-world application scenario. For this purpose, the previously outlined scenario is briefly summarized. An evaluation of the presented scenario in the context of CL will be discussed in the next section (see section 5.5).

In the current chapter, a real-world application scenario was presented. The scenario involves ML techniques, i.e., DNNs, in order to predict meta-data of network flows. The corresponding meta-data comprise information of terminated network communications, e.g., source, destination, duration, or transmitted bytes/packets. However, it is a fundamental problem of network communication that these meta-data are unknown when a communication starts. Thus, meta-information cannot be used in applications, e.g., for determining an optimal path through the network (routing). The use of ML techniques should enable the prediction of a flows' meta-data.

A supervised ML scenario is derived from this problem definition. Studying the real-world scenario involves a large-scale computer network. Thus, realistic data and requirements can be considered. In order to predict flow meta-information, an ML model needs to be derived from representative data. For this purpose, a step by step approach is presented and evaluated. Moreover, the collection of meta-data, the pre-processing and training of an ML is described. Data analyses along with the empirical results show that predicting flows' meta-data is a viable option with DNNs.

**Generalizability of the Results**  The first aspect of the discussion concerns the generalizability of the results. The implication is that a prediction of flow meta-data is possible. An investigation of the proposed approach in other computer networks could address this aspect. But without the possibility to perform a comparable investigation, generalization is very difficult.

Unfortunately, real-world data of (productive) computer networks are often inaccessible. This is due to the sensitivity of the data and the respective legal issues. At the same time, continuous service, without interference, must be ensured. Thus, access to comparable network data is challenging. This is also the reason why synthetically generated or outdated data are often used in the literature (see section 5.1.1). Statements based on generated data, however, can be generalized even less. This is due to the fact that ML models can approximate very complex functions. However, generating functions can never reproduce real-world conditions on a large scale.

The underlying network of the exemplary scenario seems to be sufficiently complex to prevent a perfect classification rate. This trend is particularly evident in the day and night rhythm. During the day, prediction becomes much more difficult, as more dynamics are created by staff members and students. Therefore, the approach can be considered as a valid procedure, even though the exact performance depends on the particular scenario.

**Quality of Results**  The obtained classification accuracy is the second aspect up for discussion. In general, an accuracy of 33 percent can be assumed as a baseline/chance for an evenly distributed three-class problem. Therefore, a classification rate above the guessed value is an improvement. Of course, the extraction of knowledge always depends on the quality of data itself. This real-world scenario is particularly interesting, because the labels for a supervised learning scenario are automatically provided. Nevertheless, the obtained data is influenced by the network structure and the exporting devices (hardware and software).

However, the question whether the achieved accuracy is sufficient for real applications remains unanswered. In order to make a valid statement, further aspects need to be considered. First, an application needs to be developed, which is based on the predictions. Second, measuring the impact of the application has to be possible. Achieving these criteria is very difficult in productive network environments. A first attempt to measure the influence of the prediction accuracy is presented by Hardegen and Rieger (2020).

On average, the empirical experiments result in accuracies above 60 percent. However, this number strongly depends on the chosen communication context. The temporal offset of training and test data is assumed to be a representative, realistic offset. Nevertheless, it can be inferred that a DNN is able to extract knowledge from the data. The possibility to learn knowledge from the data is supported by the recognized patterns as part of the conducted data analysis. Therefore, it is assumed that the prediction of network flows' meta-data is feasible with DNNs.

**Applicability of the Predictions**  The definition of class boundaries is yet another aspect that requires a discussion. As shown in section 5.3.1, network flow data is very unevenly distributed with respect to the labels. However, this unequal distribution can become problematic depending on the ML model. Accordingly, DNNs are affected.

To counteract this challenge, the available labels are used for a data-driven class division. The derived value ranges of the individual classes are, however, questionable. This in turn affects the areas of application. As a consequence, the equal-depth frequency partitioning method is merely used to demonstrate the feasibility.

Another issue requiring further research concerns the application of a replay buffer. This buffer could include hold-out samples that are continuously updated and used for under-represented classes.

Thus, application-oriented class boarders could be implemented without a data-driven analysis.

## 5.5   Real-World Continual Learning Requirements

Real-world requirements for CL scenarios can be manifold. In this section, requirements derived from the exemplary scenario are summarized. A collection of requirements will serve as an answer to the first research question, precisely *RQ 1.1*. In the course of this work, the identified requirements are used to develop an evaluation protocol. The protocol is going to implement the presented application-oriented CL requirements. At the same time, the dynamics and constraints of the scenario reveal that it is very difficult to implement a corresponding evaluation.

The presented real-world scenario focuses on the prediction of meta-data in the context of computer network communications by using DNNs. Respective meta-data of terminated network connections are collected from a productive campus network. This meta-data is referred to as network flows or in short flows. Flow data comprise both, the 5-tuple (source/destination IP address, source/destination port and the transmission protocol), and additional meta-information such as the duration or number of transmitted bytes. Thus, a potentially infinite, long stream of flow data needs to be processed. The goal is to predict certain meta-data that may be of interest for applications, for example, intelligent routing. In particular, the prediction concerns information that is not available at the beginning of a communication, e.g., its duration.

Several one-week lasting experiments validate the capability of DNNs to predict flow data. At the same time, the results show that one static ML model cannot serve as a solution for the very problem. On the contrary, it becomes obvious that the exemplary scenario needs to be classified in the area of CL. This is due to the constantly changing and dynamic context of computer networks. These changes include the influence of human behavior as well as changing applications. Based on this real-world scenario, the following constraints and requirements can be derived.

**Constraints:**   The presented constraints specify challenges implied by the exemplary scenario. The individual constraints illustrate why an evaluation within this real-world scenario is challenging. Paradoxically, an evaluation of CL models' capabilities is hard to realize in a real-world CL scenario. Based on the exemplary scenario and the conducted experiments, the following constraints are identified.

**Untraceable changing data distribution:** The dynamic nature of real-world CL scenarios particularly restrict the traceability of changing data distributions. In this exemplary scenario, data is collected in a campus network with many different components and subscribers. The unknown, changing distributions of the data result in unexpected behavior, which is difficult or impossible to estimate. Thus, the expected CL performance of a ML method cannot be estimated either. These dynamics are observable in basic experiments, as shown in section 5.4.2. The lack of traceability is an essential factor for the evaluation and comparison of ML models in the context of a CL scenario.

**Comparability of results:** Another constraint is the transfer of the present results and thus the comparability to other CL scenarios. In general, an offline dataset is created and repeatedly evaluated for the purpose of research. Accordingly, it is difficult to define a CL performance metric in a real-world scenario and to generalize it.

**Impact of prediction results:** A particular challenge in the exemplary scenario is the dependency of flows and their application. If flow prediction is used, e.g., to make intelligent routing decisions, optimizations influence future flows and therefore the subsequent training of the ML model. This results in a consecutive concatenation of effects, as the following example illustrates. Suppose that an emergent communication is correctly predicted if no optimized routing is applied. Based on the routing optimization, the resulting behavior of the communication is influenced. This could prove the approximately correct prediction to be wrong. At the same time, the exported flow with the corresponding meta-data would be used for training.

**Validation in applications:** Furthermore, the influence of an applied flow prediction should be measurable within a productive system. Intelligent routing will, again, serve as application. A challenge regards the impossible generation of a comparative quality value. The reason for this is that the data generating function cannot be reused. Notably, the generating function comprises all clients within the network including the interactions with applications. However, without an exact comparison value, only derived quality values, such as the average latency within the network can be used. Alternatively, a reduced synthetic scenario (including a controllable generating function) could be used to measure the impact of a prediction based application.

**Meaning of the classes boundaries:** The ground-truth of the data is yet another challenge (see section 6.1.1). Even if the exported flow information seems correct, details are lost due to the conversion of a regression into a classification problem. The reason for this transformation is the non-uniform distribution of the data in conjunction with the ML model. Even if this transformation converts the data into a roughly uniform distribution, its inherent meaning is questionable. At the same time, data generated by sensors or measurements are always dependent on environmental influences. In this scenario, the quality of the exported data is influenced by the resolution of hardware timers. The timer resolution of 50 ms, for example, may not be sufficiently precise.

**Interpretability of the data:** A fundamental difficulty is the interpretability of data. As part of the exemplary scenario, network flows described by 5-tuples are used. In general, it is impossible to assess the assignment of a flow to a class. The following is an exemplary 5-tuple: Source 23.53.173.33, destination 172.18.8.226, source port 443, destination port 43236, protocol 6 (TCP). Even though the label is given after the flows' termination, it can depend on various factors, e.g., the components on the path or the application. Traceability, verification and interpretation by humans are therefore questionable. This is in contrast to image data which is intended to represent a scene or a particular object, such as an image of a handwritten digit. This particularly limits the evaluability of generated data. Thus, the extent of a generated sample's validity can no longer be determined by "looking" at a sample. Accordingly, other mechanisms have to be used (e.g., see Lesort, Stoian, et al. 2019). As a consequence, the complexity of a respective evaluation increases.

**Importance of the past:** Another important constraint concerns the essential or decisive nature of knowledge. It is difficult to determine the corresponding extent. An answer or estimation cannot be derived from the data or from the classification accuracy. Basically, trends can repeat, whereby retroactive knowledge should not be lost. At the same time, however, this can be the opposite for certain knowledge. Old knowledge needs to be overwritten, as it is no longer valid. This is the case, for example, with applications whose communication pattern is changed due to an update.

**Requirements:** The CL requirements presented below are derived from the exemplary CL scenario. ML models should meet the respective requirements in order to be used in CL scenarios. At the same time, the requirements should be included in an application-oriented evaluation protocol.

**Constant processing time:** Constant processing time is a basic requirement for streaming scenarios. In order to incrementally train on an infinite stream of data the computational complexity has to be constant for each training step. This is especially true for the addition of new knowledge. The update complexity must not depend on the number of already processed samples or tasks. Assuming that each block in the exemplary scenario is considered as a task, no data stream can be processed without a constant update complexity.

**Constant memory footprint:** The memory consumption is also limited by a requirement resulting from the streaming concept. Memory consumption must be constant, otherwise an application in a streaming scenario is not feasible. If all data has to be cached for reprocessing, in theory, an infinite amount of memory has to be available. The preprocessed and uncompressed flow data, for example, correspond to a volume of about 441.67 GB. This amount was generated within one week of data collection. Even though only a subset of the data is stored, new challenges arise, e.g., which samples to save and for how long. Furthermore, the presented application scenario is only a "modest" computer network with regard to its scope. In other applications, such as image or video processing contexts, the data volumes can be much larger. A single HD video stream, for example, generates larger amounts of data, which is why codecs are used for compression.

**No foresight into future data:** In online real-world scenarios, looking at future data is impossible. Thus, an adjustment of different hyper-parameters based on future data is generally not feasible. This requirement in turn causes limitations, especially for model selection, i.e., the determination of a decent DNN architecture. In addition, other crucial hyper-parameters such as the learning rate cannot be adjusted in advance.

**Limited access to past data:** Random access to past data is also a limitation in terms of storage requirements. It mainly concerns the subsequent adjustment of hyper- or model parameters based on past data. While a certain number of samples can be cached, this is only possible if the available memory has a constant size. Thus, as soon as new data is available, memorized samples must be overwritten or individually replaced. Challenges arise related to the extent and specification of the retained data. The choice of samples is particularly challenging in streaming scenarios.

**Changeable data distribution:** According to the CL paradigm, it is crucial that a change in the data-generating distribution is assumed (e.g., section 2.3.1 or mixed forms). As a result, the ML model has be able to handle different types of drifts and shifts. The model should also be able to detect changes on its own and react accordingly. However, measurability of changes is not always easy to implement. One the one hand, changes may be insignificant (outliers), not directly detectable, or occur slowly, e.g., over weeks or even months. On the other hand, changes in the distributions can occur in very short time intervals. The results of the investigation of a short period clearly proves that (see section 5.4.2). In the presented scenario, the special task is to recognize these changes online and to react correctly.

## 5.6    Conclusion

The prediction of network flows or its meta-data is an adequate application scenario, which can be classified as *continual learning* task. The data situation is interesting, because the target values (labels) are already specified during the data collection. Thus, the exemplary scenario is predestined for supervised ML. In addition, the application of flow prediction can be used for many optimizations in the area of computer networks, e.g., optimized routing.

One drawback of the data is its uneven distribution, as shown by the data analysis in section 5.3. This is particularly problematic for DNNs. To counteract this, a transformation into a classification task is indispensable. As a consequence, the ground-truth of the data is lost (section 6.1.1). Nevertheless, the performed experiments in section 5.4 show that a predication is possible with DNNs. Although it is difficult to make a concrete statement regarding quality, the ML model seems to be able to infer knowledge from the data. Further comparative studies could attempt its measurement or develop an application scenario with comparable results.

All in all, requirements for a CL application can be derived from the presented exemplary real-world scenario. However, the chosen scenario does not cover all possible requirements that may arise in practice. In other scenarios, not all data samples may be labeled at the time of acquisition. This additional criterion could lead to further requirements as known from few-shot learning. Basically, the derived real-world requirements should be implemented in a CL investigation protocol.

Another finding is that it is difficult to carry out investigations within real-world scenarios. This is especially due to the emerging constraints. Indeterminable dynamics of the data generating function and incomprehensible changing data distributions contribute to the challenges. For these reasons, the exemplary real-world scenario is unsuitable for the investigation of ML models in the CL context.

# 6.  Application-Oriented Continual Learning Protocol

## Chapter Contents

In this chapter, an evaluation protocol is presented, which will be used to examine a *machine learning* (ML) model in the context of *continual learning* (CL). A consistent evaluation strategy is essential to establish comparability between different models (Pfülb, Gepperth, et al. 2018; Pfülb and Gepperth 2019). This includes various real-world requirements (see chapter 5). Taking these into account illustrates the applicability *catastrophic forgetting* (CF) avoidance models in CL scenarios.

The basic problem of related works is that every individual study uses a similar but often non-transparent evaluation scheme. This may allow comparability within the study itself, but prevents direct comparisons to other studies. Not only the metrics are concerned, but also the scenario for training and evaluation. For this reason, similar models that fit into a specific scheme are often used. Furthermore, the respective evaluation may not be designed for real-world requirements. Since many specialized models are mostly intended as proof-of-concept, they are only tested in fictitious scenarios. Oracles are used, for example, which support the model with additional information for training and testing. The problem is that this information is available in very few real-world applications. This leads to a shift of the problem towards the acquisition and provision of information. Thus, direct comparisons across different studies are challenging.

Another issue of related studies is that detailed evaluation procedure is not always complete or sufficiently explicit. This may be due to page limitations of the publication. Alternatively, the programming code can serve as basis to describe the evaluation scheme. However, this requires an intensive study of complex code. Moreover, there is no guarantee that the evaluation part of the code is available. To address this very problem, a detailed evaluation scheme is developed and presented in this work. It allows for a more realistic assessment regarding the CF effect and thus the CL characteristics.

**Contributions**  This chapter is supposed to answer research question RQ 1, precisely RQ 1.2 and RQ 1.3 (see section 4.1). Accordingly, an evaluation protocol, which measures the CF effect, is developed as the main contribution of this chapter. The protocol reflects requirements occurring in real-world CL applications. The first component for the development of an investigation protocol is represented by the requirements (RQ 1.1) which are derived from the real-world scenario presented in chapter 5. Second, the occurrence of the CF effect is illustrated. A measurement of the CF effect serves as an answer to RQ 1.2. The third sub-question (RQ 1.3) concerns the evaluation schemes issued in related work (see chapter 3). These schemes are included as an answer to research question RQ 1.3. To conclude, the contribution of this chapter is an evaluation protocol in the form of a flexible framework that can be used to evaluate different models in the CL context. Thus, other and new CF avoidance models can easily be added, evaluated and compared.

**Structure**  The first section of this chapter describes the datasets (mainly image datasets) used as a basis for evaluation (see section 6.1). Image datasets have a high number of input features. However, the question arises from which point onward a dataset is considered high-dimensional. A color image with the resolution $100 \times 100$ pixels (=3000 features) is nowadays assumed to be low dimensional in the context of image/video processing. Vijayakumar, D'Souza, et al. (2005), for example, designate 90 features as "high-dimensional spaces".

At the same time, a high number of dimensions may also represent information via very few features, e.g., in one pixel of an image. In order to address this issue, commonly available datasets that are applied in other studies are used. Furthermore, the chosen size of the datasets is moderate so that they can be processed by commercially available hardware. Another advantage of image datasets is related to a human's capability to interpret them. Thus, the sample itself can be validated by a human. This method is especially interesting for models that generate data samples.

The second part of this chapter describes the underlying scheme for deriving knowledge from the data. Datasets are divided into different chunks and provided to the model step by step. These predefined tasks are referred to as Sequential Learning Task (SLT). They describe, among other things, the level of difficulty of a SLT. In section 6.2, the SLTs are described in detail, and the used SLTs are summarized.

The third and last part of the present chapter refers to the training and evaluation scheme (see section 6.3). The scheme focuses on the aspect of CL while meeting the application-oriented requirements and constraints. The latter include the selection of hyper-parameters, (re-)training of models and the measurement of quality.

## 6.1  Used Datasets

In order to measure the impact of catastrophic forgetting during continual learning scenarios, publicly available datasets are used. Thus, all results can be reproduced and other models can be evaluated. Each dataset $\mathbb{X}$ consists of a collection of $n$ individual samples. Each sample $\boldsymbol{x}_n$, $n \in \{1, \ldots, N\}$ is described by an $m$-dimensional vector. Multidimensional data, such as images, are transformed into a one-dimensional form by a flattening operation. A color image with a width of 100 pixels and a height of 100 pixels (and 3 color channels) is, for example, transformed into a vector $\boldsymbol{x}$ with $m = 30000$ entries.

All of the used datasets belong to the category of classification problems, with only one single correct class assigned to each sample (section 2.2). For each sample, the target value is an element of $\{0, \ldots, C-1\}$, while $C$ is the number of different classes. The target value $y_n$, or label, is used in the one-hot format (see equation 2.1). If the original dataset is not split into training and test data, it will be split randomly into 90% training and 10% test data. By shuffling the data, an approximately even distribution within the classes is ensured for each training batch. Normalization is yet another pre-processing step that is performed on the data (not labels). Since image data and their pixels are often represented by 8 bits, min-max normalization is used. Input values are converted into the range $[-1, +1]$ (or $[0, +1]$). This range of values is beneficial because the highest possible number of fine-granular numbers can be represented. This is due to the representation of floating point numbers (Bush 2014).

**MNIST**  The Modified National Institute of Standards and Technology (MNIST) dataset is probably the most well-known dataset (see LeCun, Bottou, et al. 1998). It is often used for teaching machine learning, as it is integrated into common software frameworks. It consists of grayscale images with digits from 0 to 9 in a resolution of 28 by 28 pixels. The standard dataset consists of 60 000 training and 10 000 test images. MNIST is a modified version of the NIST dataset (Grother 1995).

In the present work, two sub-datasets are merged. They comprise the handwritten digits of employees of an American company and high school students. Details of the MNIST dataset are presented in table 6.1.

Table 6.1: Detailed information of the MNIST dataset.

| resolution | training samples | test samples | training balance | test balance | random **samples** (from classes) |
|---|---|---|---|---|---|
|  |  |  |  |  | 0 1 2 3 4 5 6 7 8 9 |
| 28×28×1 | 60 000 | 10 000 | 2.1 | 2.4 | 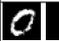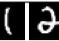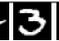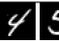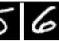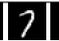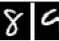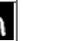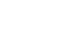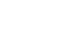 |

Even at the very first examination of the dataset, MNIST does not seem to be a challenge. The lowest error rate reported by LeCun, Bottou, et al. (1998) is 0.8% when using Support Vector Machine (SVM). Several attempts have been made to develop improved models or to find model configurations with a lower error rate. Currently, the classification quality is in the range of the last half percent to 100%. As a result, Chollet (2017) claims that MNIST should be omitted and more sophisticated dataset should be used for research:

> "*Many good ideas will not work well on MNIST (e.g. batch norm). Inversely many bad ideas may work on MNIST and no transfer to real CV.*" Chollet 2017

Looking at the respective publications shows that such statements do not necessarily lead to a lower use of the MNIST dataset. Fortunately, more and more datasets are created and used as a point of reference.

**EMNIST**    Cohen, Afshar, et al. (2017) present the Extended MNIST (EMNIST) dataset. Above all, this dataset contains additional classes. At the same time, the number of available samples is increased. Nevertheless, the format of the grayscale images is maintained (see table 6.2). This is due to the fact that MNIST is included and existing models can be evaluated without further adjustments.

There are several sub-datasets, which consist of a maximum of 62 classes with 814 255 samples. Unfortunately, unbalanced classes are included. For this reason and for the sake of comparability to other datasets, a separate subset of the dataset is constructed. The 10 classes containing the most samples are selected.

Table 6.2: Detailed information of the EMNIST dataset.

| resolution | training samples | test samples | training balance | test balance | random **samples** (from classes) |
|---|---|---|---|---|---|
|  |  |  |  |  | 0 1 2 3 4 5 6 7 8 9 |
| 28×28×1 | 345 000 | 57 900 | 2.0 | 1.9 | 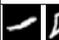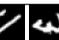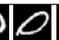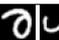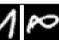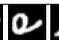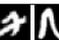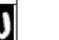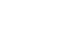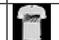 |

**FashionMNIST**    The FashionMNIST dataset attempts to address the simplicity of MNIST, as pointed out by Xiao, Rasul, et al. (2017):

> "*Moreover, Fashion-MNIST poses a more challenging classification task than the simple MNIST digits data [...]*" Xiao, Rasul, et al. 2017

The dataset is provided by the Zalando research group (see Xiao, Rasul, et al. 2017). The structure is identical to MNIST and can often be used by replacing the download URL. The resolution (dimensionality) of 28 by 28 pixels and even the size with 60 000 training and 10 000 test samples is the same. This has the advantage that models optimized for MNIST can be applied to FashionMNIST without any adjustments. Unlike MNIST, the grayscale images show representations of different types of clothing, with 10 available categories (see table 6.3).

Table 6.3: Detailed information of the FashionMNIST dataset.

| resolution | training samples | test samples | training balance | test balance | random **examples** (from classes) |
|---|---|---|---|---|---|
|  |  |  |  |  | 0 1 2 3 4 5 6 7 8 9 |
| 28×28×1 | 60 000 | 10 000 | 0 | 0 | 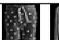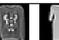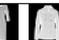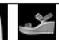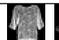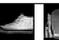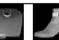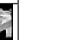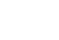 |

**SVHN**   The Street View House Numbers (SVHN) dataset (see Netzer and Wang 2011) consists of color images of house numbers extracted from Google Street View records. In this work, the cropped and centered variant is used with a resolution of $32 \times 32$ pixels. As depicted in the samples as part of table 6.4, images can be affected by interfering factors.

Table 6.4: Detailed information of the SVHN dataset.

| resolution | training samples | test samples | training balance | test balance | random **examples** (from classes) |
|---|---|---|---|---|---|
| | | | | | 0  1  2  3  4  5  6  7  8  9 |
| $32 \times 32 \times 3$ | 73 200 | 26 000 | 12.5 | 13.4 |  |

**Devanagari**   The Devanagari Handwritten Character Dataset (DHCD) consists of Indian characters. 46 elements of different character groups are included. Each contains 2 000 samples. The individual characters are extracted from handwritten documents and labeled by hand. As shown by Acharya, Pant, et al. (2016), each sample has the same resolution as samples from the MNIST dataset, but a margin of 2 pixels is added on each side. An initial test accuracy of 98.47% is reported.

In order to proportionally adjust the dataset compared to the other ones, 10 classes are randomly selected. The data is split into 90% training and 10% test samples (see table 6.5).

Table 6.5: Detailed information of the Devanagari dataset.

| resolution | training samples | test samples | training balance | training balance | random **examples** (from classes) |
|---|---|---|---|---|---|
| | | | | | 0  1  2  3  4  5  6  7  8  9 |
| $32 \times 32 \times 1$ | 18 000 | 2 000 | 0.1 | 1.6 |  |

**CIFAR-10**   The work of Vinet and Zhedanov 2011 proposes the Canadian Institute for Advanced Research (CIFAR) dataset. There are two different variants available, CIFAR-10 and CIFAR-100. Each number represents the contained number of classes. In the present work, CIFAR-10 is used. This variant consists of 60 000 samples, which are evenly distributed over the 10 classes. The color images include airplanes, automobiles, birds, cats, deer, dogs, frogs, horses, ships and trucks, which are all considered classes (see table 6.6).

Table 6.6: Detailed information of the CIFAR-10 dataset.

| resolution | training samples | test samples | training balance | test balance | random **examples** (from classes) |
|---|---|---|---|---|---|
| | | | | | 0  1  2  3  4  5  6  7  8  9 |
| $32 \times 32 \times 3$ | 50 000 | 10 000 | 0 | 0 |  |

**Fruits 360**   The Fruits 360 dataset consists of color images representing a variety of fruits. Fruits were photographed from different angles for the construction of the dataset. 55 244 samples were taken and classified into 81 different types of fruits. The images are available with a resolution of 100 by 100 pixels. The background is replaced by white color by using an automatism (see Mureşan and Oltean 2018). The 10 classes with the best representations are selected for this work (see table 6.7).

Table 6.7: Detailed information of the Fruits 360 dataset.

| resolution | training samples | test samples | training balance | test balance | random **examples** (from classes) |
|---|---|---|---|---|---|
| | | | | | 0  1  2  3  4  5  6  7  8  9 |
| $32 \times 32 \times 3$ | 7 700 | 2 600 | 3.2 | 3.1 |  |

**MADBase**   Sherif Abdelazeem 2010 provide the Modified ADBase (MADBase) dataset. ADBase consists of 70 000 samples of handwritten Arabic digits. 700 people from the field of higher education and state enterprises supported the construction of the dataset. The "Modified" version (MADBase) is an adaptation of the MNIST dataset. In particular, the format is adjusted to a uniform resolution of 28 by 28 pixels (see table 6.8).

Table 6.8: Detailed information of the MADBase dataset.

| resolution | training samples | test samples | training balance | test balance | random **examples** (from classes) | | | | | | | | | | |
|---|---|---|---|---|---|---|---|---|---|---|---|---|---|---|---|
| | | | | | 0 | 1 | 2 | 3 | 4 | 5 | 6 | 7 | 8 | 9 | |
| $28 \times 28 \times 1$ | 60 000 | 10 000 | 0 | 0 |           | | | | | | | | | |

**NotMNIST**   Even though NotMNIST is based on the MNIST dimensions, it uses publicly available fonts instead of handwriting. In addition, letters from $A$ to $J$ are depicted instead of digits (see table 6.9). The different fonts should increase the level of difficulty for ML models, see for example different $F$s: ,  or  (see Yaroslav Bulatov 2011). "*Judging by the examples, one would expect this to be a harder task than MNIST. This seems to be the case […]*" Yaroslav Bulatov 2011. Similar to many variants of MNIST, the NotMNIST dataset was created for the reason that MNIST seems to be too simple.

Table 6.9: Detailed information of the NotMNIST dataset.

| resolution | training samples | test samples | training balance | test balance | random **examples** (from classes) | | | | | | | | | | |
|---|---|---|---|---|---|---|---|---|---|---|---|---|---|---|---|
| | | | | | 0 | 1 | 2 | 3 | 4 | 5 | 6 | 7 | 8 | 9 | |
| $28 \times 28 \times 1$ | 529 100 | 18 700 | $\approx 0$ | $\approx 0$ |           | | | | | | | | | |

**ISOLET**   ISOLET is the only non-image dataset used in this work. Samples represent spoken letters recorded from 150 subjects (see Cole and Fanty 1990). It contains 7 797 samples. Each is encoded and consists of 617 features (float values). In order to establish comparability with other datasets, 10 classes are extracted. For its visualization, random samples of each class are plotted as illustrated in table 6.10.

Table 6.10: Detailed information of the ISOLET dataset.

| resolution | training samples | test samples | training balance | test balance | random **examples** (from classes) | | | | | | | | | | |
|---|---|---|---|---|---|---|---|---|---|---|---|---|---|---|---|
| | | | | | 0 | 1 | 2 | 3 | 4 | 5 | 6 | 7 | 8 | 9 | |
| $617 \times 1$ | 2 300 | 600 | 0.3 | 0 |           | | | | | | | | | |

### 6.1.1   Ground-Truth

The selection of datasets is subject to the *ground-truth*. In particular, it refers to class membership, which is crucial in the context of supervised learning. This applies to datasets whose classes and boundaries should be clearly defined. The ground-truth of a sample is given, if the class mapping is inherently correct, e.g, a subject should specifically draw the digit one for the MNIST dataset. Nevertheless, errors can occur related to the quality of samples to the assignment of labels. The selection of MNIST (test) samples in figure 6.1 is intended to show that there is a margin of interpretation, even for humans. However, this does not change the ground-truth of the data itself. Another frequently cited example is the classification of data in the area of spam detection. There is a clear separation between what is spam and what is not. The same is true for spam emails, which may be interpreted as valid emails. Likely, the ground-truth of the data remains unchanged.

However, the ground-truth of the data can change, if a regression problem is turned into a classification problem. For this purpose, the scalar value representing the label is assigned to a concrete class. Although the definition of class boundaries is fixed, their basic ground-truth may get lost. This can be due to the definition of class boundaries.

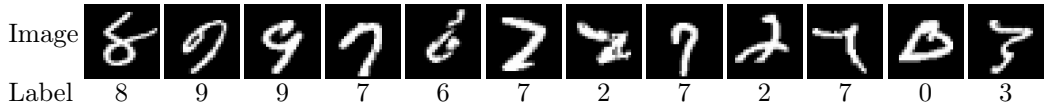

Figure 6.1: Suspicious digits of the MNIST dataset.

## 6.2  Sequential Learning Tasks

In this work, the Sequential Learning Tasks (SLTs) represent different CL scenarios. The datasets presented in section 6.1 are divided into different sub-datasets. The class labels within the datasets are used for this splitting. Since all datasets provide the same number of classes (10), any division can be applied to any dataset. The identifier of a SLT starts with a $D$ followed by the number of sub-tasks, or how many classes are contained in each sub-task. Therefore, the SLT $D_{10}$ consists of one single task which contains 10 classes. Since these ten classes represent all available classes, $D_{10}$ is referred to as the *baseline*. The experimental results of the baseline task serve as a comparative reference value for all other SLTs.

A baseline experiment corresponds to a joint training with all available data. Another example with multiple sub-tasks is $D_{3\text{-}3\text{-}3\text{-}1}$. It consists of 3 divisions with three and 1 sub-task with one class. Each sub-task within a SLT is enumerated, i.e., $T_1$, $T_2$, …, $T_x$ and corresponding classes are assigned. Thus, each sub-task is defined by one or more classes. All defined SLTs, respectively their assigned classes within the sub-task, do not overlap. They are thus disjunct, except for permuted SLTs. Permutation describes the shuffling of the samples' features according to a random, but fixed rule. In the definition of the SLT $D_{10\text{-}p10}$, the "p" indicates the permutation so that all samples are permuted in the second task. Thus, a second dataset is created, which contains the same information in a different order.

The defined SLTs (see table 6.11) are used to measure the CF effect. Different variations of SLTs are defined to test the effectiveness of the procedure. The difficulty of the tasks is represented by how many classes are added per task. Accordingly, the most difficult tasks could be the addition of one single class after another. In order to exclude the influence of the class order, variations are defined. As described in section 6.1, a training set $T_x^{\text{train}}$, as well as a test set $T_x^{\text{test}}$ is available for each sub-task $T_x$.

The permutation is based on a defined substitution operation. For this purpose, a seeded random permutation matrix is generated. This permutation matrix is applied to each sample. Thus, the order of the input features (e.g., pixel values) is swapped according to a defined pattern. The permutation applied to selected samples is illustrated in figure 6.2. A permuted dataset contains the same information and classes as the original dataset. Thus, the number of data points, classes and the degree of dimensionality are the same. Merely the order of the presented information/features changes.

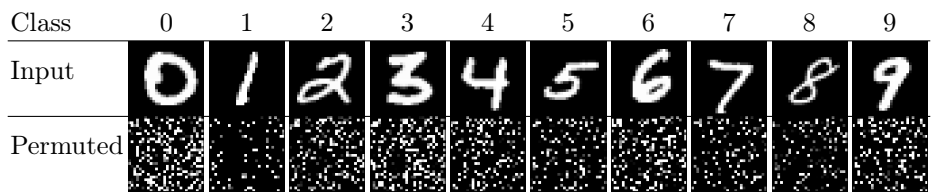

Figure 6.2: Exemplary permuted digits of the MNIST dataset.

## 6.3  Realistic Application-Oriented Evaluation Protocol

In this section, the development of an application-oriented CL evaluation protocol is presented. Above all, deep learning models are going to be investigated with this protocol, since they are affected by the CF effect. This effect leads to the (catastrophic) forgetting of all existing knowledge, as soon as an attempt is made to add new knowledge to a model. The special feature of application-oriented evaluation protocols is the implementation of real-world requirements. The newly developed investigation protocol should be used to examine CF-avoidance models in particular.

Table 6.11: Definition of SLTs and the class divisions of their sub-tasks.

| SLT | Sub-tasks |
|---|---|
| $D_{10}$ (baseline) | $T_1(0, 1, 2, 3, 4, 5, 6, 7, 8, 9)$ |
| $D_{10\text{-}p10}$ (permuted) | $T_1(0, 1, 2, 3, 4, 5, 6, 7, 8, 9)$    $T_2(0, 1, 2, 3, 4, 5, 6, 7, 8, 9)$ |
| $D_{9\text{-}1a}$ | $T_1(0, 1, 2, 3, 4, 5, 6, 7, 8)$    $T_2(9)$ |
| $D_{9\text{-}1b}$ | $T_1(0, 1, 2, 4, 5, 6, 7, 8, 9)$    $T_2(3)$ |
| $D_{9\text{-}1c}$ | $T_1(0, 2, 3, 4, 5, 6, 7, 8, 9)$    $T_2(1)$ |
| $D_{5\text{-}5a}$ | $T_1(0, 1, 2, 3, 4)$    $T_2(5, 6, 7, 8, 9)$ |
| $D_{5\text{-}5b}$ | $T_1(0, 2, 4, 6, 8)$    $T_2(1, 3, 5, 7, 9)$ |
| $D_{5\text{-}5c}$ | $T_1(0, 2, 5, 6, 7)$    $T_2(1, 3, 4, 8, 9)$ |
| $D_{5\text{-}5d}$ | $T_1(3, 4, 6, 8, 9)$    $T_2(0, 1, 2, 5, 7)$ |
| $D_{5\text{-}5e}$ | $T_1(0, 1, 3, 4, 5)$    $T_2(2, 6, 7, 8, 9)$ |
| $D_{5\text{-}5f}$ | $T_1(0, 3, 4, 8, 9)$    $T_2(1, 2, 5, 6, 7)$ |
| $D_{5\text{-}5g}$ | $T_1(0, 5, 6, 7, 8)$    $T_2(1, 2, 3, 4, 9)$ |
| $D_{5\text{-}5h}$ | $T_1(0, 2, 3, 6, 8)$    $T_2(1, 4, 5, 7, 9)$ |
| $D_{5\text{-}5i}$ | $T_1(0, 1, 2, 6, 7)$    $T_2(3, 4, 5, 8, 9)$ |
| $D_{3\text{-}3\text{-}3\text{-}1}$ | $T_1(0, 1, 2)$   $T_2(3, 4, 5)$   $T_3(6, 7, 8)$   $T_4(9)$ |
| $D_{2\text{-}2\text{-}2\text{-}2\text{-}2a}$ | $T_1(0, 1)$   $T_2(2, 3)$   $T_3(4, 5)$   $T_4(6, 7)$   $T_5(8, 9)$ |
| $D_{2\text{-}2\text{-}2\text{-}2\text{-}2b}$ | $T_1(1, 7)$   $T_2(0, 2)$   $T_3(6, 8)$   $T_4(4, 5)$   $T_5(3, 9)$ |
| $D_{1\text{-}1\text{-}1\text{-}1\text{-}1\text{-}1\text{-}1\text{-}1\text{-}1\text{-}1a}$ | $T_1(0)$   $T_2(1)$   $T_3(2)$   $T_4(3)$   $T_5(4)$   $T_6(5)$   $T_7(6)$   $T_8(7)$   $T_9(8)$   $T_{10}(9)$ |
| $D_{1\text{-}1\text{-}1\text{-}1\text{-}1\text{-}1\text{-}1\text{-}1\text{-}1\text{-}1b}$ | $T_1(7)$   $T_2(1)$   $T_3(2)$   $T_4(0)$   $T_5(6)$   $T_6(8)$   $T_7(4)$   $T_8(5)$   $T_9(9)$   $T_{10}(3)$ |

The first step towards the evaluation protocol is to summarize the requirements as prerequisite for a realistic CL scenario. Second, the selection of hyper-parameter is depicted. The resulting hyper-parameters are used for the third step, namely the training and re-training of ML models. As a last step, the metric used to compare a model is outlined.

### 6.3.1 Real-World Continual Learning Requirements

An application-oriented evaluation protocol is designed to implement real-world requirements. The presented real-world scenario is used for the collection of these requirements (see chapter 5). The resulting requirements are briefly summarized below and implemented in the evaluation protocol.

**Incremental training** A basic requirement all models have to satisfy refers to the ability of being trained incrementally (see section 2.3.2). This means that the ML models are able to add new knowledge step by step. As the used datasets consist of discriminating classification tasks, models need to be trainable in a supervised manner. However, the number of classes to be distinguished is unlimited, i.e., not only binary classification tasks can be realized.

**Constant memory consumption** Another requirement especially important for real-world scenarios concerns the constant consumption of memory. In general, it can be assumed that a certain amount of memory can be reserved by the model. However, memory consumption must not depend on the characteristics of the concrete CL scenario, e.g., the number of samples or sub-tasks. In a streaming scenario, the theoretically resulting memory requirements are infinite.

**No past data** The required constant memory consumption already indicates that no data is available for training from the past or previous sub-tasks. This is especially important in streaming scenarios. In theory, an infinite number of samples or sub-tasks would have to be stored. Accordingly, an infinite amount of memory would be required. Memory requirements are, however, very limited, especially in embedded systems.

**No future data**   Another real-world requirement concerns the use of future data. For the purpose of research and evaluation, sub-tasks with a temporal offset are introduced. In the context of realistic applications, however, access to future data is impossible. Only the data of the current sub-task and its predecessors are available, whereas the latter may be excluded due to memory limitations. This eliminates two possible training and evaluation procedures: First, at the initial training all data from the past and future are available. The access of future data would correspond to a joint training dataset for multiple CL tasks. Second, an adjustment of the current model state or hyper-parameters, based on future data, is impossible. These limitations may seem trivial at a first glance, but the corresponding information is not always implemented or explicitly presented in experimental setups (e.g., Kirkpatrick, Pascanu, et al. 2017; Goodfellow, Mirza, et al. 2013; Srivastava, Masci, et al. 2013).

**Constant update complexity**   Another requirement is the training/update complexity regarding the computational runtime. This is particularly true for streaming scenarios. If additional computing capacity is required for each new sub-task or sample, the update method is not suitable for streaming scenarios. This requirement prohibits the use of elaborate pruning or aggregation mechanisms that have to be performed after each sub-task. Therefore, the runtime complexity always has to be constant ($\mathcal{O}(1)$) and must not depend on the number or complexity of the sub-tasks. This requirement also excludes replay-based models, as long as the number of generated samples depends on the number of tasks (e.g., see Shin, Lee, et al. 2017).

**No oracles**   The use of oracles during training and testing would also affect the results. An oracle provides additional information. An oracle, for instance, indicates a sample's origin with regard to the sub-task. However, the use of oracles is prohibited in the context of the evaluation protocol for two reasons. Firstly, the chances of an oracle's existence strongly depend on the scenario. Secondly, not all ML models require information provided by oracles. One discrepancy in this regard concerns the trigger for new sub-tasks (adding new knowledge). This is referred to as task boundary. It is particularly easy to identify a possible boundary by tracing the used labels.

This information is only decisive for SLTs that have clear task boundaries at all. A changing data distribution is often unknown or fluid (see section 2.3.1), which is the case in the present real-world scenario. Therefore, a model should be able to recognize a task boundary or it has to be able to handle the respective changes.

**Continual learning objective**   In addition to the presented requirements, CL objectives are defined. The basic goal of a CL scenario is to accumulate knowledge. However, it has to and should be possible to overwrite incorrect knowledge in real applications. At a defined point in time (e.g., at the end of the training scenario), the maximum amount of knowledge should be available. In general, a loss of knowledge is assumed to occur over time. Nevertheless, the loss should take place gradually and not suddenly (i.e., catastrophically).

## 6.3.2   Hyper-Parameter Selection

The performance of ML models usually depends on hyper-parameters. The choice of the model architecture is one of the hyper-parameters (consider the XOR problem, see Brutzkus and Globerson 2019). Determining a model's architecture is referred to as model selection. In general, more artificial neurons or layers should be better, as more complex functions can be approximated. At the same time, overfitting may occur, which is a disadvantage (see section 2.2.2.5).

As many parameters are problem-dependent, they need to be adjusted, e.g., the learning rate $\epsilon$. Usually, it is difficult to infer the hyper-parameters from the problem. Instead, suitable parameters are detected via trial-and-error. Accordingly, various parameter values are tested, whereas the best result confirms the selection. Since there are often many different hyper-parameters available, a grid-search needs to be performed (see section 4.3). This means that each parameter variation is combined with each other. The experiments are conducted with the resulting combinations and evaluated as a next step. Depending on the number of hyper-parameters and values, this procedure leads to a large number of experiments. This can result in a very time and computationally intensive hyper-parameter optimization.

A challenge related to CL scenarios is the time slot and selection of hyper-parameters. In order to make a realistic assumption for CL scenarios, it is important that only the data of the first task is available for hyper-parameter selection. In other words, it is impossible to base the hyper-parameter selection on data from future tasks (or all data). This restriction has a significant influence on the chosen architecture and other hyper-parameters. In addition to the grid-search results, parameters suggested by other authors can be used, as long as they have been evaluated.

Models may be influenced by other parameters. These include the used batch size $\mathcal{B}$ and particularly the number of training iterations. A larger batch size comes with two advantages. The first one is that more representative (averaged) gradients can be determined. Compared to a batch size of 1, the parameters (weights) of the model are changed specifically for a single sample. The second advantage is that the determination of gradients for larger batches is computationally more efficient by means of ML frameworks. The limiting factor of the batch size is usually the available memory size. Despite the fact that there is usually more memory available, the batch size is set to 100 for all experiments. This choice is due to the great number of related works using the same value (see chapter 3).

The number of training iterations as a further parameter is crucial for the final performance of an ML model. In general, fewer training iterations can lead to a weak performance of the model. Since a stochastic optimization method is used for training (Stochastic Gradient Descent (SGD), see section 2.2.2.2), the crucial factors comprise the learning rate $\epsilon$, the derived gradients and the number of training iterations. In theory, one training step with a "correctly" initialized model, a "perfect" learning rate $\epsilon$ and a set/batch of "representative" data already leads to an optimal model. However, this is only possible in theory for very simple problems. In order to reduce the correlation between iterations and learning rate, a smaller learning rate has to be chosen, which requires an increased number of training iterations. If the learning rate is too small ($\epsilon \to 0$), it takes too many training iterations ($\to \infty$) for the model to converge. Since the two parameters are partially interdependent, the learning rate only is varied whereas the number of training iterations remains fixed.

Datasets do not always consist of the same number of training samples, e.g., see section 6.1. Moreover, the number of training iterations required for the obtaining of an acceptable ML model configuration is unknown. Thus, specifying a particular number of training iterations is problem-dependent. A comparable evaluation can be assured by specifying how often a complete dataset is processed. A complete run is referred to as an *epoch*. The training iterations required to process an epoch therefore depend on the batch size and the number of samples within the dataset. Regardless, the specified number of training iterations can always be a disadvantage for a model. The disadvantages result from either too many or too few iterations. At the same time, the number of iterations may also be related to other hyper-parameters or the problem itself.

*Early-stopping* is a known method to tackle the problem of stopping the training process at a "good" point. However, the implementation of early-stopping is challenging under real-world conditions due to the restriction of data access. Furthermore, it is necessary to periodically save the current progress of the model, which requires a lot of memory. The reserved memory for checkpoints must be allocated for a certain time before it can be released. Furthermore, it is impossible to exclude the possibility that a better parameter configuration can be obtained at a later training time. In general, the early-stopping approach contradicts application-oriented requirements (see section 6.3.1). In this work, the use of future or past test data to determine the best stopping point is prohibited.

In order to find a consensus regarding the number of training epochs, the following requirement should be fulfilled by the ML models. The model is not supposed to perform worse even if the training is "too long". This also applies to the overfitting effect (see section 2.2.2.5). Thus, based on empirical results, a fixed number of training epochs ($\mathcal{E}^{train} = 10$) is set for all datasets. All other hyper-parameters, such as number of hidden layers $L$, their size $S$ and other model-specific parameters are adapted by a grid-search. The varied parameters are summarized in the set $\mathcal{P}$. Concrete values for the parameter optimization are given in chapter 7.

### 6.3.3   Model Training and Re-Training

The application-oriented evaluation protocol starts with a grid-search for hyper-parameters. For this purpose, an ML model $m$ is initialized and trained with a given parameter configuration $\vec{p} \in \mathcal{P}$. In this phase, the training data from the first task ($T_1^{\text{train}} \in T_1$) of the corresponding SLTs is used. The test data $T_1^{\text{test}}$ can be used to evaluate the model at any time during training. The best configuration can be identified by evaluating the results of experiments with different parameter configurations. However, the number of measurement points and the scope of the test, e.g., the use of the entire test dataset, requires a certain effort.

The detailed procedure of all training and test steps is presented in algorithm 6.1. Line 1 represents the grid-search, which can be performed in parallel, since all models and parameters are independent for $T_1$. The training loop is shown in line 2, whereas training iterations are limited. The limit is calculated by the number of training samples $\#(T_1^{\text{train}})$, divided by the batch size and multiplied by the number of epochs ($\mathcal{E}^{train} = 10$). Basically, two operations can be distinguished: Training and testing. A training step is represented by $\text{train}(m_{\vec{p}}, T_1^{\text{train}})$ in line 3.

Due to the computational intensity of performance measurement, testing after each training step is too time-consuming and computationally expensive. Therefore, measurement points are set evenly distributed over the full dataset. Depending on the acquired number of measurements, the location of measuring points is calculated in advance (see line 4). The measurement step ($\text{test}(m_{\vec{p}}, T_1^{test})$) is more complex, as the complete test set needs to be predicted and compared with the target values/labels (see line 5). As long as the test data is representative, the determination of a realistic performance value is assured. The resulting test value ($q_{m_{\vec{p},t}}$) is stored as a quality measure for the respective model $m$, the corresponding parameter configuration $\vec{p}$ and the current state at iteration $t$ of the model $m$.

After the computationally intensive model selection phase (loop in line 1), the collected measurements are evaluated. For this purpose, the qualitatively best parameter setup $\vec{p}^*$ for a given model $m$ is specified, based on the maximum achieved value $p_{\vec{p},t}$ (shown in line 6). Succeeding this optimization step, the model is (re-)trained only by means of the following tasks ($T_2, ..., T_x$, defined by the SLT). This full evaluation process can be exemplified by a scenario in which a product is manufactured and has to acquire additional new knowledge after its delivery. As a result, a hard task boundary is imposed between production and application.

The following protocol procedure simulates the application. Moreover, new knowledge needs to be added to the model from the following tasks $T_2$ to $T_x$ (see line 7). By definition, it is still possible to optimize certain model parameters of the protocol based on ($\vec{p}^*$). However, the additional optimization is limited to a single parameter, namely the learning rate $\epsilon$ (see line line 8). The respective limitations are discussed in section 6.4.

The next step in line 9 results in a copy (checkpoint) of the best model. Subsequently, the pre-parameterized model is re-trained (lines 10 and 11) with the training data of the current task ($T_c^{\text{train}}$). Similarly, the quality of the model is measured for the current task ($T_c^{\text{test}}$) at several points during the training process. In addition, the performance on the baseline dataset is measured (see line 13). The baseline dataset $D_{10}$ contains all test samples ($T_{10}$) and is therefore considered as a reference value. However, this data is used only for evaluation purposes. It is excluded from training or other optimization steps.

The final step after each simplified parameter optimization is the selection of the best model. Two reference points can be used to identify the best model or model configuration. The maximum measured performance value is one of them. Alternatively, the last measured performance value (at the end of re-training) can constitute a more realistic reference point. The chosen value is subject of a later discussion (see section 6.4).

## 6.3.4 Model Evaluation

In this section, the precise evaluation procedure is presented. The evaluation of the experimental results is described in the last part of the protocol (see algorithm 6.1). The algorithms describe how a test step is executed and how the collected values are evaluated. A second *prescient* evaluation procedure illustrates the influence of the strategy on the results.

The smallest part of the evaluation step is the test function ($\text{test}(m, T_x^{test})$). This test function is used multiple times in algorithm 6.1, i.e., in lines 5 and 13. The test function only uses test data to perform a prediction based on the current state of the model $m$. Side effects from testing must not influence the training process or other properties of the model. As specified by the requirements (section 6.3.1), past and future test data may only be accessed for evaluation purposes.

A complete model test consists of many small test steps, since a batch size $\mathcal{B}$ is set for testing. The batch size has no influence on the test results. Theoretically, it can be as large as the number of samples in the test dataset. For reasons of simplicity, the batch size is set to the same size as the training batch size ($\mathcal{B} = \mathcal{B}^{train} = \mathcal{B}^{test} = 100$). Thus, depending on the size of the test dataset, multiple test steps have to be executed for a complete epoch ($\mathcal{E}^{test} = 1$). This procedure is comparable to the

---

Algorithm 6.1: The *realistic* application-oriented evaluation protocol.

**Data:** model $m$, SLT with sub-tasks $T_1, T_2, ...T_x$, hyper-parameter value set $\mathcal{P}$

**Result:** model with best continual learning quality for parameter configuration $q_{m_{\vec{p}^*,t,\vec{p}^+}}$

1   **forall** *parameter configuration $\vec{p} \in \mathcal{P}$* **do**    // determine accuracy for all hyper-parameters when training on $T_1$

2     **for** $t \leftarrow 0$ to $\frac{\#(T_1^{train})}{\mathcal{B}} \cdot \mathcal{E}^{train}$ **do**

3       train($m_{\vec{p}}$, $T_1^{train}$)

4       **if** $t \in measurement\ points$ **then**

5         $q_{m_{\vec{p},t}} \leftarrow$ test($m_{\vec{p}}$, $T_1^{test}$)

6   $m_{\vec{p}^*} \leftarrow$ model $m_{\vec{p}}$ with maximum $q_{m_{\vec{p},t}}$        // find best model with max. accuracy on $T_1$

7   **forall** $T_c \in (T_2, ..., T_x)$ **do**                                // subsequent tasks

8     **forall** *parameter configuration $\vec{p}^+ \subset \vec{p}^*$* **do**    // parameters to be optimized subsequently

9       $m_{\vec{p}^*,\vec{p}^+} \leftarrow m_{\vec{p}^*}$

10      **for** $t \leftarrow 0$ to $\frac{\#(T_c^{train})}{\mathcal{B}} \cdot \mathcal{E}^{train}$ **do**

11        train($m_{\vec{p}^*,\vec{p}^+}$, $T_c^{train}$)

12        **if** $t \in measurement\ points$ **then**

13          $q_{m_{\vec{p}^*,t,\vec{p}^+}} \leftarrow$ test($m_{\vec{p}^*,\vec{p}^+}$, $T_c^{test}$) `AND`   $q_{m_{\vec{p}^*,t,\vec{p}^+,D_{10}}} \leftarrow$ test($m_{\vec{p}^*,\vec{p}^+}$, $T_1^{test}$ of $D_{10}$)

14     $m_{\vec{p}^*} \leftarrow m_{\vec{p}^*,\vec{p}^+}$ with best $q_{m_{\vec{p}^*,t,\vec{p}^+}}$   // find best model with max. or max. last accuracy on $T_c$

---

individual training steps.

Many other studies in the literature use a wide variety of metrics, and choosing the right metric is critical. The decisive criterion is the underlying data and the problem. The following questions have to be considered: *Is it a regression or a classification problem? Is it a binary, single class or a multi-class classification problem? What is the goal of the prediction? Is the data evenly distributed across the classes?* For the present evaluation protocol, only classification problems are examined, which manifest as single class problems. This means that exactly one label is applicable to each sample (see section 2.2). False predictions, however, do not distinguish between the degree of falseness.

The goal of the test function is to determine whenever a prediction is correct or incorrect. In other scenarios, this is different, for example, with regard to the detection of very rare events. Accordingly, other metrics such as precision, recall, F1 score, cross entropy are used (see Handelman, Kok, et al. 2019). Since the datasets of this work have an approximately even class distribution (see section 6.1), the accuracy is used. Thus, other metrics would yield the same results. At the same time, determining the accuracy is more efficient. It simply requires the ratio of correctly and incorrectly classified samples.

As a first step, the test function uses the model to perform a predication of the test data $T_x^{test}$. Each single test sample $\boldsymbol{x}_n$ is provided with a label $\boldsymbol{y}_n$ (one-hot format). Once the predicted class label matches the true label, it is considered as a correct prediction $\hat{\boldsymbol{y}}_n = \boldsymbol{y}_n$. Otherwise, ($\hat{\boldsymbol{y}}_n \neq \boldsymbol{y}_n$) the predication is wrong.

$$accuracy = \frac{\#(\text{correct classified})}{\#(\text{tested samples})} \qquad (6.1)$$

Equation 6.1: Definition of the accuracy metric.

The procedure described above (multiple use of the test function) specifies how a single measurement point is created. The distribution of the measurement points is even over each training task. Depending on the number of training samples within the training dataset, the position of measurement points is determined. Their placement can be decisive, if there are many training iterations between the measurement points, and a possibly good model parameter configuration can be skipped. From a computational point of view, however, it is very time-consuming to perform a full test after each training step. Therefore, an ML model should be robust in different training stages to ensure the independence of test point placement. It is expected that the model converges to a suitable local minimum and stays there after several training iterations (see section 2.2.2.2). This should apply to further training steps. The same is expected in the context of overfitting (see section 2.2.2.5).

The advantage of a supervised learning scenario is that samples from the dataset can help determine the quality of the model. This is crucial for the selection of the hyper-parameters at the beginning of training. Test data constitute the basis for the evaluation of a model's performance, which allows for the determination of hyper-parameters. Experiments are performed with the different parameter configurations, and the corresponding results are evaluated.

One or more measurements can be used for the evaluation and thus for the selection of parameters. First, the *maximum measured accuracy* can be decisive for the most likely optimal parameter configuration. After the completion of the first task $T_1$, the maximum measured accuracy is identified $(\max(q_{m_{\vec{p}},t}))$. This value is assigned to a specific parameter configuration $\vec{p} \in \mathcal{P}$. Thus, the best possible parameter configuration is determined for a specific model $m$, which has been measured during the initial training phase. The disadvantage is that only one model parameter configuration may lead to this result, which cannot be considered as representative. Second, the average accuracy over all measured values is a possible procedure, which is more representative than one maximum value. Therefore, the estimated parameter configuration is more representative in terms of a constant quality of the model. The third and last measured value (or average over some of the last measuring points) is represented by the final quality of the ML model. It is a realistic quality measure with respect to application-oriented scenarios.

The determination of the best possible model $m_{\vec{p}^*}$ or parameter configuration can be realized by means of more complex procedures. However, this only applies to the first task $T_1$ and models whose initialization takes place within a development environment. The intended application may also have an influence on the used selection criterion. It is assumed that, for example, a model is initialized by a manufacturer. Thus, the first selection strategy can be used and the best possible accuracy results are expected. In general, a hyper-parameter adjustment is impossible in certain application scenarios.

The initial grid-search is followed by the presentation of the second task $T_2$. During re-training, the accuracy of the model is measured by using the test data of the second task $(T_2^{test})$. The related results can be used to re-evaluate various parameters, as performed for $T_1$. The possibility of a re-optimization of parameters, however, strongly depends on the application scenario, e.g., an adjustment of the used learning rate. Re-optimization can be impossible due to various limitations, e.g., for embedded systems with few resources. Therefore, the most restrictive evaluation approach is that no further optimization can be performed. This very strong limitation has been relaxed for the present protocol, even though its implementation is straightforward. Thus, the ML models are granted a simple optimization that only affects the learning rate.

In addition to the evaluation of the active sub-task, the overall baseline accuracy is measured. The baseline SLT $D_{10}$ consists of only one task $(T_1)$, which includes all test samples of the entire dataset. However, neither the baseline nor the test data have to be used for any optimization steps. The re-training step is repeated until all sub-tasks of an SLT are processed (see section 6.2).

For the interpretation of the resulting values, the accuracy of the baseline task $D_{10}$ is crucial. It represents how much knowledge is retained after all training tasks have been completed. An accuracy of 100 % is the optimum. Initially, it is independent of the baseline. However, this value is supposed to be unrealistic if it is not reached by the baseline experiment. Therefore, the expected values are limited to a maximum of the baseline results.

After the completion of the entire training process, the question arises which criterion is used to identify the "best" experiment. Another question is whether the last or the highest measured accuracy value is decisive (or another value). The following example illustrates the basic selection problem. A non-uniform distributed SLT is considered as a starting point, e.g., $D_{9\text{-}1}$. First, the knowledge from 9 classes has to be derived by the model. Thus, the expected accuracy for a 10-class problem is 90 % at most for the first task $T_1$. Second, the knowledge of a single class needs to be added. Thus, a maximum gain in accuracy of 10 % can be expected from $T_2$. Assume that the model is still partly subject to the CF effect and that only little knowledge of $T_1$ can be preserved. Consequently, a very low accuracy is achieved towards the end of the training process, e.g., 20 %. According to the maximum principle, the best experiment with an accuracy of 90 % is selected. The experiment competes with models that may still have 89 % accuracy at the end of the training process. For uniformly distributed SLTs (e.g., $D_{5\text{-}5}$), this information can be a critical factor, although it does not serve as a selection criterion. Finally, the application-oriented test protocol specifies that from the second task onwards, the last measured accuracy value is going to be used.

## 6.4   Discussion

In this section, the relation between the evaluation protocol and real CL scenarios is discussed. Open aspects concern the re-optimization of parameters after each task and the determination of the best parameter configuration. Furthermore, the number of training iterations and task boundaries are addressed. Finally, a comparison with a prescient evaluation protocol is conducted.

**First Hyper-parameter Optimization**   The first point up to discussion concerns the hyper-parameter optimization. The initial grid-search is quite realistic and can be considered a standard procedure. Problem-specific parameters can be determined with this procedure. Without it, an estimation of the resulting model quality is very difficult. The same is true for non-*continual learning* scenarios. If no prior knowledge about a scenario is available, it is referred to as zero-domain-knowledge scenario. However, such scenarios are rare in the domain of real-world applications. This way, an initial problem-dependent hyper-parameter optimization is justified.

**Subsequent Hyper-parameter Optimization**   The subsequent parameter optimization is questionable in the context of real-world CL applications. A second extensive optimization is problematic for products that have to operate autonomously after their assembly/delivery. Lightweight devices with very limited resources cannot perform hyper-parameter optimization. Due to memory limitations, checkpoints of an ML model cannot be stored. However, these are required for optimization in some cases. The same applies to reference data. After a comprehensive initial optimization phase, an ML model should be able to add knowledge without much effort. Moreover, an optimization should not involve the use of external services, such as data centers. This may be prohibited by either limited Internet access or data privacy issues.

In certain CL scenarios, an ML model cannot be adapted to new knowledge by a grid-search. A second grid-search is therefore only possible under specific circumstances. Nevertheless, the protocol allows for a second optimization. The implementation of a further optimization is justified by the following arguments. In general, a very costly and complex fine-tuning process is conducted for a given problem. Fine-tuning involves a fine-grained adjustment of multiple hyper-parameters and the training process. The goal is to find the best possible model configuration. Since fine-tuning cannot be performed specifically for all datasets and each ML model, further optimizations are performed as a compensation. In order not to extensively distort the evaluation results, only one parameter is optimized. The learning rate as the chosen parameter has a fundamental influence on all models and problems. As all models are affected by this parameter, a fair comparison can be conducted.

**Measurement Points**   Another aspect that requires a discussion concerns the number and positioning of measurement points. ML models can strongly depend on the number of training iterations. This is especially true for the performance measurement of CL tasks. If a model is re-trained for too long, it cannot retain the previously derived knowledge anymore. The estimation of an optimal model parameter configuration may depend on the positioning of the measurement points. In theory, strongly dependent models need to be tested after each training iteration. However, these exhaustive tests are computationally expensive.

The measuring points are associated with checkpoints. Checkpoints represent model parameter configurations with regard to a performance state for a given training iteration. However, creating a checkpoint for every measuring point would require an enormous amount of memory. Moreover, checkpoints are required to restore the best measured condition. This procedure contradicts the application-oriented requirements, especially after the first sub-task. To avoid favoring vulnerable models, a uniform placement of measurement points is used for all problems and models.

**Performance Criterion**   The previously addressed problems are related to the determination of the *best* model parameter configuration. The general question is which measurement value should be used. Even though section 6.3.4 describes conventional methods (e.g., maximum, last, average), other measurement values are possible. Many methods are conceivable, although the evaluation protocol assumes that more resources are available for the initial optimization phase. In order to ensure comparability, a uniform and realistic procedure is crucial. Thus, a selection according to the maximum performance of a model is realistic – at least with regard to the initial optimization.

For further re-training steps, the procedure needs to be adapted to the CL scenario. In general, the accuracy of the model can be tracked over time, as long as test data is available. However, due to resource constraints, the duration/capacity to store past samples is limited. At the same time, an optimization must be possible with the available resources (data, memory, computing power). If this is not the case, an optimization is not applicable at all. This applies to certain real-world scenarios.

Due to the previously discussed circumstances, the *last* measuring point is used as the selection criterion for further sub-tasks. This approach is the most restrictive one, even though early-stopping mechanism would weaken it. Since early-stopping involves additional problem-dependent parameters based on past test data, this mechanism is not applied. If a corresponding mechanism is already implemented by the ML model, the presented requirements must not be violated. The same applies to methods addressing the overfitting effect.

**Task Boundaries**  Another issue related to real-world applications concerns hard class boundaries. The presented SLTs only describe abruptly changing data distributions. At the same time, classes within the sub-tasks are always disjoint. This condition as atypical as shown in the exemplary real-world scenario (see chapter 5). There are no hard borders and the distribution seems to change arbitrarily without any chance to influence this development. Moreover, it is not recognizable whether a concept or context drift/shift has occured (see section 2.3.1). This aspect causes problems with regard to the evaluation of CF avoidance models within real-world CL scenarios. Since an evaluation in these scenarios is very difficult to interpret, SLTs should simulate a CL task. Thus, the protocol only describes hard task boundaries in the beginning.

The use of task boundaries is due to two reasons. The first one is related to the investigated models and their characteristics. In general, task boundaries are expected by many ML models and serve as a trigger for the various CF avoidance mechanisms (e.g., merging strategies). The CF avoidance mechanisms cannot be initiated without the specification of boundaries. Unfortunately, the discovery of boundaries is usually not part of the models and has to be realized by additional methods. Generally speaking, a new task can be identified easily based on the labels. This does not necessarily apply to real-world applications. Therefore, only the defined SLTs are provided for this protocol.

The second reason is the definition of a different drift scenario. Synthetic (virtual) concept/context drifts/shifts are easy to implement. Moreover, the learning objective would be the same for such scenarios. However, many procedures/models would not work without task boundaries. Therefore, the investigation of different types of changing data distributions (e.g., see section 2.3.1) is subject to future research.

**Datasets**  Another point concerns the datasets used for the evaluation. In section 6.1, various image datasets are presented. The various advantages of image datasets are discussed in the following. Image datasets have many features (high-dimensional). Moreover, the label and sample can be validated by visual inspection. These visual or similar datasets can be found in many studies, even though they are hard to assess with regard to their difficulty. Only reference values of prediction accuracies can be used for this estimation of difficulty. However, prediction results are not based on simple standard models (regarding the architecture) but on highly optimized ones, e.g., AlexNet Krizhevsky, Sutskever, et al. 2017. Convolutional Neural Networks (CNNs) are particularly interesting in image processing models, as they are biologically inspired and mimic the process of vision.

Despite "simple" model architectures, the difficulty of individual datasets can be considered in relation to the baseline experiments. Conducted experiments for the SLT $D_{10}$ (including all classes in one task $T_1$) are assumed to be the baseline, which provides an approximate upper bound with respect to a model's quality. Accordingly, the difference between the baseline and CL tasks indicates if a certain model successfully avoids the CF effect. Implications and statements are strengthened by the investigation of several different datasets (see section 6.1).

**More Realistic Evaluation Protocol**  A fundamental question is whether the protocol allows for a more realistic evaluation of CF avoidance models. To answer this question, the protocol is compared to the *prescient* protocol, which is summarized in algorithm 6.2. Algorithm 6.2 presents the compact version of the *prescient* protocol. In the first training section, an ML model with different parameter constellations is trained (i.e, hyper-parameter optimization). Training starts with the first task $T_1$ (see line 1), which is similar to algorithm 6.1. For clarity, the number of training iterations will be abbreviated by $t_{\max} = \frac{\#(T_1^{\mathrm{train}})}{\mathcal{B}} \cdot \mathcal{E}^{train}$. SLTs consisting of only two tasks ($T_1$ and $T_2$) are used. Subsequently, a model is re-trained, which is based on a reduced parameter optimization ($\vec{p}^+$) as displayed in line 2. Lines 3 and 4 represents the training loop where the re-training of the model $m_{\vec{p},\vec{p}^+}$ is performed with training data from $T_2$.

By applying this minor strategic change, a decisive difference compared to the realistic protocol (algorithm 6.1) is expected. Based on the first parameter optimization, not only the best model is re-trained anymore. Instead, all $m_{\vec{p}}$ configurations are re-trained with $T_2^{train}$. Nevertheless, the reduced parameter optimization is performed, represented by the $p^{\vec{+}}$. As a last step, the experiment with the maximum accuracy at any given time is selected from the comprehensive collection of parameter configurations.

---

**Algorithm 6.2:** The *prescient* evaluation protocol.

**Data:** model $m$, SLT with sub-tasks $T_1, T_2$, hyper-parameter value set $\mathcal{P}$
**Result:** best model $m_{\vec{p}^*}$

1  initial training of $m_{\vec{p}}$ on $T_1^{train}$ for $t_{\max}$ iterations
2  **forall** *parameter configuration $\vec{p}^+ \subset \vec{p}^*$* **do**            // parameters to be optimized subsequently
3    **for** $t \leftarrow 0$ **to** $t_{max}$ *iterations* **do**                                 // re-training of $m_{\vec{p}}$ on $T_2$
4      train($m_{\vec{p},\vec{p}^+}, T_2^{train}$)
5      $q_{m_{\vec{p},\vec{p}^+,t}} \leftarrow$ test($T_1^{test} \cup T_2^{test}, t$)
6  $m_{\vec{p}^*} \leftarrow m_{\vec{p},\vec{p}^+}$ with best $q_{m_{\vec{p},t,\vec{p}^+}}$                         // find best model with max. accuracy on $T_1 \cup T_2$

---

By applying the prescient protocol of algorithm 6.2, the results are supposed to be comparable to related studies. Results obtained by the application of the prescient protocol should at least prove whether the code base is valid. Significantly worse results are expected from the developed realistic evaluation scenario. This is particularly due to the application-oriented requirements and their implementation. As long as the conditions are comparable, models with similar results for CL tasks (SLTs) and baseline experiments can be used for applications in CL scenarios.

Moreover, it is expected that the model performance applies to all SLTs and all datasets. Considering the implementation of other requirements into the protocol, this assumption is no longer valid. Different types of changing data distributions are not implemented as part of the protocol. Furthermore, the evaluation scheme can be modified for, e.g., few-shot learning or semi-supervised approaches. Depending on the adaptation, the chosen metric as well as the determination of the "best" parameter configuration have to be modified. The same applies to the used datasets.

## 6.5   Conclusion

The presented realistic evaluation protocol is intended to measure the CL performance of CF avoidance models under application-oriented constraints. The protocol takes several real-world requirements into account (see section 6.3.1). They include, for instance, the restriction of access to past or future data, which are not available under realistic conditions. As expected, real-world requirements affect the CL performance of CF avoidance models. This contradicts the statements of previous investigations by other authors.

Based on the evaluation protocol, more precise statements regarding the avoidance of the CF effect can be expected. First of all, an ML model has to meet the collected real-world requirements. The subsequent evaluation of the protocol is supposed to reveal to what extent CF can be reduced by means of the respective avoidance models. The investigated models should achieve similar results to those in a non-continual learning scenario (referred to as baseline experiments). In this context, a model must achieve acceptable results on multiple problems (datasets, see section 6.1) and for various difficulties (Sequential Learning Tasks (SLTs), see section 6.2). If one or more ML models can demonstrate an acceptable CL performance for this protocol, additional and stricter requirements should be implemented.

# 7.   Application-oriented Evaluation of Proposed Continual Learning Methods

## Chapter Contents

In this chapter, the investigation of different *catastrophic forgetting* (CF) avoidance models is presented. The application-oriented investigation protocol described in chapter 6 constitutes the basis for the evaluation. Models from chapter 3 meeting the protocol's requirements are examined. The evaluation protocol implements various requirements that originate from real-world application scenarios in the context of *continual learning* (CL).

**Contributions**   The goal of this chapter is to investigate different CF avoidance models in order to determine their CL performance. Accordingly, the protocol proposed in chapter 6 is applied. Two research questions are answered within this chapter. The first answer responds to which CF avoidance models exist (RQ 2.1). The second one addresses which model can be harmonized with the real-world requirements from the investigation protocol (RQ 2.2). This will ultimately answer research question RQ 2 and to what extent the models help reduce the CF effect in CL applications.

First, the models were examined according to a prescient version of the investigation protocol that does not meet the real-world requirements. The results are comparable to those from related work, so that the implementation can be considered valid. However, the results according to the real-world protocol described in chapter 6 show a different outcome. Under more realistic conditions, the investigated models and techniques show that the CF effect can only be suppressed for certain tasks. In a direct comparison, Elastic Weight Consolidation (EWC) has the best CL properties. Likewise, the evaluation of a model that does not meet the requirements does not exhibit a perfect CL performance. The finding of this chapter is that none of the investigated CF avoidance models can effectively mitigate the CF effect under real-world conditions.

**Structure**   First of all, the experimental setup for the applied investigation procedure is introduced. In this context, a brief summary of chapter 6 is provided. In addition, the model-specific parameters and the associated grid-search are presented. Subsequently, the experimental results of the grid-search are summarized. The initial experiments are evaluated according to the prescient protocol. This evaluation is supposed to show the validity of the implementation. The main focus of this chapter are the conducted experiments based on the realistic protocol. Finally, the results are discussed and implications for this work are presented.

## 7.1   Experimental Setup

The experimental setup describes the procedure of the investigation. Experiments are performed by using the prescient (see section 6.4) and the realistic protocol (see chapter 6). The goal is to derive a statement concerning the occurrence of the CF effect based on the experiments. At the same time, the *machine learning* (ML) models can be verified for their CL capabilities in a realistic scenario. The general process is summarized in figure 7.1. The process describes the step by step presentation of learning tasks, so-called Sequential Learning Tasks (SLTs) (see figure 7.1a), to an ML model (see figure 7.1b).

As part of this process, different requirements from real-world scenarios have to be respected. During a training task, only data of the current sub-task is available (sub-task $T_1$ in blue or $T_2$ in green). For evaluation purposes, the baseline test data is used during the entire training process. The baseline dataset ($D_{10}$) consists of all used classes (highlighted in red). In addition to the scheme shown in figure 7.1a, other constraints limit the selection of hyper-parameters during the process. One of the constraints is that no model selection (architectural or parameter decisions) may be performed retroactively. The model selection needs to be completed after the first sub-task $T_1$.

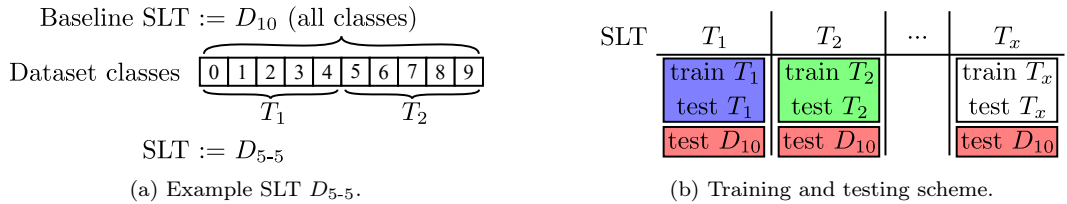

<table>
<tr><td>(a) Example SLT $D_{5\text{-}5}$.</td><td>(b) Training and testing scheme.</td></tr>
</table>

Figure 7.1: Parts of the evaluation protocol.

Before the application-oriented evaluation protocol is applied, it has to be ensured that comparable results can be achieved with the code base. For this purpose, the prescient protocol described in section 6.4 is utilized (see algorithm 6.2). After reproducing the results from related work, the code can be reused for the realistic evaluation protocol. Initially, a selection of specific SLTs is processed (see section 6.2: $D_{10}$, $D_{5\text{-}5}$, $D_{9\text{-}1}$ and $D_{10\text{-}p10}$). This list includes only SLTs, which consist of exactly two sub-tasks. As fewer sub-tasks are considered to be easier, it should at least be possible to achieve satisfactory results for this kind of SLTs.

In order to eliminate the difficulty of different problems, multiple datasets are utilized (see section 6.1, only MNIST, FashionMNIST, NotMNIST, EMNIST, Devanagari, MADBase, CIFAR-10, SVHN and Fruits 360). If all investigated parameter configurations related to a model show clear positive results towards the prevention of the CF effect, additional investigations are performed. As the exploration of many different hyper-parameters is very computationally expensive, more models can be examined. At the same time, the hyper-parameter selection ensures that a real mitigation of the CF effect is not achieved "randomly".

It is expected that all investigated models will provide comparable results to the basic tests according to the prescient protocol The realistic investigation protocol is applied to examine the occurrence of the CF effect under real-world conditions. Different variants of knowledge loss can be distinguished, which are outlined in figure 7.2. Without the application of a CF avoidance technique, the typical CF behavior can be observed, as experimentally shown in figure 7.2a. According to Goodfellow, Mirza, et al. (2013), the application of dropout (Hinton, Srivastava, et al. 2012b) should reduce the effect. Other promising mechanisms include the frequently cited EWC. The decisive factor is to what extent the effect is mitigated. A linear loss, as shown in figure 7.2b, would already be a significant improvement.

However, new challenges arise with linear forgetting. One of them is related to the identification of the best model configuration, and whether the training process has to be stopped. Early-stopping is closely related to other constraints. One of them concerns the access to reference data aiming at the determination of an optimal model configuration. In the ideal case, the CF effect is completely eliminated so that the model can be considered suitable for real-world CL scenarios (see figure 7.2c). If one or more models show the absence of the CF effect, the evaluation would be performed under more restrictive conditions and consider more complex SLTs and datasets. In this context, certain optimizations could be omitted (see section 6.4), such as adjusting the learning rate for follow-up tasks.

Thus, further vulnerabilities of the CF avoidance models can be identified.

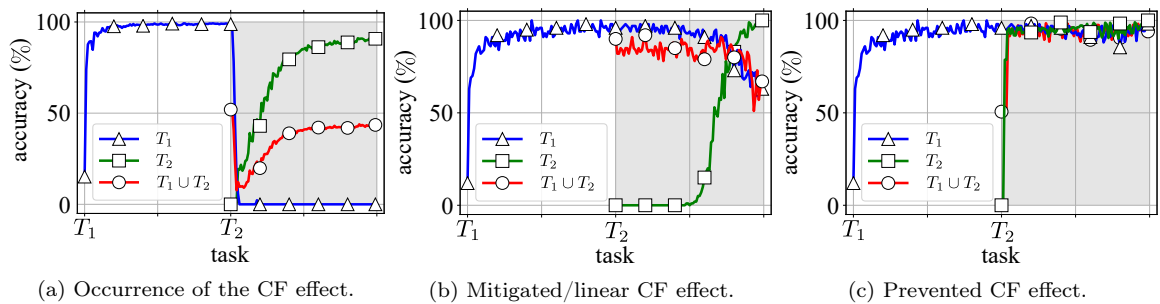

(a) Occurrence of the CF effect.    (b) Mitigated/linear CF effect.    (c) Prevented CF effect.

Figure 7.2: Different profiles of the CF effect.

### 7.1.1 Models

The models selected for evaluation are briefly introduced in section 3.2. The selection is based on the criterion whether a model can be aligned with the requirements of the evaluation protocol (see chapter 6). In addition, the author's code base has to be available in order to eliminate implementation interpretations. This is crucial, although even a faulty re-implementation cannot guarantee comparable performance. Thus, only models with a reference implementation can be examined. Details about the investigated models are presented in the following.

**Fully-Connected (FC)**  The simplest model is the one referred to as Fully-Connected (FC). It corresponds to a standard feed-forward Deep Neural Network (DNN), which is composed of several layers consisting of a fixed number of artificial neurons (see section 2.2.1). The output function is constituted by the Rectifier Linear Unit (ReLU) function, which is used for all layers. However, the last layer is terminated by a softmax readout. The DNN is optimized by minimizing the cross-entropy (see section 2.2.2.1). The number of hidden layers $L$ and the number of artificial neurons $S$ are hyper-parameters.

This simple variant does not comprise any mechanism to counteract the CF effect. As a consequence, it serves as a reference model. Moreover, it constitutes the basis for a modification (application of dropout), which is also investigated. In general, the label fully connected (FC) refers to the type of connection between the single layers (also referred to as *dense layer*). It is characterized by the fact that each neuron of the preceding layer is connected to each neuron of the subsequent layer (see figure 2.5).

**Convolutional Neural Network (CNN)**  CNNs were a decisive development step for image processing. The idea originates from biology and mimics operations of the visual cortex in living organisms. Early developments are attributed to LeCun, Haffner, et al. (1999). Since then, CNNs have become the standard mechanism in image recognition. A comprehensive description is available in Behnke (2003).

CNNs owe their performance to the operations convolution and pooling. Convolutional layers transform the input signal based on filters that correspond to the receptive fields. Each filter consist of one or multiple kernels. Kernels are defined by matrices that apply the dot-product, which realizes various image processing operations. These include, for example, the highlighting of edges or blurring. The result of each filter is a so-called feature map, whose activations are passed on to the next layer. The height and width of the receptive fields are hyper-parameters that have to be specified.

The same applies to the kernel and step size, which specifies how the filter is shifted over the input signal. Filters have to be shifted over the edges of the input data (e.g., images) which results in an undefined operation. Various padding techniques are used to define this area, e.g., set values to zero or the same value as the border pixels. In addition, pooling layers are introduced which reduce the dimensions. This simple mechanism combines several pixel values (which is also defined as an image area) by using various options. Different types of pooling operations can be applied: maximum, average or sum, etc. Max-Pooling, for example, aggregates multiple values from an image area, e.g., $2 \times 2$ pixels, on the basis of the maximum value.

The structure of a typical CNN consists of one or multiple convolutional filters and pooling layers applied alternately. Each one can be parameterized differently. A CNN is usually terminated by a linear classification layer consisting of artificial neurons. Weights and biases constitute the free parameters of the linear layer. The same is true for the kernels of the convolutional layers. In order to reduce the number of parameters, several filters share the same weights. Accordingly, layers closer to the input signal are responsible for local structural properties, such as the enhancement of edges. Deeper layers are responsible for details, e.g., textures. As a model's architecture is usually defined by hyper-parameters, complex but proven structures are deployed.

The architecture applied in this work consists of two convolutional layers. Each of them contains 32 and 64 filters of the size $5 \times 5$. The output function is ReLU. Max-pooling with a size of $2 \times 2$ pixels is introduced for the reduction of dimensions. The final linear classification layer consists of 1024 neurons. The cross-entropy loss function is minimized for training.

Although CNNs provide an enormous performance boost for image processing/classification, they are affected by CF. Similar to FC models, standard CNNs do not have any mechanism to reduce or control the CF effect. Therefore, standard CNNs serve as reference models for the application of mechanisms to avoid the CF effect.

**Elastic Weight Consolidation (EWC)**   EWC as introduced by Kirkpatrick, Pascanu, et al. (2017) is probably one of the most well-known and most cited CF avoidance models. The model is classified as regularization approach (see section 2.3.5). Changing important parameters for past tasks is punished by adding a penalty term to the loss function (see equation 7.1). With the loss function proposed in equation 7.1, regular DNN architectures can be trained. The first part of equation 7.1 ($\mathcal{L}_B(\theta)$) corresponds to the loss of the current task $B$. As the determination of the posterior probability cannot be resolved, the Laplace approximation is introduced. A Gaussian distribution is utilized to determine the posterior for the individual parameters. The mean corresponds to the parameter $\theta_A^*$. The covariance is represented by the diagonal precision matrix, which is the diagonal of the Fischer information matrix $F$. The matrix can be determined by means of the first-order derivative. It corresponds to the second derivative of the loss near a minimum. At the same time, the matrix is positive semi-definite. For each parameter $i$, the factor $\lambda$ defines the importance of the previous tasks. In this work, the TensorFlow implementation proposed by Kirkpatrick, Pascanu, et al. (2017) is integrated into the evaluation framework.

$$\mathcal{L}(\theta) = \mathcal{L}_B(\theta) + \sum_i \frac{\lambda}{2} F_i (\theta_i - \theta_{A,i}^*)^2 \tag{7.1}$$

Equation 7.1: EWC loss function (taken from Kirkpatrick, Pascanu, et al. 2017).

**Local Winner Takes All (LWTA)**   Srivastava, Masci, et al. (2013) introduce LWTA, which is classified as parameter isolation approach. The idea is derived from neuroscience and it is intended to mimic the competitive processes between neurons. This competitive process is realized by the LWTA output function (referred to as transfer function), which enables or disables a neuron's activation (or sets it to 0). This mechanism is implemented by introducing *blocks*, which in turn consist of artificial neurons. Thus, each block comprises $n$ neurons. Due to varying activations of neurons, sub-graphs are created. This state is illustrated in figure 7.3. Figure 7.3 represents blocks containing two neurons, whereas only one of them is active at a time (green). The activations of the gray neurons, however, are not passed on.

The "competition/interaction function" specifies how the neurons are de-/activated. In the work of Srivastava, Masci, et al. (2013), the "hard winner-take-all" function is specified. As shown in equation 7.2 $i$ serves as index for the block and $j$ for the neurons within the block. Accordingly, the parameters of an active neuron are adjusted or the activation signal is transmitted during the training process. LWTA can be considered as an extension of FC models allowing for the same architectures.

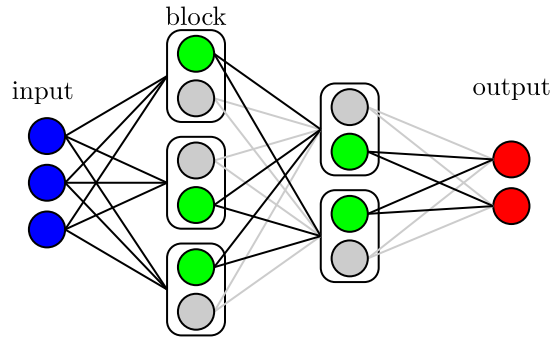

Figure 7.3: Exemplary LWTA block network with $n=2$ (compare Srivastava, Masci, et al. 2013).

$$y_i^j = \begin{cases} h_i^j & \text{if } h_i^j \geq h_i^k, \quad \forall k \in \{1, ..., n\} \\ 0 & \text{otherwise} \end{cases}$$

(7.2)

$$\text{where } h_i = f(\boldsymbol{w}_{ij}^\top \boldsymbol{x})$$

Equation 7.2: LWTA competition/interaction function described by Srivastava, Masci, et al. 2013.

**Dropout**   Hinton, Srivastava, et al. (2012b) suggest dropout as a solution to the "overfitting" problem in DNNs (see section 2.2.2.5). Dropout disables a neuron's activation (sets it to 0) based on a random factor. It is implemented as an output function (transfer function), which can be applied for each artificial neuron. For reasons of simplicity, dropout is often implemented as layer which does not contain any adjustable parameters. However, different dropout probabilities can be defined for each layer. One probability is usually defined for the input layer and another one for all other hidden layers.

As proposed by Goodfellow, Mirza, et al. (2013), dropout should mitigate the CF problem. Dropout is applied to FC-DNNs and CNNs, resulting in Dropout (D)-FC and D-CNN models. Generally speaking, dropout can be applied to all DNNs. EWC already contains the application of dropout and LWTA is an output function itself. For this reason, dropout is not integrated into these two models.

**Incremental Moment Matching (IMM)**   IMM by Lee, Kim, et al. (2017) aims at "resolving" the CF problem. It is similar to EWC and can be classified as an approach based on regularization. IMM describes the approximation of the posterior distribution of DNN parameters by means of a Gaussian distribution. Different approximation variants are proposed by Lee, Kim, et al. (2017): Mean-IMM and mode-IMM. "Mean-IMM averages the parameters of two networks in each layer, using mixing ratios $\alpha_k$ with $\sum_k^K \alpha_k = 1$" (Lee, Kim, et al. 2017). The loss function (stated in equation 7.3) is minimized based on the KL-divergence (a fundamental of Bayesian Neural Networks (BNNs)). The mean-IMM variant ignores the variances $\Sigma_{1:K}^*$.

$$\mu_{1:K}^*, \Sigma_{1:K}^* = \underset{\mu_{1:K}^*, \Sigma_{1:K}^*}{\arg\min} \sum_k^K \alpha_k KL(q_k || q_{1:K})$$

$$\mu_{1:K}^* = \sum_k^K \alpha_k \mu_k$$

(7.3)

$$\Sigma_{1:K}^* = \sum_k^K \alpha_k \big(\Sigma_k + (\mu_k - \mu_{1:K}^*)(\mu_k - \mu_{1:K}^*)^\top\big)$$

Equation 7.3: IMM loss function proposed by Lee, Kim, et al. (2017).

In contrast to the mean-IMM variant, the mode-IMM variant uses the covariance information. Similar to EWC, an approximation is utilized for the determination based on the Fisher information matrix $F$.

Without the approximation the determination of the posterior distribution would be impractical. The determination is represented in equation 7.4, where the inverse $F$ is used.

$$F_k = E\left[\frac{\partial}{\partial \mu_k} \ln p(\tilde{y}|x, \mu_k) \cdot \frac{\partial}{\partial \mu_k} \ln p(\tilde{y}|x, \mu_k)^\top\right] \tag{7.4}$$

Equation 7.4: Fisher information matrix for IMM specified by Lee, Kim, et al. 2017.

In addition to these two moment matching methods, various transfer techniques are presented by Lee, Kim, et al. (2017): Weight-transfer, L2-transfer, Drop-transfer and combinations. As stated by Lee, Kim, et al. (2017), only the weight-transfer technique is critical for the CL performance. It is thus applied as default in this work.

Weight-transfer defines the transitioning of a model's state from one task $\mu_{k-1}$ to another $\mu_k$. The transition of two solutions (model parameter configurations) is an interpolated point in between. The merging parameter $\alpha$ is varied for the evaluation of an IMM model. Although the IMM model contradicts the presented requirements from chapter 6, it is considered a state-of-the-art approach for CL applications. The approach, however, violates one of the defined real-world scenario constraints: Past data are required to adjust the merging parameter $\alpha$ and to determine an acceptable solution.

### 7.1.2   Hyper-Parameter Selection

Hyper-parameters' effects are crucial for ML model performance related to a specific problem. Therefore, a grid-search is performed to reduce the influence of problem-dependent parameters. Usually, various parameter combinations are explored and the best model (configuration) is selected based on an appropriate metric. The larger the range of parameter values and the smaller the steps, the better results can be expected. At the same time, the number of performed experiments increases with more hyper-parameters and varying values. A significant amount of computing resources is required to conduct a comprehensive investigation. For the implementation of this study, the environment presented in section 4.3 is utilized.

Not all conceivable parameter constellations can be examined, as an almost infinite number of them exists. The architecture of the DNNs, for example, belongs to these parameter constellations. Highly specialized network architectures are frequently used. They usually have been optimized for a single problem. The goal of this study is neither to design the best possible network architecture nor to determine the optimal parameters for a particular problem. The objective is rather to conduct an appropriate review of the CL performance of various CF avoidance models under real-world conditions. The hyper-parameter optimization is primarily executed to emphasize the validity of the results. This is how the independence of the chosen parameters is secured.

In the present work, only deep learning models based on DNNs are examined. The DNN architecture is one of the hyper-parameters, which has to be adapted to the corresponding problem. The architecture is composed of the number of layers $L$ and the artificial neurons $S$ in each layer. The variations of these two parameters are selected based on related work. $L \in \{2, 3\}$ and $S \in \{200, 400, 800\}$ are investigated as possible combinations.

The DNNs applied in this work consist of an input layer (Input), the optional application of Dropout (D), which is followed by the first hidden layer of Fully-Connected (FC) neurons. Again, layers are terminated with a ReLU output function. The last layer is terminated with a softmax (SM) (see section 2.2.1). For a three-layer architecture ($L = 3$), the structure is defined as follows: Input-FC1-D-ReLU-FC2-D-ReLU-FC3-SM.

CNNs are excluded from the architectural choice. Varying the number of possible model parameters would exceed the resulting number of experiments. Therefore, a predefined CNN architecture is selected which achieves reasonable results (proposed by Cireşan, Meier, et al. (2011)). In addition to the standard Fully-Connected (FC) layers, convolutional (C) and max-pooling (MP) layers are introduced. The structure is defined as follows (including the application of dropout (D)): Input-C1-MP-D-ReLU-C2-MP-D-ReLU-FC3-SM. Further hyper-parameters are set for two convolutional (C) layers $L = 2$ with 32 respectively 64 filters of the size $5 \times 5$ and a max-pooling layer of the size $2 \times 2$. The last layer consists of 1024 artificial neurons.

In addition to the model architecture, the learning rate $\epsilon$ is a crucial parameter. However, the learning rate is independently determined for different sub-tasks. For $T_1$ a selection is made from $\epsilon_{T_1} \in \{0.01, 0.001\}$. For all further re-training steps, $\epsilon_{T_c} \in \{0.001, 0.0001, 0.00001\}$ can be selected, where $c > 1$. The standard Momentum Optimizer is applied as an optimizer with the parameter $\mu = 0.99$ (see Sutskever, Martens, et al. 2013). The batch size is $\mathcal{B} = 100$ (for training and testing), and the number of training epochs is set to $\mathcal{E} = 10$ for all investigated models.

In addition to the general hyper-parameters, model-specific parameters have to be defined. In order to specify various parameters suggestions from related works are applied. For the LWTA model, the number of the LWTA blocks is set to 2 for all experiments (see Srivastava, Masci, et al. 2013). This is due to the initial investigation of SLTs with only two tasks. For dropout the probability is set to 0.5 for hidden layers and 0.2 for the input layer (see Goodfellow, Mirza, et al. 2013). To be precise, a probability of 0.5 is applied for CNNs in case of both, hidden and input layers. Kirkpatrick, Pascanu, et al. (2016) outline the importance factor $\lambda$ for EWC which is adjusted by a grid-search. At the same time, the value $\lambda = 1/\epsilon$ is specified in the publicly available code base. The approach to define $\lambda$ is used for EWC experiments, as the learning rate is varied anyway. For IMM, a vanilla Stochastic Gradient Descent (SGD) optimizer is utilized. In order to refine the posterior optimization value $\alpha$ the range between 0 and 1 with a step size of 0.01 is chosen.

### 7.1.3   Reproduction of Previous Results by the Prescient Evaluation Protocol

First of all, the results from related work will be reproduced in order to demonstrate the validity of the code base. Accordingly, the prescient evaluation protocol is utilized (see algorithm 6.2). The basis for a statement regarding the validity is a subsequent evaluation and selection of a step-wise parameter optimization. The main characteristic of the prescient protocol is that the best parameter configuration is determined in retrospective. However, the prescient evaluation protocol contradicts some real-world requirements. Therefore, the statements on obtained results cannot necessarily be transferred to all application-oriented CL scenarios. For reasons of simplicity, only one dataset is examined by means of the prescient protocol (MNIST, see section 6.1).

The results of the experiments with the prescient protocol are illustrated in table 7.1. Note that the maximum accuracy is provided depending on the corresponding SLT. Thus, a maximum of 50 % accuracy can be obtained for the investigation of the first sub-task $T_1$ for a $D_{5\text{-}5}$ SLT. The same applies for the second sub-task $T_2$ of $D_{5\text{-}5}$. The case is different for a $D_{9\text{-}1}$ SLT, where the ratio is 90 % for $T_1$ and 10 % for $T_2$. Therefore, the achievable accuracy after $T_1$ is 90 %. The rightmost column shows the test results for the permuted dataset (see definition of $D_{10\text{-}p10}$ section 6.2). The test results indicate that the permuted task is slightly affected by the CF effect. In fact, this is true for all investigated ML models. To conclude, even simple ML models such as Fully-Connected models demonstrate acceptable results.

At a first glance, the results for $D_{9\text{-}1}$ SLTs suggest that the CF effect can be controlled. The correspondance of the resulting values (about 90 %) to the maximum measured accuracy reveals that these results are obtained during the first task $T_1$. The real CL performance is evident in the results of the $D_{5\text{-}5}$ SLTs. An accuracy between 50 and 60 % means that only little knowledge is left from the first task $T_1$ or derived from $T_2$. The application of Dropout (D) does not seem to prevent the CF effect. The results for LWTA and EWC reveal a different situation. LWTA is significantly better on average than all models that do not address the CF effect. However, EWC achieves by far the best results for the prescient evaluation protocol. The superior performance of EWC for all investigated SLTs is evident. Although all tasks are performed on the same dataset, the results show a certain variance. This is due to the influence of the random initialization of the models, but also to different difficulties of the SLTs. Thus, the assignment of classes to individual tasks may be another influencing factor.

### 7.1.4   Realistic Evaluation Results

In this section, the results of the experiments with the realistic protocol are presented. This investigation protocol implements various requirements that arise in application-oriented CL scenarios. According to the protocol (represented in chapter 6), it is impossible to access future or past data after a sub-task has been processed. Therefore, a model has to be selected for the first sub-task $T_1$. After that step, a model's architecture is fixed and can no longer be changed for the following tasks.

Table 7.1: Results of the *prescient* evaluation protocol (accuracy in %).

| SLT / model | $D_{5\text{-}5}$ | | | | | | | | $D_{9\text{-}1}$ | | | $D_{10\text{-}p10}$ |
|---|---|---|---|---|---|---|---|---|---|---|---|---|
| | $D_{5\text{-}5a}$ | $D_{5\text{-}5b}$ | $D_{5\text{-}5c}$ | $D_{5\text{-}5d}$ | $D_{5\text{-}5e}$ | $D_{5\text{-}5f}$ | $D_{5\text{-}5g}$ | $D_{5\text{-}5h}$ | $D_{9\text{-}1a}$ | $D_{9\text{-}1a}$ | $D_{9\text{-}1c}$ | |
| FC | 69 | 63 | 58 | 65 | 61 | 58 | 61 | 69 | 87 | 87 | 86 | 97 |
| D-FC | 58 | 60 | 61 | 66 | 61 | 54 | 63 | 64 | 87 | 87 | 85 | 96 |
| CNN | 51 | 50 | 50 | 50 | 50 | 50 | 51 | 49 | 89 | 89 | 87 | 95 |
| D-CNN | 51 | 50 | 50 | 50 | 50 | 50 | 50 | 49 | 81 | 84 | 87 | 96 |
| LWTA | 66 | 68 | 64 | 73 | 71 | 62 | 68 | 71 | 88 | 91 | 91 | 97 |
| EWC | 92 | 92 | 91 | 93 | 94 | 94 | 89 | 93 | 100 | 100 | 100 | 100 |

The choice of the best possible parameter configuration can be determined on the basis of two criteria. First, the *best* experiment is identified by the maximum measured performance criterion. The best state of the model parameters has to be restored retroactively. However, the best criterion is usually very difficult to implement in applications consisting of multiple sub-tasks. Therefore, the best criterion is only applicable during the first optimization phase. For subsequent CL tasks, this possibility is no longer valid due to memory and computational capacities. If the duration of the training is crucial, the model has to include a procedure for early-stopping. Again, the implementation of the latter has to be possible without accessing data of past sub-tasks.

In order to conduct experiments, the code base of the investigated models is integrated into the developed evaluation framework. As part of this process, the code is adapted or extended by several interfaces. This includes, for example, the basic training- or test-step function. Based on these functions, a batch of data is presented to a model for training or testing. Parameters and sequences of function calls are performed by the framework, e.g., selecting a batch of data from a dataset for a given task. In order to obtain representative results, the different datasets are used for the evaluation purposes: MNIST, FashionMNIST, Fruits 360, MADBase, NotMNIST, Devanagari, CIFAR-10 and SVHN (see section 6.1). These datasets are divided into different types of CL tasks, which are referred to as SLTs. For the first investigation, a limited representative of SLTs is made, consisting of $D_{5\text{-}5}$ and $D_{9\text{-}1}$ tasks and the baseline task $D_{10}$: $D_{5\text{-}5a}$, $D_{5\text{-}5b}$, $D_{5\text{-}5c}$, $D_{5\text{-}5d}$, $D_{5\text{-}5e}$, $D_{5\text{-}5f}$, $D_{5\text{-}5g}$, $D_{5\text{-}5h}$, $D_{9\text{-}1a}$, $D_{9\text{-}1a}$, $D_{9\text{-}1c}$ and $D_{10\text{-}p10}$. Each of the presented SLTs consists of two CL sub-tasks ($T_1$ and $T_2$). Two learning rates are being varied for the first sub-task, and three for the second sub-task. 7 models are examined: FC, D-FC, CNN, D-CNN, LWTA, EWC and IMM. Two- and three-layer DNNs architectures are investigated with three different sizes to choose from in each layer. Approximately 100 000 parameter combinations are derived, which are performed and evaluated in the experimental setup.

In table 7.2, results are presented in an aggregated format which is due to the high number of measured values. The aggregated accuracy values (in %) are given, whereas the first value is the *best* and the second is the *last* measured value. The achieved values are color coded according to the scale given below. For the aggregation, the best experiment for each SLT ($m_{\vec{p}^*}$ see algorithm 6.1) is selected. The minimum measured accuracies from each SLT category indicate the final result. Thus, the statement should be valid with respect to the applicability of all SLT-variants (or difficulty levels). The color scheme applied to table 7.2 refers to the *best* result, even if it contradicts the application-oriented requirements. Dark or black background indicates a value below the limit of 50 % for $D_{5\text{-}5}$ (and permuted task) and respectively 90 % for $D_{9\text{-}1}$ SLTs. A bright or white background illustrates good CF avoidance capabilities and thus acceptable CL performances. It becomes obvious that most of the results are below the expected accuracy limit, despite the consideration of the *best* criterion.

The *last* criterion, second accuracy value in table 7.2, shows equivalent or significantly worse performance values. These values correspond to the last measured accuracy value after completion of a complete SLT training. It is evident that the permuted SLTs are characterized by a very high performance. Therefore, permuted tasks seem to be inappropriate for the evaluation of the CF effect. Moreover, all of the investigated models provide moderate results for all other tasks.

In order to present the experimental results in greater detail, the trends for some individual models are depicted with fewer aggregations. Each of the following illustrations summarizes all measurement points of a particular SLT type. The benefit is that the CL performance trends of a model can be visually inspected for multiple datasets at the same time.

Table 7.2: Overview of the results according to the realistic evaluation protocol (accuracy in %).

| Dataset | SLT | FC | D-FC | CNN | D-CNN | LWTA | EWC |
|---|---|---|---|---|---|---|---|
| CIFAR-10 | $D_{5\text{-}5}$ | 30/28 | 26/23 | 31/10 | 30/.18 | 31/30 | 32/20 |
| | $D_{9\text{-}1}$ | 45/10 | 37/10 | 45/10 | 48/10 | 45/10 | 36/08 |
| | $D_{10\text{-}p10}$ | 54/52 | 44/43 | 52/50 | 56/55 | 54/51 | 57/46 |
| Devanagari | $D_{5\text{-}5}$ | 49/42 | 46/26 | 49/45 | 49/11 | 11/10 | 40/23 |
| | $D_{9\text{-}1}$ | 86/10 | 84/09 | 88/10 | 89/09 | 86/09 | 88/09 |
| | $D_{10\text{-}p10}$ | 98/98 | 98/98 | 95/95 | 100/100 | 97/96 | 100/96 |
| EMNIST | $D_{5\text{-}5}$ | 50/48 | 50/48 | 50/48 | 50/48 | 50/48 | 36/08 |
| | $D_{9\text{-}1}$ | 88/09 | 88/09 | 89/09 | 89/09 | 88/09 | 92/51 |
| | $D_{10\text{-}p10}$ | 99/99 | 99/99 | 100/100 | 100/100 | 99/99 | 100/98 |
| Fashion-MNIST | $D_{5\text{-}5}$ | 46/45 | 46/44 | 47/45 | 46/46 | 46/46 | 55/47 |
| | $D_{9\text{-}1}$ | 78/10 | 77/10 | 81/10 | 81/10 | 78/10 | 85/50 |
| | $D_{10\text{-}p10}$ | 90/88 | 88/87 | 92/92 | 92/92 | 90/89 | 95/95 |
| Fruits 360 | $D_{5\text{-}5}$ | 32/14 | 46/13 | 14/09 | 14/09 | 28/11 | 34/03 |
| | $D_{9\text{-}1}$ | 34/09 | 38/21 | 14/09 | 23/09 | 38/09 | 55/13 |
| | $D_{10\text{-}p10}$ | 100/97 | 100/99 | 90/88 | 97/96 | 98/12 | 98/90 |
| MADBase | $D_{5\text{-}5}$ | 49/49 | 49/49 | 49/49 | 49/10 | 50/49 | 40/26 |
| | $D_{9\text{-}1}$ | 89/10 | 91/10 | 89/10 | 90/10 | 94/10 | 99/70 |
| | $D_{10\text{-}p10}$ | 99/99 | 99/99 | 99/99 | 99/99 | 99/98 | 100/99 |
| MNIST | $D_{5\text{-}5}$ | 49/48 | 49/47 | 48/11 | 48/15 | 10/09 | 50/31 |
| | $D_{9\text{-}1}$ | 88/10 | 88/10 | 88/10 | 88/10 | 87/10 | 99/71 |
| | $D_{10\text{-}p10}$ | 99/99 | 98/98 | 99/99 | 99/99 | 98/98 | 100/98 |
| NotMNIST | $D_{5\text{-}5}$ | 49/49 | 49/49 | 49/49 | 50/49 | 50/49 | 57/50 |
| | $D_{9\text{-}1}$ | 87/10 | 86/10 | 88/10 | 88/10 | 87/10 | 88/31 |
| | $D_{10\text{-}p10}$ | 97/97 | 97/97 | 98/98 | 98/98 | 97/97 | 99/94 |
| SVHN | $D_{5\text{-}5}$ | 30/22 | 28/08 | 20/08 | 20/08 | 40/20 | 28/16 |
| | $D_{9\text{-}1}$ | 60/07 | 35/07 | 67/07 | 58/07 | 61/07 | 26/10 |
| | $D_{10\text{-}p10}$ | 81/80 | 50/50 | 20/20 | 84/84 | 82/79 | 39/29 |

$D_{10\text{-}p10}$, $D_{5\text{-}5}$: 　50% ———— 75% ———— 100%

$D_{9\text{-}1}$: 　90% ———— 95% ———— 100%

**Dropout Experiments**　Figure 7.4 represents the trends of the CL performance for the application of dropout to a FC model. This figure illustrates the investigation of $D_{5\text{-}5}$ SLTs. The blue face on the left-hand side depicts the accuracy for the first task $T_1$ (epoch 0-10). On the right-hand side, the accuracy for task $T_2$ is shown as a green face (epoch 10-20). The red face corresponds to the accuracy values on the joined test dataset ($T_1 \cup T_2$). In addition, the maximum accuracy of the baseline experiment is depicted as a white bar on the right-hand side. The baseline serves as a reference value and shows the maximum achievable accuracy by a joint training ($D_{10}$). The order of the presented datasets is based on the baseline accuracy.

Figure 7.4 shows that on average half of the red area/face is below the green face for the D-FC model. Based on the visual representation of the realistic evaluation protocol, it is claimed that dropout does not mitigate the CF effect. As expected, ML models not addressing the CF effect show poor results in these CL scenarios. The same is true for the application of dropout to CNNs. The CNN trends are comparable to the presented FC models – the CF effect is also not mitigated.

**Permuted Experiments**　Figure 7.5 shows the results of $D_{10\text{-}p10}$ (permutation) experiments with the Fully-Connected (FC) model. FCs models represent standard DNNs in which the CF effect occurs. It is obvious that no green area/face (measurement values from $T_2$) is recognizable, as it is continuously overlaid by the combined test dataset $T_1 \cup T_2$. At the same time, the CL performance almost corresponds to the maximum measured value of the baseline experiment (white bar). Accordingly, the CF effect cannot be directly demonstrated for permuted SLTs. The same effect for permuted tasks in recognized in the work of Lee, Kim, et al. (2017) (referred to as "Shuffled MNIST"). Based on the obtained results, no more permuted tasks will be considered for future CF studies in this work. Likewise, other related works advocate against permuted tasks. Farquhar and Gal (2018) describe permuted tasks as unrealistic and as a simplification of CL tasks.

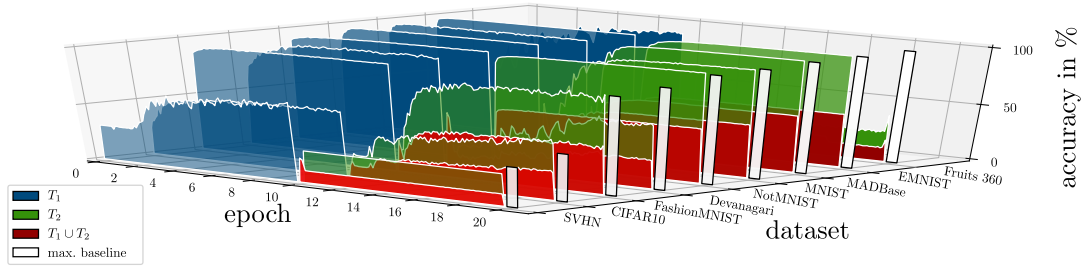

Figure 7.4: Best D-FC experiments for SLT $D_{5\text{-}5}$.

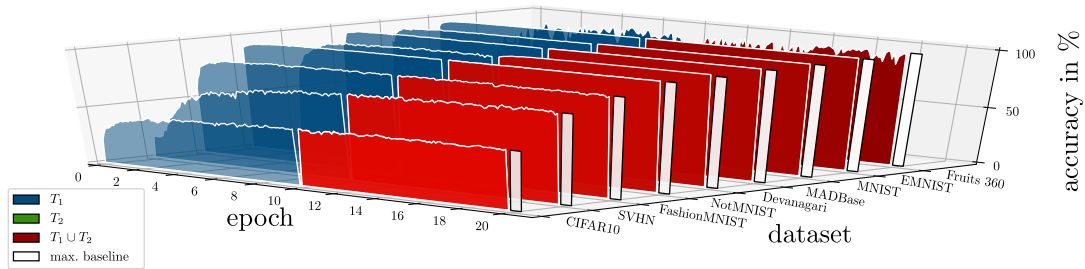

Figure 7.5: Best FC experiments for SLT $D_{10\text{-}p10}$.

**EWC Experiments**    Besides LWTA, Elastic Weight Consolidation (EWC) is a model which addresses the CF effect. Figure 7.6 illustrates the trend of EWC experiments for $D_{5\text{-}5}$ task. Again, the blue face represents the accuracy of the first sub-task $T_1$, and the green one represents the second sub-task $T_2$. However, the accuracy of the combined test data (red) is half as big of the second sub-task. Thus, EWC does not seem to completely mitigate the CF effect equally for all investigated SLT variants. The LWTA model, which is not depicted here, achieves equal or even worse CL performances. As depicted in table 7.1 (prescient protocol), EWC shows the best CL performance. But according to table 7.2, EWC seems to work only for simpler tasks such as the $D_{9\text{-}1}$ SLTs. In order to illustrate why the aggregated results show unfavorable CL performance trends, individual EWC experiments are presented in greater detail. The three visualized experiments in figure 7.7 represent an inconsistent trend. The figures have to be interpreted the same way as the previously presented ones. Again, the blue line represents the accuracy on the test data of $T_1$. The second sub-task $T_2$ (green line) starts at the 10th training epoch of the respective dataset. Red corresponds to the achieved CL performance on the joint test dataset. The dashed line represents the maximum of the baseline experiment.

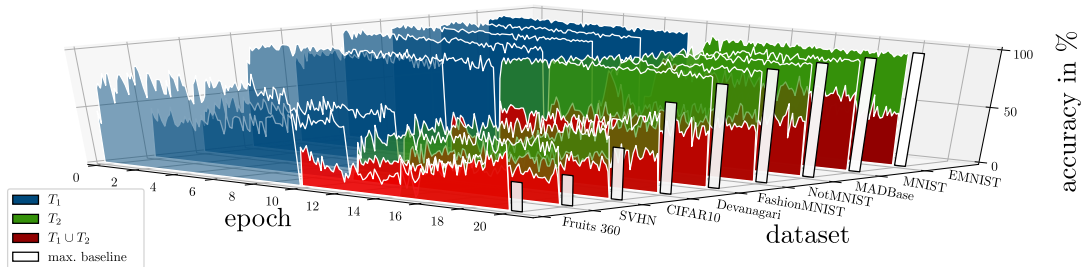

Figure 7.6: Best EWC experiments for SLT $D_{5\text{-}5}$.

EWC achieves acceptable results on the MNIST dataset in figure 7.7a, even though a linear forgetting effect can be detected. In the case of linear forgetting, the problem is to stop the training process at a training iteration with the best CL performance. According to the realistic evaluation protocol, no data from the previous task can be used for early-stopping.

A comparable but steeper forgetting is visible in figure 7.7b, which illustrates the EWC training on the EMNIST dataset. The trend shows that knowledge from $T_1$ is lost faster for the same $D_{9\text{-}1}$ SLT. It is more challenging to find an appropriate point for the determination of the re-training process. Figure 7.7c shows an even more drastic trend on the Devanagari dataset. The manifestation of the CF effect follows this pattern. The various forms of forgetting in figure 7.7b are due to the different problems/datasets. Even though each dataset is processed for the same number of epochs, the number of training iterations depends on the size of the dataset. Thus, the behavior of EWC cannot be determined beforehand as it is problem-dependent. Accordingly, the training process of EWC would have to be verified in a more fine-grained way, e.g., by adding more measurement points. At the same time, data from past tasks need to be accessed in order to stop the training process on a certain point. Especially the last requirement is very challenging, if not impossible to implement in real-world scenarios.

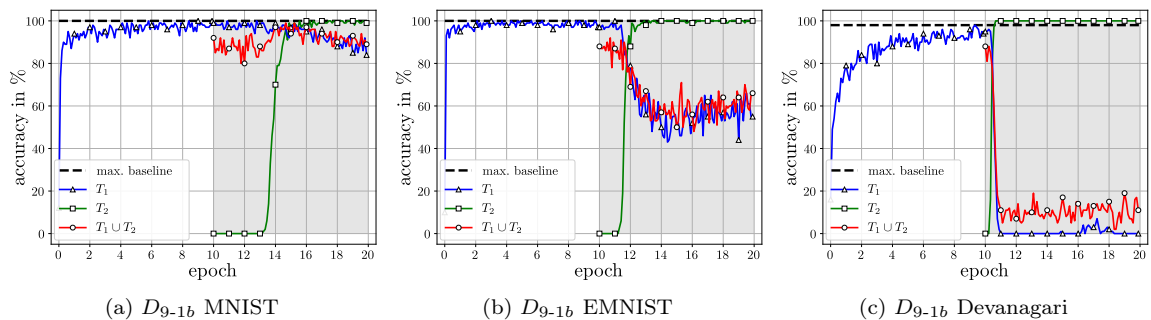

(a) $D_{9\text{-}1b}$ MNIST        (b) $D_{9\text{-}1b}$ EMNIST        (c) $D_{9\text{-}1b}$ Devanagari

Figure 7.7: EWC experiments for different datasets on $D_{9\text{-}1}$ SLTs.

In contrast to the previously illustrated $D_{9\text{-}1}$ tasks, $D_{5\text{-}5}$ SLTs are more challenging for EWC. It seems that with EWC it is more difficult to add more knowledge from the second sub-task $T_2$. This challenge becomes visible in the trend reflected in figure 7.8. In all three presented experiments (figures 7.8a to 7.8c), the final result does not meet the expected CL performance. Either the CF-effect occurs in its pure form, or forgetting starts too early in relation to the increase of knowledge. Therefore, EWC only seems effective under certain circumstances.

The behavior of EWC may again be due to the size of the datasets and the associated number of training iterations. It seems that the mechanism cannot protect the task-relevant weights long enough. Counteracting this deficit is a challenge in application-oriented scenarios.

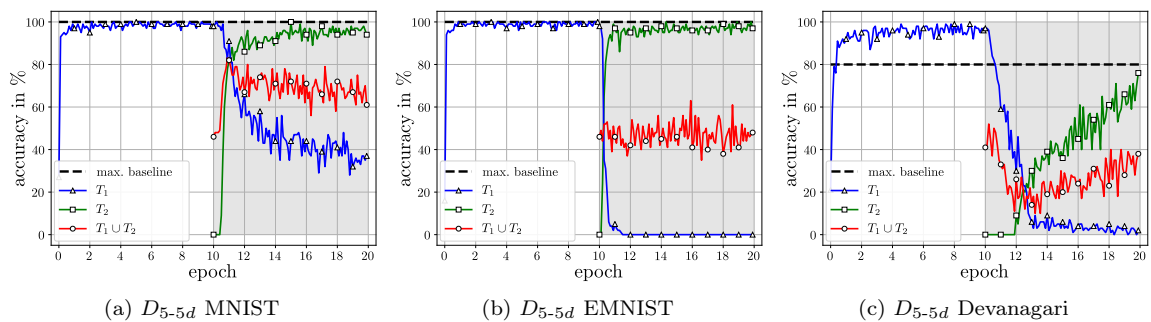

(a) $D_{5\text{-}5d}$ MNIST        (b) $D_{5\text{-}5d}$ EMNIST        (c) $D_{5\text{-}5d}$ Devanagari

Figure 7.8: EWC experiments for different datasets on $D_{5\text{-}5}$ SLTs.

**CL Performance Metric $\Omega_{all}$**  So far, the accuracy (CL metric) of the joint training dataset has been measured. In order to increase the significance of the measured accuracy values, the metric $\Omega_{all}$ proposed by Kemker, McClure, et al. (2017) is applied. $\Omega_{all}$ describes the ratio of the measured value to the baseline performance. Thus, conclusions should be drawn regardless of the difficulty of the dataset.

The *best* results from table 7.2 are combined with the baseline results. In table 7.3, both the baseline performances and the $\Omega_{all}$ values are given. The coloring is similar to table 7.2 (brighter is better) with emphasis on the $\Omega_{all}$ value. A pattern comparable to the previous one in table 7.2 is visible. Unfortunately, the CNN experiments (or the version with dropout Dropout (D)) show some anomalies related to the SVHN dataset. The reason is the poor baseline performance, which could be due to the simple architecture or possibly the difficulty of the dataset. Results with a too low baseline performance must therefore be ignored. A similar trend is apparent for the plain accuracy metric, whereas the statement regarding the CF effect remains. This is due to the fact that the baseline experiments generally provide very good results.

Table 7.3: Overview of the results according to the realistic evaluation protocol (metric: $\Omega_{all}$).

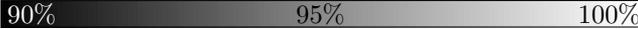

| Dataset | SLT | FC | D-FC | CNN | D-CNN | LWTA | EWC |
|---|---|---|---|---|---|---|---|
| CIFAR-10 | $D_{5\text{-}5}$ | 50/59 | 41/63 | 49/64 | 48/63 | 52/60 | 35/91 |
| | $D_{9\text{-}1}$ | 51/88 | 42/89 | 51/90 | 52/94 | 51/88 | 44/82 |
| | $D_{10\text{-}p10}$ | 53/102 | 43/101 | 54/97 | 50/112 | 52/103 | 51/112 |
| Devanagari | $D_{5\text{-}5}$ | 95/51 | 90/51 | 98/50 | 99/50 | 91/12 | 88/45 |
| | $D_{9\text{-}1}$ | 97/89 | 95/88 | 99/88 | 99/89 | 96/89 | 99/89 |
| | $D_{10\text{-}p10}$ | 97/101 | 97/100 | 12/811 | 99/100 | 96/101 | 100/100 |
| EMNIST | $D_{5\text{-}5}$ | 99/51 | 99/51 | 99/50 | 100/50 | 99/51 | 94/38 |
| | $D_{9\text{-}1}$ | 99/89 | 99/89 | 10/89 | 100/89 | 99/89 | 100/92 |
| | $D_{10\text{-}p10}$ | 99/100 | 99/100 | 10/100 | 100/100 | 99/100 | 100/100 |
| Fashion-MNIST | $D_{5\text{-}5}$ | 87/53 | 87/52 | 90/52 | 91/51 | 88/53 | 93/59 |
| | $D_{9\text{-}1}$ | 88/88 | 87/88 | 91/89 | 91/89 | 88/88 | 95/89 |
| | $D_{10\text{-}p10}$ | 89/100 | 88/100 | 91/101 | 92/100 | 89/101 | 95/100 |
| Fruits | $D_{5\text{-}5}$ | 51/62 | 96/48 | 78/17 | 88/16 | 100/28 | 63/54 |
| | $D_{9\text{-}1}$ | 52/66 | 100/38 | 79/18 | 99/23 | 99/39 | 91/60 |
| | $D_{10\text{-}p10}$ | 100/100 | 100/100 | 75/119 | 80/120 | 99/99 | 78/126 |
| MADBase | $D_{5\text{-}5}$ | 99/50 | 98/50 | 99/50 | 99/50 | 98/50 | 97/41 |
| | $D_{9\text{-}1}$ | 99/90 | 99/92 | 99/90 | 99/90 | 98/95 | 100/99 |
| | $D_{10\text{-}p10}$ | 99/100 | 99/100 | 99/100 | 99/100 | 98/100 | 100/100 |
| MNIST | $D_{5\text{-}5}$ | 98/51 | 97/51 | 99/49 | 99/49 | 95/10 | 94/53 |
| | $D_{9\text{-}1}$ | 98/90 | 98/90 | 99/88 | 99/88 | 98/88 | 100/99 |
| | $D_{10\text{-}p10}$ | 99/100 | 98/100 | 99/10 | 99/100 | 98/100 | 100/100 |
| NotMNIST | $D_{5\text{-}5}$ | 96/51 | 96/51 | 97/51 | 97/51 | 96/51 | 99/58 |
| | $D_{9\text{-}1}$ | 97/90 | 96/90 | 98/90 | 98/90 | 97/90 | 100/88 |
| | $D_{10\text{-}p10}$ | 97/100 | 97/100 | 98/100 | 98/100 | 97/100 | 99/10 |
| SVHN | $D_{5\text{-}5}$ | 74/40 | 37/76 | 20/99 | 20/100 | 77/52 | 30/93 |
| | $D_{9\text{-}1}$ | 75/79 | 41/86 | 20/331 | 20/289 | 77/79 | 30/87 |
| | $D_{10\text{-}p10}$ | 80/102 | 52/98 | 20/100 | 20/425 | 80/102 | 44/89 |

| 90% | 95% | 100% |
|---|---|---|

**IMM Experiments** Another model investigated in this work is Incremental Moment Matching (IMM) (see section 7.1.1). The IMM model is proposed by Lee, Kim, et al. (2017) and differs from the other models in two aspects. Firstly, it supposedly provides a better CL performance with respect to the CF effect. Secondly, it cannot be reconciled with the real-world requirements, which causes a problem. As a consequence, the results for IMM are listed separately and not in direct comparison with the models mentioned above.

The IMM model offers a variety of different techniques and additions that affect the CL performance. These techniques include mean-IMM and mode-IMM. The mean-IMM averages the DNN parameters layer by layer according to a given ratio $\alpha$. This is realized by the KL-divergence in the objective function. On the contrary, Mode-IMM adds the covariance information.

Three transfer techniques are proposed by Lee, Kim, et al. (2017): Weight-transfer, L2-transfer and drop-transfer. For this study, the weight-transfer technique is used due to the recommendation in the study. "[...] the use of weight-transfer was critical to the continual learning performance. For this reason, the experiments [...] use weight-transfer technique as default" (Lee, Kim, et al. 2017). IMM is convincing for more difficult SLTs such as $D_{5\text{-}5}$ tasks. This conclusion can be drawn based on the illustrated results in table 7.4.The experiments for the permuted tasks are omitted, as stated earlier in this section.

The results of the performed IMM experiments are presented in an aggregated form in table 7.4. First of all, the best experiment for a SLT ($q_{i^*}$) and dataset is determined. The best experiment constitutes the foundation for the aggregation of the different SLTs. The aggregation is then conducted based on the minimum measured CL performance for a group of SLTs. For $D_{5\text{-}5}$ tasks more than 50 % accuracy is desirable in order to evaluate the mitigation of the CF effect. The same applies to $D_{9\text{-}1}$ tasks where more than 90 % need to be achieved.

The obtained results show that in some cases knowledge can be preserved. However, the accuracy for the MNIST dataset does not correspond to the baseline experiment ($> 98\%$). At the same time, the demonstrated results proposed by Lee, Kim, et al. (2017) reveal a better CL performance. The inferior performance results in this work may be due to the model selection procedure according to the realistic protocol. In addition, the lack of an extensive fine-tuning process also affects the final result.

Table 7.4: Summary of SLTs for IMM.

| Mode \ SLT \ Dataset | CIFAR-10 | | Devanagari | | Fashion-MNIST | | MADBase | | MNIST | | SVHN | |
|---|---|---|---|---|---|---|---|---|---|---|---|---|
| | $D_{5\text{-}5}$ | $D_{9\text{-}1}$ | $D_{5\text{-}5}$ | $D_{9\text{-}1}$ | $D_{5\text{-}5}$ | $D_{9\text{-}1}$ | $D_{5\text{-}5}$ | $D_{9\text{-}1}$ | $D_{5\text{-}5}$ | $D_{9\text{-}1}$ | $D_{5\text{-}5}$ | $D_{9\text{-}1}$ |
| mode | 31 | 43 | 73 | 85 | 70 | 78 | 91 | 91 | 84 | 87 | 56 | 60 |
| mean | 30 | 43 | 67 | 85 | 62 | 78 | 82 | 92 | 82 | 88 | 50 | 59 |

The problem of IMM is the subsequent adjustment of the parameter $\alpha$, which is used to merge two model parameter configurations. The merging process with respect to the resulting accuracy is illustrated in figure 7.9. On the left-hand side, the accuracy trend of training a DNN for the first task is shown. The blue line represents the test accuracy on $T_1^{test}$ and the green line on $T_2^{test}$ for 10 epochs each. Additionally, the maximum of the baseline experiment with IMM is provided (dashed line). The right section of figure 7.9 represents the $\alpha$ tuning process. The two DNNs (one of the first and one of the second task) are "merged" in different proportions based on the $\alpha$ parameter. The test range for the $\alpha$ values lies between 0 and 1 with a step size of 0.01 (resulting in 100 tested $\alpha$ values). The two lines on the right-hand side illustrate the test accuracy on the combined test dataset $T_1 \cup T_2$. As depicted in figure 7.9, the mode variant (orange) provides slightly better results than the mean variant (red). As expected, the best $\alpha$ value is not exactly 0.5 for a $D_{5\text{-}5}$ SLT. This is clearly illustrated by the two values for mean $\alpha = 0.60$ and mode $\alpha = 0.39$. Accordingly, an optimal $\alpha$ factor cannot necessarily be calculated with the ratio of the number of classes within the sub-tasks.

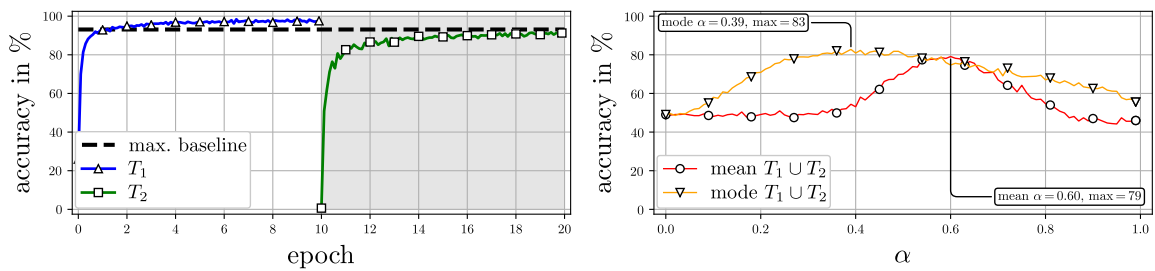

Figure 7.9: Best IMM experiment on SLT $D_{5\text{-}5f}$ for the Devanagari dataset.

As indicated before, the definition of the merging parameter $\alpha$ is not trivial for IMM. This is evident in figure 7.10, which can be interpreted the same way as figure 7.9. The left part of figure 7.10 shows the independent training process of two DNNs for $T_1$ and $T_2$. The white bar in the middle indicates the maximum of the baseline experiment. On the right-hand side, the obtained test accuracies of the varying $\alpha$ parameter are illustrated. The trends for different datasets of the mode-IMM and mean-IMM indicate that $\alpha$ has to be determined by cross-validation.

The tuning process requires data from the current as well as from the previous sub-tasks (as stated by Lee, Kim, et al. (2017)). However, the tuning contradicts the CL requirements for real-world applications. A further disadvantage of this optimization strategy is that the very computationally costly Fischer Information Matrix (FIM) has to be calculated repeatedly for each $\alpha$ value. Due to the increased cost, the optimization was not performed for all datasets. In order to determine an adequate $\alpha$ value, several hours ($> 4\,\text{h}$ for 100 different $\alpha$ values) and a large amount of memory would be required ($> 8\,\text{GB}$ of RAM). The memory issue is complicated by the use of more complex DNNs, since the determination of the FIM depends on the DNN's architecture. On the one hand, the effort of FIM determination depends on the implementation. By approximating the FIM, the computational effort can be reduced (an approximation is proposed by Gepperth and Wiech (2019)). On the other hand, the use of past data still contradicts the real-world requirements. This violation limits the applicability of IMM for various CL scenarios.

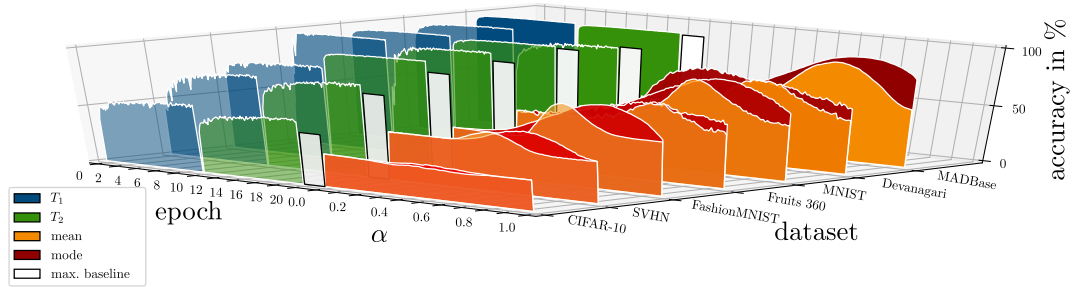

Figure 7.10: Best IMM experiments for SLT $D_{5\text{-}5b}$ for different datasets.

Figure 7.11 represents the IMM results of a $D_{9\text{-}1}$ SLT similarly to the previous figures. Comparable findings can be derived from this kind of CL tasks, although the ratio of the number of classes from $T_1$ to $T_2$ is 9:1, $\alpha = 0.1$ is not always the best merging value. Moreover, mode-IMM is not necessarily superior than mean-IMM (see figures 7.10 and 7.11). In order to select the best possible mode for a particular problem, both options need to be evaluated. In general, this exceeds the available capacities for training and evaluation.

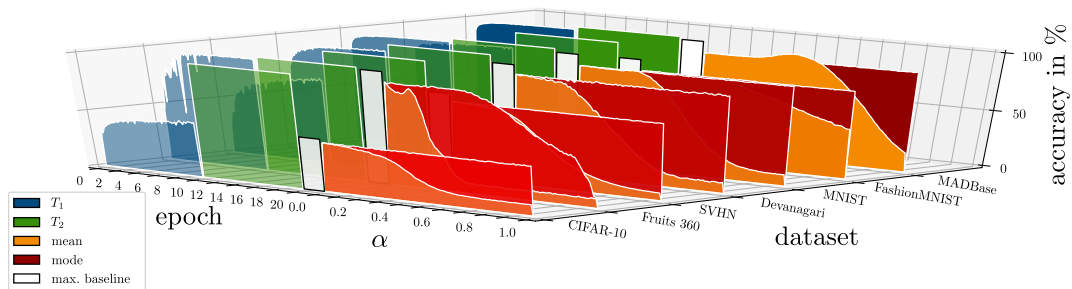

Figure 7.11: Best IMM experiments for SLT $D_{9\text{-}1b}$ for different datasets.

## 7.2 Discussion

This discussion focuses on the derived findings, as well as the impact of the real-world evaluation protocol. In particular, the findings that contradict related work are outlined. Resolving the CF problem constitutes the main subject of the discussion along with the respective conditions.

**Training Iterations and Measuring Points**    The influence of the number of training iterations on CL performance is the first aspect that is discussed. Both the number and positioning of the measuring points are crucial for the CL performance. Additionally, the mechanism of early-stopping is related to these parameters. In real-world CL scenarios, it is often challenging to determine the optimal parameters with regard to stopping the (re-)training process. This may be due to two factors.

In order to determine an accurate model parameter configuration, test data from previous tasks need to be stored and used. An appropriate model configuration can be determined with past test data by investigating the CL performance. Regular measurements need to be executed in order to stop the re-training process before too much previous knowledge is lost. At the same time, the stored test samples could be used for the re-training process, which is equivalent to a joint training. This method may be possible in some scenarios, but it contradicts the idea of CL with respect to a potentially infinite number of tasks. Consequently, it is impossible to store test samples for each task while respecting the memory constraints. Furthermore, different states of model parameter configurations (referred to as checkpoints) need to be stored to return to an optimal model parameter configuration. This, in turn, can be computationally and memory intensive depending on the underlying model.

Determining an acceptable stopping point during training is challenging. In most cases, the quality measurement of a model is based on test data. Assuming that test data is accessible, multiple questions arise: *How much test data are representative? How often the model needs to be tested?* If a test step is carried out after each training iteration, it is possible to identify an acceptable model parameter configuration. The disadvantage of this method is that it slows down training, especially for representative or larger test datasets. If a model is very sensitive in this respect, a tremendous amount of computational effort has to be expended. Depending on the selected CL scenario, the present method is impossible due to few available resources. Therefore, models must be able to determine a good model parameter configuration without much effort, or – even better – they should be robust to this effect. Ensuring that extended (re-)training does not reduce the quality of a model is a criterion that should be investigated in further studies.

**Evaluation Selection Criterion**  A major challenge is the choice of a measure when deciding on the best parameter configuration. The used metric is related to this aspect. In this context, the real-world requirements for the models presented earlier become increasingly important. The *best* and *last* criteria used in this protocol differ significantly. Depending on the CL scenario, it is decisive to what extent an evaluation based on the *best* measured value makes sense at all. Just because a model has (or can) achieve maximum performance at a random training/test point, it does not necessarily represent the optimum configuration for future tasks. This is evident in the precient evaluation protocol. Ultimately, the *last* measured value should be decisive. Otherwise, there is no guarantee with regard to quality.

The *best* model parameter configuration is related to early-stopping, the guaranteed quality protection in case of a too long training or the provision of return points (checkpoints) in case of a detected accuracy loss. The choice of a metric is relevant in this context. In the present work, the simple accuracy (see equation 6.1) is used in most cases. However, this metric is only applicable if the dataset is appropriate and thus almost balanced. If a dataset is unbalanced, or if the objective of the model is different (e.g., safety in the classification of cancer), another metric has to be used. The results of the baseline experiment are related to the problem scenario. The metric presented by Kemker, McClure, et al. (2017) includes the baseline results and thus eliminates the difficulty of the comparison of different datasets.

**Datasets**  The next aspect that is discussed concerns the used datasets (see section 6.1). In general, the utilized datasets are well established in the ML community and the respective studies. Even though a dataset like MNIST is considered to be outdated and overly simplistic, several conclusions can be derived. As MNIST is very simple, it is expected to achieve high CL performances. The CL performance is therefore expected to be located in a similar range when compared to the baseline (which is not a CL task). The case is different for more complex datasets, such as SVHN or CIFAR-10. For more complex datasets, the baseline is significantly lower, which is not clearly interpretable in the context of CL.

Moreover, it is still difficult to assess the used datasets in terms of complexity, size or number of features. Significantly larger datasets with more features are easy to find. However, more complex datasets can only be investigated with further adjustments regarding the hyper-parameters and the used model architecture. In order to draw adequate conclusions in the context of the CL paradigm, an investigation of these datasets would require a more significant investment of time and resources.

However, it is essential to take into account that the investigated image datasets are in contrast to a real-world dataset as shown in chapter 5. Real-world constraints often are manifested by their very complex influence on the data, which raises new challenges for CL models. This complication affects the desired objective within a CL scenario, e.g., obsolete knowledge needs to be specifically forgotten. Nevertheless, the applied datasets allow for conclusions regarding the absence of the CF effect.

**Sequential Learning Tasks (SLTs)**   Due to the close relation to the applied datasets, the respective SLTs are discussed in the following. An open question is how representative SLTs are compared to real-world CL problems. In fact, CL tasks with hard task boundaries, clearly disjoint classes and uniformly distributed data can be considered unusual for real-world scenarios. The application of the realistic evaluation protocol including simple SLTs results in limited CL performance for the examined models. Excellent results would, however, reinforce the need for further research into even more realistic investigation protocols. More realistic evaluation protocols include, for example, the recognition and reaction to the addition of new knowledge/classes in different changing data distributions (see section 2.3.1). Furthermore, it is important to investigate more complex task constellations, such as few-shot learning, regression problems, or selective forgetting scenarios. Assuming that an ML model can satisfy all requirements for arbitrary CL scenarios is still utopian. Moreover, the circumstances of a well-working procedure requires further specification.

**Hyper-Parameter Optimization**   Hyper-parameters can be a crucial factor for ML problems. At the same time, certain parameters can be problem-dependent. These kinds of parameters must be determined by a trial-and-error procedure. Adjusting many different values for several hyper-parameters leads to a large number of possible combinations. As a result, an extremely large number of experiments has to be performed and evaluated. In order to minimize this scope and cost, the conducted grid-search constitutes a compromise for the examination of multiple datasets. In general, more values could be explored for each problem to achieve better results. A more extensive investigation could include larger and more complex model architectures.

Despite the attempt to conduct a comprehensive study, it can be assumed that the trend will be similar for a large-scale investigation. It is assumed that better results can be expected, while the general conclusions remain the same for larger studies. The basic problem of parameter optimization is the choice of the problem-related parameter values. This concerns, for example, the model architecture where the hypothesis that more is always better prevails. Unfortunately, this does not apply to other hyper-parameters. Therefore, the question arises whether the selected values of a particular parameter are appropriate or not. CL specific models should have as few problem-dependent parameters as possible requiring adjustments. The same applies to the strategy of selecting parameter values, e.g., the more the better is a good option. This should be considered as an important property of a CF avoidance model favorable for realistic CL scenarios.

**CF Avoidance Models**   The choice of investigated CF avoidance models is the last aspect that is discussed. The main focus is on the selection of the examined models. In particular, the question arises why no further or other models were examined. The selection is due to two reasons. On the one hand, the required computational effort per model increases. On the other hand, many approaches and optimization methods inherently contradict the application-oriented requirements. The latter is the case, for example, in the examination of the IMM model.

Furthermore, the availability of a code base is fundamental for the conduction of experiments. Without a publicly available code base, researchers have to develop their own implementation based on the publications. This causes several disadvantages. Due to a lack of details or omitted critical optimization steps, conclusions regarding the CL performance can only be vague. One model which provides very good CL results is Hard Attention to the Task (HAT) proposed by Serra, Suris, et al. (2018). The code of HAT is publicly available. As the name of the model indicates, it refers to a task. What is not necessarily clear, however, is that for each sample the task information has to be given. This means that for both, training and evaluation purposes, the ID of the original task must be specified for a sample. This corresponds to the use of a training- and test time oracle.

Suppose that HAT is investigated with an SLT consisting of 10 individual sub-tasks. Applying this to the datasets used in this work would mean that each CL task represents a single class. Accordingly, the task ID for each sample would automatically be provided while representing the corresponding label. This would already correspond to a perfect solution. A code snippet specifying the signature of the evaluation function is given in listing 7.1. The `eval` function (as well as the training function) specifies the parameter `t` which expect the task ID of a sample. Thus, a perfect oracle would be available for the kind of CL tasks. However, the application of oracles contradicts the requirements of many real-world scenarios. This limitation causes the very difficult application of the approach in different CL scenarios so that it is omitted in this study. The general problem is that these severe constraints only become obvious after an intensive study of the code base.

```
160          def eval(self,t,x,y):
```

Listing 7.1: Code snippet from Serra, Suris, et al. 2018 file: `src/approaches/hat.py`

Omitting the investigation of CF avoidance models is due to yet another requirement. As described earlier, storing (hold-out) samples for training or testing contradicts memory constraints. Certain models require no past data in order to reduce the CF effect. Other ML models use the samples only to determine the optimal model parameter configuration, as it is the case for IMM. It is, however, a problem to determine how many and which samples need to be stored and for how long. Different approaches, e.g., Gradient Episodic Memory (GEM), address this very problem. A reduced number of samples is stored and reused for a joint re-training. In fact, replay methods are a common approach to circumvent the CF effect. Nevertheless, the amount of available replay memory is limited due to the requirements. If the CL scenario consists of an infinite number of tasks, this workaround is limited.

## 7.3   Conclusion

The following insights are revealed by investigating the real-world evaluations protocol. First of all, none of the investigated CF avoidance models can mitigate the CF effect to an acceptable extent under more real-world conditions. Several standard image datasets were used for the study. They consist of 10 classes that were divided into different Sequential Learning Tasks (SLTs). Each task of an SLTs consists of disjoint classes from the dataset. The protocol applied in this study differs from others with regard to its strong focus on application-oriented requirements. Due to the requirements it is impossible, for example, to retroactively set parameters after the completion of the last CL task, e.g., the model architecture. Likewise, the access to future data is prohibited. The implementation of these requirements by the investigation protocol leads to different results than those presented in related work (see chapter 3).

Considering the individual experiments of the CF avoidance models leads to the conclusion that it is possible to successfully mitigate the CF effect with EWC (under certain conditions). This is true in case the knowledge from one additional class has to be added. Unfortunately, this statement does not apply to all investigated problems. It might be due to the fact that EWC is sensitive to the number of training iterations or strongly depends on parameters, e.g., the importance parameter $\lambda$. For more complex tasks, e.g., with five new classes, the procedure seems to have more difficulties with regard to the protection of existing knowledge. The IMM model indicates a better performance, but it contradicts the requirements of real-world scenarios where data from the past task are needed. It even becomes more difficult as soon as a large number of tasks occur one after the other. For each task, more and more memory needs to be allocated in order to be able to represent the full data distribution.

In terms of the application-oriented requirements, the results imply that none of the investigated models can suppress the CF effect to a satisfactory extent. Thus, they are not applicable under the defined requirements for CL scenarios, although the evaluation protocol does not even map all conceivable requirements of CL scenarios (e.g., variants of changing data distributions). As a consequence, the question arises whether it is adequate to refer to the CF-effect as "controlled" or "solved" in the context of different CL scenarios. This question can only be answered accurately if the circumstances of a model's examination are revealed and published. To conclude, it is crucial to fully represent the circumstances in detail.

# 8.   Novel Deep Learning Model: Deep Convolutional Gaussian Mixture Models

## Chapter Contents

In this chapter, a novel deep learning approach referred to as Deep Convolutional Gaussian Mixture Models (DCGMMs) is presented. The contents of this chapter are published in Gepperth and Pfülb (2021) and Gepperth and Pfülb (2020). The *machine learning* (ML) model proposed in this work is based on Gaussian Mixture Models (GMMs), which are well-known and investigated. As GMMs are not deep, using them as universal function approximators becomes increasingly challenging. Furthermore, conventional training methods such as the Expectation-Maximization (EM) algorithm are unsuitable for *continual learning* (CL) scenarios. EM requires a data-driven initialization of the underlining GMMs, and its use in streaming scenarios is impossible without extensions. However, some properties of GMMs make them attractive for CL scenarios. GMMs are less (or at least differently) affected by the *catastrophic forgetting* (CF) effect. In contrast to Deep Neural Networks (DNNs), GMMs are unsupervised ML models (see section 2.2). In order to use these favorable properties for CL scenarios, the challenges mentioned above have to be resolved. Thus, this chapter presents the resulting adaptations and evaluates the validity of the novel approach. The CL properties of the DCGMMs are outlined in detail in the next chapter.

**Contributions**   The contribution of this chapter is a novel deep learning approach based on GMMs, which is referred to as DCGMM in the present work. Accordingly, an Stochastic Gradient Descent (SGD)-based GMMs training method is presented, which supports the use in streaming scenarios. Additionally, the transformation into a deep learning method by means of a stacking of GMMs is proposed. The evaluation of DCGMM results in similar or better performances compared to known standard methods. Moreover, the model offers various functionalities that can be used as a basis for further research.

In particular, this chapter addresses research question RQ3 and describes how a novel deep learning model can look like that is not directly subject to the CF effect. Thus, this chapter is not set in the context of the CL paradigm, but rather serves as a basis for further investigations.

**Structure** First of all, GMMs are introduced along with the associated fundamentals (see section 8.1). In the following section, the related work in the context of GMMs is presented and discussed. In the third section, a novel method for GMM training by SGD is presented (see section 8.3). Moreover, an annealing scheme is introduced, which enables the use in streaming scenarios. Section 8.4 outlines how the non-deep unsupervised learning procedure can be transformed into a deep learning model. In order to prove the validity of the new approach, a comparison to other training methods is performed in chapter 9. The chapter concludes with a discussion related to the new model.

## 8.1   Fundamentals of Gaussian Mixture Models

Gaussian Mixture Models (GMMs) are probabilistic models based on multiple Gaussian distributions. In the following, the basic elements of GMMs are introduced. The fundamentals include an overview of the Gaussian distribution and the multivariate Gaussian distribution. In addition, the EM algorithm is presented as the conventional training method for GMMs.

### 8.1.1   The Gaussian Distribution

One of the most well-known distributions is the *Gaussian* distribution (referred to as *univariate* (one-dimensional) *normal* distribution according to Bishop (1995)). Equation 8.1 shows the *probability density function* (PDF) of the Gaussian distribution. For any real number $x$, it follows that $\mathcal{N}(x|\mu, \sigma) > 0$. The value range of the function is defined by $f : \mathbb{R} \to \mathbb{R}$. The first part of equation 8.1 normalizes the resulting values and, thus, represents probabilities. The probability for a sample $x$ depends on the mean $\mu$ and the standard deviation $\sigma$.

$$\mathcal{N}(x|\mu, \sigma) := \frac{1}{\sqrt{2\pi\sigma^2}} \exp\left(-\frac{1}{2\sigma^2}(x - \mu)^2\right) \tag{8.1}$$

Equation 8.1: Probability density function of the Gaussian distribution.

Exemplary PDFs for different parameters of the Gaussian distribution are depicted in figure 8.1. The green and blue curves have the same mean $\mu = 0$. The variance $\sigma$ affects the maximum and defines the spread. The mean determines the shift on the $x$ axis, as represented by the red line ($\mu = 3$ and $\sigma = 2$). By using a PDF, the probability of a sample can be determined, e.g., for $x = 4$. If inserted into the equation 8.1, the probability of $x = 4$ for $\mathcal{N}(\mu = 3, \sigma = 2)$ is $\approx 0.2196$.

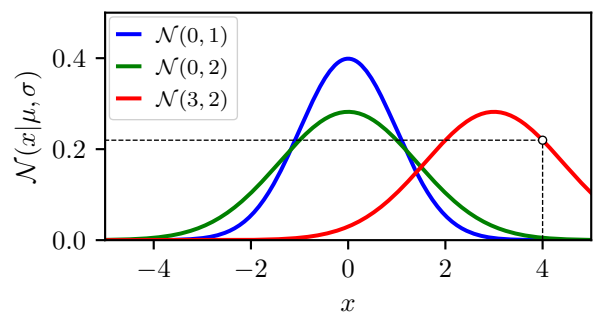

Figure 8.1: Plots of exemplary Gaussian distributions.

The parameters of a Gaussian distribution $\mu$ and $\sigma$ can be estimated from data. The expectation value $\mathbb{E}$ and its variance can be used to calculate the parameters of the PDF. The mean $\mu$ can be estimated by $\mu = \mathbb{E}(\mathbb{X}) = \frac{1}{n}\sum_{i=1}^{n} x_i$. The variance is dependent on the mean and is estimated by $\sigma = \text{Var}(\mathbb{X}) = \mathbb{E}[(\mathbb{X} - \mu)^2]$.

**Probability and Likelihood**   The difference between probability and likelihood has to be defined for this work. *Probabilities* generally describe the occurrence of a certain result or event. In contrast to that, the term *likelihood* is used if a hypothesis is included. A hypothesis can be considered as a model's parameter $\theta$. Thus, with the help of the parameters, the determination of an event's probability can be influenced. In the context of parameterized models, probability is referred to as likelihood (Bishop 1995). The present work follows the same convention.

**Log-Likelihood**   The log-likelihood is obtained as the natural logarithm of the likelihood $\mathcal{L}(x) = \log p(x|\mu, \sigma)$. Resulting from the application of the logarithm, the following statement is valid: Since the logarithm is strictly monotonically increasing, a minimum of the log-likelihood function is a minimum of the likelihood function. The same is true for the maximum.

Applying the logarithm to the likelihood function has two advantages. First, the logarithm makes the subsequent determination of multiple log-likelihood values additive. Second, it becomes easier to determine the (partial-) derivative as stated, for example, in equation 8.2. Thus, gradient-based optimization is computationally simpler, and thus more efficient to implement. Taking into account the precision of floating points, e.g., 32-bit floats, the numerical underflows are bypassed.

$$\mathcal{L}(x) = \log p(x|\mu, \sigma) = -\frac{1}{2}\log(2\pi) - \log(\sigma) - \frac{(x-\mu)^2}{2\sigma^2}$$
$$\frac{\partial \mathcal{L}(x)}{\partial \mu} = \frac{(x-\mu)^2}{\sigma^2} \qquad\qquad (8.2)$$
$$\frac{\partial \mathcal{L}(x)}{\partial \sigma} = -\frac{1}{\sigma} + \frac{(x-\mu)^2}{\sigma^3}$$

Equation 8.2: Partial derivative of the log-likelihood function for a normal distribution.

In the context of the likelihood function, the *maximum likelihood* is usually of interest. The maximum likelihood defines the highest outcome of the likelihood function $p(x|\theta)$. Accordingly, the parameters $\theta$ of a model need to be adjusted. The parameters for a Gaussian distribution are $\theta = \{\mu, \sigma\}$, as detailed in equation 8.3. The likelihood function is used for optimization by applying the negative logarithm of the likelihood function as a loss function $\mathcal{L}$ (Bishop 1995).

$$\mathcal{L}(\boldsymbol{x}|\theta) = -\frac{1}{2}\log(2\pi) - \log(\sigma) - \frac{(x-\mu)^2}{2\sigma^2} \qquad\qquad (8.3)$$

Equation 8.3: Log-likelihood function for a normal distribution.

### 8.1.2   Multivariate Gaussian Distribution

Likewise, multidimensional data $\boldsymbol{x}$ can be modeled by multivariate Gaussian distributions as shown in equation 8.4. The means are represented by a $D$-dimensional vector $\boldsymbol{\mu}$. The variances are represented by a $D \times D$ positive-definite covariance matrix $\boldsymbol{\Sigma}$. $|\boldsymbol{\Sigma}|$ corresponds to the determinant of $\boldsymbol{\Sigma}$ and $\boldsymbol{\Sigma}^{-1}$ is the inverse. Due to the positive-definiteness of the matrix, its inverse exists.

An example of a multivariate Gaussian distribution is indicated in figure 8.2. For a two-dimensional space, the mean $\boldsymbol{x}$ is a 2D vector, whereas the covariance $\Sigma$ is represented by a $2 \times 2$ matrix. The probability for samples is higher in the center than on the edges of the ellipses (see the color bar in the

$$\mathcal{N}(\boldsymbol{x}|\boldsymbol{\mu}, \boldsymbol{\Sigma}) = \frac{1}{(2\pi)^{\frac{D}{2}}} \frac{1}{\sqrt{|\boldsymbol{\Sigma}|}} \exp\left\{-\frac{1}{2}(\boldsymbol{x} - \boldsymbol{\mu})^{\top} \boldsymbol{\Sigma}^{-1}(\boldsymbol{x} - \boldsymbol{\mu})\right\} \tag{8.4}$$

Equation 8.4: Multivariate Gaussian distribution for a vector $\boldsymbol{x}$.

lower right corner). The rotation indicated by the coordinate system is explained by the covariance matrix $\Sigma$. The Gaussian curves at the bottom and right-hand side represent the marginal distributions on the axes (independent consideration of individual variables $x$ and $y$) (Bishop 1995).

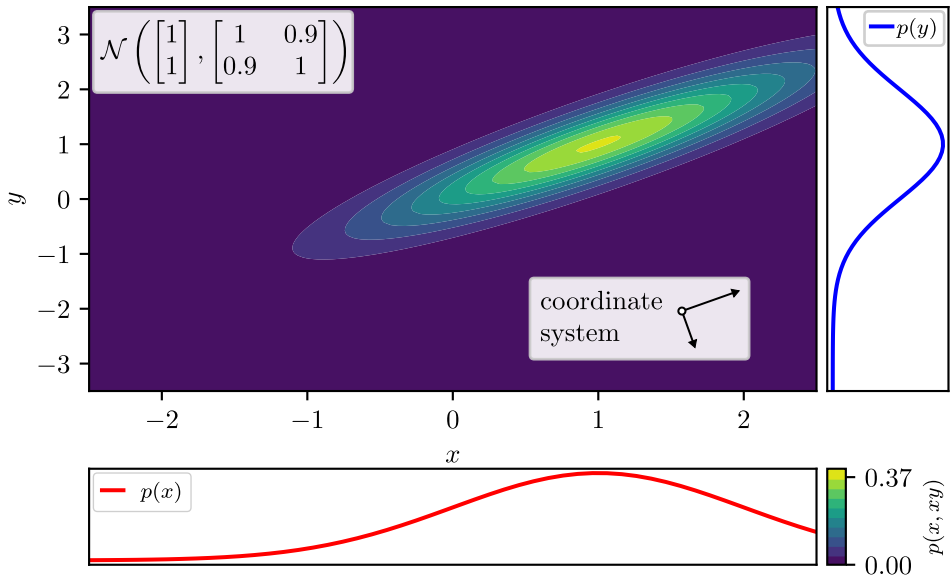

Figure 8.2: Plot of exemplary 2D multivariate Gaussian distributions.

### 8.1.3   Types of Covariance Matrices

Different types of covariance matrices can be specified for multivariate Gaussian distributions (see figure 8.3). The usual case is represented by a *full* covariance matrix $\boldsymbol{\Sigma}^{full}$. $\boldsymbol{\Sigma}^{full}$ is a symmetric, positive-definite matrix containing $D^2$ independent parameters, where $D$ represents the dimensionality of the data. As a result, the number of entries of the covariance matrix quadratically depends on the dimensionality of the data. In figure 8.3a, a Gaussian distribution with a full covariance matrix is depicted. The use of a full covariance matrix leads to the independent expansion of the distribution to the axes of the coordinate system. The main diagonal represents the spread while the off-diagonal describes the rotation of the distribution.

However, the use of the full covariance matrix may be impractical in application-oriented contexts, especially with regard to high-dimensional data. Simplifying the covariance matrix reduces the number of parameters, which in turn reduces computational complexity. However, a simplification of the covariance matrix leads to a limited degree of freedom with regard to the distribution. Thus, not all correlations from the data can be captured. Different degrees of freedom are outlined in figure 8.3. In addition, the coordinate systems illustrate the degrees of freedom. A first simplification step is constituted by the *diagonal* covariance matrix $\boldsymbol{\Sigma}^{diag} = diag(\sigma_i^2) = \Sigma_{ii}$. This reduces the number of free parameters in the covariance matrix to $D$ parameters. In figure 8.3b, only the diagonal parameter values of the covariance matrix are utilized. As a consequence, only the direction of the axes can be varied (axis-aligned ellipsoid).

The most simplified variant is the *isotropic* covariance matrix $\boldsymbol{\Sigma}^{iso} = \sigma^2 \boldsymbol{I}$. Solely concentric distributions can be modeled by the isotropic covariance type (see figure 8.3c). Thus, the height and width in the two-dimensional case is always the same. The main diagonal is represented by one single value $x$ ($\boldsymbol{\Sigma}_{ii} = x$). The advantage of the isotropic covariance is that the number of parameters of $\boldsymbol{\Sigma}^{iso}$ is 1 and, thus, lower than for the diagonal type.

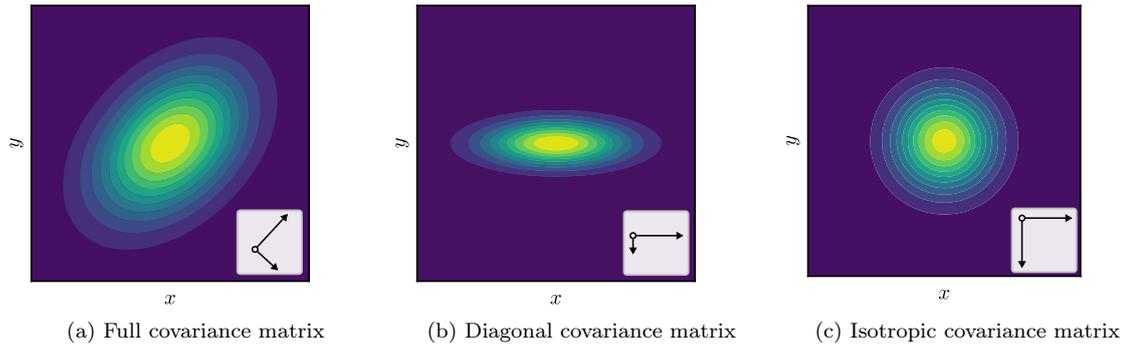

(a) Full covariance matrix       (b) Diagonal covariance matrix       (c) Isotropic covariance matrix

Figure 8.3: Three types of covariance matrix.

### 8.1.4 Gaussian Mixture Models

GMMs are probabilistic models based on a fixed number of mixed multivariate Gaussian distributions. They can be used as clustering methods, among others, which is why they can be categorized as unsupervised learning methods. By means of a weighting of $K$ Gaussian component densities, it is attempted to explain the data $\mathbb{X} = \{\boldsymbol{x}_1, \ldots, \boldsymbol{x}_n\}$. Accordingly, the $\pi_k$ are the weights in the mixture of Gaussian components, where $k \in \{1, ..., K\}$, and $\sum_{k=1}^{K} \pi_k = 1$. As a short notation, $\mathcal{N}(\boldsymbol{x}|\boldsymbol{\mu}_k, \boldsymbol{\Sigma}_k)$ is equivalent to $\mathcal{N}_k(\boldsymbol{x})$. The individual likelihoods are added up, as shown in equation 8.5.

$$p(\boldsymbol{x}_n) = \sum_{k=1}^{K} \pi_k \mathcal{N}_k(\boldsymbol{x}_n) \tag{8.5}$$

Equation 8.5: Likelihood estimation by multiple Gaussian components.

Each sample is assumed to have been created by a single component density, which is expressed by the unknown latent variable. Thus, the unobservable latent variable $z_n \in \{1, \ldots, K\}$ is re-introduced. Under this assumption, the *complete-data likelihood* of a single sample is obtained, as represented in equation 8.6.

$$\mathcal{L}(\boldsymbol{x}_n, z_n) = \pi_{z_n} \mathcal{N}_{z_n}(\boldsymbol{x}) \tag{8.6}$$

Equation 8.6: The complete-data likelihood.

The non-observability of the latent variable implies that it could be marginalized out of the complete-data log-likelihood (in equation 8.6). This assumption makes the expression suitable for optimization, and it results in the *incomplete-data likelihood* displayed in equation 8.7. As a consequence, all parameters depend on observable variables.

$$p(\boldsymbol{x}_n) = \sum_{k=1}^{K} p(\boldsymbol{x}_n, z_n = k) \tag{8.7}$$

Equation 8.7: The incomplete-data likelihood.

If all $n$ samples of the dataset $\mathbb{X}$ are considered, equation 8.8 represents the *total incomplete-data likelihood*.

As a last step, substituting from equation 8.8 the part in equation 8.5, and setting the result into the log-domain (see section 8.1.1), the *total incomplete-data log-likelihood* is obtained. The resulting

$$p(X) = \prod_n p(\boldsymbol{x}_n) = \prod_n \sum_k p(\boldsymbol{x}_n, z_n = k) \qquad (8.8)$$

Equation 8.8: The total incomplete-data likelihood.

function is suitable as a loss function $\mathcal{L}$, as it is free of non-observable parameters.

$$\mathcal{L} = \log p(\mathbb{X}) = \sum_n \log \sum_k \pi_k \mathcal{N}_k(\boldsymbol{x}_n). \qquad (8.9)$$

Equation 8.9: The total incomplete-data log-likelihood.

Assuming that not all data samples of a dataset $\mathbb{X}$ are available, the notation in equation 8.10 is introduced in order to express the likelihood for only a certain amount of data. This is relevant, for example, in the context of batch-wise processing or streaming scenarios. For reasons of numerical stability, the sum is replaced by the expectation value ($\mathbb{E}$).

$$\mathcal{L} = \mathbb{E}_n \left[ \log \sum_k \pi_k \mathcal{N}_k(\boldsymbol{x}_n) \right]. \qquad (8.10)$$

Equation 8.10: The log-likelihood loss function.

#### 8.1.4.1   Responsibilities

Since GMMs can be categorized as clustering methods, a soft membership of a sample to a data cluster or component can be determined. In other words, the (pseudo-)posterior probability indicates that a sample was generated from a particular component $k$. This is different compared to other clustering methods like k-means, whereas k-means is based on hard assignments. The main difference, however, is that GMMs indicate a soft membership of each cluster center. These memberships are referred to as *responsibility* (denoted as $\gamma$), which is determined by using the equation displayed in equation 8.11.

$$\gamma_k(\boldsymbol{x}_n) = \frac{\pi_k \mathcal{N}(\boldsymbol{x}_n | \boldsymbol{\mu}_k, \boldsymbol{\Sigma}_k)}{\sum_{j=1}^{K} \pi_j \mathcal{N}(\boldsymbol{x}_n | \boldsymbol{\mu}_j, \boldsymbol{\Sigma}_j)} \qquad (8.11)$$

Equation 8.11: Determination of soft cluster membership (responsibilities $\gamma$).

#### 8.1.4.2   Expectation-Maximization Algorithm

The Expectation-Maximization (EM) algorithm published by Dempster, Laird, et al. (1977) is used to adjust the parameters of a statistical model, e.g., a GMM. EM is a batch-type algorithm that is based on two steps. Firstly, the *Expectation* (E) step determines the "expectation" value, i.e., the responsibilities $\gamma(\boldsymbol{x})$ (see section 8.1.4.1). The E step assigns probabilities to each sample $\boldsymbol{x} \in \mathbb{X}$. Secondly, the *Maximization* (M) step changes the parameters of the underlying model in order to achieve a higher expectation value. The M step is based on the previously determined expectation values.

The two EM steps are iterated until a certain convergence criterion is fulfilled, e.g., a minimal change of the expectation value. The process usually adjusts the parameters very strongly at the beginning of the optimization, whereas the adjustment of the parameters decreases. Before the first E step can be executed, an initialization of the model has to be performed. The definition of reasonable initial parameters for the means (and covariances) is crucial. k-means is often used for the initialization of GMM components in many implementations (note that k-means also has to be initialized). The full procedure is summarized in algorithm 8.1.

---

**Algorithm 8.1: Procedure of the EM algorithm.**

---

**Data:** training data: $\mathbb{X}$, model parameter: $\theta$
**Result:** trained mixture model

**1**    *Initialize model parameter $\theta$.*
**2**    **while** *convergence criterion is not met* **do**
**3**       Expectation (E) step: Determination of the preliminary expectation value based on $\theta$.
**4**       Maximization (M) step: Fit the parameters using the maximum likelihood function.

---

The EM steps are comparable to the optimization steps used by the well-known k-means algorithm. For k-means, the number of cluster centers needs to be specified and initialized. As a next step, the centers are iteratively moved based on the corresponding "closest" and thus related data points. Finally, the data points are reassigned to the nearest cluster centers based on the distance metric (e.g., Euclidean distance). These steps are repeated until a certain number of iterations is reached or the allocation of all data points is constant at one step.

A similar algorithm is defined for training GMMs. First of all, the number of Gaussian components needs to be specified and initialized. Subsequently, the cluster centers are gradually adjusted to the data. The basic difference compared to k-means is that the assignment to a cluster center or respectively component is not hard, but gradual. The gradual assignment is illustrated in figure 8.4 by color interpolations of the samples. Gaussian components are represented by the red cross ✖ and the blue circle ●. The ellipsoids represent the covariance of the individual components. The color gradient of each sample represents the proportional assignment to each of the two components. Based on the sample's position, both, a component's position and its variance are adjusted step by step (see figures 8.4a to 8.4c).

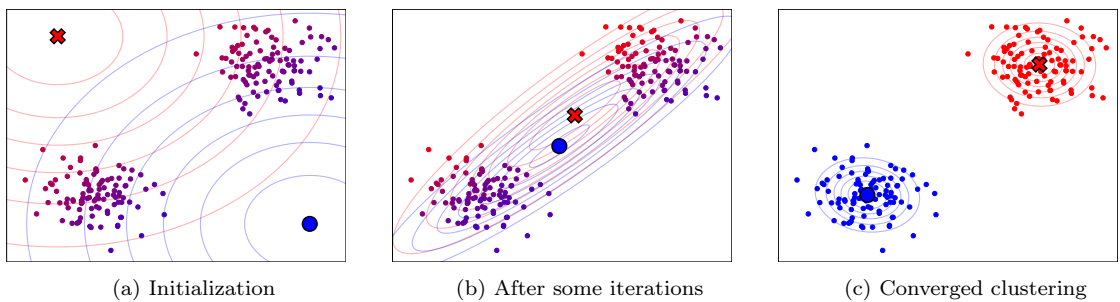

(a) Initialization       (b) After some iterations       (c) Converged clustering

Figure 8.4: Illustration of the EM algorithm.

For the comparison of k-means and GMM, the operation of the k-means algorithm is outlined in figure 8.5. Again, the red cross and the blue circle are supposed to represent cluster centers. A direct comparison shows that a sample is always assigned to exactly one center (no gradual assignment, figure 8.5 vs. figure 8.4). The distance (often Euclidian distance) is used as a basis for the assignment, which is illustrated by the lines and colors. Similarly, a corresponding converged result is obtained after several optimization iterations (see figures 8.5a to 8.5c).

**Maximum-Likelihood Estimation**    The *maximum-likelihood estimation* (MLE) describes the parameter adjustment of a probabilistic model (see Bishop 1995). In case of multiple multivariate Gaussian distributions, e.g., for a GMMs, the process is almost the same as for single Gaussian distributions. The basis for the MLE is the assumption of a convex loss function $\mathcal{L}$, which is optimized by setting the derivative to 0. By calculating the partial derivative for a single component according to $\frac{\partial}{\partial \boldsymbol{\mu}}$, the result is $\boldsymbol{\mu}^{new} = \frac{1}{N} \sum_{n=1}^{N} \boldsymbol{x}_n$ (which corresponds to the average). The derivative for $\frac{\partial}{\partial \boldsymbol{\Sigma}}$ is $\boldsymbol{\Sigma}^{new} = \frac{1}{N} \sum_{n=1}^{N} (\boldsymbol{x}_n - \boldsymbol{\mu}^{new})(\boldsymbol{x}_n - \boldsymbol{\mu}^{new})^\top$, with dependence on $\boldsymbol{\mu}^{new}$. Although the expected value is $\mathbb{E}[\boldsymbol{\mu}^{new}] = \boldsymbol{\mu}$, this is not true for $\mathbb{E}[\boldsymbol{\Sigma}^{new}] \neq \boldsymbol{\Sigma}$. Therefore, the optimization step has to be repeated iteratively until a model convergence is detected.

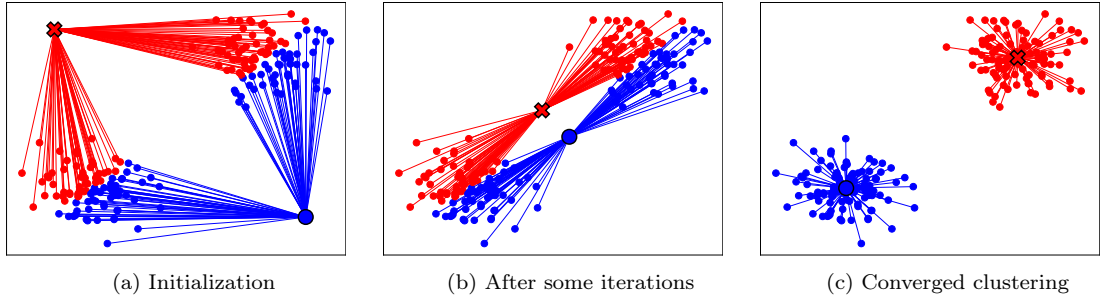

(a) Initialization                (b) After some iterations                (c) Converged clustering

Figure 8.5: Illustration of the k-means algorithm.

A further adjustment has to be performed for GMMs with $K$ multivariate Gaussian distributions. In addition to the $k$ components, the mixing coefficients $\pi_k$ have to be adapted. The procedure follows the same principle of zeroing the derivative. Additionally, weighting is applied which is represented by the responsibilities $\gamma_k$ of each sample (see section 8.1.4.1). Accordingly, the mean is initially determined by multiplying $\gamma_k(\boldsymbol{x})$, which is represented in equation equation 8.12. Due to the available $K$ components, a weighted mean is calculated depending on the responsibility $\gamma(\boldsymbol{x}_n)$ of a sample (see equation 8.12), whereby $N_k = \sum_{n=1}^{N} \gamma(\boldsymbol{x}_n)$.

$$\boldsymbol{\mu}_k = \frac{1}{N_k} \sum_{n=1}^{N} \gamma_k(\boldsymbol{x}_n) \boldsymbol{x}_n \tag{8.12}$$

Equation 8.12: Determiniation of the weighted mean for GMMs.

The same weighting procedure is applied to the determination of the variances. Again, the responsibilites are used as weighting factors (see equation 8.13).

$$\boldsymbol{\Sigma}_k = \frac{1}{N_k} \sum_{n=1}^{N} \gamma_k(\boldsymbol{x}_n)(\boldsymbol{x}_n - \boldsymbol{\mu})(\boldsymbol{x}_n - \boldsymbol{\mu})^\top \tag{8.13}$$

Equation 8.13: Determiniation of the weighted variances for GMMs.

As a last adaption step, the mixing coefficients $\pi_k$ have to be adjusted, where $\sum_{k=1}^{K} \pi_k = 1$ must be respected. The normalization is realized by adding a Lagrange multiplier (Bertsekas 1996), which results in equation 8.14, where $N_k = \sum_{n=1}^{N} \gamma_k(\boldsymbol{x}_n)$.

$$\pi_k = \frac{N_k}{N} \tag{8.14}$$

Equation 8.14: Determiniation of the mixing coefficients for GMMs.

**Properties of EM**  The EM algorithm has some inherent advantages and disadvantages which are briefly discussed in the following. A decisive advantage is that the EM algorithm does not contain any hyper-parameters that have to be adapted to the problem in a complex manner, e.g., by grid-search. This advantage constitutes a direct contrast to stochastic methods, e.g., SGD, where at least a learning rate $\epsilon$ needs to be specified. A disadvantage, however, is related to the consideration of EM as batch learning algorithm which does not allow for incremental training (see section 2.3.2). Consequently, the entire dataset has to be accessible, which makes EM unsuitable for very large datasets or streaming scenarios.

Another problem is related to an assumption of the EM algorithm. Namely, that the latent variable (un-observable) can be removed by marginalization. However, this is only the case if the loss function is truly convex, which in turn is only true if the samples are drawn from exactly one component at a time. The Jensen's inequality (Cvetkovski 2012) solely holds for the assumption of marginalization (as a lower bound), while ensuring that each E-step truly guarantees an improvement. Nevertheless, the EM method works for many types of problems. Regardless of the EM training process, a certain set of components based on the problem needs to be available.

## 8.2   Related Work

This section presents related work in the context of the novel approach, and particularly with regard to GMMs. Accordingly, the main focus is on the design of a training procedure for GMMs that can be used in the context of CL. Another main aspect is related to the transformation of the hierarchical models into a "deep" model design.

**Incremental GMM Training**   The first subject of the related work addresses the challenges raised by the standard EM algorithm (Dempster, Laird, et al. 1977). The prerequisite for the application of the EM is constituted by the complete dataset that is needed for the E-step. However, the provision of the complete dataset is impractical for today's size of datasets. Early attempts have tried to enable incremental learning (Titterington 1984). However, many processes still present challenges: "Numerical integration is often necessary and the fact that we are dealing with incomplete data will add to the complications." (Titterington 1984).

A frequently used variant is proposed by Cappé and Moulines (2009). The stochastic EM (sEM) algorithm is presented as an online variant of EM. sEM is able to replace the M-step by a stochastic approximation. An advantage compared to Titterington's (1984) proposal is that the parameter constraints (see section 8.1.4) are automatically fulfilled. The work of Chen, Zhu, et al. (2018) (variance reduced stochastic EM (sEM-vr)) illustrates that these algorithms are still the focus of current research. Another approach is based on the so-called "core-sets" as proposed by Feldman, Faulkner, et al. (2011). A sub-dataset of the entire dataset, which is presented as a data stream, is derived as a representative for the entire dataset. The sub-dataset is referred to as core-set and used for training. In the context of streaming scenarios, the core-set is extracted from each new data block and merged with the previous core-set.

Another online variant is proposed by Vijayakumar, D'Souza, et al. (2005) with Locally Weighted Projection Regression (LWPR). Their model can be considered as a GMM, which can be updated online based on gradients.

**Number of GMM Components**   Another problem is the definition of the number of Gaussian components $K$, which cannot be adjusted in retrospect. Vlassis and Likas (2002) propose a greedy EM variant in order to overcome the problem of defining the number of required GMM components . Unlike the data-driven method by Ghahramani and Hinton (1997) (Mixtures of Factor Analyzer (MFA)) or Tipping and Bishop (1999) (Principal Component Analysis (PCA)), Vlassis and Likas's (2002) approach does not rely on data for pre-defining the number of the necessary components. On the contrary, the greedy variant (based on Li and Barron' (2000) theory) successively adds further components. The work of Engel and Heinen (2010) follows a similar approach. A new component is added each time the *minimum likelihood* criterion is met by a new sample.

Pinto and Engel (2015) introduce the Incremental Gaussian Mixture Network (IGMN), which focuses on reducing the time complexity from $\mathcal{O}(NKD^3)$ to $\mathcal{O}(NKD^2)$. The precision matrix (Bernardo and Smith 2008; Bishop 1995) is used instead of the covariance matrix in order to achieve computational efficiency. The advantage of the proposed method is that the calculation of the inverse of the determinant ($\mathcal{O}(D^3)$) is replaced by an operation with a time complexity of $\mathcal{O}(D^2)$. Furthermore, the IGMN model allows for the dynamic addition of new components. Merging statistically equivalent Gaussian components is another approach presented by Song and Wang (2005). Their approach is intended to enable online data stream clustering. Taking into account the equivalence criterion ($W$ static for covariances and Hotelling's $T^2$ for means), access to past data is verified and granted if necessary.

The approach of Kristan, Skočaj, et al. (2008) comprises the addition and removal of Gaussian components (unlearning). In order to add GMM components, the second derivative of a batch of samples is calculated and the "asymptotic mean integrated squared error" is determined. The proposed mechanism is used in an iterative merge process until convergence behavior is established. The validity of their method has been proven for one-dimensional distributions (Kristan, Skočaj, et al. 2008).

**GMM Initialization**   Another fundamental problem concerns all variants based on the EM algorithm, which heavily depend on the initial state of the Gaussian components. The initialization problem is reviewed, for example, by Baudry and Celeux (2015). Different variations of the EM algorithm (among others sEM) are investigated using different initialization mechanisms. Baudry and Celeux (2015) identify the fundamental problem: "[...] the initialization issue remains and can be a most influencial factor."

**SGD Training Methods**   Hosseini and Sra (2015) describe an alternative method for training GMMs based on Riemannian manifold optimization technique. In the latest work of Hosseini and Sra (2020), a SGD based solution is presented. Their approach introduces several hyper-parameters. Besides, it circumvents the initialization problem by a variant of k-means (`k-means++`).

**Annealing**   A similar approach is presented by Verbeek, Vlassis, et al. (2005) who use the EM-algorithm in order to allow for a self-organization of the components for mixture models. A behavior similar to standard Self-Organizing Maps (SOMs) (Kohonen 1982) is achieved by adding a penalty term.

**Log-Likelihood Approximations**   Several different suggestions have been made for the approximation of the log-likelihood. Four approximations are presented by Pinheiro and Bates (1995): Linear Mixed-Effects-, Laplacian-, Importance Sampling-, and the Gaussian quadrature approximation. Ormoneit and Tresp (1998) propose the *maximum penalized likelihood* and a Bayesian approach, which both provide better results than the maximum likelihood. It aims at the avoidance of the determination problems in the high-dimensional data spaces, such as singularities or local maxima. A variant which is particularly more computational efficient is presented in the work of Dognin, Goel, et al. (2009). A similar approach is described by Van Den Oord and Schrauwen (2014). It is characterized by a "hard" assignment.

**High-Dimensional Data**   Training with high-dimensional data is a common and widely researched problem. Ge, Huang, et al. (2015) tackle the problem by using several moments ($3^{rd}$, $4^{th}$ and $6^{th}$ order) in order to characterize the Gaussian components of a GMMs. However, their approach has a disadvantage, as it is difficult to implement in streaming scenarios. The work of Richardson and Weiss 2018 shows that GMMs can be used in high-dimensional domains, such as the generation of images. Goodfellow, Pouget-Abadie, et al. (2014) compare them with a current research topic/model, namely Generative Adverserial Networks (GANs). The problem of determining the inverse of large matrices is realized by using MFA (Richardson and Weiss 2018). Another fundamental problem of high-dimensional data is circumvented by the so-called "logsumexp" trick (Nielsen and Sun 2016).

## 8.3   Stochastic Gradient Descent for Training Gaussian Mixture Models

Stochastic Gradient Descent (SGD) is an efficient method to adjust parameters within a model (see section 2.2.2.2). Models can be trained incrementally based on the SGD (see section 2.3.2 for a definition), which is a requirement for many CL scenarios. In addition, gradient descent is used as the basis for almost all deep learning approaches. Incremental learning, however, only describes the step by step extraction of knowledge from data – not the preservation of existing knowledge. The ability to learn in an incremental manner does not ensure that a model is automatically suitable for CL, which is due to the CF effect (see section 2.3.4).

This section shows how GMMs can be efficiently trained by a SGD based method. In order to enable SGD training, the GMM requirements need to be satisfied. These include, for example, the normalization of the weights $\pi$ or the positive-definiteness of the covariance matrix $\Sigma$. How these constraints are fulfilled is described in section 8.3.1. Due to numerical issues, it is challenging to use the standard log-likelihood as a loss function for the model parameter optimization. A possible solution is presented in section 8.3.2. Another crucial and already mentioned problem is related to the initialization of a GMMs, which is often avoided in related work. In general, the complete dataset and another clustering algorithm (e.g., k-means) are used and initialized before the GMM training starts. The problem is that a pre-initialization by means of the complete dataset is impossible in streaming scenarios. In order to tackle the initialization problem, an annealing scheme is introduced in section 8.3.3. The basic idea is related to the training of SOMs which is adapted to the domain of GMM training. This section concludes with a summary of the complete training process of GMMs by means of SGD (see section 8.3.4).

### 8.3.1 GMM Constraint Enforcement

In order to train GMMs via SGD, multiple requirements regarding the parameters need to be fulfilled. Usually, the EM algorithm fulfills these requirements. Meeting the GMM requirements involves the following two adjustments for each SGD based training step.

#### 8.3.1.1 Weights $\pi$

The first parameter that has to be adjusted concerns the weights $\pi$ of the $K$ GMM components. The weights describe the prior probability (up to this training point) for the sample $\boldsymbol{x}$ being drawn from the $k^{th}$ component. The following procedure is adapted from Hosseini and Sra (2015) and is comparable to the application of the Softmax function (see section 2.2.2.1). A corresponding weight $\pi_k$ is assigned to each component $k$. In order to normalize the weights, i.e., $\sum_{k=1} \pi_k = 1$, the free parameter $\xi$ is introduced. Equation 8.15 illustrates that the weights can be normalized.

$$\pi_k = \frac{\exp(\xi_k)}{\sum_j \exp(\xi_j)}.$$  (8.15)

Equation 8.15: Normalization of GMM component weights $\pi$.

#### 8.3.1.2 Covariance Matrix

For computational reasons, the covariance matrix is not used directly. Instead, its inverse matrix – the precision matrix $\boldsymbol{P} = \boldsymbol{\Sigma}^{-1}$ – is applied (see Bernardo and Smith 2008; Bishop 1995). The diagonal covariance ($\boldsymbol{\Sigma}^{diag}$) or the diagonal precision matrix respectively is introduced as yet another simplification (see section 8.1.3). This is required due to the particularly high dimensionalities of the data. Determining the full covariance matrix is too memory intensive and therefore omitted.

The diagonal matrices $\boldsymbol{D}_k$ (for each component $k$) are re-parameterized as $\boldsymbol{P}_k = \boldsymbol{D}_k^2$. By using a Cholesky decomposition/factorization (Mayers, Golub, et al. 1986), a symmetric positive-definite matrix can be decomposed into the product of a triangular matrix and its transpose. As a result, the used precisions are guaranteed to be positive-definite. For the diagonal case $\det \boldsymbol{\Sigma}_k = \det \boldsymbol{P}_k^{-1} = \left(\det(\boldsymbol{D}_k^2)\right)^{-1} = \left(\operatorname{tr}(\boldsymbol{D}_k)\right)^{-2}$ can be calculated. Due to the fact that the decomposition is still complex, it is only performed once. The elements below the diagonal need to be removed after each gradient descent step.

### 8.3.2   Max-Component Approximation

The minimization of the loss function $\mathcal{L}$ is performed by means of SGD. A problem is that underflows often occur during the determination of the loss. High data dimensionalities are responsible for the underflows. The problem causes divisions by zero, which leads to Not a Number (NaN) values in computer-aided applications. Instead of using the vanilla log-likelihood as a loss function, a lower bounded version is introduced, which is referred to as *max-component approximation*. The approximation is denoted as $\hat{\mathcal{L}}$ and illustrated in section 8.3.2. $k^*$ is defined as $\arg\max_k \pi_k \mathcal{N}_k(\boldsymbol{x}_n, \boldsymbol{\mu}_k, \boldsymbol{\Sigma}_k)$ and represents the loss value based on the best matching unit (BMU). A BMU is the component that provides the highest responsibility for a given sample.

$$\mathcal{L} = \mathbb{E}_n\left[\log \sum_k \pi_k \mathcal{N}_k(\boldsymbol{x}_n)\right],$$

$$\hat{\mathcal{L}} = \mathbb{E}_n\left[\log\left(\pi_{k^*} \mathcal{N}_{k^*}(\boldsymbol{x}_n)\right)\right] \text{ where}$$

$$\mathcal{L} \geq \hat{\mathcal{L}} \tag{8.16}$$

Equation 8.16: Max-component approximation $\hat{\mathcal{L}}$.

The max-component approximation is not tight, but constitutes a valid approximation for high data dimensionalities (see Gepperth and Pfülb 2021 for a proof). Usually, the so-called *logsumexp* trick (see McElreath 2018) is applied to handle numerical instabilities caused by exponentials. However, the application of the logsumexp only mitigates the problem but does not eliminate it. Underflows occur when the distances increase, which is usually the case in the context of large data dimensionalities. If 32-bit floats are used, for example, it is attempted to find a component probability of $\mathcal{N}_k = e^{-101}$ with the highest probability $\mathcal{N}_{k^*} = e^3$. This results in $\frac{\mathcal{N}_k}{\mathcal{N}_{k^*}} = e^{-104}$. The value exceeds the representable range of a 32-bit floating point unit and leads to an underflow, respectively NaN. The loss function $\hat{\mathcal{L}}$ presented in section 8.3.2 has the advantage of avoiding these numerical problems.

Another problem related to the standard log-likelihood $\mathcal{L}$ is known as undesirable local optima. In particular, degenerate solutions arise, which are omitted by the max-component approximation. However, degenerate solutions are characterized by all parameters (weights, centroids and covariance matrices) taking almost the same values. For this reason $\forall k$ holds that $\pi_k \approx \frac{1}{K}$, $\boldsymbol{\mu}_k = \mathbb{E}[\mathbb{X}]$ and $\boldsymbol{\Sigma}_k = \mathrm{Cov}(\mathbb{X})$, which is visualized in figure 8.6a. The component weights $\pi_k$ are indicated as part of the figure. The mentioned parameter configuration causes vanishing gradients. Even though the degenerate solution pattern is suppressed by the max-component approximation, a new problem arises, namely a single-component solution or sparse-component solution. In both of these cases, only one or a subset of GMM components is selected over and over again. This means that other components can never become the BMU and, thus, never get adapted.

An exemplary sparse-component solution is illustrated in figure 8.6b. It is noted that mainly one component is selected (with the highest $\pi$), which looks similar to components of the degenerate solution. After the initialization, a first random component is updated so that for all following steps only the same one is selected as BMU. The degenerate or sparse-component solution problem is addressed by the annealing scheme introduced in the following section (section 8.3.3).

### 8.3.3   Annealing Procedure

A fundamental problem of training GMMs without data-driven initialization is related to undesirable local optima. This problem occurs whenever the parameters of the model are updated by SGD for the standard log-likelihood $\mathcal{L}$, as well as for the max-component approximation $\hat{\mathcal{L}}$ presented in section 8.3.2. So-called sparse-component solutions occur almost every time (see figure 8.6b). They are characterized by only one or a few dominant components. These GMM components describe the complete or a very large subset of the data: $\pi_{k_i} \gg 0$, $\boldsymbol{\mu}_{k_i} = \mathbb{E}[\mathbb{X}_{k_i} \subset \mathbb{X}]$, $\boldsymbol{\Sigma}_{k_i} = \mathrm{Cov}(\mathbb{X}_{k_i} \subset \mathbb{X})$. The remaining components never become BMU, which is expressed by $\pi_k \approx 0$. These not-adjusted components (means $\mu_k$, covariance matrix $\boldsymbol{\Sigma}$ respectively the precision matrices $\boldsymbol{P}$) remain in their initial state.

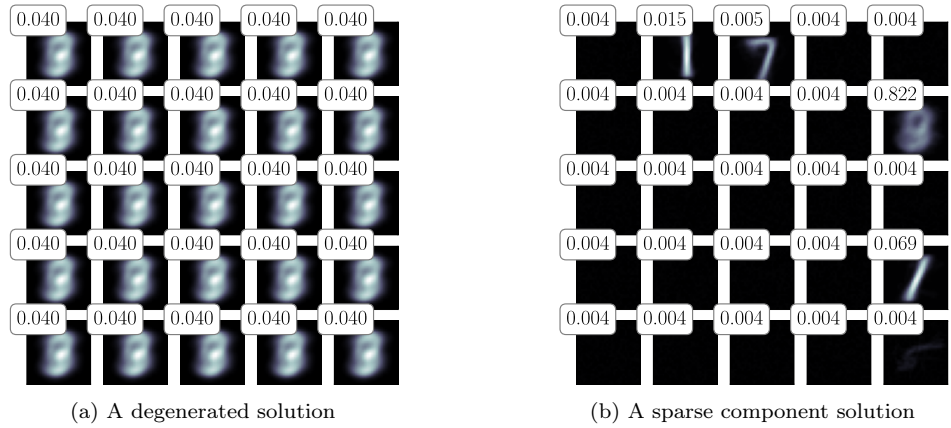

(a) A degenerated solution                    (b) A sparse component solution

Figure 8.6: Degenarated and sparse-component solutions for the MNIST dataset.

Equation 8.17 outlines that the gradients consequently vanish for $\delta_{kk^*}$. Only the parameters of the BMU components, i.e., $k^*$, are adjusted.

$$\frac{\partial \hat{\mathcal{L}}}{\partial \boldsymbol{\mu}_k} = \mathbb{E}_n \left[ \boldsymbol{P}_k \left( \boldsymbol{x}_n - \boldsymbol{\mu}_k \right) \delta_{kk^*} \right]$$

$$\frac{\partial \hat{\mathcal{L}}}{\partial \boldsymbol{P}_k} = \mathbb{E}_n \left[ \left( (\boldsymbol{P}_k)^{-1} - (\boldsymbol{x}_n - \boldsymbol{\mu}_k)(\boldsymbol{x}_n - \boldsymbol{\mu}_k)^{\top} \right) \delta_{kk^*} \right] \qquad (8.17)$$

$$\frac{\partial \hat{\mathcal{L}}}{\partial \pi_k} = \pi_k^{-1} \mathbb{E}_n \left[ \delta_{kk^*} \right].$$

Equation 8.17: Gradients for the max-component approximation $\hat{\mathcal{L}}$ loss function.

The subsequently proposed annealing approach provides a mechanism to circumvent the vanishing gradient problem and to avoid the undesirable local optima. The fundamental idea is to punish the emerging characteristic response pattern. Accordingly, a new hyper-parameter $\sigma$ is introduced, which is constantly being reduced until it disappears. The new loss function is represented by $\hat{\mathcal{L}}^{\sigma}$ and denoted as *smoothed max-component log-likelihood*. As shown in equation 8.18, a term is added which depends on the $\sigma$ parameter. The term indicates that not only the BMU is updated by SGD, but also the neighboring components. Thereby, the $\sigma$ parameter defines the intensity's influence on the neighboring components. The smoothed max-component approximation defines an effective upper bound for $\hat{\mathcal{L}}^{\sigma}$ (see Gepperth and Pfülb 2021 for a proof).

$$\hat{\mathcal{L}}^{\sigma} = \mathbb{E}_n \max_k \left[ \sum_j \boldsymbol{g}_{kj}(\sigma) \log \left( \pi_j \mathcal{N}_j(\boldsymbol{x}_n) \right) \right]$$

$$= \mathbb{E}_n \sum_j \boldsymbol{g}_{k^*j}(\sigma) \log \left( \pi_j \mathcal{N}_j(\boldsymbol{x}_n) \right). \qquad (8.18)$$

Equation 8.18: Smoothed max-component approximation $\hat{\mathcal{L}}^{\sigma}$.

In order to define a factor for influencing the neighboring components, various structures can be implemented. One possibility is to arrange the $K$ GMM component in a grid of the size $\sqrt{K} \times \sqrt{K}$. If a component $k^*$ is chosen, it influences all neighboring components. The influence value is described by an uni-modal Gaussian profile of spatial variance $\sim \sigma^2$. In order not to favor/disadvantage the components on the edges, the grid is assumed to be periodical (or spherical). The visualization in figure 8.7 illustrates the type of influence over time. Figure 8.7 is up-scaled for visualization. For $K = 25$, each grid area would consist of $5 \times 5$ pixels.

Different stages of the smoothing filters are shown in figure 8.7. On the left-hand side, an initial high value for $\sigma$ is applied. In each field, the influence of neighboring components is indicated, whereas darker faces represent a higher influence on the closest component. If the most upper left component is selected as BMU, for example, the periodic effect is clearly visible in the upper right, lower left and right corner. During the training process, $\sigma$ is reduced (see middle) until finally only $k^*$ is updated (see right). A one dimensional grid is another possible structure, as it requires the same computational effort as a 2D grid. A side effect of the applied annealing scheme is that the GMM components exhibit topological sorting. Its center depends on the choice of the first BMU and the respective sample.

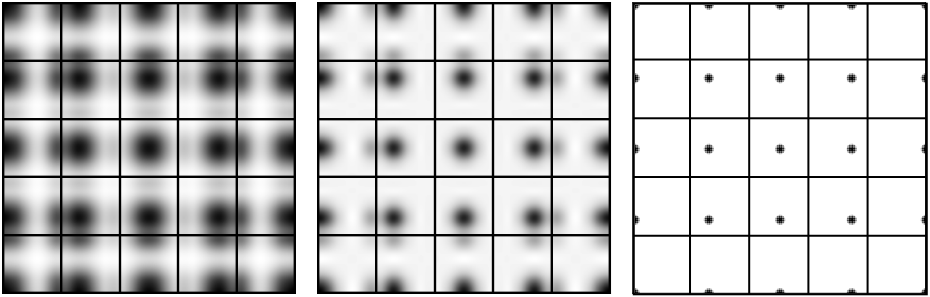

Figure 8.7: Visualization of different Gaussian smoothing filters.

Adjusting the introduced annealing parameter $\sigma$ raises a new challenge. At the beginning of the training process, the value is set to $\sigma(t) = \sigma_0$. For the reduction, $\sigma \leftarrow 0.9\sigma$ is calculated. $\sigma$ is therefore reduced asymptotically until the limit $\sigma = \sigma_\infty$ is reached. This reduction mechanism guarantees a smooth transition from $\hat{\mathcal{L}}^\sigma$ (equation 8.18) to $\hat{\mathcal{L}}$ (section 8.3.2). An applicable default value for the lower and upper $\sigma$-threshold is specified in section 8.3.4.

Due to the fact that the influence of $\sigma$ represents an upper bound with respect to the quality criterion, $\sigma$ is reduced whenever the loss does not improve any further. In order to determine the stationarity of the loss, the sliding smoothed average of the loss value is traced: $\ell(t) = (1 - \alpha)\ell(t - 1) + \alpha\hat{\mathcal{L}}^\sigma(t)$. $\alpha$ is utilized to define the time component by validating the increase of $\hat{\mathcal{L}}$ every $\frac{1}{\alpha}$ iteration. As soon as $\Delta < \delta$ is reached (see equation 8.19), $\sigma$ is reduced as described.

$$\Delta = \frac{\ell(t) - \ell(t - \alpha^{-1})}{\ell(t - \alpha^{-1}) - \hat{\mathcal{L}}^\sigma(t = 0)} \tag{8.19}$$

Equation 8.19: Definition of the $\sigma$ reduction criterion $\Delta$.

## 8.3.4  Training Procedure

In this section, all of the presented requirements, methods and adaptions used for the SGD based training of GMMs are combined. Accordingly, the universal initial values for parameters are presented. In addition, the complete SGD training process is summarized in algorithm 8.2.

First of all, SGD based training requires the definition of different parameters. These definitions are stated in line 1 of algorithm 8.2. The learning rate $\epsilon$ is always problem-dependent. A relatively small learning rate often results in longer convergence times. A too high learning rate usually inhibits reaching an acceptable local optimum. The same applies to the SGD based GMM training. The time parameter $\alpha$ (reduction criterion of $\sigma$) can be set to the same value as the learning rate ($\alpha = \epsilon$). The related parameter $\delta$ (limit for the detection of stationarity) is set to 0.05, which is a good choice for all investigated datasets. The diagonal matrix for the precision is set to $D_{\max}I$, where $D_{\max} = 20$ works for all experiments.

GMM component means $\boldsymbol{\mu}$ are initialized via small random numbers ($[-\mu^i, +\mu^i]$). Component weights $\pi$ are initialized to $\frac{1}{K}$. Regarding the number of components $K$, the bottom line is "more is always better" (depending on memory and computational limitations). The smoothed max-component log-likelihood $\hat{\mathcal{L}}^\sigma$ from section 8.3.2 is used for optimization. Initial $\sigma$ values are chosen large enough for all components to be affected by the SGD update step. A default value is $\sqrt{K}$, which offers a good initialization. $\sigma_\infty$ can be defined so that the annealing process completely vanishes or becomes neglectable.

The training loop is represented in line 2. Line 3 defines the creation or update of the annealing mask, depending on the current $\sigma$. Subsequently, parameters are updated based on the determined gradients (line 4). The following two steps describe the enforcement of the GMM constraints (see section 8.3.1). Firstly, the precision matrices are clipped in line 5 in a way that the value range of the diagonal entries remains between 0 and $D_{\max}^2$. The clipping prevents entries from growing beyond the representable range of values, which is important for data points that always remain the same (e.g., black border pixels). Secondly, line 6 represents the normalization step of the component weights.

Finally, the current loss needs to be checked for stationarity. Therefore, the current sliding average of the loss is updated (line 7). If a stationarity of the loss is detected (lines 8 and 9), the $\sigma$ is reduced (line 10). During the (step by step) training process, a smooth transition towards the max-component approximation $\hat{\mathcal{L}}$ is achieved.

---

**Algorithm 8.2:** Steps of SGD-GMM training.

**Data:** initializer values: $\mu^i$, $K$, $\epsilon$, $\sigma_0/\sigma_\infty$, $\delta$ and data $\mathbb{X}$

**Result:** trained GMM model

| | | |
|---|---|---|
| 1 | $\boldsymbol{\mu} \leftarrow \mathcal{U}(-\mu^i, +\mu^i)$, $\pi \leftarrow 1/K$, $\boldsymbol{P} \leftarrow ID_{\max}$, $\sigma \leftarrow \sigma_0$ | // initialize parameters |
| 2 | **forall** $t < T$ **do** | // training loop |
| 3 | $\quad \boldsymbol{g}(t) \leftarrow \text{create\_annealing\_mask}(\sigma, t)$ | // see section 8.3.3 |
| 4 | $\quad \boldsymbol{\mu}(t) \leftarrow \epsilon\frac{\partial \hat{\mathcal{L}}^\sigma}{\partial \boldsymbol{\mu}} + \boldsymbol{\mu}(t\text{-}1)$, $\boldsymbol{P}(t) \leftarrow \epsilon\frac{\partial \hat{\mathcal{L}}^\sigma}{\partial \boldsymbol{P}} + \boldsymbol{P}(t\text{-}1)$, $\pi(t) \leftarrow \epsilon\frac{\partial \hat{\mathcal{L}}^\sigma}{\partial \pi} + \pi(t\text{-}1)$ | |
| 5 | $\quad \boldsymbol{P}(t) \leftarrow \text{precisions\_clipping}(\boldsymbol{P}, D_{\max})$ | //see section 8.3.1 |
| 6 | $\quad \pi(t) \leftarrow \text{normalization}(\pi(t))$ | //see equation 8.15 |
| 7 | $\quad \ell(t) \leftarrow (1-\alpha)\ell(t\text{-}1) + \alpha\hat{\mathcal{L}}^\sigma(\boldsymbol{x}(t))$ | // sliding likelihood |
| 8 | $\quad$ **if** *annealing update iteration* **then** | |
| 9 | $\quad\quad$ **if** $\Delta < \delta$ **then** | // $\Delta$ see equation 8.19 |
| 10 | $\quad\quad\quad \sigma(t) \leftarrow 0.9\sigma(t\text{-}1)$ | |

---

## 8.4 Deep Convolutional Gaussian Mixture Models

The training procedure described in section section 8.3 summarizes a way to efficiently train GMMs by SGD. Vanilla GMMs are still considered unsuitable for supervised ML scenarios. The proposed GMM training method cannot be used for CL tasks as described in chapter 6. Furthermore, GMMs and their ability to approximate complex functions strongly depend on the number of Gaussian components $K$. An adequate approximation of functions is very memory and computation intensive for complex data distributions. DNNs do not have this problem, since they can approximate complex functions due to their "depth" (expressed by the universal approximation theorem).

How GMMs can be stacked like layers of artificial neurons is described in this section. The stacking should allow for the approximation of more complex functions with less computational resources (GMM components). An additional mechanism from the area of image processing is introduced, namely folding. Folding is carried out in a manner similar to Convolutional Neural Networks (CNNs). The combination of stacking GMMs and folding operations is referred to as Deep Convolutional Gaussian Mixture Model (DCGMM). Thus, DCGMMs consist of several different layers which transform it into a deep learning model. Different types of layers are introduced, among others, a folding layer and a classification layer. The latter enables the classification of samples, which allows DCGMMs to be classified as supervised learning model.

DCGMMs offer a wide range of functionalities, such as classification, density estimation and sampling. In the following, these functionalities are described and illustrated. Sampling is an interesting research area, which includes the manipulation of images and videos. At the same time, the ability to generate samples is an integral part of pseudo-rehearsal approaches in the field of CL. GMMs can generate samples themselves (independent of their training type), but with some restrictions.

In the beginning of the following section, different types of layers that can be stacked into a deep model are introduced (see section 8.4.1). Section 8.4.2 presents the architecture-level functionalities of DCGMMs. The fundamental functionalities (inference and sampling) are described by a step-by-step visualization.

## 8.4.1    Types of Layers

DCGMMs are hierarchical models consisting of different types of layers. The different layer types, as well as their input and output formats are introduced in the following. Different functionalities of each layer are provided for each direction or data flow mode, i.e., estimation and sampling. The position of a layer is indicated by $L$. An output- or input tensor of a layer is denoted according to the commonly used $NHWC$ format. The $NHWC$ format is designed for image data and generally used in conjunction with CNNs. $N$ specifies the number of processed samples per step, which usually corresponds to the batch size $\mathcal{B}$. $W$ and $H$ define the width and height of the images, $C$ the number of channels. The following applies to samples that are images: $C=1$ corresponds to a grayscale image, while $C=3$ represents colored (RGB) pictures.

Each output tensor of a layer is represented by $\mathbf{O}^{(L)} \in \mathbb{R}^4$. The forward direction is referred to as *estimation*. For an estimation, each layer with the index $L$ converts the output tensor of the previous layer $\mathbf{O}^{(L-1)}$ into its own output $\mathbf{O}^{(L)}$. The reverse direction is known as *sampling* and the input tensor is presented as $\mathbf{I}^{(L)} \in \mathbb{R}^4$. $\mathbf{I}^{(L+1)}$ is obtained from a certain layer $L$ and transformed into $\mathbf{I}^{(L)}$. Depending on the type of layer, the internal parameters are represented by $\theta^{(L)}$.

### 8.4.1.1    Folding Layer

**Forward/Estimation**   The forward step of the folding layer performs a similar transformation as the convolutional layers of CNNs. The input signal (usually an image) is decomposed by a folding layer into several sub-signals. The exact operation is determined by several parameters. In order to extract different parts/patches of an image, a filter size $f_X^{(L)}$, $f_Y^{(L)}$ and its shift (stride) $\Delta_X^{(L)}$, $\Delta_Y^{(L)}$ are defined. Thus, each extracted patch is dumped into a separate channel ($C$) in the folding layer's output. The output signal $\mathbf{O}_{NHWC}^{(L-1)}$ of the previous layer is transformed to $\mathbf{O}_{NH'W'C'}^{(L)}$ by the operation defined in equation 8.20.

$$
\begin{aligned}
H^{(L)} &= 1 + \frac{H^{(L-1)} - f_Y^{(L)}}{\Delta_Y^{(L)}} \ , \\
W^{(L)} &= 1 + \frac{W^{(L-1)} - f_X^{(L)}}{\Delta_X^{(L)}} \ \text{and} \\
C^{(L)} &= C^{(L-1)} f_X^{(L)} f_Y^{(L)}
\end{aligned}
\tag{8.20}
$$

Equation 8.20: Determination of the output dimensions of a folding layer.

Figure 8.8 illustrates the forward step for the folding layer based on an example (reading direction: left to right). For the sake of simplicity, a batch size $\mathcal{B}$ of 1 is assumed, so that $N$ can be omitted. The simplified image dataset consists of two samples with a resolution of $5 \times 5$ pixel (grayscale). Images have not been rasterized. Solely the grid is supposed to indicate the resolution as depicted in figure 8.8. The shape of the previous layer output is: $1, 5, 5, 1$. Parameters of the folding layer are: The filter size $f_Y = f_X = 3$ and the stride $\Delta_Y = \Delta_X = 2$. Thus, a filter of the size $3 \times 3$ with the size 2 in both, $x$ and $y$ direction is pushed over the input signal. The application of the filter extracts multiple patches which are dumped into the output's one-dimensional channels. The extraction is represented in figure 8.8 by different colors. The equation equation 8.20 allows the calculation of the size of the output tensor $\mathbf{O}_{NWHC}^{(L)} = 1, 2, 2, 9$. A folding layer does not have any adjustable layer parameters $\theta^{(L)}$, as merely a transformation of the signal is performed.

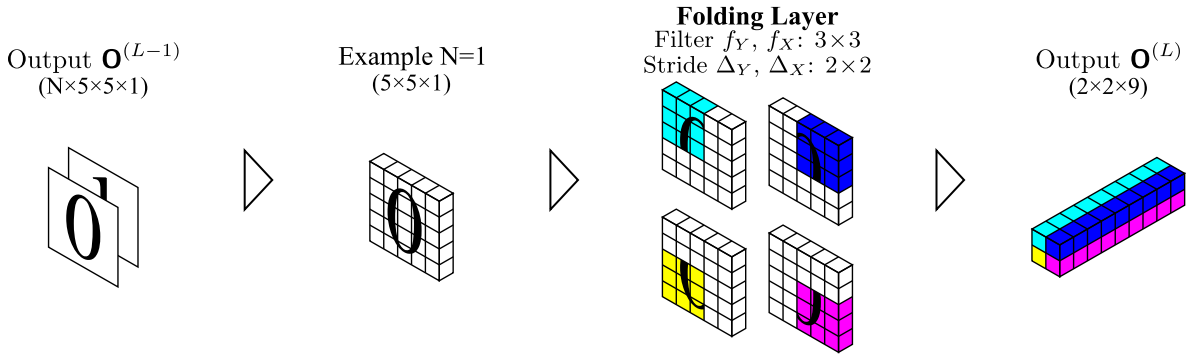

Figure 8.8: Exemplary visualization of the forward operation of a folding layer.

**Backward/Sampling** For generating samples, the folding layer has to invert the splitting process by merging the individual patches. In the simplest case, the filter size equals the stride. In this special case, merging is realized by assembling the individual patches. Merging gets more complicated when the patches overlap, as shown in figure 8.9 (reading direction: right to left). The individual patches provided as channel vectors (color coded in figure 8.9) from $\mathbf{I}^{(L+1)}$ need to be transformed back, based on the filter size $f_Y = f_X$. Various strategies can be applied for re-assembling the patches. One strategy, for example, would be to average the overlapping pixels. The merging procedure is subject to losses as illustrated on the left-hand side of figure 8.9 Therefore, a mechanism for improving the quality is presented later in section 8.4.2.5.

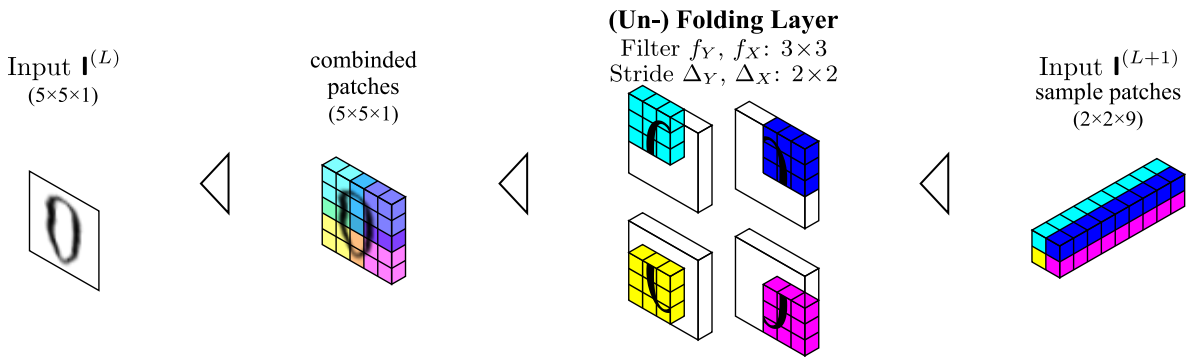

Figure 8.9: Exemplary visualization of the sampling operation of a folding layer.

### 8.4.1.2 Pooling Layer

**Forward/Estimation** The introduced pooling layer performs the same operations as the pooling layer in CNNs (see Zeiler and Fergus 2014). A standard max-pooling layer has the kernel size $k_Y^{(L)}$ and $k_X^{(L)}$ along with the stride parameters $\Delta_X^{(L)}$ and $\Delta_Y^{(L)}$. Depending on the defined parameters, the forward operation implements a aggregation/compression of several values that is subject to losses. Different methods can be applied for the reduction operation, e.g., maximum or average.

Figure 8.10 shows an example of the aggregation operation for max- and average-pooling. A $4 \times 4$ matrix is given on the left-hand side. Furthermore, a pooling layer with $k_Y^{(L)} = k_X^{(L)} = 2$ and a stride of $\Delta_X^{(L)} = \Delta_Y^{(L)} = 2$ is utilized. The resulting kernels are illustrated by different colors and the respective aggregated values on the right-hand side. Overlaps of kernels may occur, depending on the striding parameter. Especially the edges of the matrices require special treatment, e.g, padding with zeros.

**Backward/Sampling** To generate samples, the pooling operation needs to be reversed. The pooling parameters include the kernel and the stride ($k_Y^{(L)}$, $k_X^{(L)}$ and $\Delta_X^{(L)}$, $\Delta_Y^{(L)}$). A simple reverse operation for unpooling is illustrated in figure 8.11. One and the same input value is used several times depending on the pooling parameters. Again, other reverse operations could be realized. One example that requires additional internal parameters is proposed by David and Netanyahu (2016). In this case, the position of the highest value is stored and reused. In general, the information loss of the forward operation cannot be fully recovered. Besides the unpooling operations, other methods related to image processing can be utilized, e.g., the nearest-neighbor up-sampling (known as scaling).

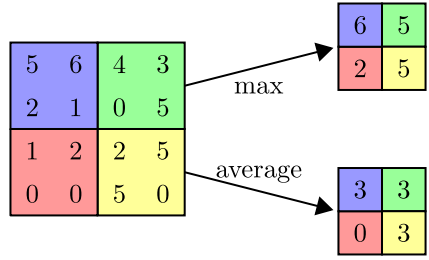

Figure 8.10: Exemplary visualization of the pooling operation.

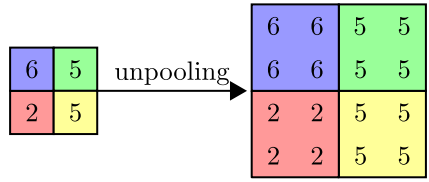

Figure 8.11: Exemplary visualization of the unpooling operation.

### 8.4.1.3   GMM Layer

**Forward/Estimation**   A single GMM layer represents an independent GMM consisting of $K$ components, as stated in section 8.1. The trainable parameters include the weights $\pi_k$, the means/centroids $\boldsymbol{\mu}_k$ and the covariances $\boldsymbol{\Sigma}_k$ where $k \in K$. A GMM layer is referred to as "convolutional", if each channel $(H \times W)$ of the previous output $\mathbf{O}^{(L-1)}_{NHWC}$ is mapped using the same parameters $\theta^{(L)}$. In figure 8.8, the output of a folding layer is illustrated. The colored channel vectors $(\mathbf{O}^{(L-1)}_{NWH:})$ correspond to the input of the convolutional GMM layer. Each component $k$ is thus able to learn from a patch of the predecessor layer and determines their individual affiliation to each component $k$. Alternatively, a non-convolutional GMM layer consists of separate parameters for each input channel, which requires more resources. However, non-convolutional GMM layers cause the number of parameters to be dependent on $W^{(L-1)} \times H^{(L-1)}$. They would thus strongly depend on the convolution parameters of the predecessor layer.

The output of a convolutional GMM layer comprises the responsibilities for each input patch. The determination of responsibilities is outlined in equation 8.21. In addition, the loss value as a separate output of a GMM layer can be valuable.

$$
\begin{aligned}
p_{NHWk}\big(\mathbf{O}^{(L-1)}\big) &= \mathcal{N}_k\big(\mathbf{O}^{(L-1)}_{NHW:}|\boldsymbol{\mu}_k, \boldsymbol{\Sigma}_k\big) \\
\mathbf{O}^{(L)}_{NHWK} &\equiv \frac{p_{NHWk}}{\sum_{i=1}^{W^{(L-1)} \times H^{(L-1)}} p_{NHWi}}
\end{aligned}
\tag{8.21}
$$

Equation 8.21: Determination of the convolutional responsibilities of a GMM layer.

The smoothed max-component approximation from section 8.3.3 is applied for training a GMM layer. The training is executed by the presented SGD-based approach (see section 8.3.4). Equation 8.22 describes the convolutional loss function $\mathcal{L}$, which takes each individual patch into account.

Figure 8.12 visualizes an exemplary GMM layer (reading direction: left to right). The output signal of the predecessor layer $\mathbf{O}^{L-1}$ corresponds to the folding operation in figure 8.8 (see section 8.4.1.1). Each color-coded channel vector represents a patch from the original input signal, e.g., an image. Due to the context of the visualization, it is assumed that the operation is performed for a single sample

$$\mathcal{L}_{HW}^{(L)} = \sum_n \log \sum_k \pi_k p_{NHWK}(\mathbf{O}^{(L-1)})$$

$$\mathcal{L}^{(L)} = \frac{\sum_{HW} \mathcal{L}_{HW}^{(L)}}{H^{(L-1)} W^{(L-1)}}$$

(8.22)

Equation 8.22: Determination of the convolutional loss $\mathcal{L}$ of a GMM layer.

($N = 1$). In figure 8.12, the GMM layer consists of $K = 3 \times 3 = 9$ components.

The GMM parameters $\theta$, as shown in figure 8.12, include the means $\boldsymbol{\mu}$ (top) and the covariances $\boldsymbol{\Sigma}$ (bottom). The weights $\pi_k$ belong to the free parameters, which are omitted here for the sake of simplicity. Each parameter ($\boldsymbol{\mu}$, $\boldsymbol{\Sigma}$) is arranged in a two-dimensional grid $3 \times 3$ (bold face), as described in section 8.3.3. The individual means and variances are transformed into $3 \times 3$ image in order to ease their illustration and interpretation.

Based on the (already initialized) internal parameters and the output of the previous layer $\mathbf{O}_{NHWC}^{(L-1)}$, the corresponding responsibilities $\gamma$ (see equation 8.22) are determined. The responsibilities are passed on as the output $\mathbf{O}^{(L)}$ of a GMM layer. The output format corresponds to the $HW$ of the previous layer and the number of components $K$. Thus, the output contains the responsibility of each component $k$ for each patch. The complete inference process for a multi-layer DCGMM is described in section 8.4.2.2.

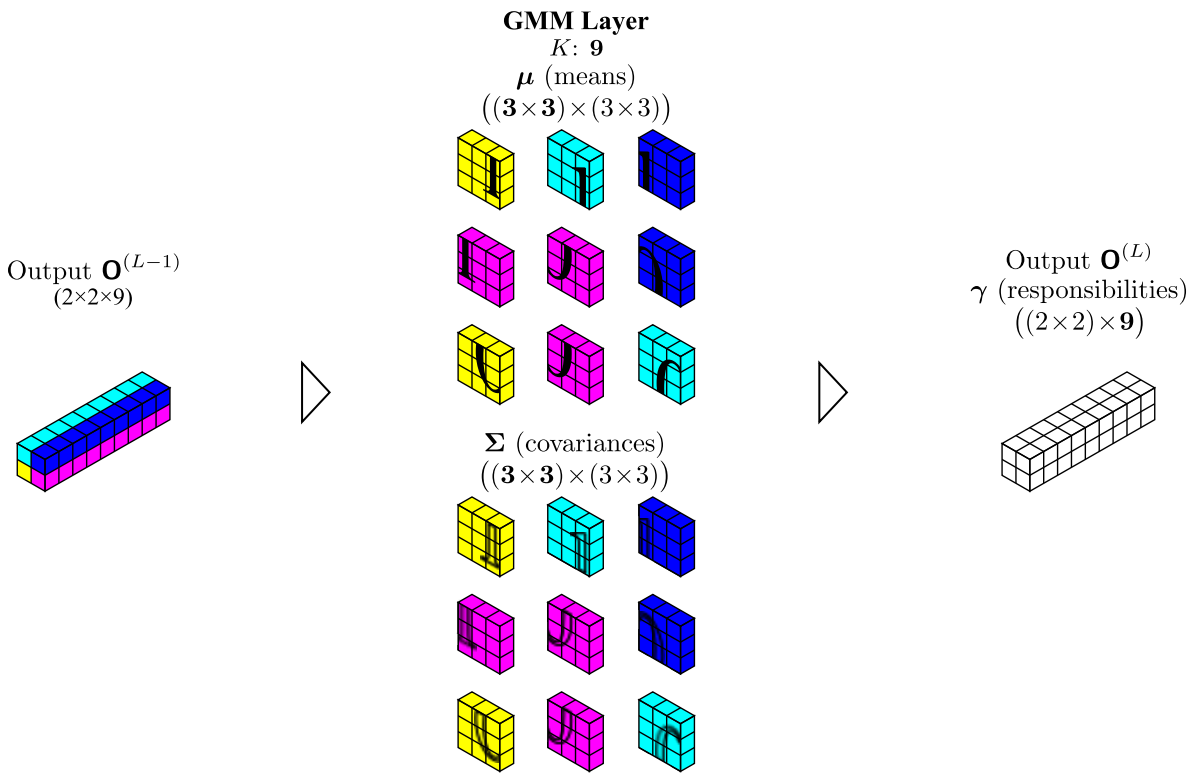

Figure 8.12: Exemplary visualization of the forward operation of a GMM layer.

**Backward/Sampling**  The backward operation of a GMM can be used to generate samples from the derived data distribution. Initially, the input $\mathbf{I}^{(L+1)}$ of the successor layer is used for component selection. The sampling process is presented in figure 8.13 (reading direction: right to left). Each channel vector of $\mathbf{I}^{(L+1)}$ defines the component out of which a sample should be drawn. For a non-folding operation, a GMM component represents complete samples. In the unsupervised case, a multinomial distribution can be used for the component selection. A combination of the multinomial distribution and the weights $\boldsymbol{\pi}$ can be considered a selection criterion.

For the case depicted in figure 8.13, the input signal $\mathbf{I}^{(L+1)}_{NHWC}$ defines for which patch a component $k$ should be chosen. The components that are supposed to be selected are highlighted in green. The transformed selection is shown below in order to match the arrangement of the representation of the internal parameters $\theta$ ($\boldsymbol{\mu}$ and $\boldsymbol{\Sigma}$). Likewise, the selection of the corresponding GMM components is outlined in green. For each $HW$ position in the input signal, a sample is generated for each patch $x_i$ by drawing samples from the different multinomial distributions ($\boldsymbol{x}_i \sim \mathcal{N}(\boldsymbol{\mu}_i, \boldsymbol{\Sigma}_i)$). The generated patches are displayed as transformed image for visualization purposes. In the presented case, the input results in the generation of an image constructed from 4 patches. The patches are passed in the $NHWC$ format as input to the next layer (e.g., (un-)folding layer shown in figure 8.8).

Various procedures can be executed for the selection of components. In the scenario presented in figure 8.13, the channels of the input signal precisely define which component has to be selected. However, a specific selection criterion is not always obvious, as the input signal consists of real numbers and, in the best case, represents a probability distribution.

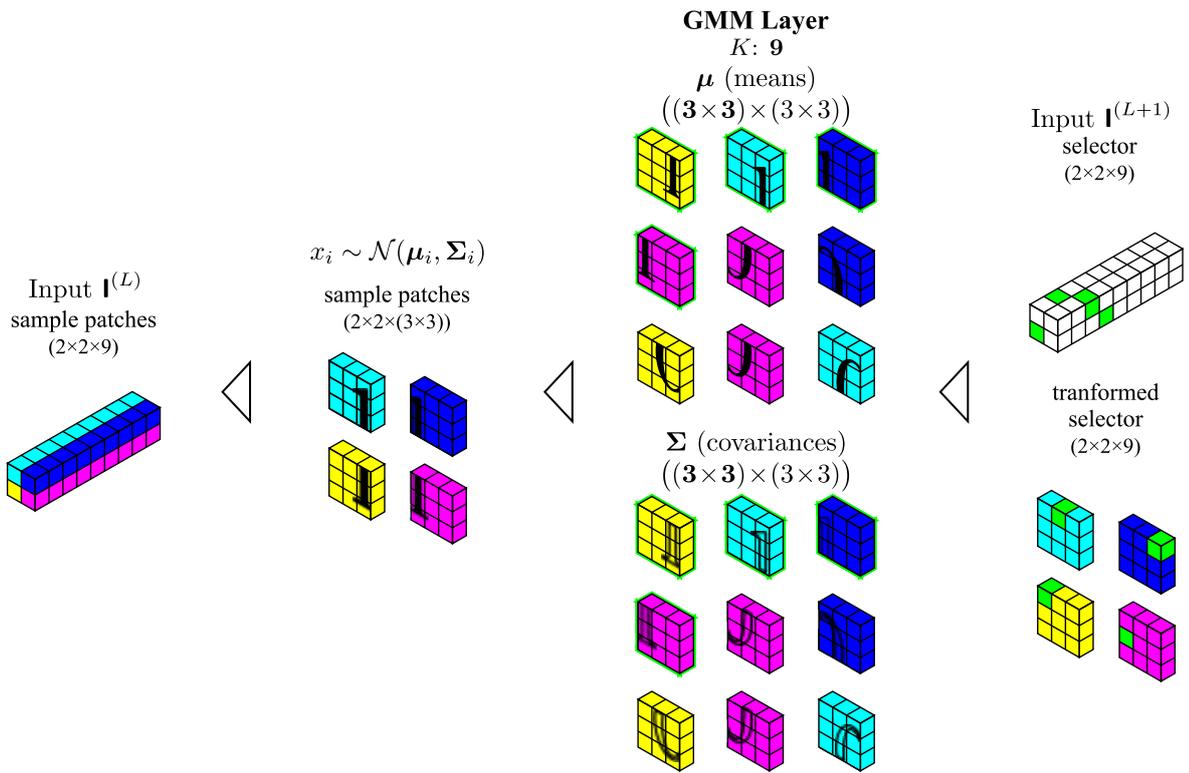

Figure 8.13: Exemplary visualization of the sampling operation of a GMM layer.

The input signal $\mathbf{I}^{(L+1)}$ is primarily used to select the components for generating samples. If the input signal represents probabilities or is converted accordingly, it serves as selection criterion. Samples can be generated from all components by using probabilities. In order to limit the selection signal, a method referred to as top-$S$-sampling is introduced. The top-$S$-sampling method describes the use of the $S$ highest activations as selection criterion. This method prevents random sampling from less likely components.

#### 8.4.1.4  Classification Layer

**Forward/Estimation**  A classification layer is defined by a conventional linear layer that allows a label's assignment to an input signal. DCGMMs can be categorized as supervised model due to their ability to classify samples. For the training of the classification layer, the labels of the samples need to be given. The internal parameters $\theta$ comprise the weights $\boldsymbol{W}$ and the bias $\boldsymbol{b}$. Therefore, linear layers consist of standard artificial neurons as used in DNNs (see section 2.2.1.2).

The linear layer is trained by optimizing the cross-entropy loss by SGD (see section 2.2.2.1). The logits (output of $L-1$, e.g., responsibilities of a GMM layer) are flattened ($\mathbf{O}^{(L-1)} \in \mathbb{R}^{N \times HWC}$) and transformed by an affine transformation ($\mathbf{O}^{(L)} = \hat{\mathbf{O}}^{(L-1)}\boldsymbol{W}^{(L)} + \boldsymbol{b}^{(L)}$). As a result, each input sample can be assigned a class label based on the output of the previous layer. The dimension and number of variables of the classification layer depends on the number of input parameter.

In figure 8.14, a binary classification problem is depicted, i.e., zero (blue) and one (green) (reading direction: left to right). In general, the linear layer can be used for multi-class classification. The previous layer $\mathbf{O}^{(L-1)}$ is assumed to be the output of a GMM layer (responsibilities $\gamma$) consisting of $K = 4$ components. Thus, the probability that a sample has been created from a certain component is the expected information value of the previous output. The linear layer was trained according to a component's responsibility for a specific class. This assumption can only be satisfied if the model is converged. Based on the readout layer and the one-hot encoding (see section 2.2), an appropriate class label can be assigned to a sample.

**Linear Layer**

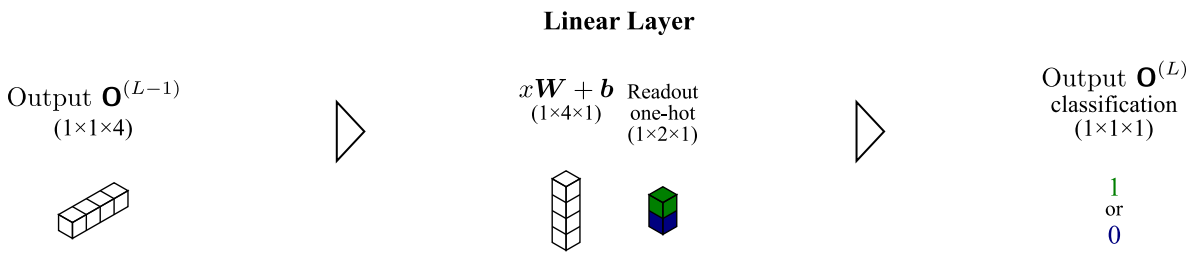

Figure 8.14: Exemplary visualization of the forward operation of a classification layer.

**Backward/Sampling**    The backward operation of the classification layer inverts the labeling process. Thus, it is possible to create samples by specifying a certain class, which is referred to as conditional sampling. For the generation of a sampling signal, the classification is reversed by $\mathbf{I}^{(L-1)} := \boldsymbol{W}^{(L)\top}\mathbf{I}^{(L)} - \boldsymbol{b}^{(L)}$. Given is a desired class as input that is encoded as a one-hot vector. Subsequently, it is converted as an input signal $\mathbf{I}^{(L-1)}$ for the predecessor layer.

The sampling process is illustrated in figure 8.15 (reading direction: right to left). For visualization purposes, it is assumed that the predecessor layer ($L{-}1$) is a GMM layer, which passes on responsibilities during training. The reverse input signal ($\mathbf{I}^{(L)}$) provides the basis for sampling by selecting a GMM component for a specific associated class. In figure 8.15, a sample from the class zero should be created. The linear layer converts the one-hot encoded class into an input signal for the layer above. A complete example for the sampling process of a full DCGMM is presented in section 8.4.2.4.

**Linear Layer**

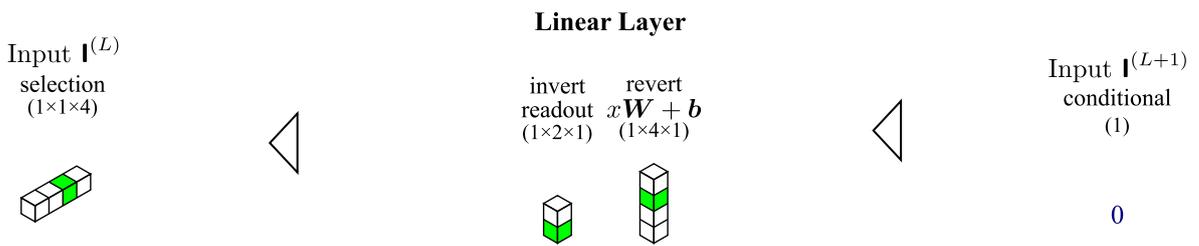

Figure 8.15: Structural visualization of the linear layer for sampling from a DCGMM.

## 8.4.2    DCGMM Functionalities

DCGMMs offer a variety of functions which are presented in this section. At first, a strategy is proposed for training DCGMMs end-to-end. Subsequently, the inference process for classification is described in more detail. Another function is the ability to detect outliers, which is related to density estimation. Finally, the process of generating samples is outlined in greater detail. The respective details refer to conditional sampling and a strategy to improve the quality of the samples (image sharpening). Especially the conditional sampling and outlier detection offer many applications in the area of CL, e.g., pseudo-rehearsal mechanism or the detection of tasks boundaries.

#### 8.4.2.1 End-to-End Training

DCGMMs can be trained by SGD, even if they are not convolutional and do not have deep structures. However, not a single objective function is used for the training process, as it is the case for DNNs or CNNs. Instead, a individual loss function $\mathcal{L}^{(L)}$ is used to optimize the parameters $\theta^{(L)}$ of each GMM layer $L$. Note that the process is not layer-by-layer but end-to-end. End-to-end means that one batch of samples is used to optimize parameters in all layers. Layer-by-layer corresponds to a layer-wise training, where always one layer after another is trained until convergence is achieved. Plain SGD is used for optimization, since advanced SGD strategies lead to problems (e.g., RMSProp (Tieleman and Hinton 2012) or Adam (Kingma and Ba 2015)). The only peculiarity of the GMM training is that for some training iterations the centroids are adjusted first, before covariances are adapted. The divided training procedure accelerates the convergence behavior.

#### 8.4.2.2 Inference

In the following, the inference process is illustrated by using a simple DCGMM model consisting of three layers (see figure 8.16, reading direction: left to right). The exemplary DCGMM architecture connects the basic layers shown in section 8.4.1, excluding the pooling layer. First of all, a folding layer with the filter $f_Y = f_X = 5$ and stride $\Delta_Y = \Delta_X = 1$ is defined. Thus, the input signal (grayscale image) with $5 \times 5$ pixels is dumped into the $1 \times 1 \times 25$ dimensional channel vector. The vector is represented by the long bar below the folding layer (see section 8.4.1).

The GMM layer consists of $K = 9$ components, whereas only the means are presented for the sake of simplicity. The model is already converged, which becomes evident by means of the drafted prototypes/centroids. In section 8.4.1, an input signal in form of an image representing the number 0 is applied. The initial folding layer only conducts a transformation into a one-dimensional vector. The vector activates the components of the GMM layer differently, whereas one component shows a very high activation (marked in green). The activations (correspond to the responsibilities) are forwarded to the classification layer. For high data dimensionalities, there is always one responsibility that is $\approx 1.0$ for an appropriately converged model. This conclusion is empirically verified in section 8.5.1.3. In addition, the transformed responsibilities are depicted in section 8.4.1. The last linear layer has "learned" from the labeled data and the respective responsibilities from the GMM layer that class zero is assigned for the active (central) component. Thus, the associated class can be assigned to the sample.

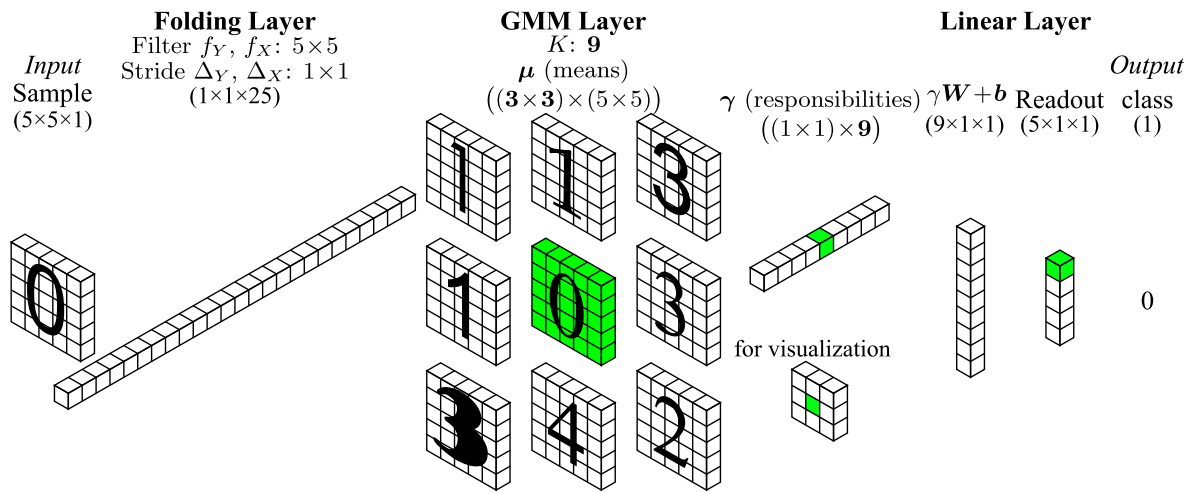

Figure 8.16: Exemplary DCGMM instance for visualizing the inference process.

#### 8.4.2.3 Outlier Detection

Outlier detection is a frequently used function of unsupervised learning/clustering methods, such as GMMs. It is utilized, for example, to identify samples that are not part of the training data, e.g., caused by measurement errors. Another example is the detection of anomalies in various areas such as attacks on IT systems. The detection of outliers is more confident when a model is in a stable phase, for example, after training. However, the definition of a sample as inlier or outlier is problem-dependent and therefore a hyper-parameter. For detection, the log-likelihood $\mathcal{L}(\boldsymbol{x})$ of a sample can be compared to the long-term average log-likelihood $\mathbb{E}_n(\mathcal{L})$ of the model. In addition, the variance $\text{Var}_n(\mathcal{L})$ of a batch can serve as an indicator for detecting outliers.

An outlier (or inlier) can be determined as specified in equation 8.24 by each GMM layer $L$ at any position of a DCGMM. The influence of the variance on the threshold can be defined by the hyper-parameter $\mathcal{C}$ (see equation 8.24), For the convolutional part, an outlier can also be determined for individual patches ($HW$) (see section 8.4.1.3). Considering the parameter configuration from section 8.4.2.2, where no convolution takes place in the first folding layer, it is specified that $H = W = 1$. Thus, the proposed configuration defines that the entire input sample is specified as inlier or outlier.

$$O_{NHW:}^{(L)} \equiv \mathbb{E}_n\left[\mathcal{L}_{NHW:}^{(L)} - \mathcal{C}\sqrt{\text{Var}_n(\mathcal{L}_{NHW:}^{(L)})}\right].$$

(8.23)

Equation 8.23: Determination of the outlier detection threshold.

The density signal $O_{NHW:}^{(L)}$ can be used in equation 8.24 to determine whether a sample is an in- or outlier. Since $\mathcal{C}$ is problem-specific, $\mathcal{C}$ can only be determined by cross-validation. The larger $\mathcal{C}$, the less restrictive the outlier detection method. The stated outlier detection procedure is easy to implement, although more intelligent definitions of the threshold can be realized.

$$f_{\text{outlier}}(\boldsymbol{x}) = \begin{cases} \text{outlier} & \text{if } \mathcal{L}_{HW}^{(L)} < O_{HW}^{(L)} \\ \text{inlier} & \text{if } \mathcal{L}_{HW}^{(L)} \geq O_{HW}^{(L)} \end{cases}$$

(8.24)

Equation 8.24: Determination of outliers and inliers.

### 8.4.2.4 Sampling

Sampling is an interesting functionality that can be used in a wide range of applications, e.g., for CL approaches based on pseudo-rehearsal mechanisms. The sampling process for a model consisting of three layers is illustrated in figure 8.17 (reading direction: right to left). Samples can be created in a conditional manner by adding a linear classification layer on top. The classification layer allows to generate an input signal for the predecessor (GMM) layer in order to specifically generate a sample from a certain component. Thus, samples can be generated selectively from a specific class.

In order to generate a sample from the class zero, the input class coded as one-hot vector needs to be inverted (marked green on the right-hand side). The resulting signal serves as a selector for a corresponding component $k$ (illustrated in a reshaped form and highlighted in green). Each GMM component $k$ is represented in figure 8.18 in a reshaped manner in order to ease interpretation. The selection strategy allows to create a sample from a certain component (center) by sampling from a multivariate normal distribution ($\boldsymbol{x}_k \sim \mathcal{N}(\boldsymbol{\mu}_k, \boldsymbol{\Sigma}_k)$). The generated sample is also reshaped in figure 8.18. Finally, the sample generated by the GMM layer needs to be reshaped into the original input format (image). The reshaping process is the inverse transformation of the folding layer (left-hand side in figure 8.17).

The inversion of the linear layer is adapted for DCGMMs which makes it suitable for the selection of a GMM component. The applied softmax function is shift-invariant which allows for the inversion based on a constant $\boldsymbol{s}$. For the inverse of the weight matrix $\boldsymbol{W}$, it is assumed that the columns are orthogonal, which is an approximation. The signal needs to be defined within the value range $[0, 1]$ and have a unit sum (like the responsibilities). Therefore, the signal represents the selection of a certain class. Thus, the sampling control signal corresponds to the expected posterior probabilities of a GMM layer for a given class. At the same time, the weights $\boldsymbol{\pi}$ are ignored for generating the samples.

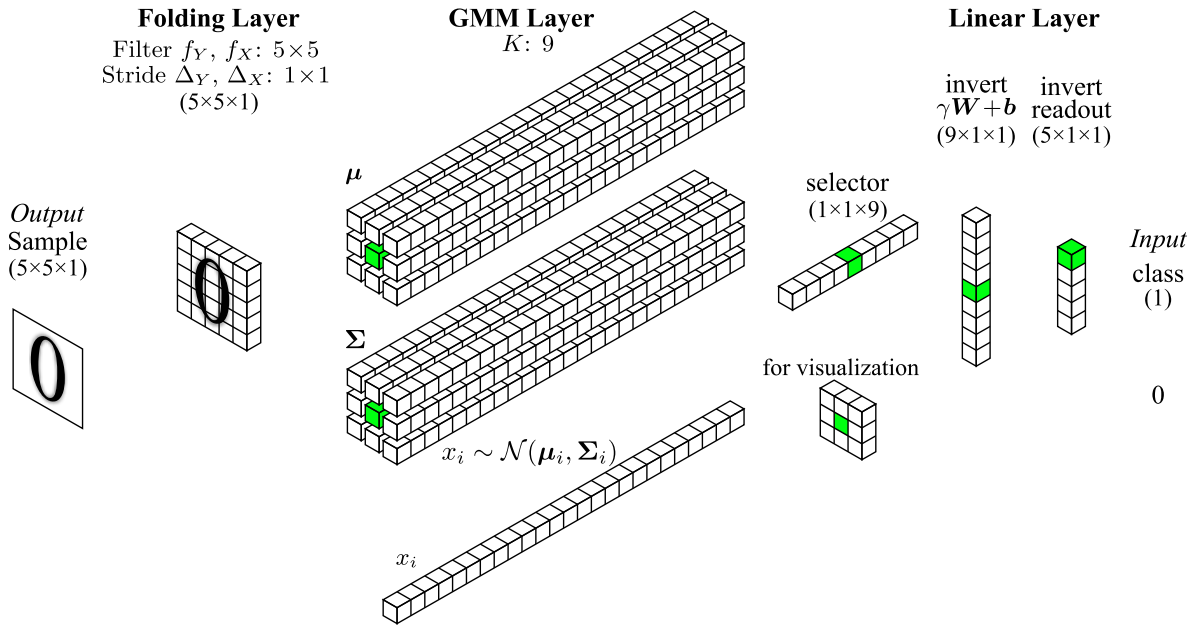

Figure 8.17: Exemplary DCGMM model for visualizing the sampling process.

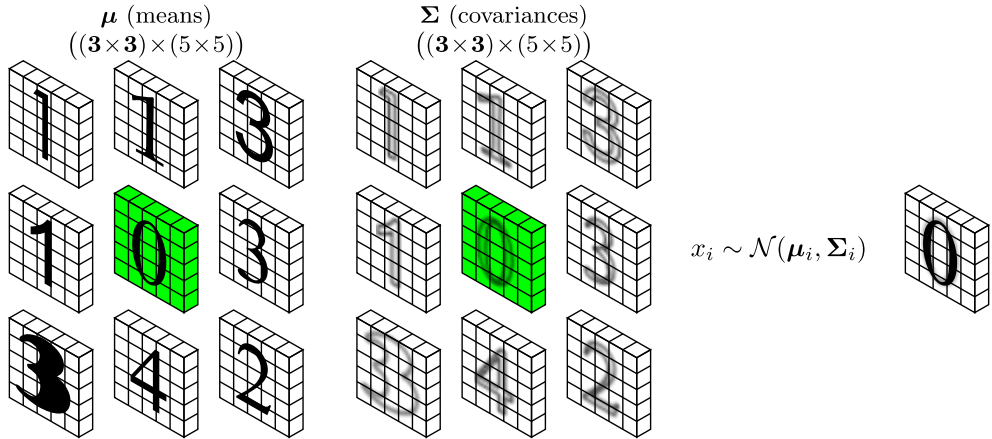

Figure 8.18: Exemplary GMM component visualization.

Nevertheless, the component weights $\boldsymbol{\pi}$ can be used in addition to the given selection criterion.

$$\mathsf{l} = \boldsymbol{s}(\boldsymbol{W}\mathsf{l} + \boldsymbol{b}) \Rightarrow \boldsymbol{i} \approx \boldsymbol{W}^{\top}\big(\boldsymbol{s}^{-1}(\mathsf{l}) + \boldsymbol{k} - \boldsymbol{b}\big) \qquad (8.25)$$

Equation 8.25: Inversion of the linear layer for conditional sampling.

### 8.4.2.5   Sharpening

When generating samples with DCGMMs, information may get lost by using folding or pooling layers. The information loss is either due to the compression of the pooling operation or the overlapping patches during folding. Sampling, as shown in section 8.4.2.4, usually starts with the highest layer, e.g., in the conditional case with the linear classification layer. The generated signal for sampling is forwarded end-to-end until the first layer is reached. The quality of the generated samples suffers from the non-invertable mapping, e.g., the composition of overlapping patches. In order to counteract the quality loss, the next higher GMM layer can adapt the resulting signal. The input signal is propagated forward (in the reverse direction). Accordingly, $g$ gradient ascent steps with a step size of $\epsilon_S$ are performed. Due to the application of the resulting gradients, the signal is improved as stated in equation 8.26. As a result, the correlations mapped by the corresponding GMM layer $L$ can be partially restored. To conclude, all GMM layers in a DCGMM architecture can be useful for the sharpening process, whereas the highest GMM layer generally contributes the most to the improvements. After sharpening, the adjusted signal $\mathbf{l}^{(L-1)}$ is passed to the layer above $L-2$.

$$\mathbf{l}^{(L-1)} \rightarrow \mathbf{l}^{(L-1)} + \frac{\partial \mathcal{L}^{(L)}}{\partial \mathbf{l}^{(L-1)}} \epsilon_{\mathcal{S}} \tag{8.26}$$

Equation 8.26: Gradient ascend step to compensate quality losses (image sharpening).

## 8.5   Evaluation

In order to validate the presented SGD-based training approach for GMMs, it is compared to a variant of the standard EM training algorithm. sEM is a modified method for training GMMs in an incremental manner introduced by Cappé and Moulines (2009). The goal is not to outperform sEM, but to obtain equivalent results that demonstrate the validity of the SGD-based GMM training approach. Various experiments are conducted which allow a direct comparison of the clustering abilities as well as the final log-likelihood values. Thus, the influence of the annealing procedure on the overall result is also investigated. An evaluation of results always needs to take the initialization process into account.

For sEM, a conventional data-driven initialization is performed by using k-means, whereas the novel approach is applied without this initialization. Considering that a data-driven initialization requires the complete dataset, the stochastic variant of EM seems redundant. Even if the initialization is beneficial for sEM, the comparison to the SGD approach is valid. The results of the SGD and sEM training are summarized in section 8.5.1. Furthermore, the experimental results for the deep convolutional variant (DCGMM) are presented in section 8.5.2. The different functionalities of the DCGMM are investigated by conducting experiments with various parameter configurations of DCGMMs.

The following conditions apply to all presented results, as each experiment is repeated ten times with the same parameter configuration. The resulting values are averaged as meta-results (including the variance). Thus, the dependency of results on the initial state (random initialization) should be neglectable. In both variants, SGD and sEM, diagonal precision matrices are used. Otherwise, the presented datasets in section 6.1 would exceed the memory and computing resources. This is due to the relatively "high" data dimensionalities of the samples. Moreover, all samples within each dataset are normalized to the value range $[0, 1]$ by a min-max normalization (see section 6.1).

### 8.5.1   SGD Experiments

For the validation of the SGD-based approaches, flat DCGMMs as simple GMMs variant are examined first in order to allow a comparison. In the following, the parameter configurations used for both training methods are specified. The number of GMM components is set to $K = 64$ and a batch size of $\mathcal{B} = 1$ is applied.

However, different procedures are used for the initialization of SGD training. Precisely, the cluster centers are initialized by $\boldsymbol{\mu} = \mathcal{U}(-\mu^i, +\mu^i)$, whereas $\mu^i = 0.1$ (if not stated otherwise). $D_{\max}$ is set to 20 for the initialization of the precision matrices $\boldsymbol{P}$.

In case of sEM training, the k-means clustering algorithm initializes the GMM components. In contrast to sEM, the SGD approach does not require a data-driven initialization, e.g., via k-means. However, sEM would yield even poorer results without a data-driven initialization (as shown later). Besides the fixed parameters, hyper-parameters were tuned by a grid-search. The following hyper-parameters were explored with regard to sEM: $\alpha \in \{0.01, 0.5\}$, $\alpha_0 \in \{0.05, 0.1\}$, $\rho_{\min} \in \{0.001, 0.0001\}$.

#### 8.5.1.1   Clustering Performance

In order to compare the clustering performance of sEM and SGD training, two clustering metrics are applied: The Davies-Bouldin score proposed by Davies and Bouldin (1979), as well as the Dunn index suggested by Dunn (1973). In general, clustering metrics are based on the results of cluster center assignment. In each case, the best results from the grid search experiments are compared. In case of the Davies-Bouldin score, less is better – whereas more is better with regard to the Dunn index. For reasons of performance (regarding the computation of the metrics), the first 1000 samples of the test dataset were utilized for each measurement. The results for both metrics and both training methods (SGD and sEM) are presented in table 8.1.

Table 8.1 shows a balanced distribution of the best values (marked in bold if better than half a standard deviation). Even though the SGD approach shows slightly better results, it can be noted that both methods are comparable with regard to their clustering performance. The SGD-based approach is therefore considered valid in terms of clustering performance.

Table 8.1: Clustering performance comparison of SGD and sEM training.

| Metric Algo. Dataset | Davies-Bouldin score | | | | Dunn index | | | |
|---|---|---|---|---|---|---|---|---|
| | SGD | | sEM | | SGD | | sEM | |
| | *mean* | *std* | *mean* | *std* | *mean* | *std* | *mean* | *std* |
| MNIST | 2.50 | 0.04 | **2.47** | 0.04 | 0.18 | 0.02 | 0.16 | 0.02 |
| FashionMNIST | **2.06** | 0.05 | 2.20 | 0.04 | 0.20 | 0.03 | 0.19 | 0.02 |
| NotMNIST | 2.30 | 0.03 | **2.12** | 0.03 | 0.15 | 0.03 | 0.14 | 0.04 |
| Devanagari | **2.60** | 0.04 | 2.64 | 0.02 | **0.33** | 0.01 | 0.27 | 0.04 |
| SVHN | **2.34** | 0.04 | 2.41 | 0.03 | 0.15 | 0.02 | 0.15 | 0.02 |

#### 8.5.1.2 Robustness of Initialization

Another investigation focuses on the robustness with respect to initialization of GMMs. In general, EM-based methods, i.e., sEM, are sensitive to initialization (see Baudry and Celeux (2015)). In order to demonstrate the robustness and the advantage of the annealing procedure, the effects of different initial states are investigated.

After initialization, GMMs are trained on 9 classes (1 to 9) for 3 epochs. For the centroid's ($\boldsymbol{\mu}$) initialization, three strategies are investigated: Random ($\mu^i \in \{0.1, 0.3, 0.5\}$), non-random (training one epoch on samples of class 0) and without annealing. The final log-likelihood values serve as a performance criterion. The initialization of the precisions remains the same, as described above. If the chosen initial precision values are too small, undesirable solutions will result as stated in figure 8.6. The outcomes of the experiments are presented in table 8.2.

The final log-likelihood values show that the training with the annealing scheme is very robust with respect to the initialization. This is especially true for random initialization. Small initialization values $\mu^i$ are considered as advantageous. Likewise, comparable results can be obtained for a non-random initialization. Compared to the switched-off annealing process, the added value becomes clearer. The switched off annealing mechanism mainly leads to sparse-component solutions, like for randomly initialized sEM trained GMMs.

Table 8.2: Different random and non-random centroid initializations on SGD training.

| Initialization Dataset | random | | | | | | non-random | | no annealing |
|---|---|---|---|---|---|---|---|---|---|
| | $\mu^i = 0.1$ | | $\mu^i = 0.3$ | | $\mu^i = 0.5$ | | init class 0 | | $\mu^i = 0.1$ |
| | *mean* | *std* | *mean* | *std* | *mean* | *std* | *mean* | *std* | *mean* |
| MNIST | 205.47 | 1.08 | 205.46 | 0.77 | 205.68 | 0.78 | 205.37 | 0.68 | 124.1 |
| FashionMNIST | 231.22 | 1.53 | 231.58 | 2.84 | 231.00 | 1.11 | 229.59 | 0.59 | 183.0 |
| NotMNIST | −48.41 | 1.77 | −48.59 | 1.56 | −48.32 | 1.13 | −49.37 | 2.32 | -203.8 |
| Devanagari | −15.95 | 1.59 | −15.76 | 1.34 | −17.01 | 1.11 | −22.07 | 4.59 | -263.4 |
| Fruits 360 | 12 095.80 | 98.02 | 12 000.70 | 127.00 | 12 036.25 | 122.06 | 10 912.79 | 1 727.61 | 331.2 |
| SVHN | 1 328.06 | 0.94 | 1 327.99 | 1.59 | 1 328.40 | 1.17 | 1 327.80 | 0.94 | 863.2 |
| ISOLET | 354.34 | 0.04 | 354.36 | 0.04 | 354.36 | 0.04 | 354.20 | 0.05 | 201.5 |

### 8.5.1.3   Density Estimation

Another functionality of GMMs is the estimation of a sample's density. Again, results of the sEM and SGD based training method are compared. The maximum achieved log-likelihood at the end of the training process serves as a reference point. Again, different datasets from section 6.1 are used for the experiments. For GMM training, each method can iterate 3 training epochs on each dataset. This should be a sufficient training time for each algorithm to converge using a batch size of $\mathcal{B}=1$.

In figure 8.19, the resulting centroids/prototypes (or means $\boldsymbol{\mu}$) from one of the conducted SGD experiments are visualized. For most datasets, the prototypes are clearly visible, e.g., handwritten digits in MNIST. However, for the SVHN dataset in figure 8.19b, the number of available prototypes ($K = 64$) is not sufficient. Similar prototypes overlap, resulting in poor log-likelihood values as visible in table 8.3. As shown in table 8.3, both approaches produce similar results, except that sEM cannot handle the dimensionality of SVHN ($3\,072$) and Fruits 360 ($30\,000$). Another difference is that the applied annealing scheme of the SGD approach leads to a topological sorting of the GMM components (similar elements are closer together, see figure 8.19).

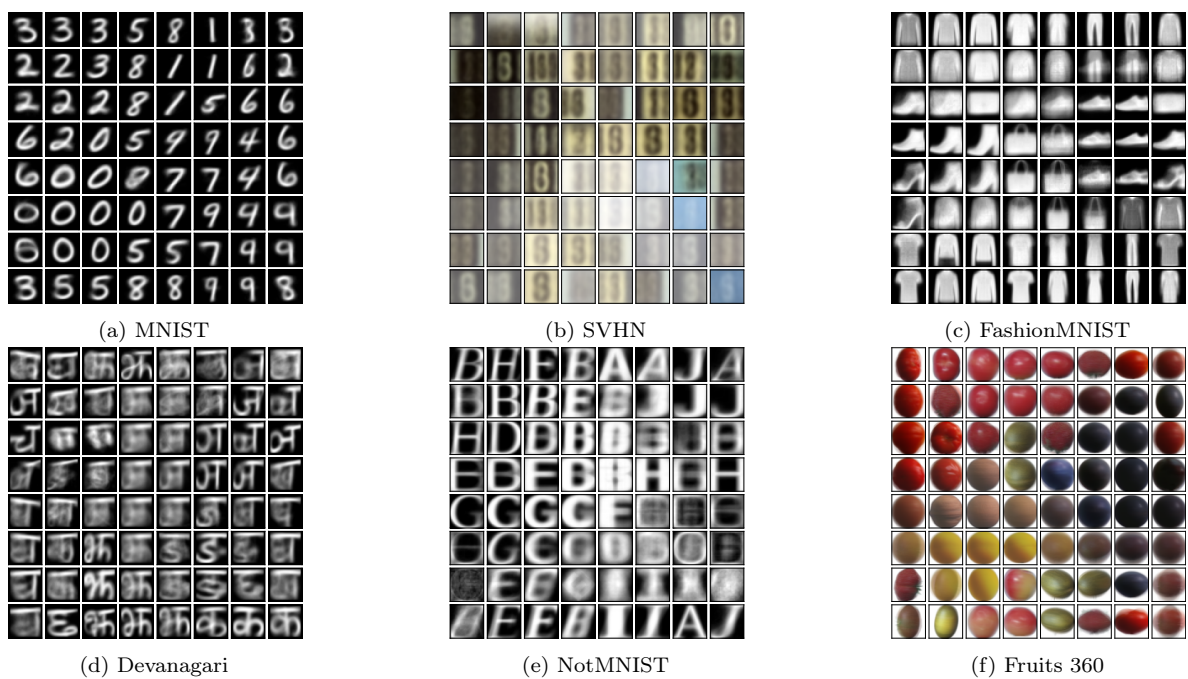

(a) MNIST                          (b) SVHN                          (c) FashionMNIST

(d) Devanagari                     (e) NotMNIST                      (f) Fruits 360

Figure 8.19: Exemplary results of centroids learned by SGD for different datasets.

The comparison of the final log-likelihood values in table 8.3 indicates the superiority of the SGD-based approach over sEM. All of the compared values that show superiority by more than half of a standard deviation are highlighted in bold. The focus is not to challenge the sEM or other GMM training algorithms, but to achieve equal values. In addition, the average responsibilities of the BMUs for the test dataset (for SGD) are presented in table 8.3. These results show that for the average of the tested samples a single BMU (GMM component) is responsible – as expected for high-dimensional data. Despite a k-means initialization that requires the complete dataset and a lot of computation time, the SGD method seems superior. The superiority of the SGD approach is especially true for the dataset with the highest dimensionalities (SVHN: $3\,072$ and Fruits 360: $30\,000$ dimensions). With the exception of SVHN, the number of components is insufficient for both training methods. Insufficient GMM components are not a fundamental problem of the training method, but of the complexity of the dataset in relation to the available number of components $K$.

In order to illustrate the problem of the sEM's poor training results, the derived centroids for the SVHN and Fruits 360 dataset are depicted in figure 8.20. It becomes obvious that a sparse-component solution has emerged for both datasets, as many components have low brightness values (on all RGB channels). The visualized solution explains the lower log-likelihood results in table 8.3 for these two datasets. Despite the k-means initialization with already meaningful prototypes being derived from the dataset, sEM training can lead to a sparse-component solutions.

Table 8.3: Final log-likelihood comparison of SGD and sEM training.

| Algo. Dataset | **SGD** | | | **sEM** | |
|---|---|---|---|---|---|
| | $\varnothing \max p_{k*}$ | *mean* | *std* | *mean* | *std* |
| MNIST | 0.992 674 | 216.6 | 0.31 | 216.8 | 1.38 |
| FashionMNIST | 0.997 609 | **234.5** | 2.28 | 222.9 | 6.03 |
| NotMNIST | 0.998 713 | **−34.7** | 1.16 | −40.0 | 8.90 |
| Devanagari | 0.999 253 | **−14.6** | 1.09 | −13.4 | 6.16 |
| Fruits 360 | 0.999 746 | **11754.3** | 75.63 | 5 483.0 | 1 201.60 |
| SVHN | 0.998 148 | **1329.8** | 0.80 | 1 176.0 | 16.91 |
| ISOLET | 0.994 069 | 354.2 | 0.01 | 354.5 | 0.37 |

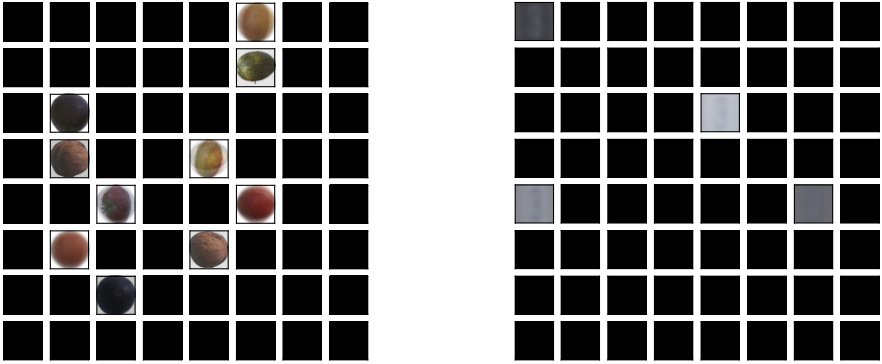

Figure 8.20: Visualization of sparse-component solutions for sEM for Fruits 360 and SVHN.

## 8.5.2 DCGMM Experiments

In this section, more complex DCGMMs architectures consisting of multiple layers are investigated. As presented in section 8.4.1, different types of layers are connected in different combinations and with various parameter configurations. An initial investigation is performed with DCGMMs consisting of two and three GMM layers. Different structures of the DCGMM architectures are defined in table 8.4.

In the following, the individual layers and their parameter configuration are defined in table 8.4. Folding layers (F) are configured by the parameters $f_Y, f_X$ (patch size) and $\Delta_Y, \Delta_X$ (stride). GMM layers (G) are specified by the number of Gaussian components $K$. (Max-)Pooling layers (P) are defined by the kernel and the striding size: $k_Y, k_X$, $\Delta_Y, \Delta_X$. Classification layers (C) are omitted, because the application and functionality is investigated in the next chapter. The number of training iterations is set to $\mathcal{E} = 25$ (epochs) for all experiments, unless stated otherwise. Likewise, sharpening is performed for $g = 1000$ iterations with the sharpening rate $\epsilon_{\mathcal{S}} = 0.1$.

For reasons of layer parameterization, dataset dependent configuration values are given in table 8.4. In order to omit the adjustment of the folding operation to different datasets, only datasets with the same structure are investigated. However, an investigation of all DCGMM architectures for all presented datasets (see section 6.1) would exceed the resources. For this reason, two structurally similar datasets are investigated (MNIST and FashionMNIST).

Table 8.4: Configurations of different DCGMM architectures.

| ID layer | 1L | 2L-a | 2L-b | 2L-c | 2L-d | 2L-e | 3L-a | 3L-b |
|---|---|---|---|---|---|---|---|---|
| 1 | F(28,28,1,1) | F(20,20,8,8) | F(7,7,7,7) | F(8,8,2,2) | F(28,28,1,1) | F(4,4,2,2) | F(3,3,1,1) | F(28,28,1,1) |
| 2 | G(25) | G(25) | G(25) | G(25) | G(25) | G(25) | G(25) | G(25) |
| 3 | | F(2,2,1,1) | F(4,4,1,1) | F(11,11,1,1) | F(1,1,1,1) | F(13,13,1,1) | P(2,2) | F(1,1,1,1) |
| 4 | | G(36) | G(36) | G(36) | G(36) | G(36) | F(4,4,1,1) | G(25) |
| 5 | | | | | | | G(25) | F(1,1,1,1) |
| 6 | | | | | | | P(2,2) | G(25) |
| 7 | | | | | | | F(6,6,1,1) | |
| 8 | | | | | | | G(49) | |
| *comment* | vanilla GMM | 1 conv. layer | 1 conv. layer | 1 conv. layer | no convolutions | 1 conv. layer | 2 conv. layers | no convolutions |

### 8.5.2.1 Interpretability of DCGMMs

Before presenting the experimental results, the interpretation of a two-layer DCGMMs is introduced. In order to support the interpretation, the individual GMM layers (2 and 4) of the 2L-a architecture (stated in table 8.4) are visualized. The model is trained with samples from the MNIST dataset (classes 0 to 4). The first folding layer ensures that 4 overlapping patches (two in $x$- and two in $y$-direction) of the size $20 \times 20$ are extracted from each input image ($\Delta_Y = \Delta_X = 8$). The left part of figure 8.21 represents the derived prototypes (each $20 \times 20$ dimensions) of the first GMM layer. On the very right, the prototypes of the second GMM layer are depicted. In the center (outlined in red), a selected prototype of the second GMM layer is shown in an enlarged form. Accordingly, the dimensionality of the selected prototype is based on the number of patches ($2 \times 2$) and the number of prototypes of the previous GMM layer ($5 \times 5$). The dimensionality of the last GMM layer is not affected by the previous folding layer (layer 3). Thus, a prototype of the second GMM layer describes which prototypes of the previous layer have a high activity (white corresponds to a high responsibility, black to low).

In case of inference, it is assumed that an input signal with a displayed 0 is applied. In the first GMM layer (see figure 8.21), the input results in an activation of the 4 highlighted patches (indicated by color). The most active ones are shown, again, on the right-hand side of the prototypes. Moreover, their positions on the grid are indicated by their coordinates. The activation combinations remain unchanged and are passed on to the second GMM layer. Therefore, the 0 input generates an activation pattern as shown in the red frame. Black indicates a low activation and white indicates a high activation. The different colored grids (gree, blue, pink and cyan) represent the positions of the input patches, whereas the initial folding is responsible for the patches. Each grid/quadrant can represent all component positions of the previous GMM layer. The second GMM layer thus learns different activation patterns from the first GMM layer. The activation of the last layer's component therefore corresponds to the pattern of the input. The final activation can be further processed, e.g. to perform classification.

In the reverse direction, i.e., for the generation of samples, the selection of the associated GMM components is reversed. Again, it is assumed that the red GMM component on the right-hand side is selected for sampling. Accordingly, the enlarged activation pattern is generated first and then passed on to the preceding layer. The activation pattern is used to select the corresponding components of the first GMM layer (gree, blue, pink and cyan on the left-hand side). The four patches in figure 8.21 correspond to the four corners of the sample that has to be generated. Finally, these patches need to be combined into one image and, if necessary, optimized by the sharpening process.

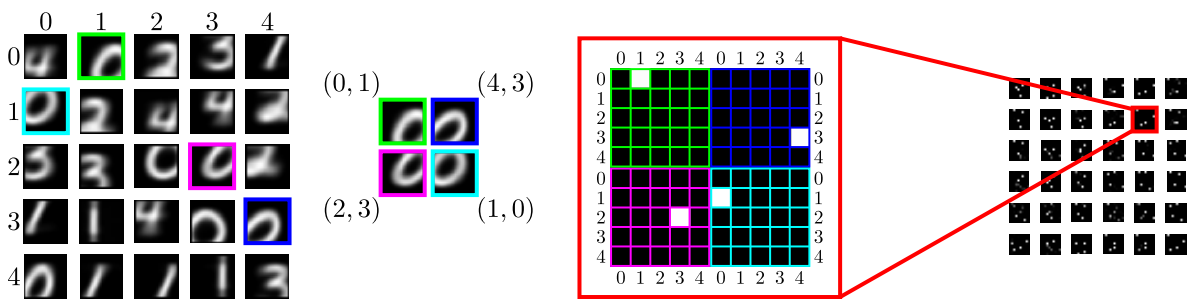

Figure 8.21: Interpretation of GMM components from the DCGMM instance (2L-a).

### 8.5.2.2 Clustering

Similarly to the experiments in section 8.5.1.1, DCGMMs are examined with respect to their clustering performance. Once again, the metrics Dunn index Dunn 1973 (more is better) and the Davies-Bouldin score Davies and Bouldin 1979 (less is better) are applied. Accordingly, samples and the assigned cluster IDs are inspected. Results of the examined architectures (see table 8.4) are summarized in table 8.5. For each of the two examined datasets, the best results for all architectures are highlighted in bold face. Mainly non-convolutional, but deeper DCGMMs imply a better clustering performance. As a consequence, non-convolutional but deep DCGMM architectures should be preferred, if a high clustering capacity is the goal.

Table 8.5: Clustering performance comparison of different DCGMM architectures.

| Dataset | Metric | 1L | 2L-a | 2L-b | 2L-c | 2L-d | 2L-e | 3L-a | 3L-b |
|---------|--------|-----|------|------|------|------|------|------|------|
| MNIST | Dunn index | 0.14 | 0.14 | 0.13 | 0.12 | **0.19** | 0.15 | 0.15 | 0.15 |
| | Davies-Bouldin score | 2.59 | 2.73 | 3.06 | 2.62 | 2.57 | 2.55 | 2.65 | **2.53** |
| FashionMNIST | Dunn index | 0.14 | 0.15 | **0.16** | 0.15 | 0.11 | 0.11 | 0.09 | 0.13 |
| | Davies-Bouldin score | 2.37 | 2.77 | 2.62 | 2.70 | 2.40 | 2.92 | 3.20 | **2.35** |

### 8.5.2.3 Outlier Detection

Outlier detection is one of the frequently used functions of clustering methods. In order to determine the capabilities of DCGMMs, the following experiments are performed. Firstly, the different DCGMM architectures from table 8.4 are trained on the first five classes (0 to 4). Secondly, test samples of the classes 0 to 4 are tested for inliers and samples from the classes 5 to 9 for outliers. For outlier/inlier detection, the log-likelihood (log-probability) value of the last/highest layer is used as described in section 8.4.2.3. In order to eliminate the choice of the parameter $\mathcal{C}$ of equation 8.23, $\mathcal{C}$ is varied for the range $[-2, 2]$. Thus, a ROC-like curve (receiver operating characteristic curve) is obtained, which represents the ratio independently of this hyper-parameter.

The curves in figure 8.22 represent the different architectures stated in table 8.4. As illustrated, convolutional architectures provide significantly better outlier detection performance. This performance contrasts non-convolutional architectures, such as $2L$-$d$ and $3L$-$b$. In the context of outlier detection, the situation is different as stated with regard to the clustering capacity in section 8.5.2.2.

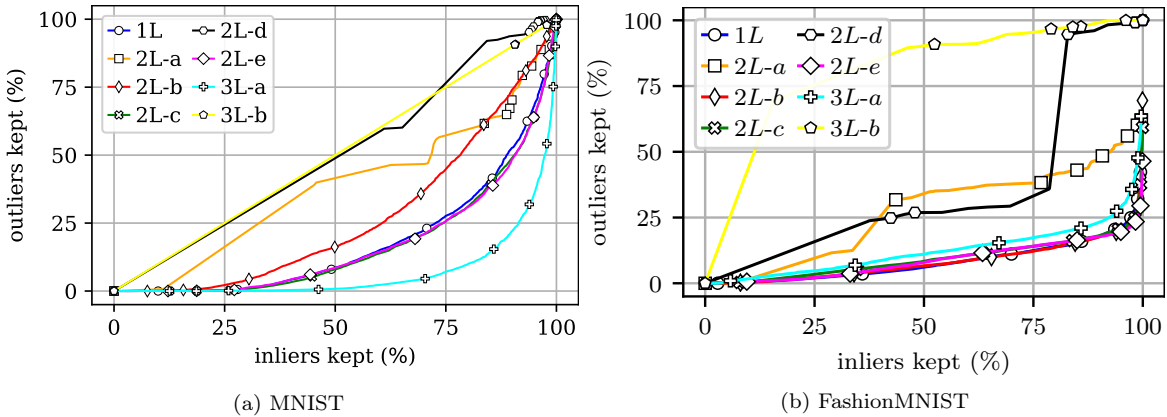

(a) MNIST                          (b) FashionMNIST

Figure 8.22: Outlier detection capabilities of different DCGMMs architectures.

### 8.5.2.4 Sampling and Sharpening

In this section, the DCGMM architecture's impact on sample generation is presented. The first 5 classes (0 to 4) of the MNIST and FashionMNIST dataset (see section 6.1) are used for this purpose. Again, the dataset selection is limited due to the specific parameters of the architectures and an eased comparison. Since the difference in quality is visible at a glance, no explicit metric is used for the evaluation. Furthermore, the influence of the top-$S$-sampling mechanism is evaluated. Finally, the sharpening method is applied on the generated samples.

**Sampling from Convolutional Architectures** DCGMMs allow the splitting of input data into partials (patches) by the application of folding layers. In the following, the influence of the folding operation on image datasets (MNIST and FashionMNIST) is presented. Therefore, different architectures (see table 8.4) are used to generate sample images. The single-layer model ($1L$), as well as the non-convolutional two-layer architecture $2L$-$d$, are used as a reference (baseline). The difference between larger and smaller patches is illustrated by the $2L$-$c$ variant (larger) and the $2L$-$e$ architecture (smaller patches).

The exemplary generated samples depicted in figures 8.23 and 8.24 illustrate that non-convolutional architectures frequently create duplicate images. Some of the duplicates are marked with red letters which are considered to be identical (see figures 8.23a and 8.23b). In contrast, identical samples are not generated with the convolutional variants (see figures 8.23c and 8.23d). The same applies for the generated samples of the FashionMNIST dataset in figure 8.24. For both image datasets, it is visible that the smaller the patches, the more diverse and sharper the generated images.

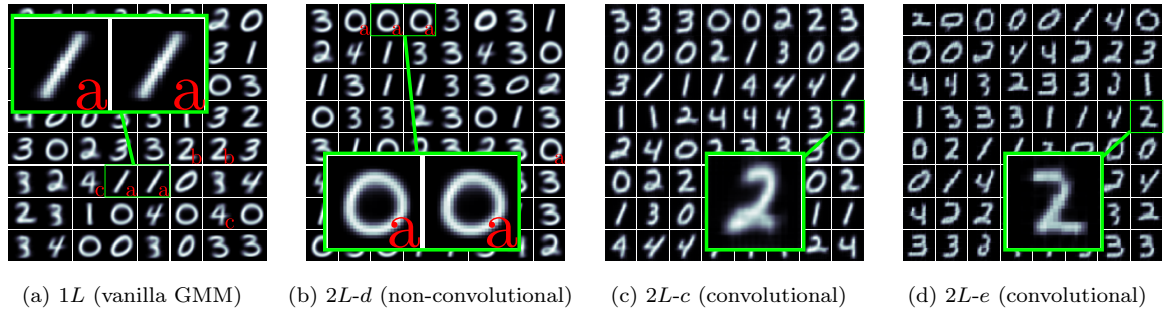

(a) 1L (vanilla GMM)   (b) 2L-d (non-convolutional)   (c) 2L-c (convolutional)   (d) 2L-e (convolutional)

Figure 8.23: The impact of convolution on sampling for MNIST.

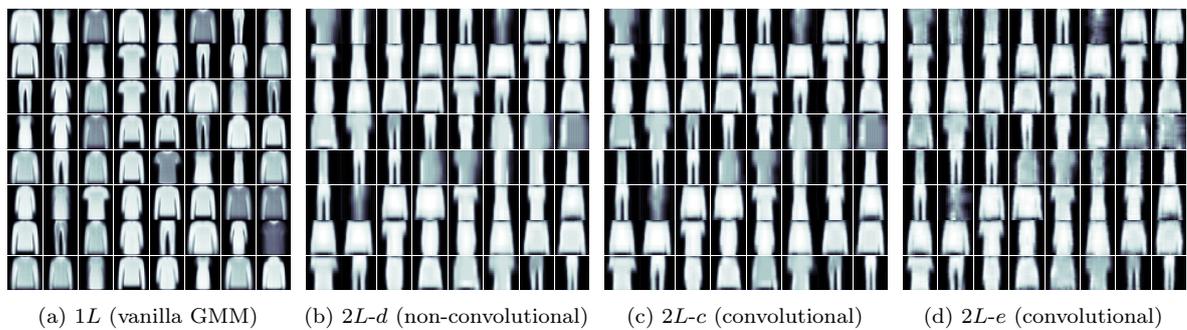

(a) 1L (vanilla GMM)   (b) 2L-d (non-convolutional)   (c) 2L-c (convolutional)   (d) 2L-e (convolutional)

Figure 8.24: The impact of convolution on sampling for FashionMNIST.

**Controlling Diversity by Top-$S$-Sampling** The top-$S$-sampling mechanism (described in section 8.4.1.3) constrains the selection of components used for sampling. In order to illustrate the effect of top-$S$-sampling, the 2L-c DCGMM architecture is used to generate samples with different configurations of $S$ (see figures 8.25 and 8.26). As highlighted in figure 8.25, a smaller $S$ leads to a reduced diversity of the generated samples. Vice versa, a larger $S$ causes more diverse samples. However, the generated samples can degenerate if one ore multiple inappropriate components are chosen. Accordingly, this observation correlates with the increasing number of generated samples showing defects. Therefore, the choice of an optimal $S$ is problem-dependent, as illustrated in figure 8.26.

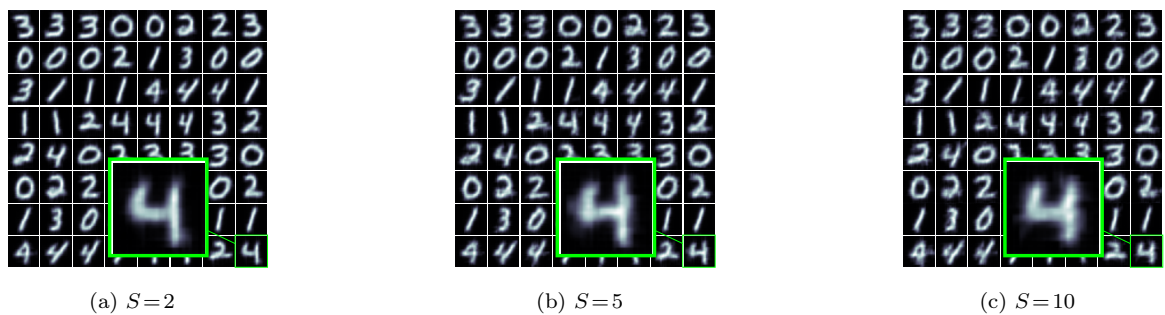

(a) $S = 2$             (b) $S = 5$             (c) $S = 10$

Figure 8.25: The impact of different values of $S$ for top-$S$-sampling for MNIST.

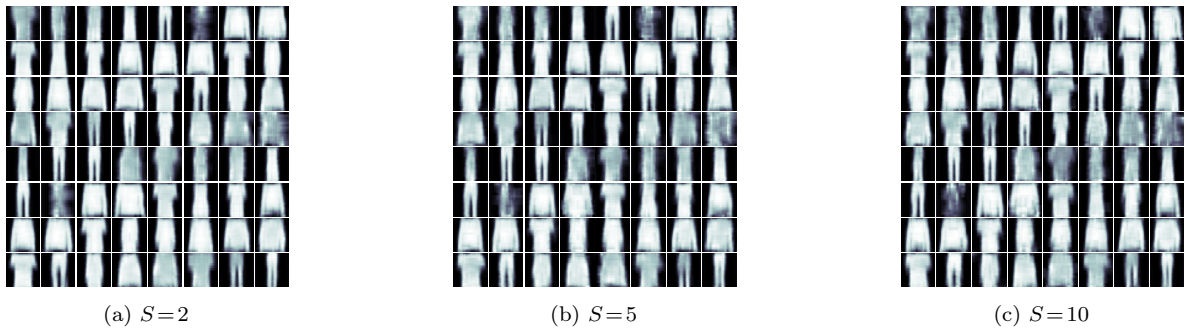

(a) $S = 2$                              (b) $S = 5$                              (c) $S = 10$

Figure 8.26: The impact of different values of $S$ for top-$S$-sampling for FashionMNIST.

**Image Sharpening**  Another functionality offered by DCGMMs is a sharpening method for samples. Sharpening is realized by gradient ascent as explained in section 8.4.2.5. In order to demonstrate the effect of sharpening, samples of the MNIST and FashionMNIST dataset are generated by a $2L$-$c$ DCGMM instance. Figures 8.27 and 8.28 illustrate samples before (left) and after the sharpening process (middle). The differences are highlighted on the right-hand side, as they are hard to recognize. The adjustments of the sharpening method are highlighted by visualizing the differences before and after the optimization.

In principle, sharpening does not change the shape of a sample, but improves its quality. Thus, hard edges resulting from the inversion of the folding process can be mitigated. The effect becomes more obvious in figure 8.28. The transitions of the individual patches become smoother. Nevertheless, the sharpening process depends on two parameters: The sharpening rate $\epsilon_S$ and the number of performed sharpening iterations $g$.

Besides improving the quality of generated samples, sharpening can be transferred for other functionalities. The sharpening function of DCGMMs can be used to reduce the noise in images. The same applies for defects, which can be identified by outlier detection and repaired by sharpening the defect part (e.g., Xie, Xu, et al. 2012). Moreover, inpainting is another functionality that may be implemented by means of the sharpening approach. Inpainting is also used to manipulate specific areas of images (or other types of data, e.g. audio or videos) based on previously learned data distributions (e.g., Yenamandra, Khurana, et al. 2021; Kim, Woo, et al. 2019).

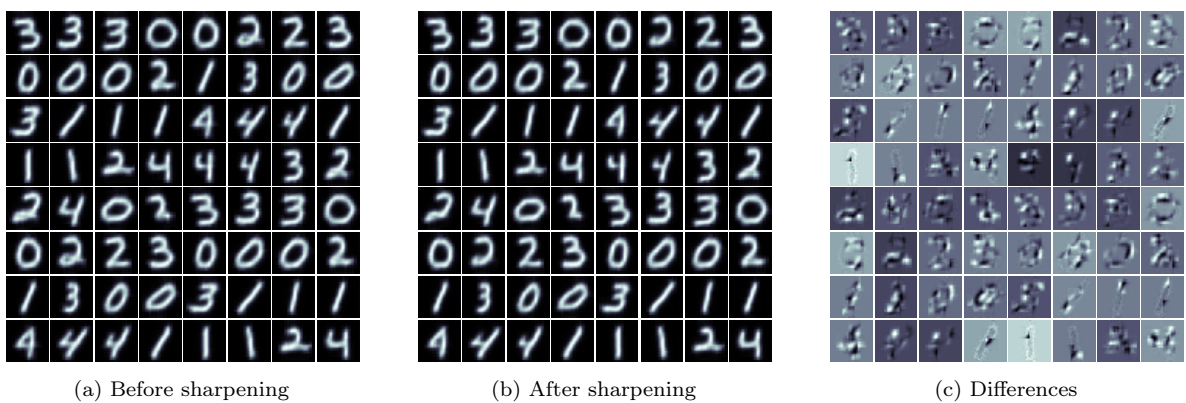

(a) Before sharpening                  (b) After sharpening                  (c) Differences

Figure 8.27: Impact of sharpening on generated samples for MNIST.

## 8.6  Discussion

Various aspects of the proposed SGD-based training approach for GMMs are discussed below. The following aspects are subject to the discussion: The novelty and the validity of DCGMMs, new emerging hyper-parameters, a justification of GMMs and the annealing scheme.

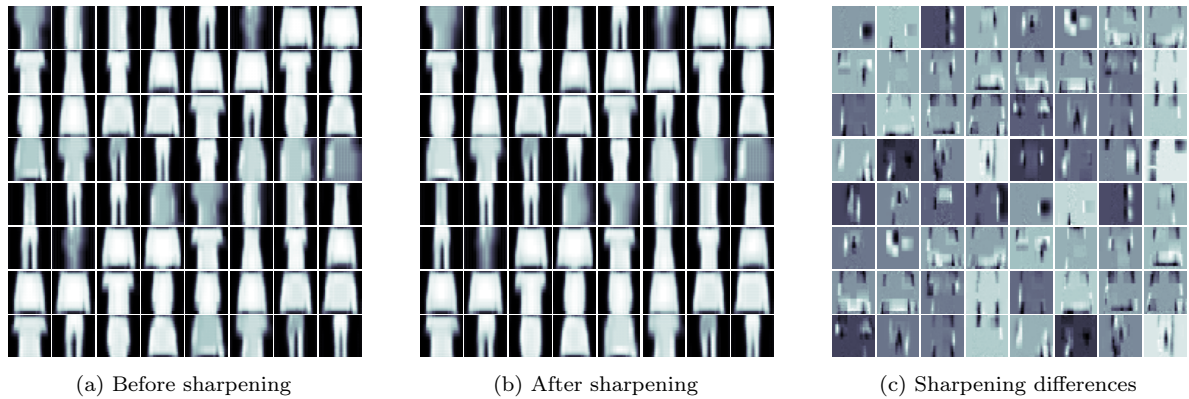

(a) Before sharpening          (b) After sharpening          (c) Sharpening differences

Figure 8.28: Impact of sharpening on generated samples for FashionMNIST.

**Novelty**   In general, both GMMs and SGD are well established and explored. Another GMM training procedure using SGD is presented by Hosseini and Sra (2015). Recent research (Hosseini and Sra 2020) implies the issue's relevance. Hosseini and Sra's (2015) method is similar to the presented SGD training scheme, although the related approach introduces multiple different hyper-parameters ($\rho$, $\kappa$, $\alpha$, $\beta$ and $\Lambda$).

As described in related work, the main difference is that the initialization problem is circumvented by using a data-driven method (e.g., k-means). k-means itself requires an initialization and is a batch type algorithm (is not applicable for streaming data).

Other research (Verbeek, Vlassis, et al. 2005) establishes a connection between mixture models and an annealing scheme that is related to SOMs. The goal of Verbeek, Vlassis, et al. (2005) is to use a mixture model as kind of a SOM. This validity of the relationship between SOMs and GMMs is proven by Gepperth and Pfülb (2020).

Van Den Oord and Schrauwen (2014) present Deep GMMs which are trained with a special variant of (hard) EM. The interpretation is based on paths through the network, whereas no folding or classification is conducted. Another *deep* version of GMMs is presented by Viroli and McLachlan (2019). They combine their deep variant with a stochastic version of the EM algorithm for training. The conceptual difference between the presented deep models and Viroli and McLachlan's (2019) proposal is that they are considered to have a one-to-one correspondence to a flat model. Since the component weights are not normalized in each layer, the layers are not independently optimized. Instead, they are intrinsically connected by paths. The decomposition of these layer correlations by the SGD approach proposed in this work means that the layers can be optimized independently, which has a significant impact on scalability. Another advantage of the separation is that additional layers can be introduced, e.g., folding, pooling and classification. Different techniques and methods from related work have inspired the novel approach introduced as part of this work.

**Validity**   In the following, the validity of the SGD based training approach for GMM is subject to discussion. The validity of the novel training approach was confirmed by comparison with a well established training method. The results of the experimental investigation indicate that the SGD based GMM training is at least equivalent to an online variant of EM, i.e., the stochastic EM (sEM) Cappé and Moulines 2009. Nevertheless, the direct comparison is unfair, as the conventional sEM is favored. In this context, the centroids for an sEM trained GMM need to be initialized. A meaningful initialization is only possible with other data-driven mechanisms, e.g., by the application of k-means. This mechanism is used in many related works. Without it, no acceptable results can be achieved by various EM based training methods. A direct comparison of the stochastic EM variant (sEM) and the SGD-based training method shows that equivalent results are achieved by using SGD. Accordingly, the SGD approach can be considered a valid procedure to train GMMs. Despite the disadvantage of a lacking initialization, the SGD approach achieves better results for high-dimensional data.

Drawing conclusions with regard to the validity of the DCGMM is challenging. A direct comparison with a similar method is not conducted. This is due to the fact that no comparable training procedure or model exists. For this reason, a comprehensive evaluation and findings on the validity of DCGMMs is considered as an objective of future research.

**Hyper-Parameters**   Different hyper-parameters emerge in the context of SGD-based training. The same applies to the novel DCGMMs. Besides the learning rate $\epsilon$, the newly introduced parameters include $\alpha$ as reduction criterion and $\delta$ as limit to detect the loss' stationarity. Furthermore, initialization parameters need to be defined: The number of Gaussian components $K$, $\boldsymbol{\mu}$, $pi$ and $\boldsymbol{\Sigma}$ as initialization for each GMM. The number of new hyper-parameters seems to be problematic at first. However, these parameters can usually be chosen or applied intuitively to all examined datasets. Section 8.3.4 provides a recommendation how to choose these hyper-parameters. The same hyper-parameters provide good results for different problems/datasets from section 6.1. This contrasts conventional training methods, e.g., sEM, where larger ranges of parameters need to be explored. Thus, the SGD-based training process does not seem to depend a lot on hyper-parameters and its definition is easier.

At the same time, some parameters can be selected interdependently, which reduces the number of new parameters. In fact, the learning rate $\epsilon$ has to be adjusted depending on the problem, as it is the case for all stochastic procedures. The enormous advantage of EM is therefore lost due to the transfer to an online variant. For the online variant sEM, the parameters need to be adjusted in a similar way as the learning rate, namely in a problem-specific manner. Even though the number of GMM components $K$ remains *the more the better*, $K$ is always resource-dependent.

Furthermore, certain hyper-parameters of the SGD training approach are easier to interpret than those of sEM. $\delta$, for example, defines the stationarity threshold and thus influences the temporal convergence behavior. Hyper-parameters of other ML methods are not as intuitively to select or adjust, e.g., for sEM: $\rho_0$, $\rho_\infty$ and $\alpha_0$.

**Justification of GMMs**   Although images do not necessarily follow a Gaussian distribution, their representation can be approximated by GMMs. Unlike the EM algorithm, the SGD based method makes no hard assumptions regarding the unobservable latent variable. Thus, SGD training does not assume that a sample is drawn from exactly one component, even though this approximation holds (see max $p_{k^*}$ responsibilities in table 8.3 of section 8.5.1.3).

The universal approximation theorem applies to DNNs with multiple hidden layers. In fact, the same applies to GMMs. For GMMs, however, it depends on the number of available Gaussian components $K$. "A Gaussian mixture model is a universal approximator [...] with enough components." (Goodfellow, Bengio, et al. 2016). Despite the universal approximation theorem (see Pinkus 1999; Hornik, Stinchcombe, et al. 1989), the question of how many components or respectively layers are needed to represent the data distribution remains unanswered. This is related to the available memory and computational resources.

The limitation of adding more GMM components is softened considerably by stacking multiple GMMs with less components, which is similar to layers in a DNN. Thus, more complex functions can be approximated by DCGMMs with fewer memory and computational resources. Accordingly, the introduction of folding and pooling layer(s) should improve the performance quality for high-dimensional data like images (as the operations in CNNs).

**Annealing Scheme**   The effect of the introduced annealing scheme is another aspect to be discussed. The basic idea of annealing is often applied to the training of SOMs. In the context of GMMs, the application of the proposed annealing strategy leads to several advantages. Annealing prevents the emergence of degenerate solutions (see figure 8.6a for an impression). Unlike other training methods, trained GMMs with the annealing scheme are not affected by the well-known "mode collapse", which is particularly known for GANs (Richardson and Weiss 2018). Mode collapse describes the problem that the generator of a GAN draws the same sample over and over again, which is due to the lacking capability of the discriminator to distinguish generated and real samples.

The main advantage of annealing is that it eliminates the need for a data-driven initialization, which is particularly beneficial for streaming scenarios. The initialization problem is often circumvented by the application of other clustering algorithms requiring the complete dataset, as well as an initialization, e.g., k-means. Disadvantages of this data-driven initialization technique comprise the introduction of additional hyper-parameters and the dependence on an initialization. The introduced hyper-parameters for the presented annealing scheme can be automatically selected and adjusted (which is at least true for all examined datasets). Based on the traced stationarity of the log-likelihood, the parameters are automatically adjusted. Thus, the adaption of suitable hyper-parameters is of minor importance.

Other advantages could be mentioned and discussed. The application of the scheme, for example,

is computationally marginal and therefore does not affect the training process. At the same time, annealing ensures that a topological sorting of the components is performed. Even though the sorting is not used in this work, further functionalities could be implemented by using it. Further research could be conducted related to neighborhood components, or other annealing patterns could be applied.

## 8.7   Conclusion and Future Work

In this chapter, a numerically stable SGD-based training method for GMMs was presented. An annealing strategy is introduced as a special feature, which ensures that no data-driven initialization is needed. In general, a data-driven initialization constitutes a huge disadvantage, which is why it is omitted in many related works. At the same time, annealing prevents the emergence of degenerate solutions. This effect can be compared to the well-known "mode collapse" in the context of GANs (see Richardson and Weiss 2018).

Based on the SGD training of GMMs, a novel ML model is proposed, which is referred to as DCGMM. DCGMMs consist of multiple GMM layers that are trained by SGD. Similarly to DNNs, DCGMMs represent deep GMMs that can be considered as universal function approximators. Furthermore, DCGMMs allow the combination of GMM layers with other types of layers, such as classification, folding and pooling layers (known from CNNs). Other layer types may be implemented, which represents a great potential in terms of research and further development. Moreover, DCGMMs offer various functionalities, such as density estimation, sampling, outlier detection, sharpening or classification. Accordingly, the novel DCGMM offers a great research potential for different application scenarios or functionalities such as inpainting. The classification of samples is addressed in the next chapter (see chapter 9).

Since DCGMMs can be considered as a novel ML model, a number of new areas of research arise. The properties associated with the data can be examined. This aspect includes, for example, how the behavior affects different data distributions within the data. Based on the density estimation, the learning rate or the annealing parameter $\sigma$ can be adjusted similar to a weighting mechanism for each individual sample. The advantage would be that the model itself determines the required weighting factor in data stream samples. Thus, outlier and inlier can be in-/excluded in the training process.

Similarly, the obtained log-likelihood values can be set in relation to the sliding log-likelihood. The resulting factor can be used in further application scenarios, e.g., CL tasks. Again, a higher or lower weighting of potentially known or unknown samples (outliers) can be utilized. The presented concept for controlling the annealing parameter $\sigma$ is only a first proposal. It describes the adjustment of $\sigma$ based on the stationarity of the log-likelihood, although more intelligent methods may be deployed. This can be particularly important for the adjustment of individual training samples.

Another optimization with regard to the diagonalized covariance matrices or precision matrices may be of interest for future research. The idea is transferred from MFAs models, which can be implemented as layers (Tang, Salakhutdinov, et al. 2012). Another open issue concerns the holistic mapping of multiple individual layer loss functions into a single one. Usually, parameters of deep learning models are optimized according to one loss function, as it is the case for DNNs or CNNs. Future research could also focus on the use of advanced optimizers such as Adam (Kingma and Ba 2015). Attempts to perform the respective optimizations did not lead to convergence.

Moreover, applying approaches for dynamic addition and removal of GMM components seems to be an interesting possibility (e.g., Song and Wang 2005; Kristan, Skočaj, et al. 2008). The integration of the weights $\pi$ into the sampling process is yet another open aspect. Even if the implementation allows the weight's inclusion, its effect remains unresolved. The deployment in other application-oriented scenarios raises new challenges. This may include applications that process related sequential data, such as speech, text or video streams.

Basically, the SGD-based training method of GMM and its transformation into layers (DCGMM) are considered as an independent, novel finding of this work. Accordingly, the results of this chapter are not necessarily related to the CL context. However, the properties and functionalities of GMMs/DCGMMs constitute the basis for further considerations and the application within CL scenarios. Therefore, the investigation of DCGMMs in the context of CL is the main focus of the next chapter.

# 9.  Continual Learning with DCGMMs

## Chapter Contents

This chapter introduces a *continual learning* (CL) learning model that is referred to as Gaussian Mixture Replay (GMR) and based on Deep Convolutional Gaussian Mixture Models (DCGMMs) (see chapter 8). The contents of this chapter have been already published in Pfülb and Gepperth 2021; Pfülb, Gepperth, et al. 2021. DCGMMs were evaluated with regard to their validity and other functionalities, reviewing the clustering capabilities and ability to generate samples in the previous chapter. This chapter specifically examines the CL performance of the GMR approach in conjunction with the classification capabilities.

GMR is classified as a pseudo-rehearsal method (see section 2.3.5), which is based on the sampling function of DCGMMs. For the investigation of CL capabilities, the investigation protocol presented in chapter 6 is applied. The protocol is characterized by the implementation of various real-world requirements. These requirements include, for example, prohibiting the use of future data. Another application-oriented requirement is that past data cannot be accessed due to memory limitations. Especially in streaming scenarios, the availability of an infinite amount of memory to store all data samples is unrealistic. Furthermore, the evaluation protocol defines CL tasks containing different classification sub-tasks. These tasks are referred to as Sequential Learning Tasks (SLTs) and they represent CL problems with hard but known task boundaries. The hard task boundaries, however, do not apply to all variants of CL scenarios. They are considered easier than gradual or unknown transitions. In particular, the indicated boundaries constitute a basic prerequisite for many CL models, e.g., Deep Neural Networks (DNNs) based models, which are purely discriminative methods. This is due to the fact that they cannot detect outliers without additional effort.

**Contributions**  The contribution of this chapter is a CL approach, referred to as Gaussian Mixture Replay (GMR), which is based on the novel DCGMM model. GMR implements a pseudo-rehearsal (replay) method, which is particularly suitable for CL scenarios. Pseudo-rehearsal methods ensure that samples from the past do not need to be stored. Therefore, samples can be generated on demand and used for replay.

A comparison of GMR and other replay-based models shows that similar CL performance can be reached. At the same time, other functionalities of the DCGMMs can be used to identify important properties of CL tasks. The detection of task boundaries is one important characteristic of CL tasks, which contrasts with other investigated models. This ability arises from the underlying Gaussian Mixture Models (GMMs), which can be used as an unsupervised clustering method. Based on the possible density estimation of one or multiple samples, class boundaries can be detected.

Compared to the well-known deep methods, e.g., DNNs, GMMs are not directly affected by the *catastrophic forgetting* (CF) effect. First results indicate that GMR exhibits comparable or better CL properties, except for the final classification performance which is not yet comparable. Therefore, the aim of the GMR model is to demonstrate the fundamental CL properties of DCGMMs. Based on this first CL proposal, further optimizations and adaptations can be explored.

**Structure** Initially, the pseudo-rehearsal procedure Gaussian Mixture Replay (GMR) is presented in section 9.1. In section 9.2, the GMR procedure is evaluated. Then the experimental setup is introduced (see section 9.2.1). The setup describes the evaluation process and the compared CL models. Subsequently, an investigation with regard to the detection of task boundaries by the GMR model is summarized in section 9.2.2. Finally, a comparison with other CL models is conducted with regard to the CL performance. Even if replay-based methods were previously excluded by real-world requirements, Generative Replay (GR) (Shin, Lee, et al. 2017) is used for comparison. Several aspects of the procedure are discussed in section 9.3. These procedures include a strategy which addresses the limitation of applying pseudo-rehearsal approaches in CL scenarios. The chapter is concluded in section 9.4, where open issues for further research are outlined.

## 9.1   Gaussian Mixture Replay

The Gaussian Mixture Replay (GMR) procedure is based on the DCGMMs presented in chapter 8. Likewise, for training GMR the proposed Stochastic Gradient Descent (SGD) training method is used (see section 8.3). In this context, the annealing scheme (see section 8.3.3) is applied as well. The application of annealing has two advantages. Firstly, it solves the problem of degenerate solutions while being numerically stable. Secondly, it allows for the training of GMMs without a data-driven initialization. Thus, a systematic comparison with other deep learning methods, e.g., based on DNNs, is possible. Another advantage of DCGMMs is the architectural arrangement of different types of layers. Stacking multiple GMM layers allows for the approximation of more complex functions. Otherwise, a single GMM is limited by the number of GMM components in terms of memory and computational effort. In addition, the use of folding layers allows to achieve better results for the respective (image) datasets. The introduction of a classification layer enables a supervised inference and thus the generation of class specific samples (conditional sample generation).

The first examination of the proposed GMR model is based on a simple DCGMM architecture consisting of three layers. More complex and deeper convolutional variants can be realized, which tend to achieve better CL performances. The first layer is a folding layer, which transforms the input data into a one-dimensional vector (see section 8.4.1.1). Accordingly, no convolution or folding is performed as it is the case in Convolutional Neural Networks (CNNs). The second layer is a GMM layer, which defines multiple Gaussian components $K$ parametrized by the prototypes $\boldsymbol{\mu}_k$ and variances $\boldsymbol{\Sigma}_k$ (see section 8.4.1.3). The exact training, inference and sampling procedure is presented in sections 8.3.4, 8.4.2.2 and 8.4.2.4.

The output of a GMM layer corresponds to the posterior distribution of a sample, referred to as responsibilities $\gamma_k$. The responsibilities are subsequently converted into class memberships by a classification layer (third and last layer). Thus, the GMR model combines both a **Learner** and a **Generator** in one and the same ML model. Functionalities such as sample generation, outlier detection and classification are combined in a single entity. The basic structure is illustrated in figure 9.1. It is possible to show that GMR has an adequate CL performance with this simple architectural structure.

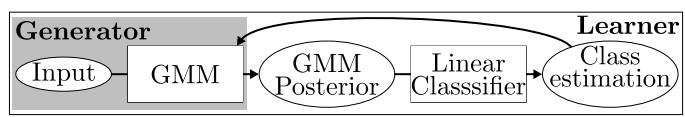

Figure 9.1: Meta-structure of the GMR model.

As stated in chapter 8, a DCGMM instance can be trained incrementally by SGD. If the training is performed in terms of the CL paradigm, the GMM layers are not or differently affected by the CF effect. The situation is different for the classification layer (linear classifier layer), which is affected by the CF effect. Due to the CF affected linear layer at the end of the DCGMMs, the problem shifts.

However, the architecture ensures that the storage of knowledge and the associated classification is separated.

Assumed is a GMM layer consisting of only 2 Gaussian components which is trained class by class. As soon as the training of the second class begins, one of the two components (the best matching unit (BMU)) is gradually adapted to the second class. The adaption can be described as gradual, as precisely one BMU (component) is adjusted for the second data distribution. In figure 9.2, different states of training are visualized for $K = 4$ Gaussian components. The training is conducted for $\mathcal{E} = 50$ training epochs, which corresponds to a very long training process for a two-class problem. Over a long period of time, one prototype after another is gradually adjusted (see intermediate states of components in figure 9.2). Accordingly, forgetting does not occur catastrophically in a GMM layer.

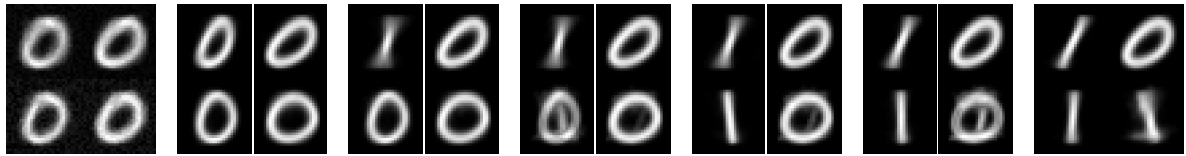

Figure 9.2: Visualization of an incrementally trained DCGMM for $\mathcal{E} = 50$.

The problem of catastrophic forgetting still occurs in the linear classification layer. However, the last classification layer is solely responsible for assigning a class label to an input signal. As a result, the linear layer only stores the knowledge about the assignment of a label to a corresponding active GMM component. The GMR procedure primarily addresses the catastrophic forgetting effect in the last layer. For this purpose, a corresponding activation of the GMM layer with an associated class label needs to be generated.

### 9.1.1   Pseudo-Rehearsal Procedure

In order to prevent the forgetting in the last layer, the pseudo-rehearsal mechanism GMR is introduced. To be precise, samples are generated from previously learned data distributions and merged with the current data distribution for a joint training. The conditional sampling strategy illustrated in section 8.4.2.4 is used to generate proportionally even samples from all classes. Interestingly, the term "proportionally" is a key issue for streaming- and CL scenarios which will be discussed later.

The replay scheme for the CL tasks (specifically for SLTs from section 6.2) is illustrated in figure 9.3. At first, samples for tasks $T_t$ for $t > 1$ have to be generated. The number of samples $\#(t)$ that need to be generated depends on the number of samples for the current task $\#(T_t^{\text{train}})$. Again, this primarily contradicts the idea of data streams, as the exact number is not known beforehand. In this context, the generation of a fixed number of samples is a shortcut for the replay approach. The goal is to demonstrate a reasonable CL performance of GMR.

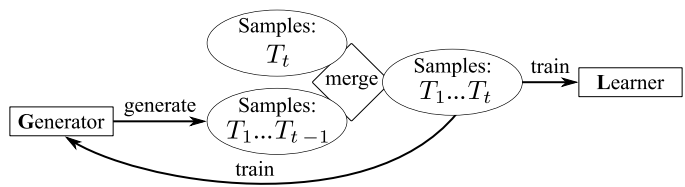

Figure 9.3: Replay scheme for GMR.

Various strategies can be used to address the challenges due to the proportional sampling constraint. Firstly, the number of generated samples can be set to a fixed number, assuring that a sufficient number of representative samples is available for re-training. As long as there are enough samples, this method ensures that the existing knowledge can be preserved. Secondly, the challenge of proportional sampling can be tackled by adjusting the importance of a sample. If generated samples are assigned a higher weighting, e.g., by a larger learning rate $\epsilon_{generated}$, a smaller number of samples needs to be generated. Thirdly, the merging strategy for generated and new samples can be a decisive factor. The factor defines the mixing ratio that is applied to specify the ratio in each training batch.

For a first investigation of GMR, the strategy of a fixed number of samples with a proportional merging ratio is implemented. Initially, the respective class labels are used to determine the ratio of generated and new samples. This can be realized simply by counting the number of learned classes and the number of new classes within a batch. This is how a balanced relationship between new and old data is achieved. A $50 : 50$ merging ratio, for example, is defined for two CL task consisting of two classes each.

## 9.2  Evaluation

In order to evaluate GMR, several experiments are conducted. First, the experimental setup is described in section 9.2.1. This includes a brief description of the applied evaluation protocol from chapter 6, as well as a presentation of the different CL models and their respective hyper-parameters. Subsequently, the ability of GMR's tasks boundary recognition is demonstrated in a purely unsupervised manner. In general, the declaration of task boundaries is often expected by conventional CL approaches, as the boundaries cannot be detected by the models themselves. Section 9.2.2 shows that the functionality to detect task boundaries is already implemented by the DCGMM procedure.

Finally, the CL capabilities of different ML models are presented in section 9.2.3. The compared methods include the Elastic Weight Consolidation (EWC) proposed by Kirkpatrick, Pascanu, et al. (2017). In order to strengthen the significance, a widely known generative approach is examined. Shin, Lee, et al. (2017) describe the pseudo-replay approach which is referred to as GR. GR can be used with different technologies for generating samples, e.g., Generative Adverserial Networks (GANs) (Goodfellow, Pouget-Abadie, et al. 2014) or Variational Autoencoders (VAEs) (see Kingma and Welling 2013). Hence, the novel GMR is compared with two state-of-the-art CL models.

### 9.2.1  Experimental Setup

For all of the presented experiments, 10 repetitions are performed for one and the same parameter configuration. The obtained results are aggregated in a so-called meta-experiment by averaging the performance values (and its variances). This ensures that the influence of initialization is reduced for certain parameter configurations.

#### 9.2.1.1  Application-Oriented Evaluation Protocol

The application-oriented evaluation protocol from chapter 6 is briefly presented in the following. The basis for the evaluation are SLTs (see section 6.2), which define a decomposition of datasets (e.g., section 6.1). Thus, the decomposition defines sub-tasks with one or multiple classes that are used for training. The subscripts indicate how many sub-tasks a SLT comprises, e.g., $D_{5\text{-}5}$ consists of two sub-tasks where each contains 5 disjoint classes. The sub-task sequence is denoted as $T_1, T_2, \ldots, T_x$.

The reference value is the result of the baseline experiment ($D_{10}$), which consists of one task that contains all 10 classes of a dataset. This non-CL measurement value is considered as a benchmark for the maximum achievable CL performance.

The definition of sub-task boundaries is one specification that is hard to reconcile with specific real-world requirements. In an application-oriented context, the specification of sub-task boundaries is not necessarily given, nor are the sub-task boundaries hard. The problem is that most of the CL models assume that this information is available. The permanent examination of the training labels could be used as task boundary, but this only applies for special CL scenarios.

In addition to the sub-tasks, the protocol defines the fulfillment of several real-world requirements. This includes, above all, the causal temporal relation, which prohibits the use of future data. Likewise, the use of past data is forbidden due to memory limitations, which is especially true for streaming scenarios. A retroactive adjustment or model selection of the architecture is also not possible. Even if all these requirements are considered as "self-evident", investigations in the CL context usually do not enforce these requirements.

In order to establish a comparable evaluation, classification accuracy is applied to measure the CL capabilities. This metric is only applicable if the data distributions of the datasets within each class are reasonably uniform. In order to exclude the influence of the baseline accuracy, only the difference to the maximum of the baseline experiment is presented. In the following, the resulting CL performance is referred to as *baseline accuracy*.

A special feature of pseudo-rehearsal methods is the generation of samples for a joint training process. After each sub-task, the model $M_t$ with a parameter configuration at the time $t$ is used to generate a sample dataset $G_t$. The generated samples are merged with samples from the current sub-task and presented to the model. The process is visualized in figure 9.4. As depicted, the model $M$ is firstly trained with samples from $T_1^{train}$. After the initial training phase is completed, the model is used to generate the dataset $G_1$ from the learned data distribution. Samples from $G_1$ are mixed in a special ratio. The mixing ratio is calculated by the proportion of number of classes in $T_1$ and $T_2$ (5:5 in this example). Thus, re-training is conducted with batches of data consisting of 50% samples from $G_1$ and 50% of $T_2^{train}$.

After processing all sub-tasks, the CL performance is measured by determining the accuracy of the combined test dataset ($D_{10}^{test}$). The final CL capacity is defined by the difference between the final accuracy achieved for a given SLT and the joint training of the model with $D_{10}^{train}$.

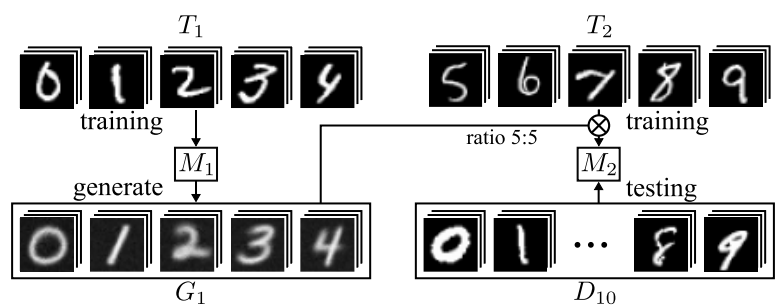

Figure 9.4: Rehearsal approach for the model training (SLT $D_{5\text{-}5a}$ on MNIST).

### 9.2.1.2   CL Models and Hyper-Parameters

Different models are used to compare their CL performance. One of the examined models is EWC (Kirkpatrick, Pascanu, et al. 2017). EWC has attracted attention as the most promising CF avoidance model from the study described in chapter 7. In order to make a comparison with another pseudo-rehearsal procedure, GR (Shin, Lee, et al. 2017) is investigated. GR can be deployed with different types of generators: GANs (Goodfellow, Pouget-Abadie, et al. 2014) and VAEs (Kingma and Welling 2013). Even though this procedure has some disadvantages which are not easy to reconcile with real-world requirements (discussed in section 9.3), this procedure serves as state-of-the-art reference model.

First of all, the different approaches (EWC, GR and GMR) are briefly introduced and their hyper-parameters are outlined. In order to conduct a fair comparison, several hyper-parameters are optimized for each model by using a grid-search. Due to the complex parameter configuration for the GR approach, only three datasets with the same dimensions are used for the present study.

**Elastic Weight Consolidation (EWC)**   One of the most cited models for CL is EWC, as presented by Kirkpatrick, Pascanu, et al. (2017). It is a typical regularization approach that is based on DNNs. EWC stores the parameters distribution of the model after completing a task $\theta^{\vec{T_t}}$. By calculating the Fischer Information Matrix (FIM) $\vec{F}^{T_t}$, the "importance" of each parameter for a respective data distribution is determined. The FIM is subsequently used in the loss function of EWC in order to penalize the change of the important parameters for the old tasks and the current sub-task $T_c$. Likewise, in EWC, as well as in conventional DNNs, the cross-entropy loss is minimized by optimizers, such as Adam (Kingma and Ba 2015). Hyper-parameters include the regularization constant $\lambda$, but also the underlying DNN architecture (number and size of layers). As suggested in the code base, $\lambda$ is set equal to $\epsilon^{-1}$, which eliminates this hyper-parameter. An alternative variant for FIM calculation is used for direct comparison, namely the Matrix of SQuares (MaSQ) (see Gepperth and Wiech 2019). This variant does not require as many resources and produces comparable results.

As only one hyper-parameter needs to be tuned in addition to EWC's architecture, the learning rate is chosen from $\epsilon \in \{0.001, 0.0001, 0.00001, 0.000001, 0.0000001\}$. The underlying DNN architecture is

$$\mathcal{L}^{EWC} = \mathcal{L}_{T_c}(\theta) + \frac{\lambda}{2} \sum_{t=1}^{c-1} \sum_i F_i^{T_t} \left( \theta_i - \theta_i^{T_t} \right)^2$$

Equation 9.1: EWC loss function (see Kirkpatrick, Pascanu, et al. 2017) adapted to the used terminology.

based on a three-layer DNN ($L = 3$) consisting of 800 neurons each ($S = 800$). This architecture is preferred, as acceptable results for the examined datasets are already obtained. The number of training epochs is set to $\mathcal{E} = 10$ epochs, because more training iterations affect the CL performance of EWC (see section 7.1.4). In order to mitigate the effect of training iterations, EWC is evaluated for the best measured accuracy instead of the last value. Nevertheless, this methodology is influenced by the number and distribution of measurement points. In general, the CL performance of an ML model should not or, if at all, marginally depend on the number of training iterations. This aspect is related to real-world constraints, where the determination of the best possible model parameter configuration is usually difficult (early-stopping).

**Generative Replay (GR)**   The second model for comparison is the generative approach Generative Replay (GR). The approach was implemented for the experiments according to Shin, Lee, et al. (2017) (as published by Gulrajani, Ahmed, et al. (2017)). In general, the replay process is the same as for GMR: After sub-task completion, samples are generated, which are then replayed into the following training process. GR is based on a generator and a separated solver. Unlike the presented GMR model, the generator and solver of GR are separate independent entities. Different types of generating or solving models can be used for each part. For the generator, two common and particularly powerful deep learning models are deployed: VAEs (Kingma and Welling 2013) or GANs (Goodfellow, Pouget-Abadie, et al. 2014) (respectively a variant referred to as Wasserstein GAN proposed by Arjovsky, Chintala, et al. (2017)). These two generating components are briefly described below. In addition to the generator, a solver is introduced. Usually, it is a DNN or CNN which is (re-)trained with the generated samples and the current data distribution. Thus, the joint training counteracts the CF effect when training the solver.

**GANs**   Goodfellow, Pouget-Abadie, et al. (2014) introduce Generative Adverserial Networks (GANs) that comprise a generating $g$ and a discriminating $d$ component. Both components ($g$ and $d$) can be considered as independent DNNs. The generating part tries to generate samples that fool the discriminating part $d$. In turn, the discriminator tries to learn how to distinguish real samples from generated ones. Generator and discriminator thus work against each other, resulting in a so-called "zero-sum game" (formulated in section 9.2.1.2). The counterpart causes the generator to learn how to generate realistic samples from the data distribution. These samples are utilized as described in section 9.1.1 for a joint re-training (see Shin, Lee, et al. 2017).

$$\arg\min_g \max_d V(d,g) = \mathbb{E}_{\boldsymbol{X} \sim p_{data}} \log\big(d(\boldsymbol{x})\big) + \mathbb{E}_{\boldsymbol{x} \sim p_{model}} \log\big(1 - d(\boldsymbol{x})\big)$$

Equation 9.2: Formulation of the *zero-sum game* in GANs (Goodfellow, Bengio, et al. 2016).

**VAEs**   Kingma and Welling (2013) propose VAEs that comprise two parts, the encoder and the decoder. The basic idea is to compress an input signal layer-by-layer by reducing the number of available artificial neurons (encoder). The decoder as a second part tries to restore the original input signal from the latent space (representation of compressed data). The main difference compared to an auto-encoder is an additional layer between the decoder and encoder. This additional layer realizes the values' transformation to probabilistic values (e.g., by a multivariate Gaussian section 8.1.2 vector). VAEs offer various functions, e.g., de-noising or (conditional) sample generation. The goal of VAEs is thus to minimize the reconstruction loss (see section 9.2.1.2, where $D_{KL}$ is the Kullback-Leibler divergence).

**Solver**   As an addition to the generating component (GAN or VAE), a further DNN is introduced by GR. The component that is responsible for classification is referred to as solver. Similarly to the

$$\mathcal{L} = \mathbb{E}_{\boldsymbol{z} \sim q(\boldsymbol{z}|\boldsymbol{x})} \log \left( p_{model}(\boldsymbol{x}|\boldsymbol{z}) \right) - D_{KL}\left( q(\boldsymbol{z}\boldsymbol{x}) || p_{model}(\boldsymbol{z}) \right)$$

Equation 9.3: Standard loss function for VAEs (Goodfellow, Bengio, et al. 2016).

generating part, different ML models may realize the solvers.

**Hyper-Parameters of GR**  Generative Replay (GR) introduces multiple entities. For all of them, different hyper-parameters need to be specified, e.g., the architectures. Usually, Keras (ML framework) components are used for implementation. In the following, the individual types of layers used for building the individual components for GR are presented:

- Fully-Connected (FC) layer which is often referred to as *dense* layer.
- LeakyReLU is the attached output function (compare ReLU section 2.2.1.2). See section 9.2.1.2 and figure 9.5 for a definition. The exact value (here 0.2) slightly varies in many implementations.

$$f_\varphi(x) = \begin{cases} x & \text{if } x > 0, \\ 0.2x & \text{otherwise} \end{cases}$$

Equation 9.4: Leaky ReLU output function.

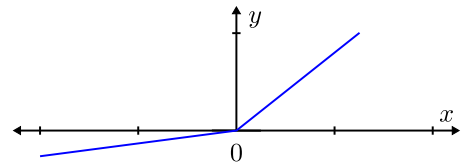

Figure 9.5: Plot of Leaky ReLU output function.

- Batch Normalization (BatchNorm) layers try to normalize the input by a transformation (Ioffe and Szegedy 2015). BatchNorm layers ensure that the neural network converges faster and therefore less training iterations are required.
- 2D convolution (Conv2D) layers represent the standard convolutional operation used by CNNs (LeCun, Haffner, et al. 1999) with a given filter and kernel size. Such layers are very advantageous in the field of image processing. The reverse direction is achieved with the transposed variant (Conv2DTrans).
- Dropout (Hinton, Srivastava, et al. 2012b) layers assure that by a defined chance certain activations are not passed to the next layer (in the present work, the dropout rate is set to 0.2 and 0.3 respectively).
- Reshape and Flatten layers are responsible for transforming the input and output respectively.

The architecture of the deployed GAN is presented in table 9.1. The architecture described in table 9.2 is used for GR experiments with VAE. For both generating methods, the architecture of the solver is specified. In addition to the architectural structure, the number of free model parameters is composed by the different layer types.

Table 9.1: Parameter configuration for Solver and GAN (generator and discriminator).

| Solver | | | Generator | | | Discriminator | | |
|---|---|---|---|---|---|---|---|---|
| Type | Shape | Parameters | Type | Shape | Parameters | Type | Shape | Parameters |
| Reshape | 28,28,1 | - | Flatten | 80 | - | Reshape | 28,28,1 | - |
| Flatten | 784 | - | FC | 12544 | 1 016 064 | Conv2D | 14,14,64 | 1 664 |
| FC | 400 | 314 000 | BatchNorm | 12544 | 50 176 | LeakyReLU | 14,14,64 | - |
| ReLU | 400 | - | LeakyReLU | 12544 | - | Dropout | 14,14,64 | - |
| FC | 400 | 160 400 | Reshape | 7,7,256 | - | Conv2D | 7,7,128 | 204 928 |
| ReLU | 400 | - | Conv2DTrans | 7,7,128 | 819 328 | LeakyReLU | 7,7,128 | - |
| FC | 400 | 160 400 | BatchNorm | 7,7,128 | 512 | Dropout | 7,7,128 | - |
| ReLU | 400 | - | LeakyReLU | 7,7,128 | - | Conv2D | 7,7,256 | 819 456 |
| FC | 10 | 4 010 | Conv2DTrans | 14,14,64 | 204 864 | LeakyReLU | 7,7,256 | - |
| Softmax | 10 | - | Batch | 14,14,64 | 256 | Dropout | 7,7,256 | - |
| | | | LeakyReLU | 14,14,64 | - | Flatten | 12544 | - |
| | | | Conv2DTrans | 28,28,1 | 1 601 | FC | 1 | 12 545 |
| | $\sum$ | 638 810 | | $\sum$ | 2 092 801 | | $\sum$ | 1 038 593 |

Table 9.2: Parameter configuration for Solver and VAE (encoder and decoder).

| **Solver** | | | **Encoder** | | | **Decoder** | | |
|---|---|---|---|---|---|---|---|---|
| Type | Shape | Parameters | Type | Shape | Parameters | Type | Shape | Parameters |
| Reshape | 28,28,1 | - | Reshape | 28,28,1 | - | Flatten | 20 | - |
| Flatten | 784 | | Conv2D | 14,14,64 | 1 088 | FC | 100 | 2 100 |
| FC | 400 | 314 000 | Conv2D | 7,7,128 | 131 200 | FC | 6272 | 633 472 |
| ReLU | 400 | - | Flatten | 6272 | - | Reshape | 7,7,128 | - |
| FC | 400 | 160 400 | FC | 100 | 627 300 | Conv2DTrans | 14,14,64 | 131 136 |
| ReLU | 400 | - | FC | 20 | 2 020 | Conv2DTrans | 28,28,1 | 1 025 |
| FC | 400 | 160 400 | FC | 40 | 840 | | | |
| ReLU | 400 | - | | | | | | |
| FC | 10 | 4 010 | | | | | | |
| Softmax | 10 | - | | | | | | |
| | $\sum$ | 638 810 | | $\sum$ | 762 448 | | $\sum$ | 767 733 |

**GR Training Process**  In contrast to the definition of a constant batch size of $\mathcal{B} = 100$, $2 \cdot \mathcal{B}$ is used for re-training ($t > 1$). Thus, merging the generated and actual training data is realized by combining generated samples and samples from the current task. For the sake of simplicity, the same amount of data is generated as contained in the new task.

For the GR approach, the generator needs to be trained first. Therefore, $\mathcal{E} = 50$ epochs of training iterations are conducted on the training data. The learning rate for the generator is tuned by a grid-search in each case (VAE and GAN), varying $\epsilon_G \in \{0.001, 0.0001\}$. After samples are generated for the past task, the solver is re-trained with the merged training data for $\mathcal{E} = 25$ epochs. The Adam optimizer (Kingma and Ba 2015) is used to train the solver with a defined learning rate of $\epsilon_S = 0.001$. After completion of a CL task, all components (generator and solver) are re-initialized, which means that they are re-trained from scratch with the merged (joint) dataset.

**Gaussian Mixture Replay GMR**  Several hyper-parameters need to be adjusted for the GMR training process. The underlying DCGMM architecture is fixed and based on three layers (as for the other models). The first layer is a folding layer which is only used to flatten (or reshape) the input signal (transform 2D images into a one dimensional vector). Since no folding is performed, no parameters are adjusted by a grid-search. The second layer is a GMM layer. The third and last layer is the linear classification layer, which allows the assignment of class labels in a supervised manner. A comparable model architecture is depicted in figure 8.16. It is also comparable to the $L1$ architecture described in table 8.4. Thus, the DCGMM used for GMR is neither convolutional nor deep. Despite the simple DCGMM architecture, limited classification accuracies and CL capacities can be expected.

One of the varied hyper-parameters is the number of Gaussian components $K \in \{6 \times 6, 8 \times 8, 10 \times 10\}$. This adaption is used to verify the statement "more is always better" for the number of Gaussian components. Moreover, a special reset method for the annealing scheme (see section 8.3.3) is introduced. The reset method describes the reset of the annealing parameter $\sigma$ after the completion of a task. It is noted that the $\sigma$ parameter defines the intensity of neighboring updates around the BMU. By means of the reset method, $\sigma$ is set proportionally to the start value $\sigma_0$ of the training. This procedure simulates a training from scratch for the underlying GMM components. The following values are investigated as reset factor by the grid-search: $\sigma_{reset} = \{-1., 0., 0.25, 0.5, 0.75, 1.\}$, whereas $-1$ disables the reset. Based on the grid-search, different loss functions for the last classification layer are evaluated (Cross-Entropy (CE) loss and Mean Squared Error (MSE), see section 2.2.2.1).

Both, the GMM and the classification layer are trained with the same learning rate $\epsilon_G = \epsilon_C = 0.01$. The classification layer is trained by the application of two different loss functions, MSE and CE loss. In order to verify that GMR is insensitive to the conducted training iterations, a high number of training epochs is defined. The first sub-task $T_1$ starts with $\mathcal{E} = 50$ training epochs. Depending on an SLT's number of sub-tasks, the number of training epochs is doubled for each task. This choice is made in order to exploit a possible linear forgetting effect. All other parameters are set as defined in section 8.3.4, and therefore not adapted to a special problem. The training and test data are normalized to the value range of $[0, 1]$.

### 9.2.2   Task Boundary Detection Experiments

The first experiment exclusively refers to the GMR model. The goal is to determine to what extent the model is suitable for the recognition of sub-task boundaries. In order to detect sub-task boundaries, it is an important functionality for CL scenarios, which is offered explicitly by only few ML models (e.g.,DNNs). In general, it is assumed that task boundaries are known and implicitly given. The outlier detection method described in section 8.4.2.3 is applied for the detection of task boundaries.

To demonstrate the detection capabilities, the GMR model is trained step by step with an SLT consisting of one class at a time ($D_{1\text{-}1\text{-}1\text{-}1\text{-}1\text{-}1\text{-}1\text{-}1\text{-}1a}$, $T_1$, ..., $T_{10}$, for the MNIST and FashionMNIST dataset). The log-likelihood value of each batch is traced, so that a new task can be recognized. If a loss value of a data batch is below 80 % of the sliding log-likelihood, a task boundary is detected. The sliding log-likelihood is traced anyway in order to implement the annealing scheme (see section 8.3.3).

A difference with regard to the batch- and sliding log-likelihood needs to be detected within 10 training iterations. After a boundary detection, the model performs 500 training iterations on the new sub-task, before the detection mechanism is re-activated. Again, for ensuring the stability of the process, long training periods with $\mathcal{E} = 200$ epochs are specified for each sub-task. The result for the three datasets is visualized in figure 9.6. For each run of the experiment (10 repetitions), a slightly transparent red bar is drawn whenever a task boundary is detected. As shown in figure 9.6, the dark/solid bars indicate the detection of 10 consecutive classes in most cases. Only in a few experiments a sub-task boundary was re-detected during the long training process.

Figure 9.6: GMR task detection capabilities for a single class SLT.

Figure 9.7 depicts the experimental result of the task detection for a $D_{2\text{-}2\text{-}2\text{-}2\text{-}2}$ SLT. A reduced number of training iterations is conducted and only the first 3 tasks and are illustrated in figure 9.7 in order to simplify the visualization. In addition to the detected task boundaries (red bars), the log-likelihood values are illustrated as blue line. It becomes obvious why the simple heuristic produces acceptable results. For each new sub-task, a significant drop in log-likelihood is noticeable.

Figure 9.7: GMR task detection capabilities for a two class SLT.

These two experimental results imply that (hard) sub-task boundaries can be detected with GMMs. Depending on the chosen strategy, boundaries can be detected after 10 training iterations on a new task. More complex and intelligent heuristics can be introduced in order to detect the boundaries.

### 9.2.3   Continual Learning Experiments

The following experiments are based on the specifications of the CF protocol described in chapter 6. A brief version of the protocol is available in section 9.2.1.1. In this section, the presented Gaussian Mixture Replay (GMR) model (see section 9.1) is investigated with regard to its CL performance. The well-known Elastic Weight Consolidation (EWC) model proposed by Kirkpatrick, Pascanu, et al. (2017) is used for comparison. Elastic Weight Consolidation (EWC) is categorized as regularization approach (see sections 7.1.1 and 9.2.1.2). Since GMR is considered a pseudo-rehearsal approach, it is compared to another replay approach referred to as Generative Replay (GR) (Shin, Lee, et al. 2017) method.

The following conclusions are drawn from the experimental results which are supported by a grid-search with 4 380 performed experiments (see section 9.2.1.2). Ten repetitions of an experiment with the same specific parameter configurations are aggregated as meta-results, whereas the measurement values are averaged. In addition to the average, the standard deviation is given for an improved assessment of the results.

The last measured performance value is always the measurement criterion, except for EWC. The best (maximum) measured performance value is used for EWC in order to compensate a too long training time, as EWC is sensitive in this respect. The baseline experiment (SLT $D_{10}$, see section 6.2) which combines all CL sub-tasks in one is used as a reference value. Experiments are conducted for a subset of datasets. Otherwise additional architectures for the different models would have to be introduced. This is especially true for the GR approach, as sophisticated architectures for the generator are deployed. Thus, only the architectures proposed in related works are investigated.

The results of the conducted CL experiments are summarized in table 9.3. The best accuracy values of the baseline experiment (first line), as well as the differences (diff) to the individual SLTs are presented. The greater the difference, the worse the CL performance of each model. In turn, smaller *diff* values indicate a better CL performance. Averaged values are listed for the GR variants with the different generator types (VAE and GAN). For the VAE variant, the standard deviation is maximally 7 % for all SLTs. This variance implies a constant performance over all repetitions for all SLTs. For GANs, the standard deviation is at most 35 %. An explanation of the high variance is presented later.

As indicated by the results in table 9.3, none of the models provides a perfect CL performance for the investigation protocol. It becomes obvious that EWC shows the greatest CL performance loss. This is due to the sensitivity of the placement of measurement points and the number of training iterations. In case more test-/measurement-points could be carried out, better results can be expected. However, this would tremendously increase the investigation effort, which is hard to realize in certain application-oriented scenarios.

For the investigation of the GR approach, two variants of generators are applied, where VAE is superior to GAN. This is due to the sensitivity of the GAN training. Without an extensive adaptation of the scenario, and respectively the problem and hyper-parameters, an even worse CL performance can be observed. This is implied by the high standard deviation of 35 % for GR using GANs. VAEs seem to be less vulnerable to hyper-parameters, so that acceptable CL performance values are obtained.

The new GMR approach shows the lowest level of CL performance loss, but a poor baseline performance (in contrast to the other models). This is due to the very simple DCGMM architecture (non-convolutional and non-deep). Initial experiments with non-deep and convolutional DCGMMs result in a classification accuracy of up to 97 % for the baseline task.

Regarding the GMR hyper-parameters, the evaluation of the experimental results confirms that more Gaussian components result in a better performance. As expected, GMR experiments with $K = 100$ components consistently demonstrate the best CL performances. This conclusion is similar to the reset method, which influences the $\sigma$ value after each training task. Experiments without resetting the $\sigma$ value always provide the best CL performances. No reset corresponds to a continual training process without adjusting $\sigma$ at task boundaries. In comparison to a complete/full reset of $\sigma$ to $\sigma_0$, a new topological sorting of the prototypes can be established. At the same time, the final CL performance is slightly worse than without a reset. Another finding is that the CE loss function for training the classification layer always provides the best CL results.

In order to provide more insights into the training/evaluation process, individual trends of the (meta-)experiments are presented in figure 9.8. The developments are visualized for different datasets, whereas the trends are similar for all investigated datasets. The various colored lines represent the averaged accuracy of the meta-experiments (10 experiment repetitions) for the best parameter configuration. Baseline experiments with the SLT $D_{10}$ (containing all 10 classes) are conducted in order to calculate the reference value. The relation between the reference value and the measured CL performance is illustrated in figure 9.8. It is referred to as baseline accuracy. The standard deviation is visualized by means of the face/area. For each task, 10 evenly distributed quality measurements are performed during each task training.

Table 9.3: Results of the CL experiments for GMR, EWC and GR.

| model / dataset SLT | GMR |  |  |  |  |  | EWC |  |  |  |  |  | GR |  |  |  |  |  |
|---|---|---|---|---|---|---|---|---|---|---|---|---|---|---|---|---|---|---|
|  | MNIST |  | FashionMNIST |  | Devanagari |  | MNIST |  | FashionMNIST |  | Devanagari |  | MNIST |  | FashionMNIST |  | Devanagari |  |
|  | acc. | std | acc. | std | acc. | std | acc. | std | acc. | std | acc. | std | acc. | std | acc. | std | acc. | std |
| $D_{10\ baseline}$ | 87.4 | 0.5 | 73.9 | 0.2 | 74.1 | 0.7 | 97.6 | 0.2 | 87.6 | 0.4 | 95.6 | 0.5 | 99.3 | 0.1 | 99.3 | 0.4 | 99.1 | 0.3 |
|  | diff. | std | diff. | std | diff. | std | diff. | std | diff. | std | diff. | std | VAE | GAN | VAE | GAN | VAE | GAN |
| $D_{9-1a}$ | 1.3 | 0.5 | 2.7 | 0.2 | 3.2 | 0.7 | 41.8 | 0.2 | 9.6 | 0.4 | 56.6 | 0.5 | 3.6 | 62.2 | 33.2 | 40.6 | 16.0 | 88.1 |
| $D_{9-1b}$ | 3.5 | 2.1 | 1.5 | 0.8 | 1.4 | 0.8 | 50.7 | 7.7 | 20.1 | 2.5 | 29.7 | 13.3 | 3.3 | 44.0 | 28.2 | 40.8 | 14.8 | 84.7 |
| $D_{5-5a}$ | 0.6 | 1.5 | 1.2 | 1.5 | 6.8 | 1.3 | 35.3 | 6.6 | 32.7 | 4.2 | 46.0 | 15.3 | 3.0 | 17.4 | 28.3 | 34.1 | 8.8 | 51.7 |
| $D_{5-5b}$ | 1.3 | 1.9 | 1.9 | 0.4 | 4.7 | 1.5 | 35.0 | 1.8 | 36.0 | 2.7 | 47.1 | 0.1 | 2.3 | 9.0 | 28.0 | 34.7 | 9.3 | 48.5 |
| $D_{2-2-2-2-2a}$ | 9.5 | 3.8 | 8.5 | 0.9 | 22.5 | 2.7 | 72.2 | 7.4 | 55.6 | 4.1 | 72.1 | 2.7 | 23.8 | 63.3 | 30.7 | 54.3 | 56.0 | 79.0 |
| $D_{2-2-2-2-2b}$ | 10.4 | 5.2 | 5.7 | 2.3 | 14.7 | 2.9 | 72.6 | 3.2 | 57.3 | 5.0 | 73.2 | 2.3 | 22.9 | 75.1 | 35.1 | 60.2 | 49.9 | 79.3 |
| $D_{1-1-1-1-1-1-1-1-1-1a}$ | 23.3 | 4.1 | 19.2 | 1.8 | 27.5 | 7.5 | 74.5 | 3.8 | 55.2 | 5.6 | 71.1 | 2.3 | 50.8 | 80.3 | 48.6 | 69.8 | 73.5 | 88.6 |
| $D_{1-1-1-1-1-1-1-1-1-1b}$ | 16.3 | 2.1 | 19.6 | 2.0 | 35.6 | 2.7 | 76.6 | 3.4 | 56.0 | 4.2 | 77.2 | 2.7 | 50.8 | 82.3 | 49.4 | 81.3 | 60.9 | 79.4 |

The EWC results (red) in all sub-figures demonstrate that the longer the training lasts, the more CL performance is lost. What does not become obvious is that a more fine-grained investigation with more measuring points would be beneficial for EWC. However, there is no implemented mechanism to stop the training procedure at the best possible model parameter configuration. The stopping criterion should be realized without violating the requirements of real-world applications, e.g., by using samples from the past.

The GR approach shows two different results with respect to the CL performance. The deployment of GANs as generator for the GR procedure reveals an inferior CL performance in contrast to VAEs. In general, GANs have the ability to generate samples that are at least as good as VAE. The problem is that the chosen model architecture, hyper-parameters or the training process do not seem sufficiently optimized for these datasets and SLTs. In contrast to that, GR with VAEs yields significantly better CL performance. The obtained VAE results are comparable to the CL performances of GMR (taking into account the lower baseline of GMR). These trends become obvious for all SLTs and datasets presented in figure 9.8.

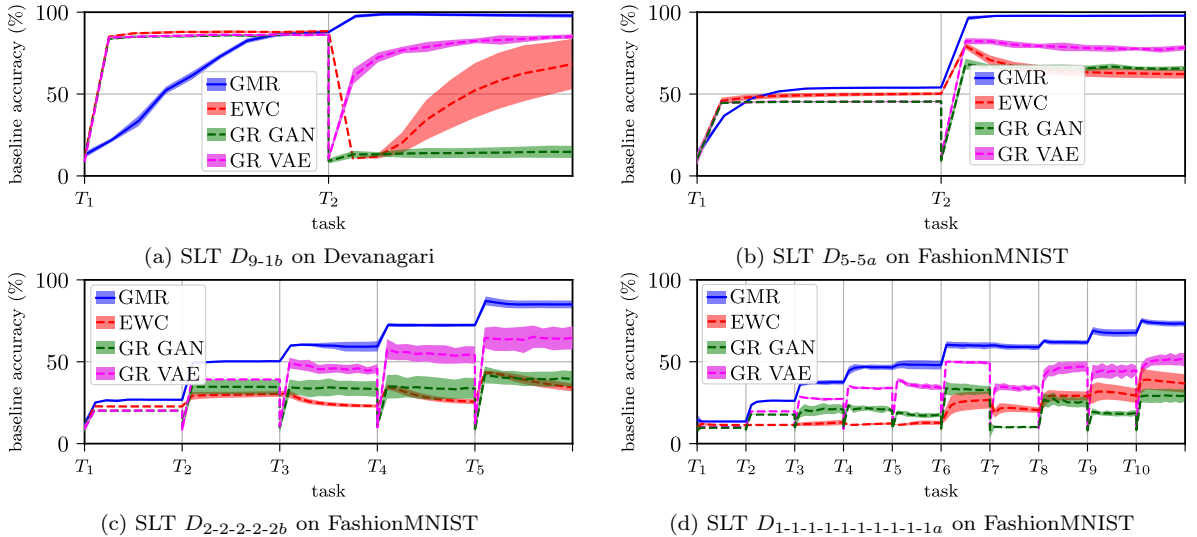

(a) SLT $D_{9-1b}$ on Devanagari

(b) SLT $D_{5-5a}$ on FashionMNIST

(c) SLT $D_{2-2-2-2-2b}$ on FashionMNIST

(d) SLT $D_{1-1-1-1-1-1-1-1-1-1a}$ on FashionMNIST

Figure 9.8: Trends for different SLT experiments for GMR, EWC and GR.

In order to offer more details of the GMR method, the prototypes $\boldsymbol{\mu}$, variances $\boldsymbol{\Sigma}$, as well as the generated samples are illustrated in figure 9.9. The MNIST dataset in combination with the $D_{5-5a}$ SLT is applied, whereas the SLT defines a split of the 10 class dataset into two tasks with five classes each.

Figures 9.9a and 9.9d (top left and bottom) show the derived prototypes ($\boldsymbol{\mu}$) and variances ($\boldsymbol{\Sigma}$) after training $T_1$ for the classes 0, 1, 2, 3 and 4. The Gaussian components represent the model parameter configuration after 50 epochs of training, although convergence has already been achieved much earlier.

The GMM layer has $10 \times 10$ components that are represented as a grid. Again, the topological sorting that gradually passes over the edges (left right and bottom top) becomes visible.

After completing the training of $T_1$, samples $G_1$ are generated as shown in figure 9.9b. The placement corresponds to the (repeated) sampling vector consisting of all 5 classes $(0, 1, 2, 3, 4, 5)$. The diversity of the individual samples is limited by the chosen architecture and the top-$S$-sampling strategy. Moreover, the applied variances become visible, which can be interpreted as noise.

For the second sub-task, the training data of $T_2^{train}$ is merged with the generated samples $G_1$ ($\otimes$ corresponds to a mixing ratio of $1 : 1$). Certain components can be preserved in the long-term by means of the replay mechanism. The prototypes in figure 9.9c and the variances in figure 9.9e represent the state after another 50 epochs of training. Comparing the prototypes of $T_1$ with $T_2$ indicates that the positions of the retained components for $T_1$ have not changed. Only some components have been "overwritten" in order to reflect/address the data distribution from $T_2$.

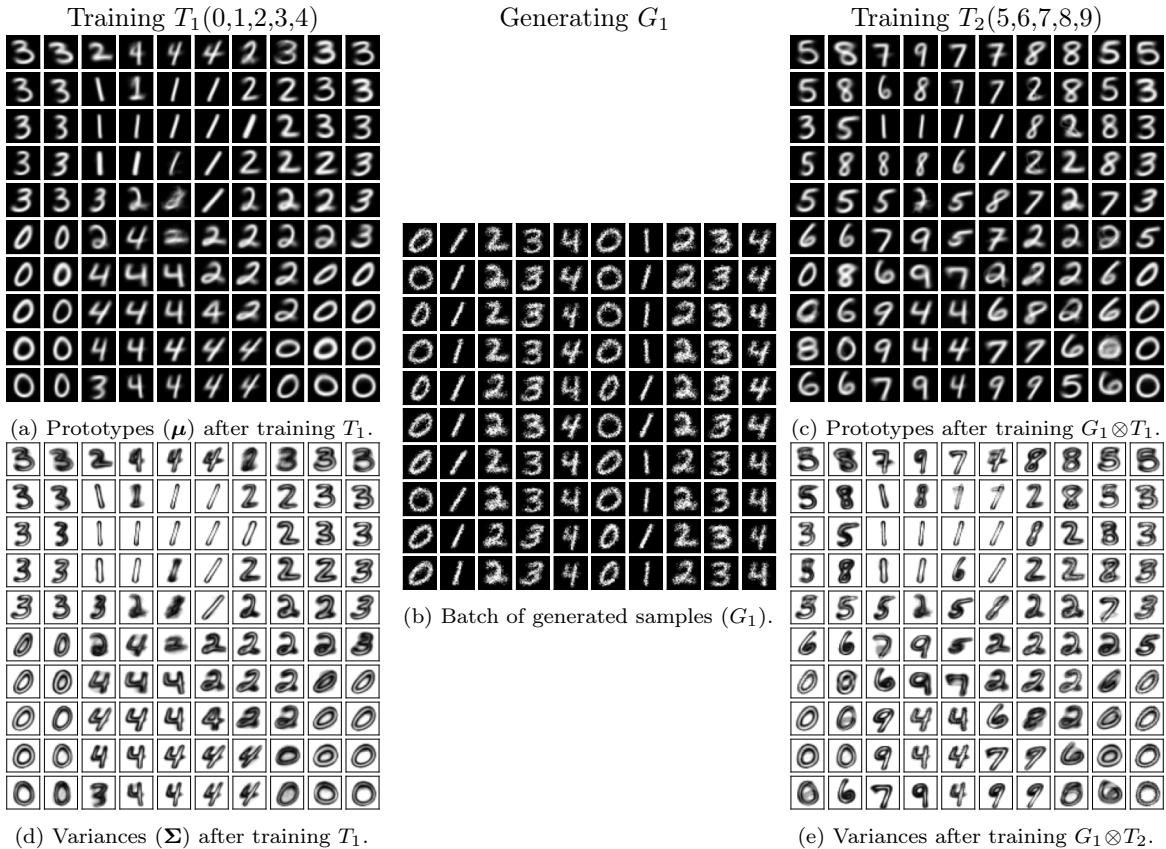

Figure 9.9: Visualization of GMM $\boldsymbol{\mu}$, $\boldsymbol{\Sigma}$ and generated samples $G_1$ for SLT $D_{5\text{-}5a}$.

## 9.3 Discussion

In this section, various aspects of the proposed GMR approach are discussed. One aspect is the sampling quality, which is depicted in figure 9.9. Furthermore, it is discussed why the comparative performance values differ from conclusions drawn in related work. Another point up to discussion is related to the interpretation of the findings. Moreover, advantages of the GMR are outlined, specifically with focus on the functionalities. In this context, the reconciliation with the real-world requirements is discussed, as well as the role of replay approaches for the CL paradigm.

**Sampling Quality**    Replaying generated samples is not a novel CL approach, especially with regard to mitigating the CF effect in deep learning models. VAEs and GANs offer enormous potential concerning the samples' quality and variability (e.g., Wang, Liu, et al. 2018; Karras, Aila, et al. 2018). In general, the pseudo-rehearsal process is strongly dependent on the quality and complexity of the generator. If the generated samples do not fully represent the problem, an inappropriate data distribution may be learned, which leads to a loss of previous knowledge.

The GMR method is based on DCGMMs introduced in chapter 8. DCGMMs represent an SGD-based training technique and layerification of GMMs. The generated samples in figure 9.9b indicate that the quality is not convincing with respect to the original data distribution (MNIST). It is noted that the simplest model architecture is deployed (a non-convolutional folding layer, a GMM and a classification layer).

In the context of CL, the sampling quality is sufficient for obtaining a comparative CL performance. The sampling quality contrasts the other generative models, which can only be re-trained with high-quality and representative samples. In relation to the GR model, both solver and generator need to be independently re-trained again and again for each CL sub-task. This procedure is mandatory, since all components are subject to the CF effect, and previously learned knowledge is lost by further retraining.

The occurrence of the CF effect is different for the GMR approach. The basic knowledge is mainly stored in an unsupervised manner within the GMM layer. The effect of slow forgetting in GMM layers is demonstrated in section 9.1. As a consequence, the linear classification layer is primarily affected by CF. Therefore, the pseudo-rehearsal mechanism addresses knowledge prevention in the last linear classification layer. Thus, by replaying generated samples, the assignment of a class to a Gaussian component is refreshed. This effect becomes obvious in the position of the prototypes in figure 9.9a and figure 9.9c, as well as in the trends of figure 9.8. For this reason, the quality of the generated samples is less relevant.

As the quality of the generated samples is not crucial, other replay mechanisms can be realized. Another approach could comprise the freezing of particularly representative components and the associated class. In this case, the activation pattern only would have to be replayed. Therefore, no samples would need to be generated.

**Comparative Performance Values**   Another point up to discussion concerns the low performance values of the reference CL model in the context of related work. This may be due to several factors. In general, determining suitable hyper-parameters is time- and resource consuming. Even though a certain number of parameters is varied for the investigation (see section 9.2.1.2), specifying them for diverse model types, CL tasks (SLTs) and problems (datasets) is challenging. This is especially true if models or their CL performance highly depend on the choice of hyper-parameters. The deployed architecture is one of the hyper-parameters which is difficult to define for different types of tasks and problems.

The applied evaluation strategy is another crucial factor that influences the resulting CL performance. As mentioned before, EWC seems to be more sensitive to the number of re-training iterations, respectively the position of measurement points (see figure 9.8). In figure 9.8, a peak in the CL performance is noted after a task change. If more measurements are performed, a higher maximum performance can be expected. However, this would lead to the disadvantage that the perfect time or model parameter specification needs to be determined, which can be difficult under certain conditions.

Furthermore, GANs as sample generators cause various challenges. The mutual training process of generator and discriminator leads to dependencies. If one of both components is not well configured, the entire training process is affected. Furthermore, GANs are subject to "mode collapse". In this case, the GAN solely generates identical samples from a single class, which cannot be distinguished from real samples by the discriminator. Thus, the variability disappears and so do the samples from other classes. Alternatively, the discriminator dominates, whereupon the generator learns nothing (caused by vanishing gradients). In this context, the relationship between the alternating training of generator and discriminator is particularly important. These kinds of problems are addressed by many different GAN variants, e.g., Wasserstein GAN (Arjovsky, Chintala, et al. 2017), DCGAN (Radford, Metz, et al. 2016), BEGAN (Li, Xiao, et al. 2018). However, even the adapted variants need further optimization (see Gulrajani, Ahmed, et al. 2017).

Another problem of GANs is that they are affected by the CF effect as outlined by Thanh-Tung and Tran (2018). Therefore, GANs always need to be trained with the joint training dataset by using self-generated samples. Furthermore, the determination and adjustment of GAN parameters seems to be a fundamental issue since the training process is very sensitive. This problem manifests in the experimental results, where no satisfying parameter configuration was identified. For the application of VAEs, the parameterization seems to be easier with respect to the resulting CL performance.

**Interpretation of Results**   The results of the CL investigation can be interpreted as follows. It is noted that the GMR approach exhibits valuable CL capabilities. Moreover, the model is less complex and the stored knowledge is easier to interpret, i.e., prototypes, variances, etc. The examined GMR architecture has a lower number of free parameters and is therefore more memory efficient. The GR variants contain approx. $3\,000\,000$ free parameters on average. In contrast, the used GMR model contains $2Kd + 10K + 10 = 201010$ (for components $K = 100$, data dimensionality $d = 1000$ and 10 classes).

The current disadvantage of GMR is the significantly lower baseline performance. The baseline performance of $87.4\,\%$ on the MNIST dataset is too weak. In contrast, the DNN used for EWC presents a baseline performance of $97.5\,\%$ which is due to the architecture of the DNN. The EWC, however, has no influence on the baseline performance. State-of-the-art performance for MNIST is presented by the solver network ($99.3\,\%$) of the GR approach (without fine tuning). Regardless, the obtained results are in the error range of $10^{-1}$ within the world rankings.

Further optimizations regarding the architecture and fine-tuning of parameters can lead to better CL performances. However, the trends of the different models will remain the same. The problem of the poor baseline performance of GMR can be addressed by using more complex DCGMM architectures. First experiments imply that a folding variant of the same DCGMM may achieve a baseline performance of up to $97.5\,\%$. This value still does not equal the common $99\,\%$ on MNIST. It is nevertheless acceptable for the conduction of further research. Considering that the different components of the GR approaches use convolutional layers, the CL performance of the GMR is quite acceptable.

**Functionalities**   Another difference between GMR and the other investigated models is the range of proposed functionalities. GMR owes its functionalities to the underlying DCGMM. Discriminative methods such as DNNs are not able to perform density estimation. Thus, it is difficult to distinguish outliers from inliers. This function is a significant advantage with respect to a reaction to task boundaries.

Most CL models are based on the definition of task boundaries. In order to resolve CL problems (SLTs), all models require and use the task boundary trigger, e.g., to start a merging mechanism or start generating samples. Without specifying the boundaries, an additional model needs to be introduced, even though it adds complexity. The same applies to the separation of generator and learner, as it is the case for the GR approach. In contrast to that, GMR combines generator and solver in one entity. Other functions such as sampling, inpainting or variation generation can only be performed with generative methods. Unfortunately, EWC does not belong to this type of methods (which is not its goal).

**Reconcile with Real-World Requirements**   Another point that is discussed addresses the fulfillment of real-world requirements for CL applications. One requirement for all investigations is that neither data from the past nor from the future can be accessed for optimization purposes. This is specified as part of the investigation protocol (see chapter 6). Both, GMR and GR generate samples from past data distributions and therefore do not require additional memory for hold-out buffers. Likewise, EWC does not require samples from the past.

However, another requirement is related to a constant time complexity and the number of previous sub-tasks. This requirement can only be fulfilled, if the same number of samples is generated for each new sub-task. If the number needs to be proportional to the number of training samples in the new sub-task, this requirement is particularly difficult to realize. This particularly applies to streaming scenarios, as the number of samples is not known beforehand. As a consequence, a constant number of generated replay samples has to be sufficient for all replay models, regardless of the number of sub-tasks. This constraint was not implemented as part of the present investigation, although GMR is hardly affected by the number of generated samples. A replay experiment with a fixed number of samples (5000) per sub-task achieves the same CL performance. This is due to the location where the forgetting takes place, namely the classification layer.

Since EWC does not demand any past or future data for suppressing the CF effect, defining a point to stop the training process is crucial. A mechanism or criterion to stop the training process is missing in EWC and would have to be added. At the same time, no samples from the past can be used to determine an appropriate point to stop the training process. This, in turn, would contradict the memory constraints for a potentially infinite number of sub-tasks.

Another constraint prohibits the access of future data. Future data is sometimes used to adjust the model architecture or hyper-parameters in advance. This methodology is often applied, although it is technically unfeasible in real-world applications. The same is true for the architecture of the generator and solver for the GR approach used in this study. Problem independence with regard to the hyper-parameters seems to be an advantage of the GMR model. The parameter optimization always results in the same parameter configuration for all problems/datasets – at least in this study. Accordingly, the deployment of GMR could be realized in zero-knowledge scenarios.

**Generative-Replay and Real-World Constraints**  Another issue related to generative-replay approaches and CL is of a general nature. The ratio of samples from new to old classes must be adhered to, even if a fixed number of samples is generated. This is due to the sensitivity of the DNNs in terms of data distribution and the number of classes. If generated samples from 99 classes are mixed with a single new class, the resulting ratio is 99:1. For a batch of data with the size $\mathcal{B} = 100$, 99 samples are generated and only one sample of the stream can be processed at a time. This constantly increases the training time by each CL task's number of new classes. For this reason, replay approaches were excluded from the initial investigation in chapter 7.

The basic problem is that data needs to be replayed in order to enable joint training. In fact, replay approaches contradict the idea of *continual learning*. The following method addresses this problem, but solely mitigates the arising problem of constant update complexity. The underlying idea is that the model itself determines whether or not forgetting begins. DCGMMs provides the required functionalities (i.e., density estimation) to detect forgetting. Implementing the detection of forgetting and proper reactions is considered a subject for further research, even though it merely shifts the problem. If the forgetting rate increases, more knowledge needs to be retrained and the more complex the training becomes for new tasks. For this reason, the question arises whether a replay approach constitutes the best possible solution for CL tasks.

## 9.4   Conclusion and Future Work

In this chapter, the Gaussian Mixture Replay (GMR) model is introduced which represents a pseudo-rehearsal approach for CL based on DCGMMs (see chapter 8). The method describes a replay mechanism that generates samples of the learned data distribution. Generated samples are utilized to realize a joint training with past and current data. The main difference to conventional replay approaches, e.g., GR (Shin, Lee, et al. 2017), is that forgetting mostly takes place in the last linear classification layer. Therefore, the main part of knowledge is stored in the unsupervised GMM layer. Generated samples are used to mitigate the CF effect by a joint training of the liner layer which is responsible for mapping a label to an active Gaussian component.

For the evaluation, different datasets are divided into SLTs, which represent various CL tasks with hard task boundaries. Moreover, the evaluation protocol presented in chapter 6 implements requirements from real-world applications. Access to data from future tasks is prohibited for optimization purposes. In order to establish comparability, different CL models were subject to the same evaluation protocol. This includes the well-known EWC (Kirkpatrick, Pascanu, et al. 2017), as well as state-of-the-art pseudo-rehearsal methods such as GR (Shin, Lee, et al. 2017). The latter can operate with different types of generators: VAEs (Kingma and Welling 2013) and GANs (Goodfellow, Pouget-Abadie, et al. 2014).

In summary, the CL investigation results indicate that the GMR model obtains comparable CL performance even with the simplest architecture. It is claimed that the GMR model is affected differently by the CF effect compared to other deep learning models (e.g., DNNs). A limitation, however, is the lower baseline performance, which is about $10\,\%$ worse compared to the other models. However, this can be compensated with more complex convolutional and deeper architectures. Initial investigations indicate accuracies of up to $97\,\%$, which is acceptable as a basis for future research.

The problem of the lower baseline performance can be addressed by the use of deeper and/or convolutional DCGMMs. A study in the context of the CL paradigm could be a first step to verify a causal correlation. A better quality of the generated samples could have a particular positive impact on the CL performance. Investigating other problem domains is another area for future research. The study of discrete features could be part of these problem domains. In this context, an adaptation of the sampling process could be beneficial.

Another open research questions concerns the gradual transition between the tasks. A more intelligent adaptation to different types of changing data distributions is important with regard to real-world scenarios. The basis could be provided by monitoring the log-likelihood of streamed samples. Various reactions would be an option, such as the adjustment of the learning rate $\epsilon$ or the annealing parameter $\sigma$. The decision when, which and how many samples need to be generated by the model itself is a further optimization of the replay process.

In the CL context, other mechanisms can be realized in order to protect prototypes from modifications. One approach could be the utilization of BMU counters or component weights $\boldsymbol{\pi}$ for the decision whether a prototype should be protected or adapted. "Selective forgetting" can be considered as a young research area for DNNs (see Hayase, Yasutomi, et al. 2020). In this context, existing knowledge needs to be explicitly removed or changed. Assuming that the individual GMM components store the derived knowledge, the presented model constitutes an interesting opportunity for selective forgetting. Various mechanisms for adding and removing GMM components (Kristan, Skočaj, et al. 2008) can be realized in order to address the selective forgetting or the CL paradigm.

# 10.  Findings and General Discussion

## Chapter Contents

This chapter highlights the relevant conclusions in order to answer the research questions of the present work. This summary includes the findings and challenges that were obtained in the course of this work. Likewise, the research questions are revisited and answered.

**Structure**  The structure of this chapter corresponds to the sequence of the research questions presented in chapter 4. For each section, a brief summary of the corresponding chapter is presented. It is followed by a holistic discussion of the rising challenges, before the research questions *RQs* are answered.

First of all, the presented evaluation protocol is discussed in section 10.1. The evaluation protocol is applied in a study of different deep learning models. Findings of the investigation with regard to the *continual learning* (CL) performance are discussed in section 10.2. An independent part of this work focuses on the development of a novel deep learning model – referred to as Deep Convolutional Gaussian Mixture Model (DCGMM). DCGMMs and their functionalities are discussed in section 10.3. This chapter is concluded with a discussion of Gaussian Mixture Replay (GMR), which transfers the presented model into the CL context. Again, the evaluation protocol is used in section 10.4 in order to establish comparability.

## 10.1  Application-Oriented CL Evaluation Protocol

The evaluation protocol proposed in chapter 6 is intended to detect the occurrence of the *catastrophic forgetting* (CF) effect in deep learning models. The CL performance is related to the CF effect due to the ability to protect past knowledge. In order to develop an application-oriented evaluation protocol, various real-world requirements are derived from the real-world scenario presented in chapter 5.

The protocol describes the training and evaluation process of *machine learning* (ML) step by step. The protocol implements various application-oriented requirements, as it aims at drawing conclusions regarding the applicability of a ML model. The requirements are derived from an exemplary real-world scenario, where CL capabilities constitute a basic prerequisite. Unfortunately, it has become obvious that an evaluation within the exemplary scenario is challenging. The dynamics and properties of the data would lead to results that are difficult to interpret.

The comparison of the proposed evaluation protocol to others from related works reveals several aspects. In many cases, the appropriate level of detail is lacking or its specification is missing. However, these "small" details can be decisive for the applicability of CL models in real-world application scenarios. The proposed protocol distinguishes between two quality measurement criteria: "Best" and "last". The *best* quality criterion is defined by the best possible model parameter configuration that can be achieved during the training process. Quite often, this criterion can only be implemented for the initial model training, which corresponds to the product development or manufacturing phase. In this phase, sufficient computational resources are available for the optimization. The same applies to a computational optimization (grid-search) regarding hyper-parameters, e.g., a model's architecture.

During the initialization phase, data from future tasks are not accessible, which corresponds to the application in a real-world scenario. A comparable scenario is the manufacturing of a robot and its delivery, although knowledge needs to be added after the delivery. The constraints and requirements of the realistic scenario are implemented as part of the proposed evaluation protocol. Specifically prepared datasets with a high number of features are supposed to allow for the investigation of CL capabilities of various CL avoidance models. In order to derive CL tasks, the datasets are divided into multiple sub-tasks. The sub-tasks consist of different disjoint class combinations, where knowledge needs to be extracted sequentially task by task. It is obvious that this type dataset division does not represent all possible CL tasks, i.e., different types of changing data. Nevertheless, first conclusions can be drawn with regard to the extraction and preservation of knowledge from the entire dataset. Accordingly, the so-called Sequential Learning Tasks (SLTs) represent the easiest type of CL task which is due to the indicated hard task boundaries.

The adaptation of problem-dependent hyper-parameters is crucial for the performance of ML models. The CL evaluation protocol specifies an initial hyper-parameter optimization for several parameters. The standard optimization procedure is based on a training dataset where multiple experiments are performed with different combinations of hyper-parameters. After conducting multiple experiments, the best parameter constellation is determined according to the experimental results. In this optimization phase, the model's architecture is selected.

For the following CL sub-tasks, a retroactive optimization with respect to a model's architecture is unrealistic. Some hyper-parameters can be adapted, although this is very computationally intensive. This kind of optimization would be challenging for embedded services due to limited resources, e.g., memory, processing or power supply. Nevertheless, the protocol allows the subsequent optimization of one important hyper-parameter, i.e, the learning rate. The main purpose of a second optimization is to counteract the lack of an extensive fine-tuning process in the initial phase. Thus, this second and smaller optimization should compensate the lacking fine-tuning.

**Real-World Requirements for CL Scenarios *RQ 1.1*** *RQ 1.1* addresses the question of real-world requirements for CL scenarios. An exemplary scenario described as part of chapter 5 serves as a basis for deriving CL requirements. The CL scenario describes a challenge from the area of computer networks which should be tackled by the application of an ML technique. Unknown metadata at the beginning of a network connection is the information that is supposed to be predicted, e.g, the duration or number of transferred bytes. The challenge is to train an ML model and enable the prediction in a data stream.

Metadata of completed communications are provided as a stream of data in order to train an ML model. For this reason, the corresponding ML model has to fulfill different scenario-specific requirements, e.g., incremental training. Deep Neural Networks (DNNs) constitute a suitable method, since they have a constant update complexity and derive complex functions from data. The goal is to predict metadata (e.g., bit-rate of a communication) that can be used, for example, in more advanced applications such as optimized routing technologies. In the context of the present work, the exemplary scenario is discussed from the perspective of the CL paradigm. Accordingly, various requirements for the CL scenario were derived. Considering the dynamics and influence of experimental trends in all aspects, e.g., human variations, application updates and many more, continual adaptation is essential for success.

The CL evaluation within the real-world scenario seems very challenging. The unknown change of the data distribution leads to untraceable dynamics, causing obstacles related to the specification of a learning objective. To conclude, the derivation of various requirements and their implementation by the evaluation protocol seem to be the only option.

**CF Effect in DNNs *RQ 1.2*** The second sub-research question *RQ 1.2* addresses the manifestation of the CF effect in DNNs. After training a DNN with an excellent performance, additional knowledge should be added. The new knowledge is presented in form of a new/different data distribution. The problem of an abrupt knowledge loss arises due to further training steps on the new data distribution. The previously derived knowledge is lost after a few training steps. Thus, the CF prevents/complicates the use of DNNs in CL scenarios.

It is obvious that this behavior is not new and has been described in many other works. The effect is proven by training DNNs with a dataset that is divided into two tasks of disjoint classes. Each sub-task represents a different data distribution. By measuring the prediction accuracy on test samples of each task, the occurrence of the CF effect becomes observable. At the same time, the separation of training and test data ensures a valid conclusion regarding the memorization of training data. Considering the performance measures obtained for the individual sub-tasks, the rapid drop in knowledge with regard to the previous sub-tasks becomes clear.

**Existing CF Evaluations Protocols *RQ 1.3*** Research question *RQ 1.3* refers to the investigations carried out in related works. Various scientific studies define applied evaluation schemes in an inconsistent manner. The experimental setups of related work usually describe various CL tasks and hyper-parameter configurations for evaluation purposes. In addition, the parameters leading to the best possible performance are presented. But the determination of the the parameter constellation or the description of a dedicated evaluation procedure is often omitted. This is not necessarily due to the authors' intentions, but rather to self-evident requirements or page limits. Even if a code base is available, the scheme and evaluation strategy is frequently omitted. The absence may be due to the fact that the focus of research is often on a novel CF avoidance model. Tracing the code base of an implemented protocol is usually very time-consuming. These circumstances lead to the challenging comprehensibility of results and classification of the findings.

**Application-Oriented CL Protocol RQ 1** **RQ 1** is answered by the evaluation protocol presented in chapter 6. The protocol implements several real-world requirements resulting from the answers of the sub-research question *RQ 1.1*. Based on the conclusion of *RQ 1.2*, the protocol investigates CL tasks that allow to measure the occurrence of the CF effect. The answer to *RQ 1.3* helps identify the deficits of the existing protocols, which supports the specification of a precise evaluation protocol.

Several different ML models can be investigated under uniform conditions by means of the developed evaluation protocol. These conditions allow for a comparison of the CF avoidance models and a classification of the methods with regard to the requirements. Conclusions and findings drawn by the CL investigation protocol always need to be set into the context of the implemented requirements of the protocol. Accordingly, findings regarding the CF avoidance capabilities of a deep learning model can never be absolute.

## 10.2   Continual Learning Investigation

The proposed CL evaluation protocol outlined in chapter 6 is used to examine various CF avoidance models. Various existing models were examined for the investigation. According to the respective authors, all of them "avoid", "mitigate" or at least "weaken" the CF effect. The present study comprises standard models such as Fully-Connected (FC)-DNNs and Convolutional Neural Networks (CNNs), which do not address the CF effect and therefore are considered as reference values (baselines).

Furthermore, the avoidance effect of dropout (Hinton, Srivastava, et al. 2012b) applied to the two standard models as proposed by Goodfellow, Mirza, et al. (2013) is evaluated. Special attention is dedicated to different CF avoidance models: Local Winner Takes All (LWTA) (Srivastava, Masci, et al. 2013), Elastic Weight Consolidation (EWC) (Kirkpatrick, Pascanu, et al. 2016) and Incremental Moment Matching (IMM) (Lee, Kim, et al. 2017). These models can be reconciled with the real-world requirements of the investigation protocol, with the exception of IMM. Even though IMM needs data from a past task, the model was examined as it is considered a state-of-the-art model. The contradiction with the presented real-world CL requirements concerns many other CF avoidance models presented in chapter 3, e.g, the introduction of oracles specifying the task ID of a sample as for the Hard Attention to the Task (HAT) model (Serra, Suris, et al. (2018)).

In order to realize the empirical study, the code bases of the ML models were integrated into the developed investigation framework. The framework implements the evaluation protocol and enables both, a uniform execution of experiments and the evaluation itself. The experimental results indicate that none of the investigated models provides a satisfying CL performance for all examined datasets and all CL tasks while complying with the application-oriented requirements. The obtained results contrast the conclusions and findings by the corresponding models' authors. However, significantly better CL performance (similar to the results of related work) can be obtained if the models are evaluated according to a less application-oriented protocol.

Nevertheless, some models achieve an acceptable CL performance for a few CL tasks. By applying EWC, a "catastrophic" forgetting can sometimes be avoided, whereas a "linear" forgetting process was observed. This particular trend can be determined for CL tasks where less knowledge needs to be added (e.g., single class). Unfortunately, this does not apply to CL tasks where more knowledge is added (e.g., from five classes). Also, the linear forgetting effect depends on the number of performed training iterations. Conversely, IMM performs better on tasks with more knowledge value, respectively more classes. It is shown that IMM obtains acceptable results with a subsequent adjustment of the hyper-parameter $\alpha$. In order to tune the parameter that defines the importance of the individual tasks, past data has to be used, even though this is a contradiction to the real-world requirements.

**Existing CF Avoidance Methods *RQ 2.1***     *RQ 2.1* refers to deep learning models addressing the CF problem. This question is answered by a systematic literature review and the resulting list of various methods (see section 3.2). Various existing models try to circumvent the problem in different ways (see section 2.3.5). A fundamental problem is that an implementation is not provided for all CF avoidance models. Furthermore, not all CF avoidance models comply with the defined real-world requirements. This includes, for example, the Hard Attention to the Task (HAT) model proposed by Serra, Suris, et al. (2018). In addition to the features of a sample, the task ID must be specified. For CL tasks with one new class each, this would be equivalent to the real sample's label.

A great number of new approaches or optimized methods is constantly being published. This trend is accompanied by new workshops and categories at high ranked conferences (e.g., ICLR and CVPR) addressing CL. Current developments indicate that the CF problem can still not be considered as fully solved in the CL context.

**CL Investigation Framework *RQ 2.2***     The second sub-research question *RQ 2.2* relates to the evaluation of CL models. This question is answered by providing an investigation framework for the examination of ML models according to the uniform evaluation protocol. The framework requires certain interfaces that need to be met. The manual model integration into the framework needs special attention in order to comply with the requirements. If a model stores samples, it is only allowed to allocate a limited amount of memory. This condition has to ensured by an inspection of the code or monitoring the allocated memory.

In addition, the framework developed as part of this work includes multiple functionalities and satisfies different prerequisites. The framework also provides different datasets and the functionality to implement various types of CL tasks. Furthermore, the framework comprises a mechanism to perform large-scale parameter optimization with all relevant components.

**CL Investigation *RQ 2***     *RQ 2* can be answered by interpreting the CL performance results. The question is to what extent the investigated models can control the CF effect under application-oriented constraints. None of the studied models can completely eliminate or mitigate the forgetting effect for all CL tasks and all datasets. Only for certain problems and for some particular tasks the CF effect can be mitigated under the application-oriented conditions. Therefore, conclusions can solely be drawn if the exact circumstances of the avoidance or mitigation of the CF effect are provided.

## 10.3    Novel Deep Learning Model DCGMM

A novel deep learning model referred to as Deep Convolutional Gaussian Mixture Models (DCGMMs) is proposed in chapter 8. DCGMMs are based on Gaussian Mixture Models (GMMs), which are not considered as deep learning models. GMMs are usually trained by variants of the Expectation-Maximization (EM) algorithm (see Dempster, Laird, et al. 1977), and they belong to the unsupervised learning methods. EM has some disadvantages that complicate its use in a data stream. Even though some adaptations, such as a stochastic version referred to as stochastic EM (sEM) (Cappé and Moulines 2009), allow an incremental training, the initialization problem of EM and sEM remains.

The Stochastic Gradient Descent (SGD) based training method presented in section 8.3 allows the incremental training of GMMs and solves the data-driven initialization problem. In order to denote GMMs as "deep" models, a layer arrangement is suggested in section 8.4. In addition, various supplementary layer types are introduced which support the use of GMM layers. On the one hand, classification layers transform DCGMMs into a supervised model. On the other hand, folding/convolutional layers perform a comparable transformation which empowers CNNs in the context of image processing. The layerification realizes many different functionalities: density estimation, classification, (conditional) sampling and outlier detection, just to name a few. In order to verify the validity of the SGD-based training approach, various experiments prove that equivalent results as compared to conventional standard training algorithms (EM, sEM respectively) can be achieved. As the focus is not solely on CL applications, the novel model is also suitable for other research areas.

The new training procedure is in no way intended to present a superior approach, e.g., compared to EM or sEM. Nonetheless, no data-driven initialization is required for training GMMs, which is a clear advantage compared to the standard procedures. Even though the training procedure introduces multiple hyper-parameters, the parameters are often intuitively chosen or do not seem to be very problem-dependent. All of these advantages advocate for further exploring this approach.

The associated research question RQ 3 consists of two sub-questions, which can be considered as alternatives to each other. Either existing approach can be extended, or multiple CF avoidance techniques can be combined. The alternative research question focuses on the development of a novel deep learning model in the context of the CL paradigm. The latter is the case in the present work.

**Existing CF Avoidance Techniques *RQ 3.1***   Sub-research question *RQ 3.1* addresses how CF avoidance models suppress the forgetting effect. The basic idea is to understand existing CF avoidance techniques in order to combine or extend them. The aim is thus to develop an improved or novel CF avoidance technique.

Categorizing existing deep learning methods with regard to their approach towards CF reflects their underlying concept (see section 2.3.5). The corresponding question can therefore be answered by a literature review (see section 3.2). The best model in this study is the regularization approach EWC (see section 7.1.1). EWC is based on a penalty term in the loss function that prevents task-specific model parameters from being changed. However, determining the importance of parameters for different tasks is challenging. Another problem is related to stopping the training process in order to get the best possible model parameter configuration. The determination of the best possible stopping point becomes more challenging in application-oriented scenarios.

Parameter isolation constitutes yet another CF avoidance category – LWTA is one of them. Different sections of the underlying DNNs are identified and assigned to a certain sub-task. The basic problem is the selection of a section that is responsible for a certain sub-task or sample. If this additional information is unavailable, the choice becomes challenging. The same is true for the implementation of multiple readout layers. Usually, oracles provide guidance for the selection. As oracles are rare in real-world scenarios, they are prohibited by the application-oriented requirements. By specifying a sample's task ID, a CL task consisting of one class per task is already solved.

Another approach that suppresses the CF effect is replay. Both, rehearsal and pseudo-rehearsal ensure that a joint training with a data distribution of past and current sub-tasks can be performed. This way, knowledge can be derived from all combined data distributions. Replay processes simulate a joint training with all data samples.

Replay mechanisms do not directly address the CF effect, but they frequently achieve the best CL performance. Storing samples from old tasks violates application-oriented memory constraints. In order to consistently store samples, a potentially infinite amount of memory needs to be provided. As this is impossible in data streams, the number of samples that can be stored is limited. This leads to new challenges, e.g., which samples are most representative and how a potentially infinite number of tasks can be processed. As an alternative to storing samples, the sample's influence on the model is stored (e.g., the gradients). Storage capacity, however, is still limited, so that the same challenge arises again, sooner or later.

Another method for the realization of rehearsal is the generation of samples from the derived data distribution. Generative models like Variational Autoencoders (VAEs) (Kingma and Welling 2013) or Generative Adverserial Networks (GANs) (Goodfellow, Pouget-Abadie, et al. 2014) are considered to be excellent generators that can be used for pseudo-rehearsal procedures. Generators can be used to circumvent memory constraints by simply generating samples that represent the past data distribution. Thus, representative samples can be generated and mixed with samples from new sub-tasks in order to provide a joint training dataset.

However, new challenges arise with regard to the quality of the generated samples. If generated samples do not completely represent the data distribution of a previous sub-task, knowledge may get lost. For each re-training step, the previously generated samples need to be used for the derivation of the joint data distribution. Furthermore, it is often difficult to measure the quality of the generated samples. In addition, the complexity of the training process of the generators is another challenge. Other open questions concern the required number of generated samples and the update complexity.

The fundamental problem of using generative approaches is that it contradicts the idea of continual learning. By simulating a joint training, the CF is not suppressed, but bypassed. Once new knowledge is added, old knowledge needs to be relearned. This leads to an increasing training complexity with each new task. Thus, it is no longer possible to efficiently add further knowledge after a certain amount of knowledge has been processed. For this reason, replay approaches were excluded from the CF avoidance investigation.

**Non-Deep Learning Methods RQ 3.2**   *RQ 3.2* relates to ML models that are not inherently deep, but not directly subject to the CF effect. In this context, many different ML models/algorithms are worth to be considered. Examining various deep learning techniques in the CL context drew attention to GMMs. IMM (Lee, Kim, et al. (2017)) models the importance of parameters by Mixture of Gaussians (MoG). Research in the area of CL shows that GMMs are investigated and used in CL scenarios, e.g., by Kristan, Skočaj, et al. (2008) (or see section 8.2). Therefore, an attempt is made to address the open issues of the standard learning procedure and develop a new GMM training technique based on SGD.

GMMs are inherently unsupervised learning methods. Without the ability to perform classification, a comparison to discriminative ML models (e.g., DNNs) is impossible. DNNs consist of $L$ numbers of hidden layers and $S$ numbers of artificial neurons. The more layers and artificial neurons, the more complex functions can be approximated (universal approximation theorem, see Cybenko 1989). The same applies to GMMs, but their ability to approximate complex functions depends on the number of Gaussian components $K$ (see Goodfellow, Bengio, et al. 2016). Since the computational cost linearly depends on the number of components $K$, layerification would be beneficial for GMMs. Thus, the objective is to develop a deep learning model based on the SGD-based optimization technique and a layerized version of GMMs.

**Novel Deep Learning Method RQ 3**   A part of the answer to research question **RQ 3** is the developed novel deep learning approach referred to as Deep Convolutional Gaussian Mixture Model (DCGMM). The novel procedure is based on a SGD training process for (deep) GMMs. The annealing scheme is very useful (outstanding), as it allows GMM training without a data-driven initialization. Similarly, DNNs can be trained without data-driven initialization.

Moreover, layering GMMs supports their combination with conventional types of layers known from other deep learning models. Therefore, DCGMMs can be denoted as deep learning models. By adding various operations and layers, DCGMMs can be used to discriminate (classify) or to (conditionally) generate samples. In order to answer the research question **RQ 3** completely, the CL properties of DCGMMs need to be evaluated. A CL investigation of DCGMMs is part of an extension and referred to as Gaussian Mixture Replay (GMR). GMR is a replay mechanism for DCGMMs (see section 10.4).

## 10.4   Gaussian Mixture Replay

GMR is a pseudo-rehearsal approach for CL based on DCGMMs (see chapter 8). The replay method makes use of the conditional sampling functionality of DCGMMs in order to generate samples that are replayed for training new sub-tasks. The basic difference to conventional deep learning models (like DNNs) is that the knowledge is mainly stored inside the GMM layers. GMMs belong to the unsupervised learning methods. DCGMMs obtain their classification ability by an additional standard linear classification layer. The linear layer is responsible for mapping an activation of one or multiple GMM components to a single class. Therefore, the replay mechanism is mainly used to protect knowledge in the last layer.

In order to evaluate the GMR approach, the evaluation protocol from chapter 6 is applied. Furthermore, EWC (Kirkpatrick, Pascanu, et al. 2016) and Generative Replay (GR) (Shin, Lee, et al. 2017) as another replay approach is investigated and compared to GMR. The obtained results indicate that the CL performance of EWC is very dependent on the number of performed training iterations and measurement points. Thus, EWC achieved the lowest CL performance according to the evaluation protocol.

GR was deployed and evaluated with two different types of generators: VAEs (Kingma and Welling 2013) and GANs (Goodfellow, Pouget-Abadie, et al. 2014). A CNN served as a solver for the GR approach. The investigation indicates that the CL performance of GR strongly depends on the quality of the generator. Despite the large number of introduced hyper-parameters and the definition of the training scheme, GAN training seems to be unstable in the conducted experiments, unlike VAEs. Therefore, the CL performance of GR including VAEs is considered as representative. A satisfactory CL performance of GR can be recognized, especially for CL tasks consisting of two sub-tasks. Nevertheless, the more CL sub-tasks are included, the lower the resulting CL performances.

The solver (CNN) of GR, as well as the DNN used for EWC outperforms the baseline performance of GMR. The inferior performance of GMR can be attributed to the applied DCGMM architecture – a single GMM layer ($K = 100$) and a classification layer. Similar classification rates to those of simple DNNs (as used for EWC) can be obtained with a convolutional DCGMM. Considering the baseline accuracy for the MNIST dataset (about 97 %) implies that the obtained performance is not equivalent to the state-of-the-art for this dataset. Nevertheless, the goal is to provide a CL performance that is acceptable for an application-oriented context. This is exactly the case for the GMR approach. A CL performance comparable to other generative replay methods is achieved with respect to the baseline accuracy.

**Reconcile GMR with Real-World Requirements**  The main challenge is to reconcile a pseudo-rehearsal mechanism with application-oriented real-world requirements. First of all, the fundamental problem of replay approaches is introduced. Basically, the challenge with replay approaches is that they violate the requirement of constant update complexity. This is due to the generation of "past" samples. It may not be possible to create enough samples for a representative data distribution due to a fixed number of generated samples. Thus, generating 20 samples would be sufficient, as long as 20 classes need to be distinguished. The number of generated samples can be increased step-wise. Even though this strategy is problem-dependent, it is the only possibility for streaming scenarios. A second and more difficult challenge is related to the joint training dataset. In this context, the particular dependence on strongly varying data distributions within the classes is especially problematic for DNNs. The more knowledge/classes have been learned in past tasks, the more samples need to be generated relative to the new knowledge/classes. A weighting mechanism can be considered as a first attempt to address this problem, but it only works for a particular number of knowledge/classes.

All replay processes are affected by this uneven data distribution problem, including the presented GMR. GMR, respectively DCGMMs, offer several possibilities to address this problem. Estimating and monitoring the densities of training samples or generated samples can be accomplished by the model itself. By means of these mechanisms, forgetting can be detected and counteracted, e.g., by weighting or generating samples for a joint training. In this context, best matching unit (BMU) counter or component weights $\pi$ can become involved in order to protect or release GMM components for further training. Therefore, the GMR model can be considered as a first CL approach incorporating DCGMMs.

**CL with GMRs RQ 3**  The replay procedure GMR is the second part of the answer to research question **RQ 3**. Accordingly, the presented GMR is another attempt to address the CF effect for deep learning models. The proposed DCGMM is used as basis for the replay mechanism. A first investigation indicates that GMR provides various properties suitable for CL scenarios. The CF effect is addressed by the proportional shifting of knowledge into unsupervised GMMs. A layered version of GMMs is differently affected by forgetting compared to DNNs. Thus, the novel approach – the DCGMM – and the replay method GMR answer research question **RQ 3**.

**Comparability with Other CL Methods RQ 4**  Research question **RQ 4** relates to the CL performance of GMR and other CL models, especially deep learning models addressing the CF effect. In general, the investigated DCGMM architecture achieves a lower baseline performance in non-CL scenarios than other deep learning models (e.g., CNNs). A competitive value of >99 % accuracy on the MNIST dataset compared to the initial 87 % is a too weak result. Nevertheless, classification accuracies of about 97 % are achieved with a convolutional variant of the same DCGMM architecture.

The investigation of the GMR's CL performance indicates similar results when compared to other replay approaches, such as GR (Shin, Lee, et al. 2017). However, considering the baseline performance, the GMR approach clearly illustrates the functionalities and capabilities required in different CL scenarios. The superiority of a simpler parameterization is another important advantage. The CL results indicate that a simple replay mechanism (GMR) based on DCGMMs leads to an acceptable CL performance. Thus, GMR and respectively DCGMMs serve as a basis for further investigations in many other research areas, even beyond CL.

# 11. Conclusion and Outlook

This chapter summarizes the conclusions and prospects of the present work. The work is considered on the whole for the purpose of outlining the lessons learned. Furthermore, a collection of contributions is presented. Finally, newly emerging issues are addressed, which imply a need for further research.

## 11.1 Summary and Conclusion

The ability to continuously accumulate knowledge has been a decisive factor for the evolution of the human species. The same applies to the development of weak and strong *artificial intelligence*s (AIs) and many application-oriented scenarios. A branch of AI research is concerned with a deep learning approach referred to as Deep Neural Networks (DNNs). Biologically inspired *machine learning* (ML) models have enormous potential and offer many advantages compared to traditional ML models. Nevertheless, DNNs suffer from an effect referred to as *catastrophic forgetting* (CF). The CF effect manifests in an abrupt forgetting of existing knowledge, as soon as new knowledge is added. In order to overcome the problem, the complete training process usually has to be repeated with old and new data samples. A disadvantage of this procedure is that, on the one hand, old training data must be kept available. On the other hand, the energy- and computationally-intensive process has to be entirely repeated.

The CF problem has been investigated for quite some time. As a result of this research, various CF avoidance models and methods have been published. In general, other CF avoidance approaches ensure that the effect is mitigated or that the problem is completely solved. Accordingly, this is experimentally proven in the related work, where each new model achieves even better *continual learning* (CL) performances. As various types of CF avoidance techniques are frequently compared with each other, a superiority of each novel model is demonstrated.

However, the comparison of deep learning models addressing the CF effect is not universally specified. As a first step of this work, a detailed and generally valid evaluation protocol with respect to CL scenarios is introduced. The presented evaluation protocol provides a general scheme for investigating ML models, especially deep learning models, i.e., DNNs. As a special feature of the protocol, it respects different requirements of application oriented real-world scenarios. In order to develop a valid protocol, requirements were derived from an exemplary CL scenario. The result of this first step is an evaluation framework that allows the investigation of ML models in CL scenarios.

The second step of this work describes the investigation of different CF avoidance models in order to quantify the obtained CL performance. The investigation required the code-wise embedding of the models into the evaluation framework. The result of the evaluation indicates that none of the examined ML models can achieve an acceptable CL performance. This conclusion contrasts the findings of other authors. The weak CL performance can be explained by means of the application-oriented requirements that are implemented by the investigation protocol.

Nevertheless, individual CL tasks obtain satisfactory CL performances. The comprehensive investigation mainly implies that CF avoidance models in general can only "control", "mitigate" or "suppress" the CF effect under certain circumstances and under a given evaluation protocol. General assertions, such as "Here, we propose incremental moment matching (IMM) to resolve the catastrophic forgetting problem." (Lee, Kim, et al. 2017), can thus be regarded as inappropriate. A further point of criticism concerns the frequent lack or only partial indication of the investigation/evaluation scheme used in other research. Usually, the code of the CF avoidance model is given, whereby the implementation of the evaluation is omitted. Therefore, many published research results are not clearly reproducible.

The third part of this work focused on the development of a novel deep learning model, which helps to make the forgetting effect more controllable. The novel model is based on Gaussian Mixture Models (GMMs). The novelty of this work's approach is related to the ability to train GMMs by using Stochastic Gradient Descent (SGD), solve the data-driven initialization problem and realize a "deep" arrangement in layers. The novel model is referred to as Deep Convolutional Gaussian Mixture Model (DCGMM). The main difference compared to standard DNNs is that the major part of knowledge is stored in GMM layers and only the assignment of a class label is performed by a conventional linear layer. An advantage is that DCGMMs inherently support many important functionalities, e.g., density estimation or the ability to generate samples.

As the last step of this work, the newly developed deep learning model is transferred into the CL context. For this purpose, the replay approach to avoid CF is combined with DCGMMs, which is referred to as Gaussian Mixture Replay (GMR). GMR implements a pseudo-rehearsal procedure using the functionality of DCGMMs in order to generate samples and realize a joint training. Similar CL capabilities can be achieved when comparing GMR with other replay processes. Thus, GMR offers basic functionalities that can help perform further research on the fulfillment of remaining application-oriented CL requirements.

In sum, the main contributions of this work include the following aspects:

- A detailed description of the investigation protocol for CL scenarios, which includes application-oriented requirement extracted from an exemplary real-world scenario, is provided.

- The results of the investigation of several CF avoidance models indicate that the CF effect can only be considered as "solved" under certain conditions that are challenging to reconcile with application oriented CL requirements.

- A novel deep learning model is introduced referred to as Deep Convolutional Gaussian Mixture Model (DCGMM). DCGMMs are based on GMMs that are arranged in layers and trained by SGD without any data-driven initialization.

- Gaussian Mixture Replay (GMR) is proposed as a CL approach, which is based on DCGMMs. GMR is classified as a (pseudo-rehearsal) replay approach with an acceptable CL performance in application-oriented scenarios.

## 11.2  Prospects

The findings of this work can serve as a basis for future research. Examining the CL investigation protocol offers a first starting point for enhancements. Possible refinements include the possibility of describing and implementing more complex CL tasks. The investigation of further ML models is another open issue. In this context, the assessment of possible application scenarios for different models seems interesting. Another fundamental need for further research concerns the novel deep learning model DCGMM and its application in various scenarios. Furthermore, the transfer of DCGMMs to other contexts such as reinforcement learning or the processing of sequences can constitute the focus of future research.

In the following, some open issues and promising ideas for further research are presented. They are dedicated to the three major findings of this work:

- Investigation protocol and study:

  - Adding further real-world requirements related to a changing data distribution.
  - The addition of other datasets, including (more non-visual) problems with higher dimensionalities.
  - Further development of the evaluation framework towards an independence of programming languages and ML frameworks.
  - Implementation of a requirements catalog for a transparent comparison of different ML models.
  - Integration of an automated runtime and memory analysis.
  - Comparison of other (novel) CF avoidance models based on the uniform investigation protocol.

- Deep Convolutional Gaussian Mixture Models (DCGMMs):

  – Optimization of the DCGMM model with regard to functionalities and implementation, e.g., Mixtures of Factor Analyzer (MFA) layer.

  – Investigation of new mechanisms for outlier detection or regularization of the annealing parameter $\sigma$.

  – Investigating the possibility of targeted/intentional forgetting with regard to deletion or adaptation of specific knowledge.

  – Examination of the robustness known from the field of adversarial attacks.

  – Applying DCGMMs as a basis for other ML methodologies, e.g., reinforcement or transfer learning, and other areas such as unsegmented or connected tasks (like Recurrent Neural Networks (RNNs)).

  – Adapting the functionalities into new areas such as inpainting or noise suppression.

  – Investigating the suitability for other application areas.

- CL methodology (GMR):

  – Customization of the GMR replay procedure in order to meet the requirements of constant re-training complexity.

  – Using density estimation for the development of a novel method without a replay approach.

  – Further investigation of similar CL scenarios, e.g., with changing data distributions.

As a final note, the following thought serves as a reminder of continual learning's importance. Computer science is actually trying to imitate what nature has produced over several millions of years of evolution: The human brain. In just a few decades, machine learning algorithms have become increasingly powerful and sometimes even frightening to many people within our society. After all, the brain's capabilities constitute the "Holy Grail" in the context of continual learning. The ability to continuously accumulate, delete, and selectively correct knowledge can still be considered a challenge in the field of machine learning. Biologically inspired machine learning models attempt to mimic the brain's complex processes, which are often inexplicable, even for neuroscience. Moreover, only highly specialized and customized algorithms can solve abstract problems. But they fail as soon as a small change of a single parameter occurs. Accordingly, the outstanding capabilities of biological (continual) learning have not yet been achieved.

# A.   Listings

## Chapter Contents

# List of Figures

# List of Tables

# List of Algorithms

# List of Equations

# List of Abbreviations

| | |
|---|---|
| **AI** | *artificial intelligence* |
| **ANN** | Artificial Neural Network |
| **ASN** | Autonomous System Number |
| **BMU** | best matching unit |
| **CE** | Cross-Entropy |
| **CF** | *catastrophic forgetting* |
| **CL** | *continual learning* |
| **CNN** | Convolutional Neural Network |
| **CSV** | Comma-separated values |
| **DCGMM** | Deep Convolutional Gaussian Mixture Model |
| **DNN** | Deep Neural Network |
| **DNS** | Domain Name System |
| **D** | Dropout |
| **EM** | Expectation-Maximization |
| **EWC** | Elastic Weight Consolidation |
| **FC** | Fully-Connected |
| **FIM** | Fischer Information Matrix |
| **GAN** | Generative Adverserial Network |
| **GMM** | Gaussian Mixture Model |
| **GMR** | Gaussian Mixture Replay |
| **GR** | Generative Replay |
| **IMM** | Incremental Moment Matching |
| **JSON** | JavaScript Object Notation |
| **LWTA** | Local Winner Takes All |
| **MFA** | Mixtures of Factor Analyzer |
| **MLE** | *maximum-likelihood estimation* |
| **MLP** | Multi Layer Perceptron |
| **ML** | *machine learning* |
| **MSE** | Mean Squared Error |
| $NHWC$ | Number of samples $\times$ Height $\times$ Width $\times$ Channels |
| **NN** | Neural Network |
| **NaN** | Not a Number |
| **PCA** | Principal Component Analysis |
| **PDF** | *probability density function* |
| **QoS** | Quality of Service |
| **RNN** | Recurrent Neural Network |
| **ReLU** | Rectifier Linear Unit |
| **ReST** | Representational State Transfer |
| **SDN** | Software Defined Network |
| **SGD** | Stochastic Gradient Descent |
| **SLT** | Sequential Learning Task |
| **SOM** | Self-Organizing Map |
| **SSH** | Secure Shell |
| **SVM** | Support Vector Machine |
| **TCP** | Transmission Control Protocol |
| **UDP** | User Datagram Protocol |
| **VAE** | Variational Autoencoder |
| **VLAN** | Virtual Local Area Network |
| **sEM** | stochastic Expectation-Maximization (EM) |
| **t-SNE** | t-distributed Stochastic Neighbor Embedding |

# List of References

Acharya, Shailesh, Ashok Kumar Pant, and Prashnna Kumar Gyawali (2016). "Deep learning based large scale handwritten Devanagari character recognition". In: *SKIMA 2015 - 9th International Conference on Software, Knowledge, Information Management and Applications*. DOI: 10.1109/SKIMA.2015.7400041.

Adel, Tameem, Cuong V. Nguyen, Richard E. Turner, Zoubin Ghahramani, and Adrian Weller (2019). "Interpretable Continual Learning". In: URL: https://openreview.net/forum?id=S1g9N2A5FX.

Agarwal, S, M Kodialam, and T Lakshman (2013). "Traffic engineering in software defined networks". In: *IEEE pp* 2211.January.

Alberts, B., D. Bray, J. Lewis, M. Raff, K. Roberts, and J.D. Watson (2002). *Molecular Biology of the Cell*. 4th. Garland.

Aljundi, R., P. Chakravarty, and T. Tuytelaars (2017). "Expert Gate: Lifelong Learning with a Network of Experts". In: *2017 IEEE Conference on Computer Vision and Pattern Recognition (CVPR)*, pp. 7120–7129. DOI: 10.1109/CVPR.2017.753.

Aljundi, Rahaf, Min Lin, Baptiste Goujaud, and Yoshua Bengio (2019). "Gradient based sample selection for online continual learning". In: *Advances in Neural Information Processing Systems* 32.NeurIPS. ISSN: 10495258. arXiv: 1903.08671.

Arbib, Michael A and James J Bonaiuto (2016). *From neuron to cognition via computational neuroscience*. MIT Press.

Arjovsky, Martin, Soumith Chintala, and Léon Bottou (2017). "Wasserstein GAN". In: arXiv: 1701.07875. URL: http://arxiv.org/abs/1701.07875.

Awduche, Daniel O (1999). "MPLS and Traffic Engineering in IP Networks". In: *IEEE Communications Magazine*.

Azzouni, Abdelhadi, Raouf Boutaba, and Guy Pujolle (2017). "NeuRoute: Predictive Dynamic Routing for Software-Defined Networks". In: *13th International Conference on Network and Service Management*.

Baudry, Jean Patrick and Gilles Celeux (2015). "EM for mixtures: Initialization requires special care". In: *Statistics and Computing* 25.4, pp. 713–726. ISSN: 15731375. DOI: 10.1007/s11222-015-9561-x.

Behnke, Sven (2003). *Hierarchical neural networks for image interpretation*. Vol. 2766, pp. 1–220. ISBN: 3540407227.

Benson, Theophilus, Aditya Akella, and David A Maltz (2010). "Network Traffic Characteristics of Data Centers in the Wild". In: *10th ACM SIGCOMM conference on Internet measurement*.

Bernardo, José M. and Adrian F.M. Smith (2008). "Bayesian Theory". In: *Bayesian Theory*, pp. 1–595. DOI: 10.1002/9780470316870.

Bertsekas, Dimitri P. (1996). *Constrained Optimization and Lagrange Multiplier Methods (Optimization and Neural Computation Series)*. 1st ed. Athena Scientific. ISBN: 1886529043.

Bishop, Christopher M. (1995). *Neural Networks for Pattern Recognition*. USA: Oxford University Press, Inc. ISBN: 0198538642.

Boutaba, Raouf, Mohammad A. Salahuddin, Noura Limam, Sara Ayoubi, Nashid Shahriar, Felipe Estrada-Solano, and Oscar M. Caicedo (2018). "A Comprehensive Survey on Machine Learning for Networking: Evolution, Applications and Research Opportunities". In: *Journal of Internet Services and Applications*.

Brim, Scott W. and Brian E. Carpenter (2002). *Middleboxes: Taxonomy and Issues*. RFC 3234.

Brownlee, Jason (2019). *Probability for machine learning: Discover how to harness uncertainty with Python*. Machine Learning Mastery.

Brutzkus, Alon and Amir Globerson (2019). "Why do larger models generalize better? A theoretical perspective via the XOR problem". In: *36th International Conference on Machine Learning, ICML 2019* 2019-June, pp. 1310–1318. arXiv: 1810.03037.

Buda, Mateusz, Atsuto Maki, and Maciej A Mazurowski (2018). "A systematic study of the class imbalance problem in convolutional neural networks". In: *Neural Networks* 106, pp. 249–259.

Bush, Bruce M. (2014). "The Perils of Floating Point". In: *Lahey Computer Systems, Inc.*, p. 8. URL: http://www.lahey.com/float.htm.

Buzzega, Pietro, Matteo Boschini, Angelo Porrello, Davide Abati, and Simone Calderara (2020). "Dark Experience for General Continual Learning: a Strong, Simple Baseline". In: *Advances in Neural Information Processing Systems 33: Annual Conference on Neural Information Processing Systems 2020, NeurIPS 2020, December 6-12, 2020, virtual*. URL: https://proceedings.neurips.cc/paper/2020/hash/b704ea2c39778f07c617f6b7ce480e9e-Abstract.html.

Cappé, Olivier and Eric Moulines (2009). "On-Line Expectation-Maximization Algorithm for Latent Data Models". In: *Journal of the Royal Statistical Society. Series B (Statistical Methodology)* 71.3, pp. 593–613. ISSN: 13697412, 14679868. URL: http://www.jstor.org/stable/40247590.

Cauchy, Augustin et al. (1847). "Méthode générale pour la résolution des systemes d'équations simultanées". In: *Comp. Rend. Sci. Paris* 25.1847, pp. 536–538.

Chaudhry, Arslan, Puneet K Dokania, Thalaiyasingam Ajanthan, and Philip H S Torr (2018). "Riemannian Walk for Incremental Learning: Understanding Forgetting and Intransigence Arslan". In: 1, pp. 1–22. arXiv: arXiv:1801.10112v3.

Chaudhry, Arslan, Albert Gordo, Puneet K Dokania, Philip Torr, and David Lopez-paz (2021). "Using Hindsight to Anchor Past Knowledge in Continual Learning". In: arXiv: arXiv:2002.08165v2.

Chaudhry, Arslan, Ranzato Marc'Aurelio, Marcus Rohrbach, and Mohamed Elhoseiny (2019). "Efficient lifelong learning with A-GEM". In: *7th International Conference on Learning Representations, ICLR 2019*, pp. 1–20. arXiv: 1812.00420.

Chen, Jianfei, Jun Zhu, Yee Whye Teh, and Tong Zhang (2018). "Stochastic expectation maximization with variance reduction". In: *Advances in Neural Information Processing Systems* 2018-Decem.NeurIPS, pp. 7967–7977. ISSN: 10495258.

Chen, Zhiyuan and Bing Liu (2016). "Lifelong Machine Learning". In: *Synthesis Lectures on Artificial Intelligence and Machine Learning* 10.3, pp. 1–27. ISSN: 19394616. DOI: 10.2200/S00737ED1V01Y201610AIM033.

Chollet, François (2017). *François Chollet Twitter*. URL: https://twitter.com/fchollet/status/852594987527045120 (visited on 03/25/2021).

Cireşan, Dan C., Ueli Meier, Jonathan Masci, Luca M. Gambardella, and Jürgen Schmidhuber (2011). "Flexible, high performance convolutional neural networks for image classification". In: *IJCAI International Joint Conference on Artificial Intelligence*, pp. 1237–1242. ISSN: 10450823. DOI: 10.5591/978-1-57735-516-8/IJCAI11-210. eprint: arXiv:1011.1669v3.

Cisco (2008). *Flexible NetFlow Technology White Paper*.

Claise, B. (2004). *Cisco Systems NetFlow Serv. Export V9*. RFC 3954.

Cohen, Gregory, Saeed Afshar, Jonathan Tapson, and Andre Van Schaik (2017). "EMNIST: Extending MNIST to handwritten letters". In: *Proceedings of the International Joint Conference on Neural Networks* 2017-May, pp. 2921–2926. DOI: 10.1109/IJCNN.2017.7966217. eprint: 1702.05373.

Cole, Ronald and Mark Fanty (1990). "Spoken letter recognition". In: pp. 385–390. DOI: 10.3115/116580.116725.

Cvetkovski, Zdravko (2012). "Convexity, Jensen's Inequality". In: *Inequalities: Theorems, Techniques and Selected Problems*. Berlin, Heidelberg: Springer Berlin Heidelberg, pp. 69–77. ISBN: 978-3-642-23792-8. DOI: 10.1007/978-3-642-23792-8_7. URL: https://doi.org/10.1007/978-3-642-23792-8_7.

Cybenko, G. (1989). "Approximation by superpositions of a sigmoidal function". In: *Mathematics of Control, Signals, and Systems* 2.4, pp. 303–314. ISSN: 0932-4194. DOI: 10.1007/BF02551274. URL: http://link.springer.com/10.1007/BF02551274.

David, Omid E. and Nathan S. Netanyahu (2016). "DeepPainter: Painter Classification Using Deep Convolutional Autoencoders". In: pp. 20–28.

Davies, David L. and Donald W. Bouldin (1979). "A Cluster Separation Measure". In: *IEEE Transactions on Pattern Analysis and Machine Intelligence* PAMI-1.2, pp. 224–227. ISSN: 01628828. DOI: 10.1109/TPAMI.1979.4766909.

Dayan, Peter, Laurence F Abbott, et al. (2003). "Theoretical neuroscience: computational and mathematical modeling of neural systems". In: *Journal of Cognitive Neuroscience* 15.1, pp. 154–155.

Delange, M., R. Aljundi, M. Masana, S. Parisot, X. Jia, A. Leonardis, G. Slabaugh, and T. Tuytelaars (2021). "A continual learning survey: Defying forgetting in classification tasks". In: *IEEE Transactions on Pattern Analysis and Machine Intelligence*, pp. 1–1. ISSN: 1939-3539. DOI: 10.1109/TPAMI.2021.3057446.

Dempster, A. P., N. M. Laird, and D. B. Rubin (1977). " Maximum Likelihood from Incomplete Data Via the EM Algorithm ". In: *Journal of the Royal Statistical Society: Series B (Methodological)* 39.1, pp. 1–22. DOI: 10.1111/j.2517-6161.1977.tb01600.x.

Dognin, Pierre L., Vaibhava Goel, John R. Hershey, and Peder A. Olsen (2009). "A fast, accurate approximation to log likelihood of Gaussian mixture models". In: *ICASSP, IEEE International Conference on Acoustics, Speech and Signal Processing - Proceedings* 3, pp. 3817–3820. ISSN: 15206149. DOI: 10.1109/ICASSP.2009.4960459.

Duch, Włodzisław and Norbert Jankowski (1999). "Survey of neural transfer functions". In: *Neural Computing Surveys* 2, pp. 163–212.

Duchi, John, Elad Hazan, and Yoram Singer (2010). "Adaptive subgradient methods for online learning and stochastic optimization". In: *COLT 2010 - The 23rd Conference on Learning Theory* 12, pp. 257–269.

Dunn, J. C. (1973). "A fuzzy relative of the ISODATA process and its use in detecting compact well-separated clusters". In: *Journal of Cybernetics* 3.3, pp. 32–57. ISSN: 00220280. DOI: 10.1080/01969727308546046.

Engel, Paulo Martins and Milton Roberto Heinen (2010). "Incremental Learning of Multivariate Gaussian Mixture Models". In: *Advances in Artificial Intelligence – SBIA 2010*. Berlin, Heidelberg: Springer Berlin Heidelberg, pp. 82–91. ISBN: 978-3-642-16138-4.

Eykholt, Kevin, Ivan Evtimov, Earlence Fernandes, Bo Li, Amir Rahmati, Chaowei Xiao, Atul Prakash, Tadayoshi Kohno, and Dawn Song (2017). "Robust Physical-World Attacks on Deep Learning Models". In: arXiv: 1707.08945. URL: http://arxiv.org/abs/1707.08945.

Fadlullah, Zubair Md, Fengxiao Tang, Bomin Mao, Nei Kato, Osamu Akashi, Takeru Inoue, and Kimihiro Mizutani (2017a). "State-of-the-Art Deep Learning: Evolving Machine Intelligence Toward Tomorrow's Intelligent Network Traffic Control Systems". In: *IEEE Communications Surveys and Tutorials* 19.4, pp. 2432–2455. ISSN: 1553877X. DOI: 10.1109/COMST.2017.2707140.

Fadlullah, Zubair Md, Fengxiao Tang, Bomin Mao, Nei Kato, Osamu Akashi, Takeru Inoue, and Kimihiro Mizutani (2017b). "State-of-the-Art Deep Learning: Evolving Machine Intelligence Toward Tomorrow's Intelligent Network Traffic Control Systems". In: *IEEE Communications Surveys & Tutorials* 19.4, pp. 2432–2455.

Farquhar, Sebastian and Yarin Gal (2018). "Towards robust evaluations of continual learning". In: *arXiv*. ISSN: 23318422. arXiv: 1805.09733.

Fausett, Laurene (1994). *Fundamentals of Neural Networks: Architectures, Algorithms, and Applications*. USA: Prentice-Hall, Inc. ISBN: 0133341860.

Feldman, Dan, Matthew Faulkner, and Andreas Krause (2011). "Scalable Training of Mixture Models via Coresets". In: *Advances in Neural Information Processing Systems*. Ed. by J. Shawe-Taylor, R. Zemel, P. Bartlett, F. Pereira, and K. Q. Weinberger. Vol. 24. Curran Associates, Inc., p. 9. URL: https://proceedings.neurips.cc/paper/2011/file/2b6d65b9a9445c4271ab9076ead5605a-Paper.pdf.

Fernando, Chrisantha, Dylan Banarse, Charles Blundell, Yori Zwols, David Ha, Andrei A. Rusu, Alexander Pritzel, and Daan Wierstra (2017). "PathNet: Evolution Channels Gradient Descent in Super Neural Networks". In: ISSN: 1701.08734. arXiv: 1701.08734. URL: http://arxiv.org/abs/1701.08734.

Flach, Peter (2012). *Machine learning: the art and science of algorithms that make sense of data*. Cambridge university press.

French, Robert M. (1997). "Pseudo-recurrent Connectionist Networks: An Approach to the 'Sensitivity-Stability' Dilemma". In: *Connection Science* 9.4, pp. 353–380. ISSN: 09540091. DOI: 10.1080/095400997116595.

Ge, Rong, Qingqing Huang, and Sham M. Kakade (2015). "Learning mixtures of gaussians in high dimensions". In: *Proceedings of the Annual ACM Symposium on Theory of Computing* 14-17-June-2015, pp. 761–770. ISSN: 07378017. DOI: 10.1145/2746539.2746616. arXiv: 1503.00424.

Gepperth, Alexander and Benedikt Pfülb (2020). "A Rigorous Link Between Self-Organizing Maps and Gaussian Mixture Models". In: *Artificial Neural Networks and Machine Learning - ICANN 2020 - 29th International Conference on Artificial Neural Networks, Bratislava, Slovakia, September 15-18, 2020, Proceedings, Part II*. Vol. 12397. Lecture Notes in Computer Science. Springer, pp. 863–872. DOI: 10.1007/978-3-030-61616-8\_69. URL: https://doi.org/10.1007/978-3-030-61616-8\_69.

Gepperth, Alexander and Benedikt Pfülb (2021). "Gradient-based training of Gaussian Mixture Models in High-Dimensional Spaces". In: *Neural Processing Letters*. ISSN: 1573-773X. DOI: 10.1007/s11063-021-10599-3. eprint: 1912.09379.

Gepperth, Alexander and Florian Wiech (2019). "Simplified Computation and Interpretation of Fisher Matrices in Incremental Learning with Deep Neural Networks". In: *Lecture Notes in*

*Computer Science (including subseries Lecture Notes in Artificial Intelligence and Lecture Notes in Bioinformatics)* 11728 LNCS, pp. 481–494. ISSN: 16113349. DOI: 10.1007/978-3-030-30484-3_39.

Ghahramani, Zoubin and Geoffrey E. Hinton (1997). "The EM Algorithm for Mixtures of Factor Analyzers". In: *Compute*, pp. 1–8. ISSN: 0003-0023. DOI: 173068. arXiv: 173068.

Glorot, Xavier and Yoshua Bengio (2010). "Understanding the difficulty of training deep feedforward neural networks". In: *Proceedings of the Thirteenth International Conference on Artificial Intelligence and Statistics*. Ed. by Yee Whye Teh and Mike Titterington. Vol. 9. Proceedings of Machine Learning Research. Chia Laguna Resort, Sardinia, Italy: PMLR, pp. 249–256. URL: http://proceedings.mlr.press/v9/glorot10a.html.

Glorot, Xavier, Antoine Bordes, and Yoshua Bengio (2011). "Deep Sparse Rectifier Neural Networks". In: *Proceedings of the Fourteenth International Conference on Artificial Intelligence and Statistics*. Ed. by Geoffrey Gordon, David Dunson, and Miroslav Dudík. Vol. 15. Proceedings of Machine Learning Research. Fort Lauderdale, FL, USA: PMLR, pp. 315–323. URL: http://proceedings.mlr.press/v15/glorot11a.html.

Goodfellow, Ian, Yoshua Bengio, and Aaron Courville (2016). *Deep Learning.* http://www.deeplearningbook.org. MIT Press.

Goodfellow, Ian, Jean Pouget-Abadie, Mehdi Mirza, Bing Xu, David Warde-Farley, Sherjil Ozair, Aaron Courville, and Yoshua Bengio (2014). "Generative Adversarial Nets". In: *Advances in Neural Information Processing Systems 27*.

Goodfellow, Ian J., Mehdi Mirza, Da Xiao, Aaron Courville, and Yoshua Bengio (2013). "An Empirical Investigation of Catastrophic Forgetting in Gradient-Based Neural Networks". In: ISSN: 1751-8113. DOI: 10.1088/1751-8113/44/8/085201. eprint: 1312.6211. URL: http://arxiv.org/abs/1312.6211.

Goodfellow, Ian J., Jonathon Shlens, and Christian Szegedy (2015). "Explaining and harnessing adversarial examples". In: *3rd International Conference on Learning Representations, ICLR 2015 - Conference Track Proceedings*, pp. 1–11. arXiv: 1412.6572.

Graf, Hans Peter, Eric Cosatto, Leon Bottou, Igor Durdanovic, and Vladimir Vapnik (2005). "Parallel support vector machines: The cascade SVM". In: *Advances in Neural Information Processing Systems.* ISSN: 10495258.

Graupe, Daniel (2013). *Principles of artificial neural networks.* Vol. 7. World Scientific.

Grother, P. (1995). "NIST Special Database 19 Handprinted Forms and Characters Database". In.

Gulrajani, Ishaan, Faruk Ahmed, Martin Arjovsky, Vincent Dumoulin, and Aaron Courville (2017). "Improved Training of Wasserstein GANs". In: arXiv: 1704.00028. URL: http://arxiv.org/abs/1704.00028.

Hagan, Martin T, Howard B Demuth, and Mark Beale (1997). *Neural network design.* PWS Publishing Co.

Han, Song, Huizi Mao, Enhao Gong, Shijian Tang, William J. Dally, Jeff Pool, John Tran, Bryan Catanzaro, Sharan Narang, Erich Elsen, Peter Vajda, and Manohar Paluri (2017). "DSD: Dense-sparse-dense training for deep neural networks". In: *5th International Conference on Learning Representations, ICLR 2017 - Conference Track Proceedings.* arXiv: 1607.04381.

Handelman, Guy S., Hong Kuan Kok, Ronil V. Chandra, Amir H. Razavi, Shiwei Huang, Mark Brooks, Michael J. Lee, and Hamed Asadi (2019). "Peering Into the Black Box of Artificial Intelligence: Evaluation Metrics of Machine Learning Methods". In: *American Journal of Roentgenology* 212.1. PMID: 30332290, pp. 38–43. DOI: 10.2214/AJR.18.20224. eprint: https://doi.org/10.2214/AJR.18.20224. URL: https://doi.org/10.2214/AJR.18.20224.

Hardegen, Christoph, Benedikt Pfülb, Sebastian Rieger, and Alexander Gepperth (Dec. 2020). "Predicting Network Flow Characteristics Using Deep Learning and Real-World Network Traffic". In: *IEEE Trans. on Netw. and Serv. Manag.* 17.4, 2662–2676. ISSN: 1932-4537. DOI: 10.1109/TNSM.2020.3025131. URL: https://doi.org/10.1109/TNSM.2020.3025131.

Hardegen, Christoph, Benedikt Pfülb, Sebastian Rieger, Alexander Gepperth, and Sven Reißmann (2019). "Flow-based Throughput Prediction using Deep Learning and Real-World Network Traffic". In: *Proceedings of the 15th International Conference on Network and Service Management.*

Hardegen, Christoph and Sebastian Rieger (2020). "Prediction-based Flow Routing in Programmable Networks with P4". In: *16th International Conference on Network and Service Management, CNSM 2020, 2nd International Workshop on Analytics for Service and Application Management, AnServApp 2020 and 1st International Workshop on the Future Evolution of Internet Protocols, IPFuture 2020*, pp. 3–7. DOI: 10.23919/CNSM50824.2020.9269072.

Hayase, Tomohiro, Suguru Yasutomi, and Takashi Katoh (2020). "Selective Forgetting of Deep Networks at a Finer Level than Samples". In: arXiv: 2012.11849. URL: http://arxiv.org/abs/2012.11849.

Hayes, Tyler L., Ronald Kemker, Nathan D. Cahill, and Christopher Kanan (2018). "New Metrics and Experimental Paradigms for Continual Learning". In: *2018 IEEE/CVF Conference on Computer Vision and Pattern Recognition Workshops (CVPRW)*, pp. 2112–2123. DOI: 10.1109/cvprw.2018.00273.

Haykin, Simon (2009). *Neural networks and learning machines, 3/E*. Pearson Education India.

Heaton, Jeff (2015). "Artificial intelligence for humans, volume 3: Deep learning and neural networks; heaton research". In: *Inc.: St. Louis, MO, USA*.

Hinton, Geoffrey, Nitish Srivastava, and Kevin Swersky (2012a). *Lecture 6.5 - RmsProp: Divide the gradient by a running average of its recent magnitude*.

Hinton, Geoffrey E., Nitish Srivastava, Alex Krizhevsky, Ilya Sutskever, and Ruslan R. Salakhutdinov (2012b). "Improving neural networks by preventing co-adaptation of feature detectors". In: pp. 1–18. ISSN: 9781467394673. DOI: arXiv:1207.0580. arXiv: 1207.0580. URL: http://arxiv.org/abs/1207.0580.

Hornik, Kurt, Maxwell Stinchcombe, and Halbert White (1989). "Multilayer feedforward networks are universal approximators". In: *Neural Networks* 2.5, pp. 359–366. ISSN: 08936080. DOI: 10.1016/0893-6080(89)90020-8.

Hosseini, Reshad and Suvrit Sra (2015). "Matrix Manifold Optimization for Gaussian Mixtures". In: *Advances in Neural Information Processing Systems*. Ed. by C. Cortes, N. Lawrence, D. Lee, M. Sugiyama, and R. Garnett. Vol. 28. Curran Associates, Inc. URL: https://proceedings.neurips.cc/paper/2015/file/dbe272bab69f8e13f14b405e038deb64-Paper.pdf.

Hosseini, Reshad and Suvrit Sra (2020). "An alternative to EM for Gaussian mixture models: batch and stochastic Riemannian optimization". In: *Mathematical Programming* 181.1, pp. 187–223. ISSN: 14364646. DOI: 10.1007/s10107-019-01381-4. eprint: 1706.03267. URL: https://doi.org/10.1007/s10107-019-01381-4.

Hsu, Yen Chang, Yen Cheng Liu, Anita Ramasamy, and Zsolt Kira (2018). "Re-evaluating continual learning scenarios: A categorization and case for strong baselines". In: *arXiv* Nips. ISSN: 23318422. arXiv: 1810.12488.

Huszár, Ferenc (2018). "Note on the quadratic penalties in elastic weight consolidation". In: *Proceedings of the National Academy of Sciences of the United States of America* 115.11, E2496–E2497. ISSN: 10916490. DOI: 10.1073/pnas.1717042115. arXiv: arXiv:1712.03847v1.

Ioffe, Sergey and Christian Szegedy (2015). "Batch Normalization: Accelerating Deep Network Training by Reducing Internal Covariate Shift". In: *Proceedings of the 32nd International Conference on Machine Learning*. Ed. by Francis Bach and David Blei. Vol. 37. Proceedings of Machine Learning Research. Lille, France: PMLR, pp. 448–456. URL: https://proceedings.mlr.press/v37/ioffe15.html.

Iyengar, Jana, Costin Raiciu, Sebastien Barre, Mark J. Handley, and Alan Ford (2011). *Architectural Guidelines for Multipath TCP Development*. RFC 6182.

Japkowicz, Nathalie and Shaju Stephen (2002). "The class imbalance problem: A systematic study". In: *Intelligent data analysis* 6.5, pp. 429–449.

Joseph, K. J. and Vineeth N. Balasubramanian (2020). "Meta-Consolidation for Continual Learning". In: *Advances in Neural Information Processing Systems*. Ed. by H. Larochelle, M. Ranzato, R. Hadsell, M. F. Balcan, and H. Lin. Vol. 33. Curran Associates, Inc., pp. 14374–14386. URL: https://proceedings.neurips.cc/paper/2020/file/a5585a4d4b12277fee5cad0880611bc6-Paper.pdf.

Jurkiewicz, Piotr, Grzegorz Rzym, and Piotr Borylo (2018). "Flow Length and Size Distributions in Campus Internet Traffic". In: *Computing Research Repository*.

Kamra, Nitin, Umang Gupta, and Yan Liu (2017). "Deep Generative Dual Memory Network for Continual Learning". In: *CoRR* abs/1710.10368. arXiv: 1710.10368. URL: http://arxiv.org/abs/1710.10368.

Karahoca, Adem (2012). *Advances in data mining knowledge discovery and applications*. BoD–Books on Demand.

Karras, Tero, Timo Aila, Samuli Laine, and Jaakko Lehtinen (2018). "Progressive growing of GANs for improved quality, stability, and variation". In: *6th International Conference on Learning Representations, ICLR 2018 - Conference Track Proceedings*, pp. 1–26.

Kelleher, John D (2019). *Deep learning*. MIT press.

Keller, James M, Derong Liu, and David B Fogel (2016). *Fundamentals of computational intelligence: neural networks, fuzzy systems, and evolutionary computation*. John Wiley & Sons.

Kemker, Ronald, Marc McClure, Angelina Abitino, Tyler Hayes, and Christopher Kanan (2017). "Measuring Catastrophic Forgetting in Neural Networks". In: ISSN: 1091-6490. DOI: 10.1073/pnas.1611835114. eprint: 1708.02072. URL: http://arxiv.org/abs/1708.02072.

Kim, Dahun, Sanghyun Woo, Joon Young Lee, and In So Kweon (2019). "Deep video inpainting". In: *Proceedings of the IEEE Computer Society Conference on Computer Vision and Pattern Recognition* 2019-June, pp. 5785–5794. ISSN: 10636919. DOI: 10.1109/CVPR.2019.00594. eprint: 1905.01639.

Kim, Hyo-Eun, Seungwook Kim, and Jaehwan Lee (2018). "Keep and Learn: Continual Learning by Constraining the Latent Space for Knowledge Preservation in Neural Networks". In: arXiv: 1805.10784. URL: http://arxiv.org/abs/1805.10784.

Kingma, Diederik P. and Jimmy Lei Ba (2015). "Adam: A method for stochastic optimization". In: *3rd International Conference on Learning Representations, ICLR 2015 - Conference Track Proceedings*, pp. 1–15. eprint: 1412.6980.

Kingma, Diederik P and Max Welling (2013). "Auto-Encoding Variational Bayes". In: Ml, pp. 1–14. arXiv: 1312.6114. URL: http://arxiv.org/abs/1312.6114.

Kirkpatrick, James, Razvan Pascanu, Neil Rabinowitz, Joel Veness, Guillaume Desjardins, Andrei A. Rusu, Kieran Milan, John Quan, Tiago Ramalho, Agnieszka Grabska-Barwinska, Demis Hassabis, Claudia Clopath, Dharshan Kumaran, and Raia Hadsell (2017). "Overcoming catastrophic forgetting in neural networks". In: *Proceedings of the National Academy of Sciences* 114.13, pp. 3521–3526. ISSN: 0027-8424. DOI: 10.1073/pnas.1611835114. eprint: https://www.pnas.org/content/114/13/3521.full.pdf. URL: https://www.pnas.org/content/114/13/3521.

Kirkpatrick, James, Razvan Pascanu, Neil C. Rabinowitz, Joel Veness, Guillaume Desjardins, Andrei A. Rusu, Kieran Milan, John Quan, Tiago Ramalho, Agnieszka Grabska-Barwinska, Demis Hassabis, Claudia Clopath, Dharshan Kumaran, and Raia Hadsell (2016). "Overcoming catastrophic forgetting in neural networks". In: *CoRR* abs/1612.00796. arXiv: 1612.00796. URL: http://arxiv.org/abs/1612.00796.

Kohonen, Teuvo (1982). "Self-organized formation of topologically correct feature maps". In: *Biological Cybernetics* 43.1, pp. 59–69. ISSN: 03401200. DOI: 10.1007/BF00337288.

Kreutz, Diego, Fernando MV Ramos, Paulo Verissimo, Christian Esteve Rothenberg, Siamak Azodolmolky, and Steve Uhlig (2015). "Software-Defined Networking: A Comprehensive Survey". In: *Proceedings of the IEEE*.

Kristan, Matej, Danijel Skočaj, and Ales Leonardis (2008). "Incremental learning with Gaussian mixture models". In: 1, pp. 25–32.

Krizhevsky, Alex, Ilya Sutskever, and Geoffrey E. Hinton (2017). "ImageNet Classification with Deep Convolutional Neural Networks". In: *Communications of the ACM* 60.6, pp. 84–90. ISSN: 0001-0782. DOI: 10.1145/3065386. URL: https://dl.acm.org/doi/10.1145/3065386.

Kubat, Miroslav, Stan Matwin, et al. (1997). "Addressing the curse of imbalanced training sets: one-sided selection". In: *Icml*. Vol. 97. 1. Citeseer, p. 179.

LeCun, Yann, Léon Bottou, Yoshua Bengio, and Patrick Haffner (1998). "Gradient-based learning applied to document recognition". In: *Proceedings of the IEEE* 86.11, pp. 2278–2323. ISSN: 00189219. DOI: 10.1109/5.726791. eprint: 1102.0183.

LeCun, Yann, Patrick Haffner, Léon Bottou, and Yoshua Bengio (1999). "Object Recognition with Gradient-Based Learning". In: *Shape, Contour and Grouping in Computer Vision*. Berlin, Heidelberg: Springer Berlin Heidelberg, pp. 319–345. ISBN: 978-3-540-46805-9. DOI: 10.1007/3-540-46805-6_19. URL: https://doi.org/10.1007/3-540-46805-6_19.

LeCun, Yann A., Léon Bottou, Genevieve B. Orr, and Klaus-Robert Müller (2012). "Efficient BackProp". In: *Neural Networks: Tricks of the Trade: Second Edition*. Berlin, Heidelberg: Springer Berlin Heidelberg, pp. 9–48. ISBN: 978-3-642-35289-8. DOI: 10.1007/978-3-642-35289-8_3. URL: https://doi.org/10.1007/978-3-642-35289-8_3.

Lee, Sang-Woo, Jin-Hwa Kim, Jaehyun Jun, Jung-Woo Ha, and Byoung-Tak Zhang (2017). "Overcoming Catastrophic Forgetting by Incremental Moment Matching". In: Nips, pp. 1–16. eprint: 1703.08475. URL: http://arxiv.org/abs/1703.08475.

Lesort, Timothée, Massimo Caccia, and Irina Rish (2021). "Understanding Continual Learning Settings with Data Distribution Drift Analysis". In: pp. 1–9. arXiv: 2104.01678. URL: http://arxiv.org/abs/2104.01678.

Lesort, Timothée, Hugo Caselles-Dupré, Michael Garcia-Ortiz, Andrei Stoian, and David Filliat (2019). "Generative Models from the perspective of Continual Learning". In: *2019 International Joint Conference on Neural Networks (IJCNN)*, pp. 1–8.

Lesort, Timothée, Andrei Stoian, Jean François Goudou, and David Filliat (2019). "Training discriminative models to evaluate generative ones". In: *Lecture Notes in Computer Science (including subseries Lecture Notes in Artificial Intelligence and Lecture Notes in Bioinformatics)* 11729 LNCS, pp. 604–619. ISSN: 16113349. DOI: 10.1007/978-3-030-30508-6_48. arXiv: 1806.10840.

Li, Der-Chiang, Susan C Hu, Liang-Sian Lin, and Chun-Wu Yeh (2017). "Detecting representative data and generating synthetic samples to improve learning accuracy with imbalanced data sets". In: *PloS one* 12.8, e0181853.

Li, Jonathan Q. and Andrew R. Barron (2000). "Mixture density estimation". In: *Advances in Neural Information Processing Systems* x, pp. 279–285. ISSN: 10495258.

Li, Yanchun, Nanfeng Xiao, and Wanli Ouyang (2018). "Improved boundary equilibrium generative adversarial networks". In: *IEEE Access* 6.c, pp. 11342–11348. ISSN: 21693536. DOI: 10.1109/ACCESS.2018.2804278.

Li, Zhizhong and Derek Hoiem (2018). "Learning without Forgetting". In: *IEEE Transactions on Pattern Analysis and Machine Intelligence* 40.12, pp. 2935–2947. ISSN: 19393539. DOI: 10.1109/TPAMI.2017.2773081. arXiv: arXiv:1606.09282v3.

Lopez-Paz, David and Marc'Aurelio Ranzato (2017). "Gradient Episodic Memory for Continual Learning". In: *Proceedings of the 31st International Conference on Neural Information Processing Systems*. NIPS'17. Long Beach, California, USA: Curran Associates Inc., 6470–6479. ISBN: 9781510860964.

Maaten, Laurens van der (2014). "Accelerating t-SNE using Tree-Based Algorithms". In: *Journal of Machine Learning Research*.

Maaten, Laurens van der and Geoffrey Hinton (2008). "Visualizing Data using t-SNE". In: *Journal of Machine Learning Research*.

Mallya, Arun and Svetlana Lazebnik (2018). "PackNet: Adding Multiple Tasks to a Single Network by Iterative Pruning". In: *Proceedings of the IEEE Computer Society Conference on Computer Vision and Pattern Recognition*, pp. 7765–7773. ISSN: 10636919. DOI: 10.1109/CVPR.2018.00810. arXiv: 1711.05769.

MaxMind (2019). "MaxMind: IP Geolocation and Online Fraud Prevention". In: URL: https://dev.maxmind.com/geoip/geoip2/geolite2/.

Mayers, David F., Gene H. Golub, and Charles F. van Loan (1986). *Matrix Computations*. Vol. 47. 175, p. 376. ISBN: 9781421407944. DOI: 10.2307/2008107.

McClelland, James L, Bruce L McNaughton, and Randall C O'Reilly (1995). "Why there are complementary learning systems in the hippocampus and neocortex: insights from the successes and failures of connectionist models of learning and memory." In: *Psychological review* 102.3, pp. 419–457. ISSN: 0033-295X. DOI: 10.1037/0033-295X.102.3.419. URL: http://eutils.ncbi.nlm.nih.gov/entrez/eutils/elink.fcgi?dbfrom=pubmed&id=7624455&retmode=ref&cmd=prlinkshttp://www.ncbi.nlm.nih.gov/pubmed/7624455.

McCloskey, Michael and Neal J. Cohen (1989). "Catastrophic Interference in Connectionist Networks: The Sequential Learning Problem". In: *Psychology of Learning and Motivation - Advances in Research and Theory* 24.C, pp. 109–165. ISSN: 00797421. DOI: 10.1016/S0079-7421(08)60536-8.

McCulloch, Warren S. and Walter Pitts (1943). "A logical calculus of the ideas immanent in nervous activity". In: *The Bulletin of Mathematical Biophysics* 5.4, pp. 115–133. ISSN: 0007-4985. DOI: 10.1007/BF02478259. URL: http://link.springer.com/10.1007/BF02478259.

McElreath, Richard (2018). "Statistical rethinking: A bayesian course with examples in R and stan". In: *Statistical Rethinking: A Bayesian Course with Examples in R and Stan*, pp. 1–469. DOI: 10.1201/9781315372495.

McKeown, Nick, Tom Anderson, Hari Balakrishnan, Guru Parulkar, Larry Peterson, Jennifer Rexford, Scott Shenker, and Jonathan Turner (2008). "OpenFlow: Enabling Innovation in Campus Networks". In: *ACM SIGCOMM Computer Communication Review*.

Mermillod, Martial, Aurélia Bugaiska, and Patrick BONIN (2013). "The stability-plasticity dilemma: investigating the continuum from catastrophic forgetting to age-limited learning effects". In: *Frontiers in Psychology* 4, p. 504. ISSN: 1664-1078. URL: https://www.frontiersin.org/article/10.3389/fpsyg.2013.00504.

Mitchell, Tom M. (1997). *Machine Learning*. New York: McGraw-Hill. ISBN: 978-0-07-042807-2.

Mohammed, A. R., S. A. Mohammed, and S. Shirmohammadi (2019). "Machine Learning and Deep Learning Based Traffic Classification and Prediction in Software Defined Networking". In: *2019 IEEE International Symposium on Measurements Networking*.

Mori, Tatsuya, Masato Uchida, Ryoichi Kawahara, Jianping Pan, and Shigeki Goto (2004). "Identifying Elephant Flows Through Periodically Sampled Packets". In: *4th ACM SIGCOMM Conference on Internet Measurement*.

Mureşan, Horea and Mihai Oltean (2018). "Fruit recognition from images using deep learning". In: *Acta Universitatis Sapientiae, Informatica* 10.1, pp. 26–42. ISSN: 2066-7760. DOI: 10.2478/ausi-2018-0002. eprint: 1712.00580. URL: http://arxiv.org/abs/1712.00580.

Nagy, David G and Gergo Orban (2016). "Episodic memory for continual model learning". In: Nips, pp. 1–5. arXiv: 1712.01169v1.

Najafabadi, Maryam M, Flavio Villanustre, Taghi M Khoshgoftaar, Naeem Seliya, Randall Wald, and Edin Muharemagic (2015). "Deep learning applications and challenges in big data analytics". In: *Journal of big data* 2.1, pp. 1–21.

Nesterov, Yu (2013). "Gradient methods for minimizing composite functions". In: *Mathematical Programming* 140.1, pp. 125–161. ISSN: 00255610. DOI: 10.1007/s10107-012-0629-5.

Netzer, Yuval and Tao Wang (2011). "Reading digits in natural images with unsupervised feature learning". In: *Neural Information Processing Systems (NIPS)*, pp. 1–9.

Nguyen, Cuong V., Yingzhen Li, Thang D. Bui, and Richard E. Turner (2018). "Variational continual learning". In: *6th International Conference on Learning Representations, ICLR 2018 - Conference Track Proceedings* Vi, pp. 1–18. arXiv: 1710.10628.

Nguyen, Giang, Stefan Dlugolinsky, Martin Bobák, Viet Tran, Alvaro Lopez Garcia, Ignacio Heredia, Peter Malík, and Ladislav Hluchý (2019). "Machine learning and deep learning frameworks and libraries for large-scale data mining: a survey". In: *Artificial Intelligence Review* 52.1, pp. 77–124.

Nguyen, Thuy TT and Grenville J Armitage (2008). "A Survey of Techniques for Internet Traffic Classification using Machine Learning". In: *IEEE Communications Surveys*.

Nielsen, Frank and Ke Sun (2016). "Guaranteed bounds on information-theoretic measures of univariate mixtures using piecewise log-sum-exp inequalities". In: *Entropy* 18.12. ISSN: 10994300. DOI: 10.3390/e18120442.

Ormoneit, Dirk and Volker Tresp (1998). "Averaging, maximum penalized likelihood and Bayesian estimation for improving Gaussian mixture probability density estimates". In: *IEEE Transactions on Neural Networks* 9.4, pp. 639–650. ISSN: 10459227. DOI: 10.1109/72.701177.

Parisi, German I., Ronald Kemker, Jose L. Part, Christopher Kanan, and Stefan Wermter (2019). "Continual lifelong learning with neural networks: A review". In: *Neural Networks* 113, pp. 54–71. ISSN: 18792782. DOI: 10.1016/j.neunet.2019.01.012. arXiv: 1802.07569.

Pfülb, B., A. Gepperth, S. Abdullah, and A. Kilian (2018). "Catastrophic Forgetting: Still a Problem for DNNs". In: vol. 11139. Lecture Notes in Computer Science. Cham: Springer International Publishing, pp. 487–497. ISBN: 978-3-030-01417-9. DOI: 10.1007/978-3-030-01418-6_48. URL: http://link.springer.com/10.1007/978-3-030-01418-6.

Pfülb, Benedikt and Alexander Gepperth (2019). "A Comprehensive, Application-Oriented Study of Catastrophic Forgetting in DNNs". In: *7th International Conference on Learning Representations, ICLR 2019* 1.c, pp. 1–14. arXiv: 1905.08101.

Pfülb, Benedikt and Alexander Gepperth (2021). "Overcoming Catastrophic Forgetting with Gaussian Mixture Replay". In: *International Joint Conference on Neural Networks (IJCNN)*, p. 9. eprint: 2104.09220.

Pfülb, Benedikt, Alexander Gepperth, and Benedikt Bagus (2021). "Continual Learning with Fully Probabilistic Models". In: *2nd CLVISION CVPR Workshop (Accepted as Findings)*, p. 10. eprint: 2104.09240. URL: http://arxiv.org/abs/2104.09240.

Pfülb, Benedikt, Christoph Hardegen, Alexander Gepperth, and Sebastian Rieger (2019). "A Study of Deep Learning for Network Traffic Data Forecasting". In: *Artificial Neural Networks and Machine Learning – ICANN 2019: Text and Time Series*. Cham: Springer International Publishing, pp. 497–512. ISBN: 978-3-030-30490-4.

Pinheiro, Jose C. and Douglas M. Bates (1995). "Approximations to the log-likelihood function in the nonlinear mixed-effects model". In: *Journal of Computational and Graphical Statistics* 4.1, pp. 12–35. ISSN: 15372715. DOI: 10.1080/10618600.1995.10474663.

Pinkus, Allan (1999). "Approximation theory of the MLP model in neural networks". In: *Acta Numerica* 8, pp. 143–195. ISSN: 14740508. DOI: 10.1017/S0962492900002919.

Pinto, Rafael Coimbra and Paulo Martins Engel (2015). "A Fast Incremental Gaussian Mixture Model". In: *PLoS ONE* 10.10. ISSN: 19326203. DOI: 10.1371/journal.pone.0139931. eprint: 1506.04422.

Poupart, Pascal, Zhitang Chen, Priyank Jaini, Fred Fung, Hengky Susanto, Yanhui Geng, Li Chen, Kai Chen, and Hao Jin (2016). "Online Flow Size Prediction for Improved Network Routing". In: *IEEE 24th International Conference on Network Protocols (ICNP)*.

Qian, Ning (1999). "On the momentum term in gradient descent learning algorithms". In: *Neural Networks* 12.1, pp. 145–151. ISSN: 08936080. DOI: 10.1016/S0893-6080(98)00116-6.

Radford, Alec, Luke Metz, and Soumith Chintala (2016). "Unsupervised representation learning with deep convolutional generative adversarial networks". In: *4th International Conference on Learning Representations, ICLR 2016 - Conference Track Proceedings*, pp. 1–16. arXiv: 1511.06434.

Raj, Rajendra, Mihaela Sabin, John Impagliazzo, David Bowers, Mats Daniels, Felienne Hermans, Natalie Kiesler, Amruth N. Kumar, Bonnie MacKellar, Renée McCauley, Syed Waqar Nabi, and Michael Oudshoorn (2021). "Professional Competencies in Computing Education: Pedagogies and Assessment". In: *Proceedings of the 2021 Working Group Reports on Innovation and Technology in Computer Science Education*. New York: ACM, 133–161. ISBN: 9781450392020. DOI: 10.1145/3502870.3506570.

Rasmussen, CE. and CKI. Williams (Jan. 2006). *Gaussian Processes for Machine Learning*. Adaptive Computation and Machine Learning. Cambridge, MA, USA: MIT Press, p. 248.

Rastegarfar, Houman, Madeleine Glick, Nicolaas Viljoen, Mingwei Yang, John Wissinger, Lloyd LaComb, and Nasser Peyghambarian (2016). "TCP Flow Classification and Bandwidth Aggregation in Optically Interconnected Data Center Networks". In: *Journal of Optical Communications and Networking 8*.

Ratcliff, Roger (1990). "Connectionist Models of Recognition Memory: Constraints Imposed by Learning and Forgetting Functions". In: *Psychological Review* 97.2, pp. 285–308. ISSN: 0033295X. DOI: 10.1037/0033-295X.97.2.285.

Rebuffi, Sylvestre Alvise, Alexander Kolesnikov, Georg Sperl, and Christoph H. Lampert (2017). "iCaRL: Incremental classifier and representation learning". In: *Proceedings - 30th IEEE Conference on Computer Vision and Pattern Recognition, CVPR 2017* 2017-Janua, pp. 5533–5542. DOI: 10.1109/CVPR.2017.587. arXiv: arXiv:1611.07725v2.

Reis, J., M. Rocha, T. K. Phan, D. Griffin, F. Le, and M. Rio (2019). "Deep Neural Networks for Network Routing". In: *2019 International Joint Conference on Neural Networks*.

Richardson, Eitan and Yair Weiss (2018). "On GANs and GMMs". In: *Advances in Neural Information Processing Systems* 2018-Decem.NeurIPS, pp. 5847–5858. ISSN: 10495258. eprint: 1805.12462.

Riemer, Matthew, Ignacio Cases, Robert Ajemian, Miao Liu, Irina Rish, Yuhai Tu, and Gerald Tesauro (2019). "Learning to learn without forgetting by maximizing transfer and minimizing interference". In: *7th International Conference on Learning Representations, ICLR 2019*, pp. 1–31. arXiv: 1810.11910.

Rusek, Krzysztof, José Suárez-Varela, Albert Mestres, Pere Barlet-Ros, and Albert Cabellos-Aparicio (2019). "Unveiling the Potential of Graph Neural Networks for Network Modeling and Optimization in SDN". In: *Proceedings of the 2019 ACM Symposium on SDN Research*. ACM.

Rusu, Andrei A., Neil C. Rabinowitz, Guillaume Desjardins, Hubert Soyer, James Kirkpatrick, Koray Kavukcuoglu, Razvan Pascanu, and Raia Hadsell (2016). "Progressive Neural Networks". In: arXiv: 1606.04671. URL: http://arxiv.org/abs/1606.04671.

Schlag, Sebastian, Matthias Schmitt, and Christian Schulz (2019). "Faster support vector machines". In: *Proceedings of the Workshop on Algorithm Engineering and Experiments* January.340506, pp. 199–210. ISSN: 21640300. DOI: 10.1137/1.9781611975499.16. arXiv: 1808.06394.

Scholkopf, Bernhard and Alexander J. Smola (2001). *Learning with Kernels: Support Vector Machines, Regularization, Optimization, and Beyond*. Cambridge, MA, USA: MIT Press. ISBN: 0262194759.

Schwarz, Jonathan, Jelena Luketina, Wojciech M. Czarnecki, Agnieszka Grabska-Barwinska, Yee Whye Teh, Razvan Pascanu, and Raia Hadsell (2018). "Progress & compress: A scalable framework for continual learning". In: *35th International Conference on Machine Learning, ICML 2018* 10, pp. 7199–7208. arXiv: 1805.06370.

Serra, Joan, Dídac Suris, Marius Mirón, and Alexandras Karatzoglou (2018). In: *35th International Conference on Machine Learning, ICML 2018* 10, pp. 7225–7234. arXiv: 1801.01423.

Shadmehr, Reza and Sandro Mussa-Ivaldi (2012). *Biological learning and control: how the brain builds representations, predicts events, and makes decisions*. Mit Press.

Sherif Abdelazeem, Ezzat El-Sherif (2010). *The Arabic Handwritten Digits Databases ADBase & MADBase*. URL: http://datacenter.aucegypt.edu/shazeem/ (visited on 06/13/2018).

Shi, Hongtao, Hongping Li, Dan Zhang, Chaqiu Cheng, and Wei Wu (2017). "Efficient and Robust Feature Extraction and Selection for Traffic Classification". In: *Computer Networks* 119.

Shin, Hanul, Jung Kwon Lee, Jaehong Kim, and Jiwon Kim (2017). "Continual learning with deep generative replay". In: *Advances in Neural Information Processing Systems* 2017-Decem.Nips, pp. 2991–3000. ISSN: 10495258. eprint: 1705.08690.

Solinas, M., S. Rousset, R. Cohendet, Y. Bourrier, M. Mainsant, A. Molnos, M. Reyboz, and M. Mermillod (2021). "Beneficial effect of combined replay for continual learning". In: *ICAART 2021 - Proceedings of the 13th International Conference on Agents and Artificial Intelligence* 2.Icaart, pp. 205–217. DOI: 10.5220/0010251202050217.

Song, Mingzhou and Hongbin Wang (2005). "Highly Efficient Incremental Estimation of Gaussian Mixture Models for Online Data Stream Clustering". In: *Intelligent Computing: Theory and Applications III* 5803, p. 174. DOI: 10.1117/12.601724.

Srivastava, Rupesh Kumar, Jonathan Masci, Sohrob Kazerounian, Faustino Gomez, and Jürgen Schmidhuber (2013). "Compete to Compute". In: *Nips* 26, pp. 2310–2318. ISSN: 10495258. URL: https://proceedings.neurips.cc/paper/2013/file/8f1d43620bc6bb580df6e80b0dc05c48-Paper.pdf.

Sutskever, Ilya, James Martens, George Dahl, and Geoffrey Hinton (2013). "On the importance of initialization and momentum in deep learning". In: *ICASSP, IEEE International Conference on Acoustics, Speech and Signal Processing - Proceedings* 2010, pp. 8609–8613. ISSN: 15206149. DOI: 10.1109/ICASSP.2013.6639346. eprint: arXiv:1301.3605v3.

Tang, Yichuan, Ruslan Salakhutdinov, and Geoffrey Hinton (2012). "Deep mixtures of factor analysers". In: *Proceedings of the 29th International Conference on Machine Learning, ICML 2012* 1, pp. 505–512. eprint: 1206.4635.

Thanh-Tung, Hoang and Truyen Tran (2018). "On Catastrophic Forgetting and Mode Collapse in Generative Adversarial Networks". In: arXiv: 1807.04015. URL: http://arxiv.org/abs/1807.04015.

Thrun, Sebastian (1996a). *Explanation-Based Neural Network Learning: A Lifelong Learning Approach*. ISBN: 9781461285977.

Thrun, Sebastian (1996b). "Is Learning The n-th Thing Any Easier Than Learning The First?" In: *Advances in Neural Information Processing Systems*, p. 7. ISSN: 1049-5258. URL: http://citeseer.ist.psu.edu/viewdoc/summary?doi=10.1.1.44.2898.

Tieleman, T. and G. Hinton (2012). *Lecture 6.5 - RmsProp: Divide the gradient by a running average of its recent magnitude*. COURSERA: Neural Networks for Machine Learning.

Tipping, Michael E. and Christopher M. Bishop (1999). "Mixtures of Probabilistic Principal Component Analyzers". In: *Neural Computation* 11.2, pp. 443–482. ISSN: 0899-7667. DOI: 10.1162/089976699300016728. URL: https://direct.mit.edu/neco/article/11/2/443-482/6238.

Titterington, D. M. (1984). "Recursive Parameter Estimation Using Incomplete Data". In: *Journal of the Royal Statistical Society. Series B (Methodological)* 46.2, pp. 257–267. ISSN: 00359246. URL: http://www.jstor.org/stable/2345509.

Trappenberg, Thomas (2009). *Fundamentals of computational neuroscience*. OUP Oxford.

Tsymbal, A (2004). "The Problem of Concept Drift: Definitions and Related Work". In: *Computer Science Department, Trinity College Dublin* May.

Valadarsky, Asaf, Michael Schapira, Dafna Shahaf, and Aviv Tamar (2017). "Learning to Route". In: *Proceedings of the 16th ACM Workshop on Hot Topics in Networks - HotNets-XVI*, pp. 185–191. DOI: 10.1145/3152434.3152441. URL: http://dl.acm.org/citation.cfm?doid=3152434.3152441.

Van Den Oord, Aäron and Benjamin Schrauwen (2014). "Factoring variations in natural images with deep Gaussian mixture models". In: *Advances in Neural Information Processing Systems* 4.January, pp. 3518–3526. ISSN: 10495258.

Ven, Gido M. van de and Andreas S. Tolias (2019). "Three scenarios for continual learning". In: pp. 1–18. arXiv: 1904.07734. URL: http://arxiv.org/abs/1904.07734.

Verbeek, J. J., N. Vlassis, and B. J.A. Kröse (2005). "Self-organizing mixture models". In: *Neurocomputing* 63.SPEC. ISS. Pp. 99–123. ISSN: 09252312. DOI: 10.1016/j.neucom.2004.04.008.

Vijayakumar, Sethu, Aaron D'Souza, and Stefan Schaal (2005). "Incremental online learning in high dimensions". In: *Neural Computation* 17.12, pp. 2602–2634. ISSN: 08997667. DOI: 10.1162/089976605774320557.

Vinet, Luc and Alexei Zhedanov (2011). "A 'missing' family of classical orthogonal polynomials". In: *Journal of Physics A: Mathematical and Theoretical* 44.8, pp. 1–60. ISSN: 17518113. DOI: 10.1088/1751-8113/44/8/085201. eprint: 1011.1669.

Viroli, Cinzia and Geoffrey J. McLachlan (2019). "Deep Gaussian mixture models". In: *Statistics and Computing* 29.1, pp. 43–51. ISSN: 15731375. DOI: 10.1007/s11222-017-9793-z. eprint: 1711.06929.

Vlassis, Nikos and Aristidis Likas (2002). "A greedy EM algorithm for Gaussian mixture learning". In: *Neural Processing Letters* 15.1, pp. 77–87. ISSN: 13704621. DOI: 10.1023/A:1013844811137.

Wang, Heng and Zubin Abraham (2015). "Concept drift detection for streaming data". In: *2015 international joint conference on neural networks (IJCNN)*. IEEE, pp. 1–9.

Wang, Mowei, Yong Cui, Xin Wang, Shihan Xiao, and Junchen Jiang (2018). "Machine Learning for Networking: Workflow, Advances and Opportunities". In: *IEEE Network*.

Wang, Shenghui, Stefan Schlobach, and Michel Klein (2010). "What Is Concept Drift and How to Measure It?" In: *Lecture Notes in Computer Science (including subseries Lecture Notes in Artificial Intelligence and Lecture Notes in Bioinformatics)*. Vol. 6317 LNAI. March 2014, pp. 241–256. ISBN: 3642164374. DOI: 10.1007/978-3-642-16438-5_17. URL: http://link.springer.com/10.1007/978-3-642-16438-5_17.

Wang, Ting-Chun, Ming-Yu Liu, Jun-Yan Zhu, Andrew Tao, Jan Kautz, and Bryan Catanzaro (2018). "High-Resolution Image Synthesis and Semantic Manipulation with Conditional GANs". In: *Proceedings of the IEEE Conference on Computer Vision and Pattern Recognition*.

Wani, M Arif, Farooq Ahmad Bhat, Saduf Afzal, and Asif Iqbal Khan (2020). *Advances in deep learning*. Springer.

Widmer, Gerhard and Miroslav Kubát (2004). "Learning in the presence of concept drift and hidden contexts". In: *Machine Learning* 23, pp. 69–101.

Xiao, Han, Kashif Rasul, and Roland Vollgraf (2017). "Fashion-MNIST: a Novel Image Dataset for Benchmarking Machine Learning Algorithms". In: pp. 1–6. eprint: 1708.07747. URL: http://arxiv.org/abs/1708.07747.

Xiao, Peng, Wenyu Qu, Heng Qi, Yujie Xu, and Zhiyang Li (2015). "An Efficient Elephant Flow Detection with Cost-sensitive in SDN". In: *1st International Conference on Industrial Networks and Intelligent Systems (INISCom)*.

Xie, Junyuan, Linli Xu, and Enhong Chen (2012). "Image denoising and inpainting with deep neural networks". In: *Advances in Neural Information Processing Systems* 1, pp. 341–349. ISSN: 10495258.

Yao, H., T. Mai, C. Jiang, L. Kuang, and S. Guo (2019). "AI Routers & Network Mind: A Hybrid Machine Learning Paradigm for Packet Routing". In: *IEEE Computational Intelligence Magazine*.

Yaroslav Bulatov (2011). *Machine Learning, etc notMNIST dataset*. URL: http://yaroslavvb.blogspot.com.br/2011/09/notmnist-dataset.html (visited on 09/03/2018).

Yenamandra, Sriram, Ansh Khurana, Rohit Jena, and Suyash P. Awate (2021). "Learning Image Inpainting from Incomplete Images using Self-Supervision". In: pp. 10390–10397. DOI: 10.1109/ICPR48806.2021.9413049.

Zeiler, Matthew D. (2012). "ADADELTA: An Adaptive Learning Rate Method". In: arXiv: 1212.5701. URL: http://arxiv.org/abs/1212.5701.

Zeiler, Matthew D. and Rob Fergus (2014). "Visualizing and understanding convolutional networks". In: *Lecture Notes in Computer Science (including subseries Lecture Notes in Artificial Intelligence and Lecture Notes in Bioinformatics)* 8689 LNCS.PART 1, pp. 818–833. ISSN: 16113349. DOI: 10.1007/978-3-319-10590-1_53. arXiv: 1311.2901.

Zenke, Friedemann, Ben Poole, and Surya Ganguli (2017). "Continual learning through synaptic intelligence". In: *34th International Conference on Machine Learning, ICML 2017* 8, pp. 6072–6082. ISSN: 2640-3498. arXiv: 1703.04200.

Zhang, Jiao, Tao Huang, Shuo Wang, and Yun-jie Liu (2019). "Future Internet: Trends and Challenges". In: *Frontiers of Information Technology & Electronic Engineering*.

Zhuang, Z., J. Wang, Q. Qi, H. Sun, and J. Liao (2019). "Toward Greater Intelligence in Route Planning: A Graph-Aware Deep Learning Approach". In: *IEEE Systems Journal*.

Zintgraf, Luisa M., Taco S. Cohen, Tameem Adel, and Max Welling (2017). "Visualizing Deep Neural Network Decisions: Prediction Difference Analysis". In: *5th International Conference on Learning Representations, ICLR 2017 - Conference Track Proceedings*, pp. 1–12. arXiv: 1702.04595.

Žliobaitė, Indrė (2010). "Learning under concept drift: an overview". In: *arXiv preprint arXiv:1010.4784*.

# B.  Statutory Declaration

The submitted doctoral dissertation on the subject "Continual Learning with Deep Learning Methods in an Application-Oriented Context" is my own work and to the rules of proper scientific conduct. I did not seek unauthorized assistance of a third party and I have employed no other sources or means except the ones listed. I clearly marked any direct and indirect quotations derived from the works of others. I did not yet present this doctoral dissertation or parts of it at any other higher education institution in Germany or abroad. I hereby confirm the accuracy of the affirmation above. I am aware of the significance of this affirmation and the legal ramifications in case of untrue or incomplete statements. I affirm in lieu of oath that the statements above are to the best of my knowledge true and complete. I agree that for the purpose of assessing plagiarism the dissertation may be electronically forwarded, stored and processed.

_______________________________          _______________________________
                Place, Date                                              Signature